\documentclass[10pt,twocolumn,letterpaper]{article}
\usepackage[pagenumbers]{wacv} 

\usepackage{graphicx}
\usepackage{amsmath}
\usepackage{amssymb}
\usepackage{booktabs}
\usepackage{wrapfig}
\usepackage{colortbl}

\usepackage[pagebackref,breaklinks,colorlinks]{hyperref}

\usepackage[capitalize]{cleveref}
\crefname{section}{Sec.}{Secs.}
\Crefname{section}{Section}{Sections}
\Crefname{table}{Table}{Tables}
\crefname{table}{Tab.}{Tabs.}

\usepackage{comment}
\usepackage{multirow}

\DeclareMathOperator*{\protoidx}{m}
\DeclareMathOperator*{\groupidx}{n}

\begin{document}

\title{Multi-Scale Grouped Prototypes for Interpretable Semantic Segmentation}

\author{Hugo Porta\textsuperscript{1} \hspace{1em} Emanuele Dalsasso\textsuperscript{1} \hspace{1em} Diego Marcos\textsuperscript{2,3} \hspace{1em} Devis Tuia\textsuperscript{1}\\
\textsuperscript{1}EPFL \hspace{1em} \textsuperscript{2}Inria \hspace{1em} \textsuperscript{3}University of Montpellier\\
\tt\small \{hugo.porta, emanuele.dalsasso, devis.tuia\}@epfl.ch \\
\tt\small diego.marcos@inria.fr
}

\maketitle

\begin{abstract}
   Prototypical part learning is emerging as a promising approach for making semantic segmentation interpretable. The model selects real patches seen during training as prototypes and constructs the dense prediction map based on the similarity between parts of the test image and the prototypes. This improves interpretability since the user can inspect the link between the predicted output and the patterns learned by the model in terms of prototypical information. In this paper, we propose a method for interpretable semantic segmentation that leverages multi-scale image representation for prototypical part learning. First, we introduce a prototype layer that explicitly learns diverse prototypical parts at several scales, leading to multi-scale representations in the prototype activation output. Then, we propose a sparse grouping mechanism that produces multi-scale sparse groups of these scale-specific prototypical parts. This provides a deeper understanding of the interactions between multi-scale object representations while enhancing the interpretability of the segmentation model. The experiments conducted on Pascal VOC, Cityscapes, and ADE20K demonstrate that the proposed method increases model sparsity, improves interpretability over existing prototype-based methods, and narrows the performance gap with the non-interpretable counterpart models. Code is available at \href{https://github.com/eceo-epfl/ScaleProtoSeg}{github.com/eceo-epfl/ScaleProtoSeg}.
\end{abstract}

\vspace{-1em}
\section{Introduction}
\label{sec:intro}
\begin{figure}[ht!]
    \centering
    \includegraphics[width=\columnwidth]{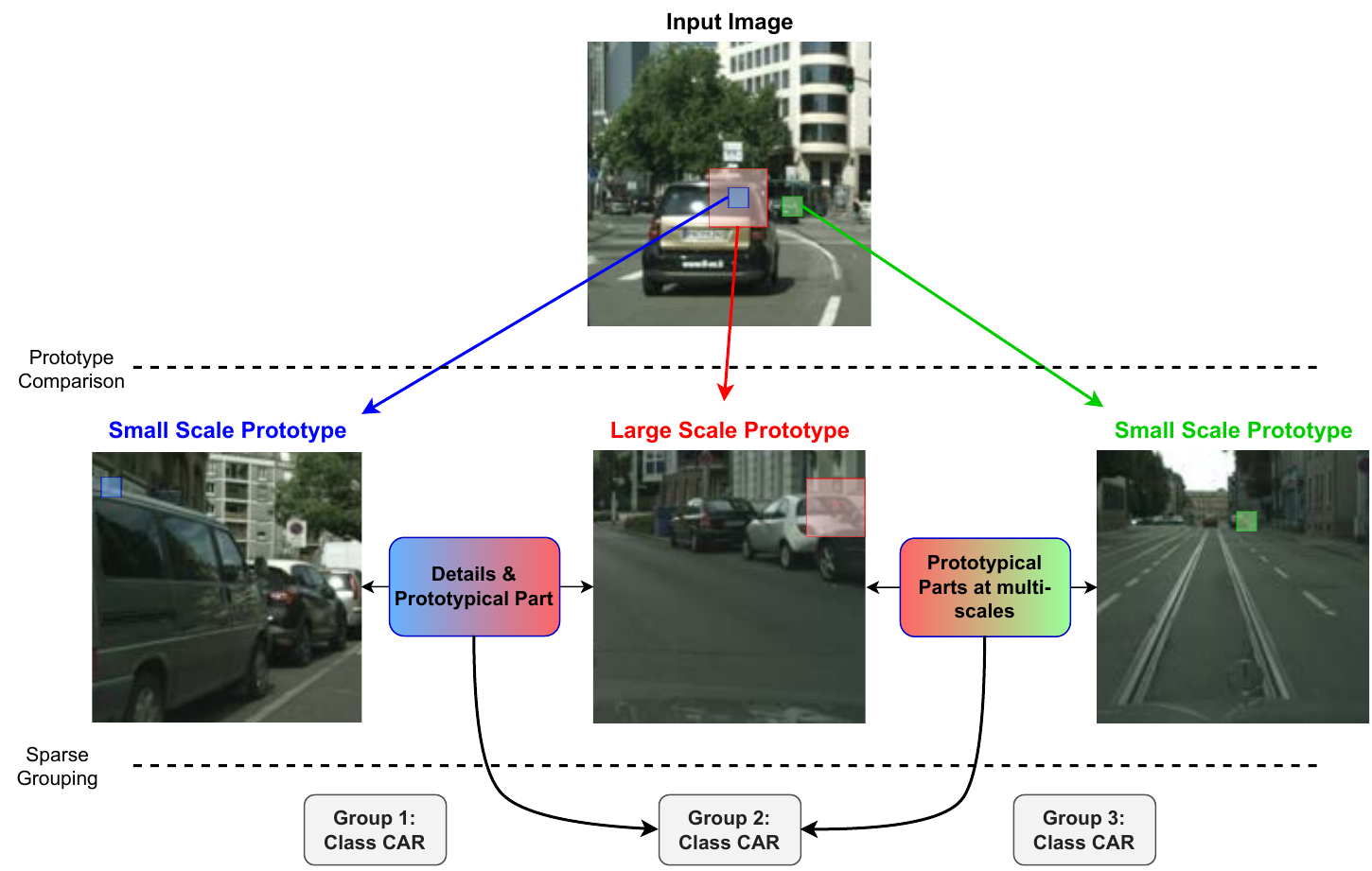}
    \caption{ScaleProtoSeg learns scale-specific prototypes at multiple scales and a sparse prototype grouping to extract patterns referring to different levels of details or scales.\vspace{-2em}}
    \label{fig:group}
\end{figure}

In the last few years, deep learning-based semantic segmentation has seen rapid adoption in numerous fields, from industrial use cases such as autonomous driving~\cite{feng2020deep, siam2018comparative, treml2016speeding} to the environmental sciences~\cite{rashkovetsky2021wildfire, belenguer2019burned, kussul2017deep}. This expansion was driven by its increasing performance on a multitude of tasks and benchmarks, often coinciding with an increase in model complexity~\cite{xie2021segformer, ranftl2021vision, cheng2022masked, wang2023internimage}, which favors un-interpretable black-box models, to the detriment of their explainability.

The models' lack of interpretability is particularly harmful in high-stakes applications~\cite{rudin2019stop, tuia2021toward} and sometimes prevents the wider adoption of deep learning models in applied fields involving high-stakes decision making~\cite{petch2022opening,stiglic2020interpretability,ghassemi2021false}. Indeed, deep learning models can rely on spurious cues, such as Clever-Hans predictors, often symptoms of data contamination~\cite{lapuschkin2019unmasking, adebayo2021post}. This is detrimental in real use cases due to the potential lack of generalization and opaqueness of the decision process. Issues in the generalization of deep learning models are also illustrated by adversarial examples, which can be engineered via small, imperceptible perturbations of images~\cite{kurakin2018adversarial} or be inherent in natural images~\cite{hendrycks2021natural}.

The field of eXplainable Artificial Intelligence (XAI) aims at alleviating the risks associated with the lack of model interpretability by presenting some aspects of the models' decision process into a form understandable to humans~\cite{doshi2017towards}. To produce explainable results, the model design often requires some simplifications that reduce the performance of the original, more complex black-box counterpart~\cite{linardatos2020explainable}. Most methods focus on classification and regression tasks~\cite{kim2018interpretability, goyal2019counterfactual, selvaraju2017grad, koh2020concept, agarwal2021neural}, including prototype-based approaches~\cite{chen2019looks}. A few works aim at making semantic segmentation models interpretable~\cite{vinogradova2020towards, santamaria2020towards, sacha2023protoseg, hoyer2019grid}. However, no approach considers questions about the relationship between prototypes across semantic classes and scales, despite instances of objects appearing at different positions in the image or at multiple range of distances. Accounting for these redundancies allows for learning more diverse prototypes for a given object across scales, in turn leading to more explicit interpretability~\cite{marcos2020contextual}.

In this paper, we tackle these gaps and propose a method for multi-scale interpretable semantic segmentation, ScaleProtoSeg (see Figure~\ref{fig:group}), that \textit{(i)} explicitly learns prototypes at several scales and \textit{(ii)} groups the scale-specific prototypes thanks to a sparse grouping mechanism that provides information on the interaction between prototypical parts at multiple scales, while reducing the number of active prototypes contributing to the decision. Multi-scale prototype learning disentangles the information at different scales so that the model and the users have access to different levels of contextual information in the interpretable decision process. The sparse grouping mechanism allows a transparent understanding of the interaction of scale-specific representations such as object details and parts or similar parts at multi-scale (see Figure~\ref{fig:group}), while maintaining parts correspondences via regularization, avoiding altogether prototype pruning. We are the first to jointly leverage multi-scale representation and prototype learning in an interpretable semantic segmentation model.

We test our methods on three semantic segmentation benchmarks: Pascal VOC 2012~\cite{pascal-voc-2012}, Cityscapes~\cite{cordts2016cityscapes}, and ADE20K \cite{zhou2017scene}, by considering DeepLabv2~\cite{chen2017deeplab} as our base model architecture. We first show that multi-scale prototype learning improves the performance of single-scale prototype-based interpretable semantic segmentation methods (with similar amounts of prototypes) across all considered benchmarks. Moreover, thanks to the sparse grouping mechanism, we demonstrate that constraining the decision process to a small group of prototypes per class enforces interpretability while retaining competitive performance. The contributions of our paper can be summarized as follows:
\begin{itemize}
    \item we propose a multi-scale prototype layer that enforces the model to focus on the prototypical parts' representations at multiple scales;

    \item we define a grouping procedure that learns sparse combinations of the scale-specific prototypes across all scales and increases the interpretability of the decision process;

    \item not only we show the superiority of our ScaleProtoSeg method in three popular datasets in semantic segmentation over the prototype-based method~\cite{sacha2023protoseg}, but also highlight its improved interpretability measured in terms of \textit{stability}, \textit{consistency} and \textit{sparsity}.
\end{itemize}

\section{Related works}
\label{sec:reference}

\paragraph{Semantic segmentation.} 
Fully Convolutional Networks (FCN)~\cite{long2015fully} are widely used in semantic segmentation methods. They are based on an encoder-decoder architecture, where the encoder extracts discriminative features from the input image and the decoder converts the learned semantic representation into per-pixel predictions. Following FCN, researchers focused on improving different aspects of the semantic segmentation methods such as enlarging the model receptive field while limiting the parameters increase ~\cite{yu2015multi, zhao2017pyramid, chen2018encoder}, specifying boundary information~\cite{takikawa2019gated, yu2018learning, ding2019boundary}, or providing contextual information~\cite{zhang2018context, lin2017refinenet, he2019adaptive}. Some methods proposed specific modules learning pixel affinities or attention~\cite{liu2017learning, fu2019dual, huang2019ccnet, zhao2018psanet} to allow the network to base its prediction also on similar pixels that do not lie in the direct vicinity of the pixel at hand. 
Furthermore, there has been a growing interest in how multi-scale information could be extracted. A common pattern in decoder architectures is the concatenation of scale-specific feature maps at multiple scales~\cite{ranftl2021vision, xie2021segformer, ronneberger2015u}. For instance, Chen et al.~\cite{chen2017deeplab} introduce atrous spatial pyramid pooling (ASPP) to learn in parallel a multi-scale representation from a single-scale feature map. We focus on providing an interpretable version of this model relying on its widely employed decoder multi-scale architecture.
\paragraph{Explainable artificial intelligence.} XAI methods can be split into \textit{post-hoc} vs \textit{by-design} approaches. Post-hoc methods aim to explain black-box models after training by using an auxiliary method to generate explanations, while by-design approaches enforce interpretability in the model itself. Our proposed method falls into the latter category. Some examples of interpretable by-design approaches are concept bottleneck models~\cite{koh2020concept, jeyakumar2022automatic, marcos2020contextual, oikarinen2023label}; attention modules~\cite{zhou2016learning, zheng2017learning, rigotti2021attention, serrano2019attention} which point to critical parts of each input sample; generalized additive models~\cite{hastie1987generalized, agarwal2021neural}; and prototype learning introduced in~\cite{chen2019looks, li2018deep}, where part of an encoded input image is compared to a set of class-specific prototypes, represented by training samples. Several extensions followed the original paper \cite{chen2019looks}, aiming at enforcing orthogonality in the prototype construction~\cite{wang2021interpretable, donnelly2022deformable}, reducing the number of prototypes~\cite{rymarczyk2021protopshare}, or even leveraging label taxonomy via a hierarchical structure~\cite{hase2019interpretable}. Prototype learning can be extended to other tasks beyond image classification such as sequence learning~\cite{ming2019interpretable} and time series analysis~\cite{gee2019explaining} and, most recently, semantic segmentation~\cite{sacha2023protoseg}. In this work, we rely on prototype learning for interpretable semantic segmentation.
\paragraph{Interpretable semantic segmentation.} Among the few existing methods tackling this problem, extensions of Grad-CAM to semantic segmentation have been explored \cite{vinogradova2020towards,hoyer2019grid} for post-hoc methods. By-design approaches can provide explainable results by leveraging symbolic language \cite{santamaria2020towards}, through the use of a semantic bottleneck \cite{losch2019interpretability} or by exploiting the attention mechanism~\cite{gu2020net}. Prototypical parts learning was proposed recently in ProtoSeg~\cite{sacha2023protoseg} for interpretable semantic segmentation, extending the classification-based method proposed in~\cite{chen2019looks}. In this work, we aim to investigate multi-scale prototype learning to break the performance/interpretability trade-off present in ProtoSeg. This is achieved by explicitly leveraging the multi-scale nature of semantic segmentation representations (via scale-specific prototypes) and by encouraging sparsity (via sparse grouping across prototypes and scales).
\paragraph{Prototypes and semantic segmentation.} Prototype learning based on parametric or non-parametric prototypes has seen an increase of interest in image classification \cite{mettes2019hyperspherical, yang2018robust}, few-shot and zero-shot \cite{snell2017prototypical, jetley2015prototypical, xu2020attribute}, unsupervised \cite{wu2018unsupervised} and self-supervised learning \cite{li2020prototypical}. For semantic segmentation, parametric prototypes are used to represent unique class embeddings, both in an explicit \cite{strudel2021segmenter, jain2021scaling, cheng2021per} or implicit way \cite{zhou2022rethinking}. Non-parametric prototypes were also leveraged, first in the few-shot learning context \cite{dong2018few, wang2019panet} and then in more classic semantic segmentation models \cite{wang2022visual, zhou2022rethinking, hwang2019segsort}. 
The methods in \cite{wang2022visual, zhou2022rethinking} are the closest to our proposition: both compute multiple class-specific prototypes via online clustering and leverage metric learning losses to enforce a compact embedding space. The pixel-wise classification is then performed via nearest-prototype assignment. Despite those similarities, those works compute prototypes in the final embedding space, while we focus on enhanced interpretability via i) prototypical parts through specific regularization and ii) multi-scale object representation. These core differences limit the applicability of direct performance comparisons. Finally,  \cite{tang2023holistic, tang2023prototransfer} leverage non-parametric prototypes for modality and domain alignment for multi-modal semantic segmentation (e.g. video and point clouds), but do not explore issues related to interpretability or multi-scale prototype learning as our proposed method.

\section{Method}
\label{sec:method}

\begin{figure*}[h]
 \captionsetup[subfigure]{justification=centering}
  \begin{subfigure}{\textwidth}
		\includegraphics[width=\textwidth]{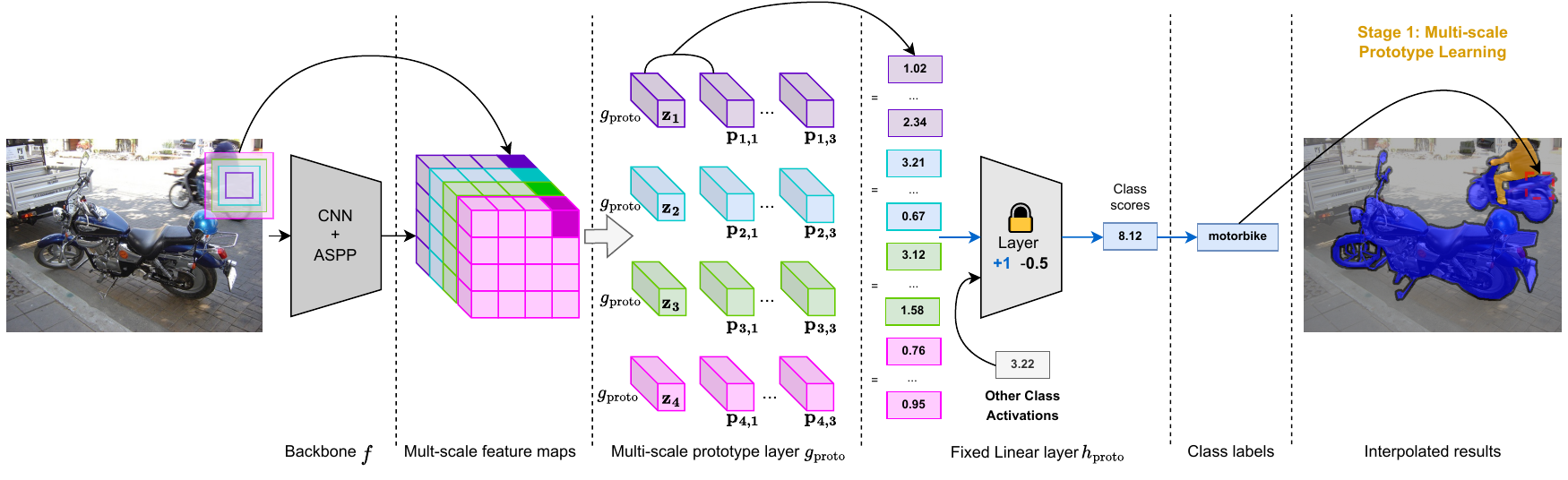}
  \subcaption{\textbf{Stage 1:} Multi-scale prototype learning.}
\end{subfigure}
\begin{subfigure}{\textwidth}
		\includegraphics[width=\textwidth]{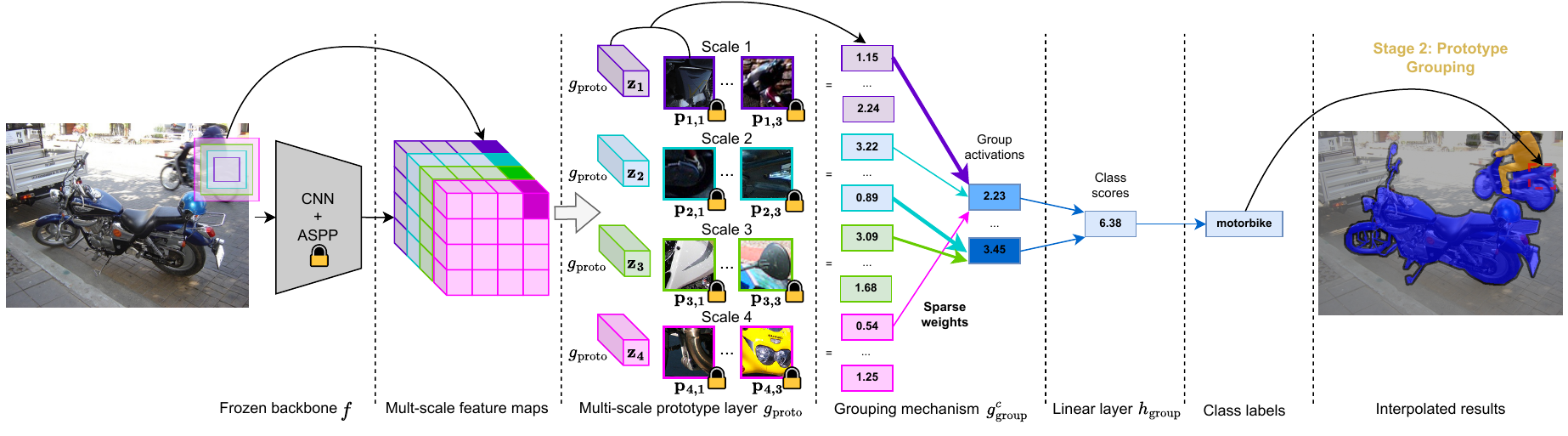}
  \subcaption{\textbf{Stage 2:} Prototype grouping.}
    \end{subfigure}
  \caption{Overall architecture of ScaleProtoSeg. Each color in the feature maps and following layers corresponds to a specific scale ($S=4$ and $M = 3$ in this illustration).\vspace{-1em}}
    \label{fig:method}
\end{figure*}

In this section, we present our multi-scale grouped prototypes method for interpretable semantic segmentation: ScaleProtoSeg (Figure~\ref{fig:method}). We first introduce the multi-scale prototype architecture (Section~\ref{sec:multi-scale}). Then we describe the proposed grouping mechanism to extract sparse groups of prototypes across scales (Section~\ref{sec:group}). Lastly, we detail the multi-stage training procedure used to learn both prototypes and groups (Section~\ref{sec:training-proc}).

\subsection{Multi-scale prototype learning}
\label{sec:multi-scale}

Our model architecture for multi-scale prototype learning is presented in Figure~\ref{fig:method}: \textbf{Stage 1}. It is composed of a backbone network $f$, a multi-scale prototype layer $g_\text{proto}$, and a linear layer $h_\text{proto}$. For an input RGB image $\mathbf{x} \in \mathbb{R}^{H \times W \times 3}$, $f$ outputs a multi-scale feature map $f(\mathbf{x}) \in \mathbb{R}^{H_{r} \times W_{r} \times S \times d}$, representing scale-specific features at $S$ scales, each one with $d$ dimensions. The scalar $r$ is a size reduction factor. In practice, we modify the ASPP layer from~\cite{chen2017deeplab} to concatenate the scale-specific feature maps instead of summing them. At each scale $s \in \mathcal{S}$, let $\mathbf{z}_{s} \in \mathbb{R}^{d}$ be a vector from the scale-specific feature map $f_{s}(\mathbf{x})$. As illustrated in Figure \ref{fig:method}, this vector represents the features extracted from an area of the input image corresponding to the receptive field at scale $s$. The multi-scale prototype layer $g_\text{proto}$ is composed of $M$ learnable scale-dependent prototypes, where the $\protoidx_\text{th}$ scale-dependent prototype (with $\protoidx \in [1,\dots,M]$) is described by the vector $\mathbf{p}_{s, \protoidx} \in \mathbb{R}^{d}$, which is randomly initialized and learned through gradient descent. For each feature vector $\mathbf{z}_{s}$, the prototypes' activations are computed following ProtoPNet~\cite{chen2019looks}:
\begin{equation}
    g_\text{proto}(\mathbf{z}_{s}, \mathbf{p}_{s, \protoidx}) = \log \left(\frac{\|\mathbf{z}_{s} - \mathbf{p}_{s, \protoidx}\|_{2}^{2} + 1}{\|\mathbf{z}_{s} - \mathbf{p}_{s, \protoidx}\|_{2}^{2} + \epsilon} \right)
\end{equation}

with $\epsilon \ll 1$ a constant for numerical stability. Then, for each scale-specific feature vector $\mathbf{z}_{s}$, the $M$ activation scores $\left[ g_\text{proto}(\mathbf{z}_{s}, \mathbf{p}_{s, 1}),\dots, g_\text{proto}(\mathbf{z}_{s}, \mathbf{p}_{s, M})\right]$ are concatenated across the $S$ scales indexed by $s$. The linear layer $h_\text{proto}$, with weight matrix $\mathbf{w}_{h_\text{proto}} \in \mathbb{R}^{C \times M\cdot S}$, learns a mapping from those prototype activations to the $C$ output classes probabilities. The classification probabilities map, obtained after processing all feature vectors $\mathbf{z} = \left[\mathbf{z}_1,\dots,\mathbf{z}_S\right]$ in parallel, is of dimension $H_{r} \times W_{r} \times C$. To upscale it to the original image resolution and produce the final segmentation map, the classification probability is linearly interpolated.

For semantic segmentation, the objective is to learn prototypes that are assigned to a specific class $c \in \mathcal{C}$. This is enforced by initializing the weights $\mathbf{w}_{h_\text{proto}}$ of the linear layer $h_\text{proto}$ following~\cite{chen2019looks}, and replicating the same assignment to a specific class across all scales, such as with:
$$  w_{h_\text{proto}}^{(c,\protoidx)} =
    \left \{
    \begin{aligned}
    &1 && \text{if}\ \mathbf{p}_{\protoidx} \in P_c \\
    &-0.5 && \text{otherwise}
  \end{aligned} \right.$$
where $P_{c}$ is the set of prototypes that we assign to class $c \in \mathcal{C}$, across all scales and $\protoidx \in [1, \dots, |P_{c}|]$. Those weights are frozen for the majority of the training procedure (see Section~\ref{sec:training-proc}) to maintain the steering of the prototypes toward class-specific patterns.

\subsection{Prototype grouping}
\label{sec:group}
Our model architecture for prototype grouping is presented in Figure~\ref{fig:method}: \textbf{Stage 2}.
Once the scale-specific prototypes $\mathbf{p}_{s, \protoidx}$ are learned through the multi-scale learning stage as described in Section~\ref{sec:multi-scale}, we group them into sparse groups across scales. For this purpose, we use class-specific grouping functions: $g_{c}(\mathbf{z}) = g_\text{group}^{c}(g_\text{proto}(\mathbf{z}, P_{c}))$, which group prototypes assigned to the same class and compute the groups' activations $g_{c}(\mathbf{z}) \in \mathbb{R}^{N_{c}}$,
where $N_{c}$ is the number of groups per class. Following~\cite{marcos2020contextual}, those functions are parametrized by sparse non-negative weight matrices $\mathbf{w}_{g, c} \in [0, 1]^{N_{c} \times |P_{c}|}$, where each row is constrained on the probability simplex: ${w}_{g, c}^{(\groupidx, \protoidx)} \geq 0$ and $\sum_{\protoidx}{{w}_{g, c}^{(\groupidx, \protoidx)}} = 1, \forall \groupidx \in [1, \dots, N_{c}]$. The output group activation for the group $g_{c, \groupidx}(\mathbf{z})$ that combines the prototypes $\mathbf{p}_{\protoidx} \in P_{c}$ at multiple scales is computed as follows:

\begin{equation}
    g_{c, \groupidx}(\mathbf{z}) = \prod_{\protoidx=1,\dots,|P_{c}|}{g_\text{proto}(\mathbf{z}, \mathbf{p}_{\protoidx})^{{w}_{g, c}^{(\groupidx, \protoidx)}}}.
\end{equation}

The projection on the simplex and the weighted geometric mean are used to learn sparse weight matrices, as group activations can be high only if all prototypes in the group are strongly activated. The linear layer $h_\text{group}$ is adapted to accommodate groups through a weight matrix $\mathbf{w}_{h_\text{group}} \in \mathbb{R}^{C \times N}$, where $N=N_c\times C$ is the total number of groups, each one containing a sparse combination of the set of $M\times S$ prototypes. The layer $h_\text{group}$ follows the same initialization process as in Section \ref{sec:multi-scale}, but with group assignment.

\subsection{Multi-stage training procedure}
\label{sec:training-proc}
In order to train ScaleProtoSeg, we resort to a two-stage training procedure: first, we learn the multi-scale prototypes and project them to the training set patches, without any grouping. Then we learn the prototypes grouping functions with the prototypes fixed. The two stages are illustrated in Figure~\ref{fig:method} and detailed below.

\paragraph{Stage 1: Multi-scale prototype learning.} For the multi-scale prototype learning we apply a training procedure similar to~\cite{chen2019looks, sacha2023protoseg}, which consists of three steps. First, in a warm-up step, the ASPP~\cite{chen2017deeplab} and the scale-specific prototypes are trained while freezing the rest of the backbone $f$ and the last linear layer $h_\text{proto}$. Second, we run a joint optimization stage where all the model is trained except the last linear layer $h_\text{proto}$. Third, for all scales $s \in \mathcal{S}$, the prototypes are projected to their nearest vector $\mathbf{z}_{s}$ from the training set, and duplicates are removed. The fine-tuning stage from~\cite{sacha2023protoseg} is not necessary for ScaleProtoSeg, as we replace the last layer $h_\text{proto}$ with $h_\text{group}$ in the second stage below. Moreover, contrary to~\cite{chen2019looks, sacha2023protoseg}, we do not need to run their pruning algorithm, as the sparse grouping mechanism will also enforce a natural decrease in the number of prototypes used.

In all those steps, we apply as a regularization loss the diversity loss~\cite{sacha2023protoseg}, which aims to prevent the prototypes from activating the same region of an object.
To enforce diversity among scale-specific prototypes, the diversity loss is evaluated independently at each scale $s \in \mathcal{S}$.
Indeed, the multi-scale prototype layer aims to represent similar parts across scales with different contextual information. For each class $c \in \mathcal{C}$ and scale $s \in \mathcal{S}$, we note $P_{s, c}$ as the set of prototypes assigned to $c$ and $s$. Furthermore, for the scale-specific feature map $f_{s}(\mathbf{x})$ we note $y_{\mathbf{z}} \in \{1, \dots, C \}$ as the ground truth labels for each vector $\mathbf{z}$ across all scales.\\
First, for the diversity loss, we define $v(f_{s}(\mathbf{x}), \mathbf{p}_{s,\protoidx})$ as the softmax vector of distances between each vector assigned to a class $c$, and a prototype $\mathbf{p}_{s,\protoidx} \in P_{s, c}$:
\begin{align}
    v(f_{s}(\mathbf{x}), \mathbf{p}_{s,\protoidx}) = & \text{softmax}(\|\mathbf{z}_{s} - \mathbf{p}_{s,\protoidx}\|^2 : \nonumber \\ &  \forall \mathbf{z}_{s} \in f_{s}(\mathbf{x}), y_{\mathbf{z}} = c)
\end{align}
Next we use the Jeffrey similarity $S_{J}(U_{1}, \dots , U_{l})$ as a measure of similarity between distributions $(U_{1}, \dots, U_{l})$, following ProtoSeg \cite{sacha2023protoseg}:
\begin{equation}
        S_{J}(U_{1}, \dots , U_{l}) = \frac{1}{\binom{l}{2}} \sum_{i < j}\exp(-D_{J}(U_{i}, U_{j}))
\end{equation}
with $D_{J}(U, V)$ the Jeffrey's divergence defined in~\cite{jeffreys1998theory}. The diversity loss is then computed as follows for a given class $c$ and scale $s$:
\begin{align}
    L_{J}(f_{s}(\mathbf{x}), P_{s, c}) = & S_{J}(v(f_{s}(\mathbf{x}), \mathbf{p}_{s, 1}), \dots , \nonumber \\ & v(f_{s}(\mathbf{x}), \mathbf{p}_{s, M}))
\end{align}
and the total loss across all classes and scales is:
\begin{equation}
    L_{J} = \frac{1}{C \cdot S} \sum_{c \in \mathcal{C}} \sum_{s \in \mathcal{S}}L_{J}(f_{s}(\mathbf{x}), P_{s, c})
\end{equation}
The final loss term for the multi-scale prototype learning (\textbf{Stage 1}) becomes:
\begin{equation}
    L_\text{proto} = L_\text{CE} + \lambda_{J} \cdot L_{J}
\end{equation}
with $L_\text{CE}$ the per patch cross-entropy loss and  $\lambda_{J}$ a hyperparameter controlling the weight of the regularization.

\paragraph{Stage 2: Prototype grouping mechanism.} The multi-scale prototype grouping mechanism is applied to the learned scale-specific prototypes after they are projected to the training set patches. We proceed with two steps: we first run a warm-up step to train the class-specific grouping functions $g_\text{group}^{c}$ while keeping the rest of the model frozen, namely the backbone $f$ and the prototypes, so that we maintain interpretability. Then, in the second step, we run a joint training phase where we finetune both the grouping functions and the last linear layer $h_\text{group}$.
During the learning of the groups, we apply a sparsity regularization term promoting that only a limited subset of prototypes are active for a given group. For this purpose, we define an entropic loss term, $L_\text{ent}$, on the weight matrices $\mathbf{w}_{g,c} \in [0, 1]^{N \times |P_{c}|}$. Indeed, as we are projecting each row of those weight matrices $\mathbf{w}_{g, c}^{(\groupidx, :)}$ on the probability simplex it is possible to directly compute and minimize information theory measures such as the entropy of those rows, to enforce a sparse combination of prototypes among each group. The entropy loss is computed as follows:
\vspace{-1ex}
\begin{equation}
    L_\text{ent} = \frac{1}{C \cdot N_{c}}\sum_{c \in \mathcal{C}}\sum_{\groupidx = 1}^{N_{c}}\sum_{\protoidx = 1}^{M} - w_{g, c}^{(\groupidx, \protoidx)}\log(w_{g, c}^{(\groupidx, \protoidx)})
\end{equation}
Then, in the second step, we fine-tune also the linear layer $h_{\text{group}}$. To enforce more sparsity, following the training protocol from~\cite{chen2019looks}, we apply an L1-norm loss term on $h_{\text{group}}$ solely on the weights $w_{h_{\text{group}}}^{c, \groupidx}$ where the group $g_{\groupidx}$ is not in the set of groups $G_{c}$ assigned to class $c$. The total loss term for the grouping mechanism (Figure \ref{fig:method}: \textbf{Stage 2}) becomes:
\vspace{-1ex}
\begin{equation}
    L_\text{proto} = L_\text{CE} + \lambda_\text{ent} \cdot L_\text{ent} + \lambda_\text{L1} \cdot \sum_{c \in \mathcal{C}} \sum_{\groupidx: g_{\groupidx} \notin G_{c}} |w_{h_{\text{group}}}^{(c,\groupidx)}|
\label{eq:loss}\end{equation}
where $\lambda_\text{ent}$ and $\lambda_\text{L1}$ are hyperparameters controlling the weight of the regularization terms. Once the whole architecture is trained, a final post-processing stage is done: the weights of the grouping function matrices $\mathbf{w}_{g,c}$ below a certain threshold $\alpha$ are set to $0$. This enforces even more sparsity within the groups.


\section{Experiments}
\label{sec:exp}

\subsection{Experimental setup}
\label{sec:set-up}

In all the experiments presented in the section \ref{sec:result}, we use DeepLabv2~\cite{chen2017deeplab} with ResNet-101~\cite{he2016deep} pre-trained on ImageNet as the backbone. For the multi-scale prototype learning, we assign $M=3$ scale-dependent prototypes per scale to each class and $S=4$ scales, so $12$ prototypes per class in total. Moreover, we set the number of groups per class to $N_{c}=3$ for the grouping mechanism. We evaluate ScaleProtoSeg on \textbf{(i)} Pascal VOC 2012~\cite{pascal-voc-2012} (made of $1464$ train, $1449$ validation, and $1446$ test images with $21$ classes; the Pascal VOC training set is extended to $10582$ images following~\cite{hariharan2011semantic}), \textbf{(ii)} Cityscapes~\cite{cordts2016cityscapes} (composed of $2975$ train, $500$ validation, and $1525$ test images of street scenes, with $19$ classes) and \textbf{(iii)} ADE20K~\cite{zhou2017scene} (a scene-parsing dataset with 150 fine-grained semantic classes split in $\sim 20000$ training and $2000$ validation images; for the training of ProtoSeg~\cite{sacha2023protoseg} on ADE20K we extend the number of prototypes to $12$ for direct comparison with ScaleProtoSeg). A detailed description of the experimental setup is available in the supplementary materials in Section~\ref{sec:exp-set-up-det}.

\subsection{Results and discussion}
\label{sec:result}

\paragraph{Method performance.}
In Table~\ref{tab:main-perf}, we present the performance of our interpretable semantic segmentation method ScaleProtoSeg. Due to the well-known lack of faithfulness of saliency-based methods~\cite{rudin2019stop, adebayo2018sanity} and the difficulty to compare against other by-design methods \underline{not} based on prototypes (different benchmark datasets, different backbones and/or lack of public code repositories)~\cite{losch2019interpretability,santamaria2020towards}, we compare ScaleProtoSeg against ProtoSeg~\cite{sacha2023protoseg} (the closest previous methods in the literature aimed at prototype-based interpretability for semantic segmentation) and the non-interpretable counterpart DeepLabv2~\cite{chen2017deeplab}. It is worth mentioning that, as interpretability comes at the price of constrained and regularized training (namely through prototype projection to real training samples, the use of the diversity loss, and the sparse grouping mechanism), an interpretable model aims to close the gap with the non-interpretable counterpart, which acts as an upper-bound. In this regard, ScaleProtoSeg showcases a substantial improvement over ProtoSeg in terms of mIoU for Cityscapes and ADE20K, namely of $\sim 1.5\%$ and $4\%$. For ADE20K, the proposed ScaleProtoSeg goes beyond its original goal of enabling interpretability by surpassing the performance of its black-box counterpart.
The Cityscapes and ADE20K datasets present more natural images with numerous objects at various scales, unlike in Pascal, where our method provides less of an advantage (results are on par with those of ProtoSeg). Indeed, we hypothesize that learning explicitly scale-specific prototypes at multiple scales is more advantageous in segmentation tasks with a large depth of field. In the supplementary materials, we further demonstrate the transferability of ScaleProtoSeg {(a)} to the medical domain, {(b)} to another segmentation architecture and (c) to a larger benchmark dataset (COCO-Stuff) in Section~\ref{sec:em} and~\ref{sec:coco}.

This observation is confirmed by the results presented in Table~\ref{tab:breakdown}, which showcases the high improvement brought by the projection step of ScaleProtoSeg, performed during the proposed multi-scale prototype learning (see Figure~\ref{fig:method}: \textbf{Stage 1}). This improvement comes with an increase of solely two prototypes per class from 190 to 228 prototypes on Pascal VOC and 210 (201 after deduplication) to 252 on Cityscapes.
Furthermore, in Table~\ref{tab:breakdown} we observe that, through the grouping mechanism and thresholding, we reduce the total number of prototypes used by our model by $51.3\%$, $48.0\%$, and $37.1\%$ for Cityscapes, Pascal, and ADE20K respectively, compared to $32.6\%$, $36.7\%$, and $35.5\%$ for ProtoSeg after deduplication and pruning. As a result, ScaleProtoSeg uses fewer prototypes than ProtoSeg despite a higher initial count except for ADE20K (see Section~\ref{sec:set-up}). The proposed grouping mechanism (see Figure~\ref{fig:method}: \textbf{Stage 2}) trades better interpretability and simplified decision process (interaction of only 3 groups per class) with only $0.4\%$ to $0.8\%$ mIoU loss.

In the supplementary materials, we detail the capability of controlling the sparsity-performance trade-off via the thresholding of the grouping weights matrices and the entropy regularization in Section~\ref{sec:sparse} and~\ref{sec:ent}.

\begin{table}[t!]
\centering
\resizebox{\columnwidth}{!}{
\begin{tabular}{l|cc|cc|c}
& \multicolumn{2}{c|}{\begin{tabular}[c]{@{}c@{}}Cityscapes\\ mIoU\end{tabular}} & \multicolumn{2}{c|}{\begin{tabular}[c]{@{}c@{}}Pascal\\ mIoU\end{tabular}} & \begin{tabular}[c]{@{}c@{}}ADE20K\\ mIoU\end{tabular} \\ \hline
Method                  & \multicolumn{1}{c}{val}               & \multicolumn{1}{c|}{test}              & \multicolumn{1}{c}{val}             & \multicolumn{1}{c|}{test}  & val       \\ \hline
DeepLabv2~\cite{chen2017deeplab} &  71.40                                & 70.40                                  & 77.69                               & 79.70            &  34.00                    \\ \hline
ProtoSeg~\cite{sacha2023protoseg} &   67.54$\pm 0.22$             & 67.04                                  & \textbf{71.98}$\pm 0.11$                               & \textbf{72.92}
& 29.67$\pm 0.23$ \\
ScaleProtoSeg & \textbf{68.97}$\pm 0.25$                        & \textbf{68.52}                                & 71.80$\pm 0.38$                      & 72.35      & \textbf{34.18}$\pm 0.18$                           
\end{tabular}}
\caption{ScaleProtoSeg mIoU performance on Pascal VOC, Cityscapes, and ADE20K. On the validation sets, we report the results over 3 runs, while for the test sets the results are based on ~\cite{sacha2023protoseg} and our best ScaleProtoSeg validation run.}\vspace{-1em}
\label{tab:main-perf}
\end{table}

\begin{table*}[h]
\centering
\resizebox{\textwidth}{!}{
\begin{tabular}{|c|cccc|cccc|cccc|}
\hline
\textbf{Dataset}    & \multicolumn{4}{c|}{Cityscapes}                                            & \multicolumn{4}{c|}{Pascal}                                                & \multicolumn{4}{c|}{ADE20K}                                                \\ \hline
\textbf{Method}     & \multicolumn{2}{c|}{ProtoSeg}         & \multicolumn{2}{c|}{ScaleProtoSeg} & \multicolumn{2}{c|}{ProtoSeg}         & \multicolumn{2}{c|}{ScaleProtoSeg} & \multicolumn{2}{c|}{ProtoSeg}         & \multicolumn{2}{c|}{ScaleProtoSeg} \\ \hline
\textbf{Step}      & project & \multicolumn{1}{c|}{pruned} & project           & grouped        & project & \multicolumn{1}{c|}{pruned} & project           & grouped        & project & \multicolumn{1}{c|}{pruned} & project          & grouped         \\ \hline
\textbf{Prototypes} & 190     & \multicolumn{1}{c|}{128}    & 228               & \textbf{111}   & 201     & \multicolumn{1}{c|}{133}    & 252               & \textbf{131}   & 1756    & \multicolumn{1}{c|}{1161}   & 1800             & \textbf{1132}   \\ \hline
\textbf{mIoU}       & 67.24   & \multicolumn{1}{c|}{67.23}  & \textbf{70.01}    & 69.22          & 72.00   & \multicolumn{1}{c|}{72.05}  & \textbf{72.94}    & 72.26          & 30.89   & \multicolumn{1}{c|}{29.97}  & \textbf{34.72}   & 34.32           \\ \hline
\end{tabular}}
\caption{Methods performance at different training steps based on the results from~\cite{sacha2023protoseg} and our best ScaleProtoSeg run. The number of prototypes indicated in the \textit{project} step (during \textbf{Stage 1}) is after the removal of the duplicates. The number of prototypes in the \textit{grouped} step (during \textbf{Stage 2}) corresponds to a threshold of $0.05$ on the group matrices.}
\label{tab:breakdown}\vspace{-1em}
\end{table*}

\begin{figure}[h]
 \captionsetup[subfigure]{justification=centering}
  \begin{subfigure}{0.48\textwidth}
		\includegraphics[width=\textwidth]{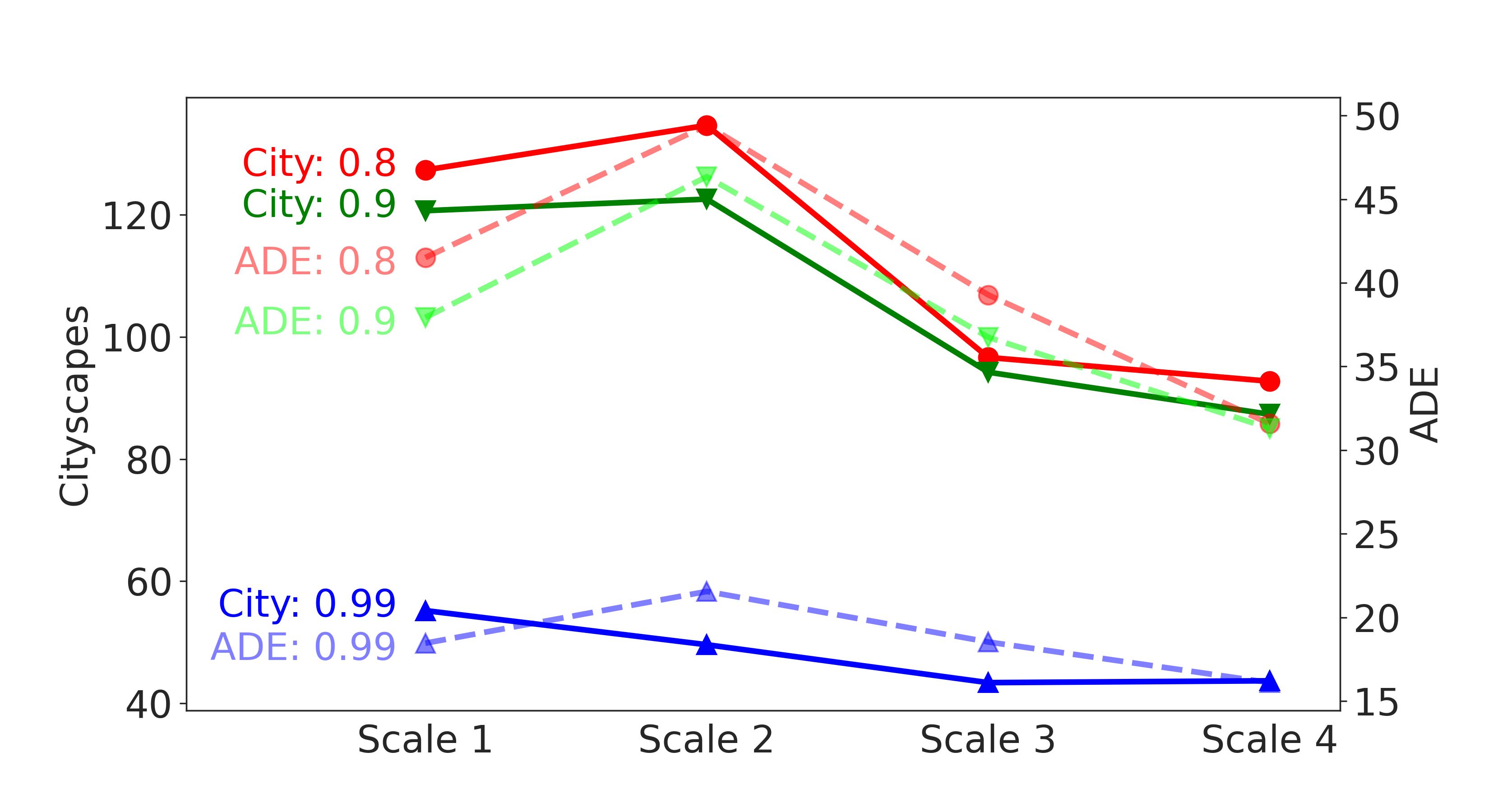}
  \subcaption{Per-scale average number of connected components}
\end{subfigure}
\begin{subfigure}{0.48\textwidth}
		\includegraphics[width=\textwidth]{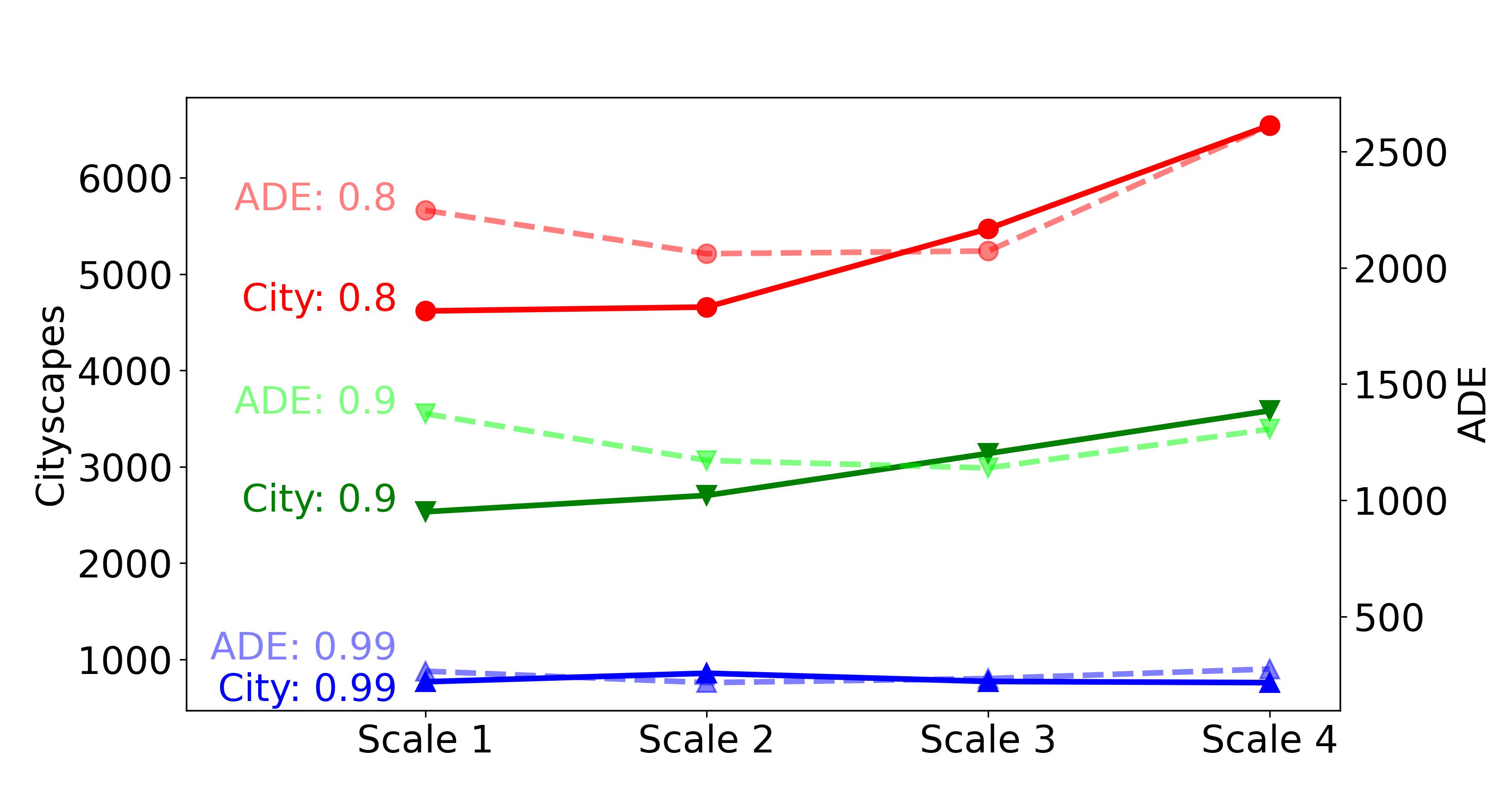}
  \subcaption{Per-scale average pixel area of the connected components}
    \end{subfigure}
  \caption{Analysis of the binarized prototype activations at multiple percentile thresholds $p_{th} \in \{0.8, 0.9, 0.99\}$ on Cityscapes and ADE20K validation sets.\vspace{-2em}}
		\label{fig:comp}
	\end{figure}
 
\paragraph{Multi-scale prototypes analysis.}
After the prototype projection at the end of the multi-scale prototype learning stage, different patterns of prototype activations emerge across scales. In Figure \ref{fig:comp}, we present a quantitative analysis of the prototype activations scale-specific patterns on Cityscapes and ADE20K. For this purpose, we first binarize the activation maps of the multi-scale prototype layer from ScaleProtoSeg on the validation set of both datasets: we do so with multiple percentiles $p_{th} \in \{0.8, 0.9, 0.99\}$. Then we measure the number of connected components in the binarized maps as well as their average area in number of pixels. The average results per scale across all scale-specific prototypes are displayed in Figure~\ref{fig:comp}.

We compute the connected components with the standard algorithm from \textit{opencv} with an 8-connectivity relation between non-zero elements. We observe an overall decrease in the number of connected components when going from \textit{Scale 1} to \textit{Scale 4} across all thresholds and datasets and an increase in the average pixel area of the detected components (especially for ADE20K, which is less focused on images with large depth of field). In practice, this can be interpreted as a consequence of the lower field of view of the feature maps specific to \textit{Scale 1} compared to those of \textit{Scale 4} in the ASPP \cite{chen2017deeplab}, where each scale-specific feature map vector represents a smaller receptive field on the image with less context. As such, the prototypes specific to \textit{Scale 1} can be activated by smaller-scale prototypical parts and texture components. This analysis reflects the advantage of learning explicitly scale-specific prototypes across multiple scales: not only ScaleProtoSeg shows a performance improvement (see Table~\ref{tab:breakdown}), but it also learns diverse object representations across scales, highlighting the multi-scale nature of semantic segmentation. In the supplementary material, we also analyzed if the prototypes present equivariance properties across scales in Section~\ref{sec:scale-analysis}.

\begin{figure*}[h]
    \centering
    \includegraphics[width=\textwidth]{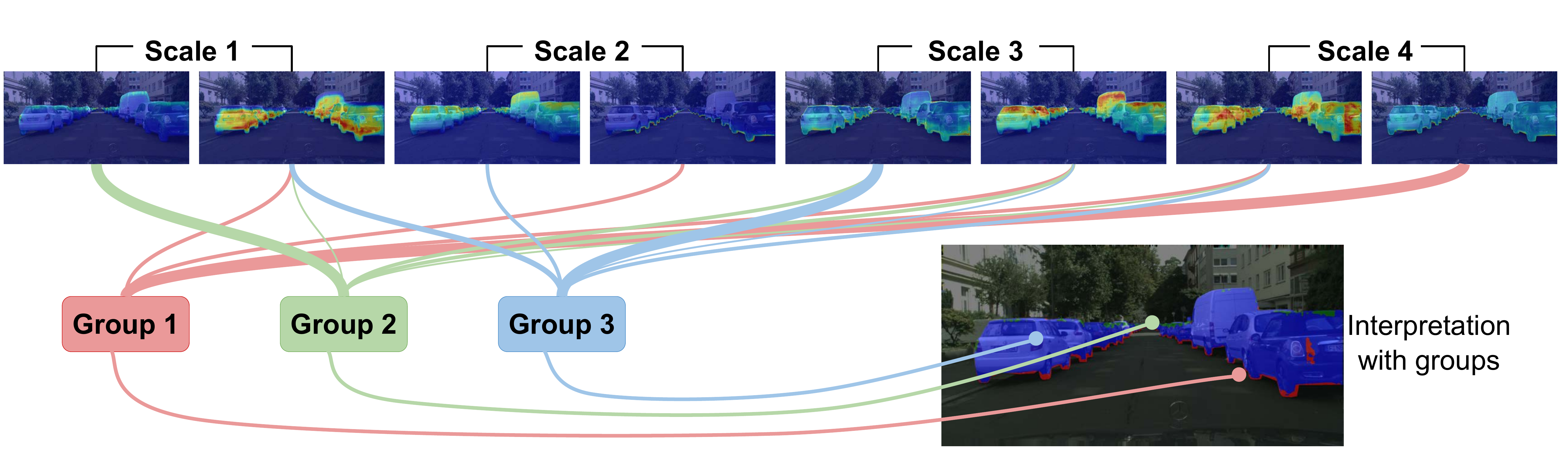}
    \caption{ScaleProtoSeg provides the interpretation of a segmentation through the analysis of groups of prototypes. For the example of the class \textit{car} on Cityscapes, $2$ prototypes per scale (whose activations are displayed \textbf{at the top} of the figure) are used by the model across the $3$ learned groups shown \textbf{at the bottom right}. For this class, groups correspond to the bottom part, the main part or the upper part of the car.}
    \label{fig:group-ex}
\end{figure*}

\vspace{-1em} \paragraph{Quantitative analysis of interpretability.}
The quantitative evaluation of interpretability is an intrinsic problem in the field of XAI~\cite{doshi2017towards, adebayo2018sanity} and especially in the context of per-design methods such as prototypical-parts learning~\cite{kim2021hive, xu2023sanity}. Several studies proposed human-based evaluations of explainable methods for classification tasks~\cite{davoodi2023interpretability, colin2022cannot}. However, scaling those evaluation scenarios to semantic segmentation raises concerns due to the complexity of the necessary human feedback compared to classification. We propose to quantitatively assess the degree of interpretability in terms of \textit{consistency}, \textit{stability}, and \textit{sparsity}. The first two metrics have been proposed in \cite{huang2023evaluation} to measure the interpretability of a classification model.
We leverage the part-annotations extension of Pascal and Cityscapes~\cite{degeus2021panopticparts,meletis2020panopticparts} covering respectively 16 and 5 classes, and propose an extension of these metrics to the task of semantic segmentation. Given an input image $\mathbf{x}$ and a prototype $\mathbf{p}_{\protoidx}$, the prototype activation $g_\text{proto}(\mathbf{z}, \mathbf{p}_{\protoidx}), \: \forall \mathbf{z} \in f(\mathbf{x})$ is binarized using multiple percentile thresholds: $\{70^{th}, 80^{th}, 90^{th}\}$ instead of a fixed size bounding box, as multiple objects of the same class can be present in $\mathbf{x}$. We then compute the average consistency and stability scores across the multiple thresholds on the part-annotated validation sets for 3 runs per method, similarly to~\cite{huang2023evaluation}. We also introduce a global sparsity metric that measures the number of active prototypes per class after thresholding the last linear layer absolute values with $\tau_{th} = 0.005$, which is similar to the sparsity ratio used as a measure of interpretations' compactness in \cite{nauta2023pip}. The measured sparsity is linked to several properties of interpretability: transparency \cite{gautam2022protovae}, understandability \cite{colin2022cannot, davoodi2023interpretability}, and simplicity \cite{ming2019interpretable}. Despite this metrics not being exhaustive, our assessment shows that ScaleProtoSeg displays overall enhanced interpretability over ProtoSeg (see Table~\ref{tab:inter-pascal} and ~\ref{tab:inter-city}).

\begin{table}[h]
\centering
\resizebox{\columnwidth}{!}{
\begin{tabular}{l|ccc|}
\cline{2-4}
\textbf{}                               & \multicolumn{3}{c|}{Pascal}                                                                                             \\ \hline
\multicolumn{1}{|l|}{\textbf{Methods:}} & \multicolumn{1}{c|}{Consistency $\uparrow$}     & \multicolumn{1}{c|}{Stability $\uparrow$}       & Sparsity $\downarrow$      \\ \hline
\multicolumn{1}{|l|}{ProtoSeg}          & \multicolumn{1}{c|}{35.05 $\pm 1.44$} & \multicolumn{1}{c|}{73.45 $\pm 0.45$}          & 157.67 $\pm 3.40$            \\ \hline
\multicolumn{1}{|l|}{ScaleProtoSeg}     & \multicolumn{1}{c|}{\textbf{38.78} $\pm 1.68$}          & \multicolumn{1}{c|}{\textbf{76.30} $\pm 0.26$} & \textbf{23.34} $\pm 5.22$ \\ \hline
\end{tabular}}
\caption{Consistency, stability, and sparsity scores on Pascal part-annotated sets.}\vspace{-2em}
\label{tab:inter-pascal}
\end{table}

\begin{table}[h]
\centering
\resizebox{\columnwidth}{!}{
\begin{tabular}{l|ccc|}
\cline{2-4}
\textbf{}                               & \multicolumn{3}{c|}{Cityscapes}                                                                                             \\ \hline
\multicolumn{1}{|l|}{\textbf{Methods:}} & \multicolumn{1}{c|}{Consistency $\uparrow$}     & \multicolumn{1}{c|}{Stability $\uparrow$}       & Sparsity $\downarrow$      \\ \hline
\multicolumn{1}{|l|}{ProtoSeg}          & \multicolumn{1}{c|}{\textbf{34.48} $\pm 1.74$} & \multicolumn{1}{c|}{27.00 $\pm 1.58$}          & 134 $\pm 2.45$            \\ \hline
\multicolumn{1}{|l|}{ScaleProtoSeg}     & \multicolumn{1}{c|}{31.11 $\pm 1.66$}          & \multicolumn{1}{c|}{\textbf{32.54} $\pm 2.36$} & \textbf{12.91} $\pm 1.26$ \\ \hline
\end{tabular}}
\caption{Consistency, stability, and sparsity scores on Cityscapes part-annotated sets.}\vspace{-2em}
\label{tab:inter-city}
\end{table}

\paragraph{Qualitative analysis of grouping mechanism interpretability.} 

In terms of interpretability, the groups can be first represented through specific visualization for each sample, as shown in Figure~\ref{fig:group-ex}, where the images on the graph display the prototype activations for an input sample. Those activations are grouped per scale and provide local explanations for the model. Moreover, the edges in those visualizations are static for all samples assigned to the class \textit{car} in Cityscapes, and so provide global explanations on the model behavior. In Figure~\ref{fig:group-ex}, we can easily observe, despite the large number of prototypes compared to the model mean of $3.39$, that different patterns emerge for each group. The first group focuses mainly on a prototype in \textit{Scale 4} activated on the bottom part of the cars and another similar at \textit{Scale 2}. The second group focuses on a prototype at \textit{Scale 1}, which activates the top of the cars with another one similar at \textit{Scale 3}. Lastly, the third group focuses on multiple parts of the cars and, will activate on the main body of them. Groups $1$ and $2$, by combining similar prototypes across two scales, show transparently the use of multi-scale information in the model. Figure~\ref{fig:group-ex} demonstrates that the grouping mechanism in our method can be representative of both local and global interpretability. The example of group assignments for the class \textit{car} illustrates the identified parts in the group visualization. Interestingly, the second group seems more activated on cars further in the depth field, as its main activated prototype is from \textit{Scale 1} with a smaller field of view. This clearly illustrates the role of scale in the identification of prototypical parts.

\begin{figure}[t]
\centering
\setlength\tabcolsep{1.5pt} 
\resizebox{\columnwidth}{!}{
 \begin{tabular}{lccc}
 
 \rotatebox{90}{\quad\quad Original} & 
 \includegraphics[width=0.32\textwidth,  height=3cm]{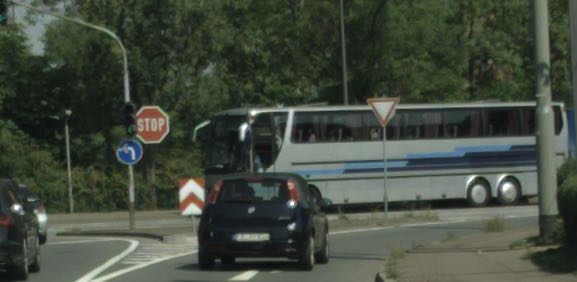} & 
 \includegraphics[width=0.32\textwidth, height=3cm]{images/plot_img_ADE_val_00000139_original.jpg} & 
 \includegraphics[width=0.32\textwidth,height=3cm]{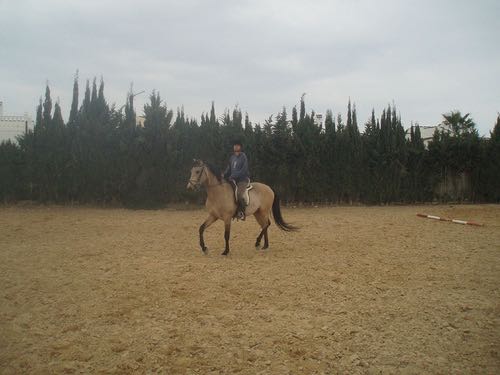} \\
 \rotatebox{90}{\parbox{3cm}{\centering ProtoSeg \cite{sacha2023protoseg} prototypes}} & 
 \includegraphics[width=0.32\textwidth,height=3cm]{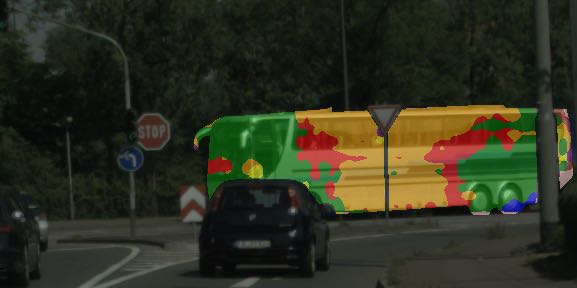} & 
 \includegraphics[width=0.32\textwidth,height=3cm]{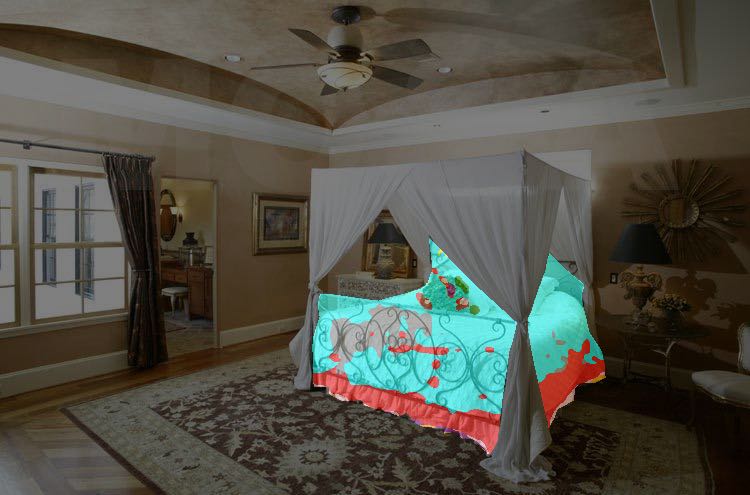} & 
 \includegraphics[width=0.32\textwidth, height=3cm]{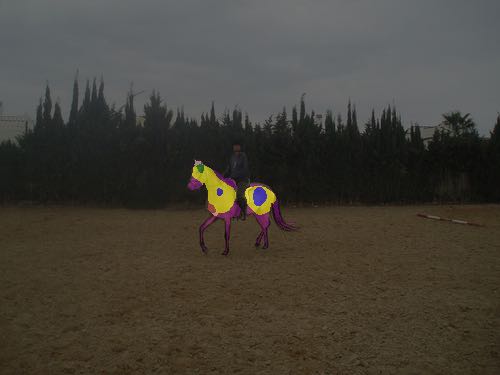} \\
 \rotatebox{90}{\parbox{3cm}{\centering ScaleProtoSeg groups}} & 
 \includegraphics[width=0.32\textwidth, height=3cm]{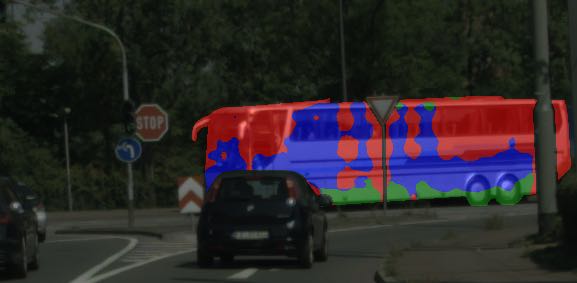} & 
 \includegraphics[width=0.32\textwidth,height=3cm]{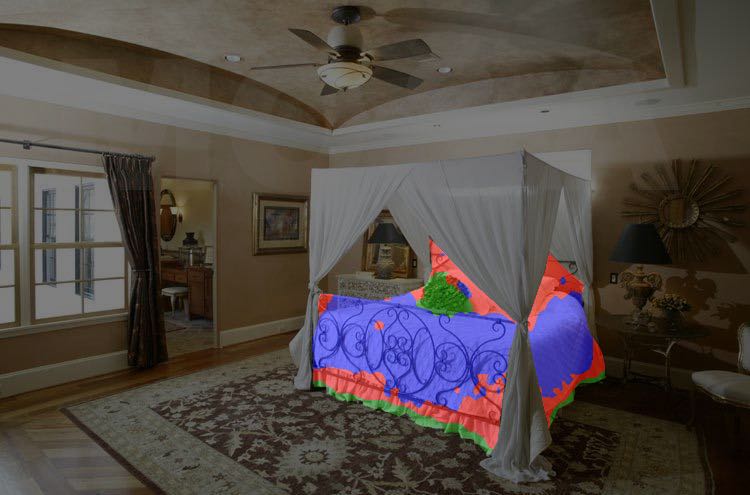} & 
 \includegraphics[width=0.32\textwidth, height=3cm]{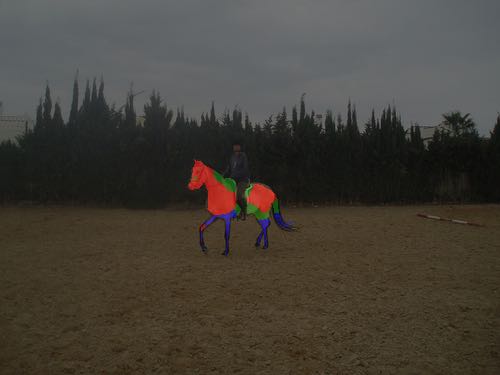}
 \end{tabular}}
 \caption{Model prototype and group assignments for the class \textit{bus}, \textit{bed}, and \textit{horse} on Cityscapes, ADE20K, and Pascal.\vspace{-2em}}
 \label{fig:dec-proc}
\end{figure}

In Figure~\ref{fig:dec-proc}, we show the final decision process based on the prototype assignments for ProtoSeg (row $2$) and the group assignments for ScaleProtoSeg (row $3$). The ProtoSeg outputs were computed based on a rerun of the model after pruning. In all three examples of Figure~\ref{fig:dec-proc}, our groups identify one or more prototypical parts each, for the class \textit{bus}, \textit{bed}, and \textit{horse}. For instance, in the bus image, the first group in red corresponds to both the front and back top of the bus, a pattern that can be also seen in the other two images. Moreover, we see through those examples that, due to the limited number of groups $(N_{c}=3)$ in ScaleProtoSeg, the interpretation of the final decision process is more constrained compared to the baseline method ProtoSeg. Indeed, even if we consider the prototype interactions leading to the group and their assignment, our method still uses fewer prototypes compared to ProtoSeg, as described in Table~\ref{tab:breakdown}. Lastly, as detailed in the supplementary Section~\ref{sec:overhead}, our method enforces stronger sparsity regularization on the final classification layer inducing a smaller computation overhead on DeepLabv2~\cite{chen2017deeplab} compared to ProtoSeg~\cite{sacha2023protoseg}.

Overall, through the sparsity of the groups, their simple visualizations, their small number, and the sparse final linear layer, we advocate that our method enables the user to investigate more transparently the semantic segmentation model while improving upon the state-of-the-art ProtoSeg.

\section{Conclusion}
\label{sec:conclusion}

In this paper, we present an interpretable semantic segmentation model, ScaleProtoSeg, that leverages multi-scale representations for prototype learning and introduces a novel grouping mechanism to learn prototype interactions across scales. We showed that our model results are particularly advantageous in complex datasets presenting many objects at different depths: Cityscapes and ADE20K. Moreover, through our analysis, we evaluated ScaleProtoSeg interpretability across $3$ quantitative metrics: \textit{consistency}, \textit{stability}, and \textit{sparsity} and inspected the different patterns learned by the prototypes across scales. This showcases the potential of multi-scale prototype learning to provide a deeper understanding of the effect of scales on object representations. Lastly, our novel grouping mechanism provides a clear representation of the final decision process: it shows how the model learns to group the prototypes across scales by limiting the number of active groups, allowing for their easy and sparse visualization for all images.

{\small
\bibliographystyle{ieee_fullname}
\bibliography{egbib}
}

\twocolumn[{%
    \begin{center}
        \section*{Supplementary: Multi-Scale Grouped Prototypes for Interpretable Semantic Segmentation}
    \end{center}
}]

\vspace{20em}

\section{Details on experimental set-up}
\label{sec:exp-set-up-det}In this section, we provide the details of the experimental set-up briefly described in Section~\ref{sec:set-up} to support reproducing the results. 

Firstly, across all training stages, the prototypes are the same size as the output feature maps of the non-concatenated ASPP in DeepLabv2~\cite{chen2017deeplab}: $D = 64$. For both multi-scale prototype training and the grouping mechanism we leverage augmentation techniques, such as random horizontal flipping, cropping, and scaling images by a factor between $[0.5, 1.5]$ for Cityscapes~\cite{cordts2016cityscapes} and Pascal VOC~\cite{pascal-voc-2012}, and $[0.5, 2]$ for ADE20K~\cite{zhou2017scene}. The batch size used in all the training stages is $10$ and we select the Adam optimizer with a weight decay of $5e^{-4}$, $\beta_1 = 0.9$, and $\beta_2 = 0.999$ for our experiments. Moreover, as the batch size in our experiments is limited, we freeze the batch normalization parameters of the ResNet-101 backbone due to its impact on performance.

For the multi-scale prototype training stage, we set the weights of the loss terms to $\lambda_{L1} = 1e^{-4}$ and $\lambda_{J} = 0.25$ following ProtoSeg on both Cityscapes and Pascal VOC, and $\lambda_{L1} = 1e^{-5}$ for ADE20K due to the large number of classes. For the datasets: Cityscapes and Pascal VOC, we run first the warm-up step for $3000$ batch iterations with a fixed learning rate of $2.5e^{-4}$. The joint training step is run for $30000$ batch iterations with an initial learning rate of $2.5e^{-5}$ for ResNet-101 and $2.5e^{-4}$ for the prototype and ASPP layers. For this step, we leverage a learning rate scheduler following the polynomial learning rate policy with $power = 0.9$. Lastly, the fine-tuning step is done over $2000$ batch iterations extended to $6000$ for Pascal with a fixed learning rate of $1e^{-5}$. For ADE20K we leverage the same learning rate and learning rate scheduler as for the other two datasets but we double the number of batch iterations compared to Cityscapes on all steps: 6000 batches for warm-up, 60000 batches for joint training, and 4000 batches for fine-tuning.

The training of ProtoSeg on ADE20K follows the same experimental set-up as the multi-scale prototype training stage described above and as mentioned in Section~\ref{sec:set-up}: 12 prototypes per class were used to match ScaleProtoSeg.

During the training of the grouping mechanism, we set the weights of the loss terms to $\lambda_{L1} = 1e^{-3}$ for Cityscapes and Pascal VOC, and $\lambda_{L1} = 1e^{-4}$ for ADE20K. Moreover, we set $\lambda_{ent} = 0.05$ for Cityscapes and Pascal VOC, and $\lambda_{ent} = 0.25$ for ADE20K. We run the warm-up stage to train the group projections for $2000$ batch iterations with a fixed learning rate of $2.5e^{-4}$ for all datasets. Then, we run the fine-tuning stage for $30000$ batches with a learning rate of $2.5e^{-4}$ and the same optimizer and scheduling policy as in the prototype joint training phase for all datasets.

We evaluate ScaleProtoSeg on Pascal VOC for which we set the training image resolution to $321 \times 321$ and the testing one to $513 \times 513$. Moreover, we use multi-scale inputs~\cite{chen2017deeplab} with scales $\{0.5, 0.75, 1\}$ during training. We also evaluate ScaleProtoSeg on Cityscapes, for this dataset, we do not use MSC input training, we set the training resolution to $513 \times 513$ and the testing one to $1024 \times 2048$. Lastly, we evaluate ScaleProtoSeg and ProtoSeg on ADE20K with no MSC input, a training resolution of $512 \times 512$, and at test time we resize the smallest size of the image to the training resolution and keep the aspect ratio. Lastly, the performance of DeepLabv2 on ADE20K was extracted from~\cite{huang2017semantic}.

The interpretability evaluation of our method is done on the parts annotated validation sets from Cityscapes and PASCAL-Context~\cite{degeus2021panopticparts, meletis2020panopticparts}. In particular, to align to the classes and images covered during training for Pascal, we tested on the overlap images between PASCAL VOC and PASCAL-Context and so used the semantic annotations from PASCAL VOC and the part annotations from~\cite{degeus2021panopticparts, meletis2020panopticparts} for the 16 classes covered. This represents 925 validation images for Pascal. Moreover, the stability and consistency metrics require part keypoints in their formulation, for this purpose we leverage the standard algorithm from opencv with an 8-connectivity relation between non-zero elements to compute the centroids of every connected component in the part annotations, similar to Section~\ref{sec:result}. The prototype activations are computed using the images in their native size for both datasets and interpolated to their corresponding part annotation size. We leverage multiple binarization thresholds on the prototype activations as mentioned in Section~\ref{sec:result} to improve the robustness of those metrics in terms of variance across runs and to avoid a strong dependency on a fixed hyperparameter contrary to the window size used in~\cite{huang2023evaluation}.

\section{Transferability of ScaleProtoSeg}
\label{sec:em}

To demonstrate the transferability of the proposed ScaleProtoSeg method beyond the domain of natural images, we run experiments on a medical dataset. In particular, we use the EM segmentation challenge dataset from ISBI 2012~\cite{arganda2012segmentation}, containing 30 images of size $512\times512$ with 2 classes that are randomly split 2 times in 20 training and 10 validation images, similar to the experiment in ProtoSeg~\cite{sacha2023protoseg}. The architecture used as the backbone for this experiment is the original U-Net~\cite{ronneberger2015u}. The applicability of ScaleProtoSeg to another segmentation backbone is straightforward as it simply requires stacking the ASPP at the network output before the classification layer, allowing the extraction of multi-scale feature maps. The prototype layer contains 10 and 12 prototypes per class for ProtoSeg and ScaleProtoSeg respectively, so 3 prototypes per scale and class for ScaleProtoSeg and similarly 3 groups per class.  We run 3 experiments per method in the same set-up as in ProtoSeg with pruning and grouping for each method respectively. For the prototype training stage, we skip the warm-up step and run the joint training and fine-tuning step for 10000 batches iterations each with a batch size of 2. The learning rate for the joint step is fixed at $1e^{-4}$ and we use the same learning scheduler as in Section~\ref{sec:exp-set-up-det}. The learning rate for the fine-tuning step is constant and fixed at $1e^{-5}$. The weights of the loss terms for both methods during the prototype training stage are set to $\lambda_{L1} = 1e^{-4}$ and $\lambda_{J} = 0.25$. In the grouping stage for ScaleProtoSeg, we also skip the warm-up step and run the joint training step for 10000 batch iterations with a learning of $5e^{-5}$ following the same scheduler as in Section~\ref{sec:exp-set-up-det}. The weights of the loss terms during the grouping stage are set to $\lambda_{L1} = 1e^{-4}$ and $\lambda_{ent} = 0.25$. The optimizer and augmentation pipeline used across all stages and methods is the same as in Section~\ref{sec:exp-set-up-det}. We report the mIoU in Table~\ref{tab:medical}. Results show that \textit{ScaleProtoSeg can transfer to another backbone and image modality}. Indeed, not only ScaleProtoSeg outperforms ProtoSeg in datasets with complex scenarios (CityScapes and ADE20K), but it also yields marginally better performance in the considered medical dataset despite the fact that the images do not present multi-scale features (similarly to the PASCAL VOC benchmark).

\begin{table}[h]
\centering
\resizebox{\columnwidth}{!}{
\begin{tabular}{|l|c|c|}
\hline
\textbf{Method}                               & \textbf{EM Split 1}  & \textbf{EM Split 2}           \\ \hline
ProtoSeg (U-Net + ASPP)      & 77.63 $\pm 0.29$ & 78.92 $\pm 0.19$ \\ \hline
ScaleProtoSeg (U-Net + ASPP) & \textbf{77.97} $\pm 0.16$ & \textbf{79.29} $\pm 0.17$ \\ \hline
\end{tabular}}
\caption{IoU performance of ScaleProtoSeg compared against ProtoSeg. Results demonstrate the effective transferability of ScaleProtoSeg to another segmentation architecture (U-Net) and the medical domain (ISBI 2012 dataset)~\cite{arganda2012segmentation}.}
\label{tab:medical}
\end{table}

\section{Extension to large dataset}
\label{sec:coco}
In order to test our method in a use case close to real-world applications we extend our evaluation to a large benchmark: COCO-Stuff~\cite{caesar2018coco}, containing 182 classes among 11 are not present in the dataset with 118k training and 5k validation samples. We leverage a similar setup to ADE20K for the model hyperparameters, except that in the multi-scale prototype training stage, we iterate for a total of 110k iterations: 6000 warm-up steps, 100k joint steps, and 4000 finetuning steps. Moreover, the iterations are done directly on a batch of size 10. We also use multi-scale inputs with scales $\{0.5, 0.75, 1\}$ and a training resolution of $321 \times 321$, while testing at the native resolution. The objective is to align to the set-up for the DeepLabv2 baseline provided at this repository: \href{https://github.com/kazuto1011/deeplab-pytorch}{github.com/kazuto1011/deeplab-pytorch}. The results for our method: ScaleProtoSeg are presented in Table~\ref{tab:perf-ext}, and showcase that our method despite lower performance stays competitive with DeepLabv2 on a larger dataset while providing interpretability. Moreover, \textbf{ProtoSeg trained in a similar setup has a performance of 32.97 mIoU without pruning for one run}, so ScaleProtoSeg slightly outperforms it.

\begin{table}[t!]
\centering
\begin{tabular}{|l|c|}
\hline
\textbf{Method} & \textbf{\begin{tabular}[c]{@{}c@{}}COCO-Stuff\\ mIoU\end{tabular}} \\ \hline
DeepLabv2                             & 39.7                                                                   \\ \hline
ScaleProtoSeg                         & 34.50 $\pm 0.19$                                                                 \\ \hline
\end{tabular}
\caption{ScaleProtoSeg mIoU performance on COCO-Stuff validation set. We report the results over 3 runs for our method.}
\label{tab:perf-ext}
\end{table}

\section{Sparsity Regularization}
\label{sec:sparse}

In the grouping mechanism, it is also possible to control for the sparsity-performance trade-off via the thresholding of the grouping weights in the matrices $\mathbf{w}_{g, c}$ defined in Section~\ref{sec:group}. This is demonstrated in Table~\ref{tab:threshold}. The best results are obtained for threshold $\alpha = 0.05$ on all datasets. At this threshold, the groups are sparse with an average of $3.70$, $3.39$, and $4.08$ active prototypes per group for Pascal, Cityscapes, and ADE20K, which leads to better interpretability performance for the sparsity metric compared to ProtoSeg.
Moreover, we analyze the effect of the entropy regularization on the grouping mechanism as shown in Table~\ref{tab:ablation}. This regularization enables our method to use fewer prototypes in total and per group while providing similar performance, with up to $20$ total prototypes dropped and $2.5$ prototypes per group less for Cityscapes. 

\begin{table}[h]
\centering
\resizebox{\columnwidth}{!}{
\begin{tabular}{|l|c|c|c|c|}
\hline
\textbf{Dataset}            & \textbf{Threshold} & \textbf{Prototypes} & \textbf{Group Avg} & \textbf{mIoU} \\ \hline
\multirow{3}{*}{Pascal}     & $\alpha = 0.$                 & 133                 & 4.00                   & 72.11         \\ 
                            & $\alpha = 0.05$               & 131                 & 3.70                   & \textbf{72.26}         \\ 
                            & $\alpha = 0.1$                & \textbf{126}                 & \textbf{3.21}                   & 72.12         \\ \hline
\multirow{3}{*}{Cityscapes} & $\alpha = 0.$                 & 114                 & 3.68                   & 69.20         \\
                            & $\alpha = 0.05$               & 111                 & 3.39                   & \textbf{69.22}         \\ 
                            & $\alpha = 0.1$                & \textbf{109}                 & \textbf{3.10}                   & 69.03         \\ \hline
\multirow{3}{*}{ADE20K} & $\alpha = 0.$                 & 1168                 & 4.53                   & 34.03         \\
                            & $\alpha = 0.05$               & 1132                 & 4.08                   & \textbf{34.32}         \\
                            & $\alpha = 0.1$                & \textbf{1067}                 & \textbf{3.48}                   & 34.16         \\ \hline
\end{tabular}}
\caption{Ablation study of the effect of thresholding on the grouping mechanism with $\lambda_\text{ent} = 0.05$ on our ScaleProtoSeg best performing run.}
\label{tab:threshold}
\end{table}

\begin{table}[h]
\centering
\resizebox{\columnwidth}{!}{
\begin{tabular}{|l|c|c|c|c|}
\hline
\textbf{Dataset}            & \textbf{Regularization}     & \textbf{Prototypes} & \textbf{Group Avg} & \textbf{mIoU}  \\ \hline
\multirow{2}{*}{Pascal}     & $\lambda_\text{ent} = 0$    & 146                 & 5.59               & 72.22          \\
                            & $\lambda_\text{ent} = 0.05$ & \textbf{131}                 & \textbf{3.70}      & \textbf{72.26} \\ \hline
\multirow{2}{*}{Cityscapes} & $\lambda_\text{ent} = 0$    & 132                 & 5.79               & \textbf{69.22} \\
                            & $\lambda_\text{ent} = 0.05$ & \textbf{111}                 & \textbf{3.39}      & \textbf{69.22} \\ \hline
\multirow{2}{*}{ADE20K}     & $\lambda_\text{ent} = 0$    & 1405                 & 6.06                & 33.40            \\
                            & $\lambda_\text{ent} = 0.25$ & \textbf{1132}                & \textbf{4.08}               & \textbf{34.32}          \\ \hline
\end{tabular}}
\caption{Ablation study on the effect of the entropy regularization on the grouping mechanism with $\alpha = 0.05$ on our ScaleProtoSeg best performing run.}
\label{tab:ablation}
\end{table}

\section{Group Overlap}
\label{sec:ent}

In this section, we extend the analysis on the effect of the entropy loss regularization on the grouping mechanism presented above, as entropy regularization also impacts the overlap between group activations and supports identifying prototypical parts.

\begin{table}[h]
\centering
\begin{tabular}{|l|c|c|}
\hline
\textbf{Dataset}            & \textbf{Regularization}     & \textbf{mIoU Groups} \\ \hline
\multirow{2}{*}{Pascal}     & $\lambda_\text{ent} = 0$    & 47.45                \\
                            & $\lambda_\text{ent} = 0.05$ & \textbf{28.29}       \\ \hline
\multirow{2}{*}{Cityscapes} & $\lambda_\text{ent} = 0$    & 61.65                \\
                            & $\lambda_\text{ent} = 0.05$ & \textbf{43.48}       \\ \hline
\multirow{2}{*}{ADE20K}     & $\lambda_\text{ent} = 0$    & 50.25                \\
                            & $\lambda_\text{ent} = 0.25$ & \textbf{42.74}       \\ \hline
\end{tabular}
\caption{Analysis of the effect of the entropy regularization on the group activations overlap measured via mIoU with a threshold $\alpha = 0.05$ on our ScaleProtoSeg best performing run.}
\label{tab:overlap}
\end{table}

\begin{figure}[h]
 \centering
 \captionsetup[subfigure]{justification=centering}
  \begin{subfigure}{0.48\columnwidth}
		\includegraphics[width=\textwidth,  height=2.9cm]{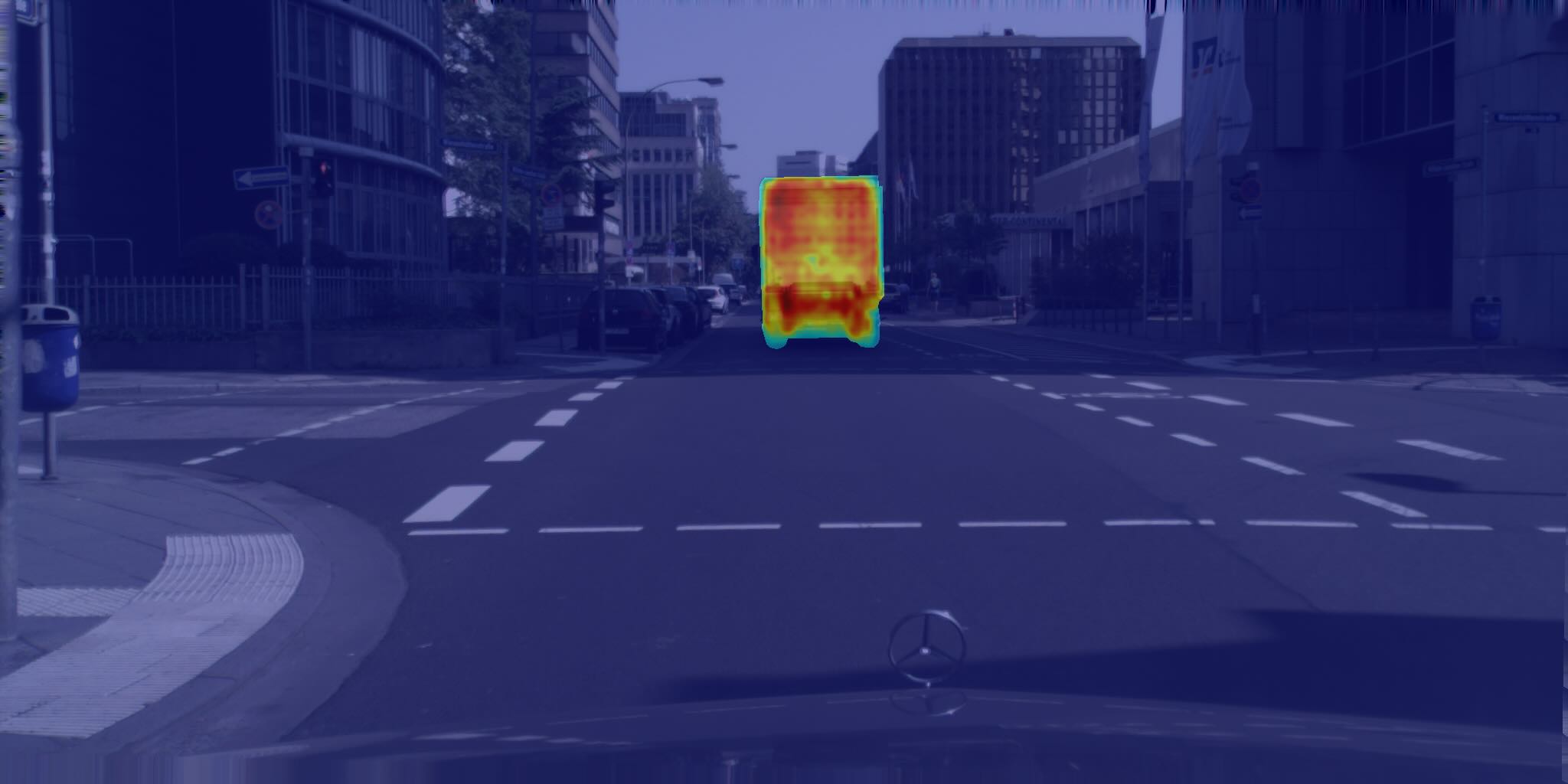}
		\includegraphics[width=\textwidth,  height=2.9cm]{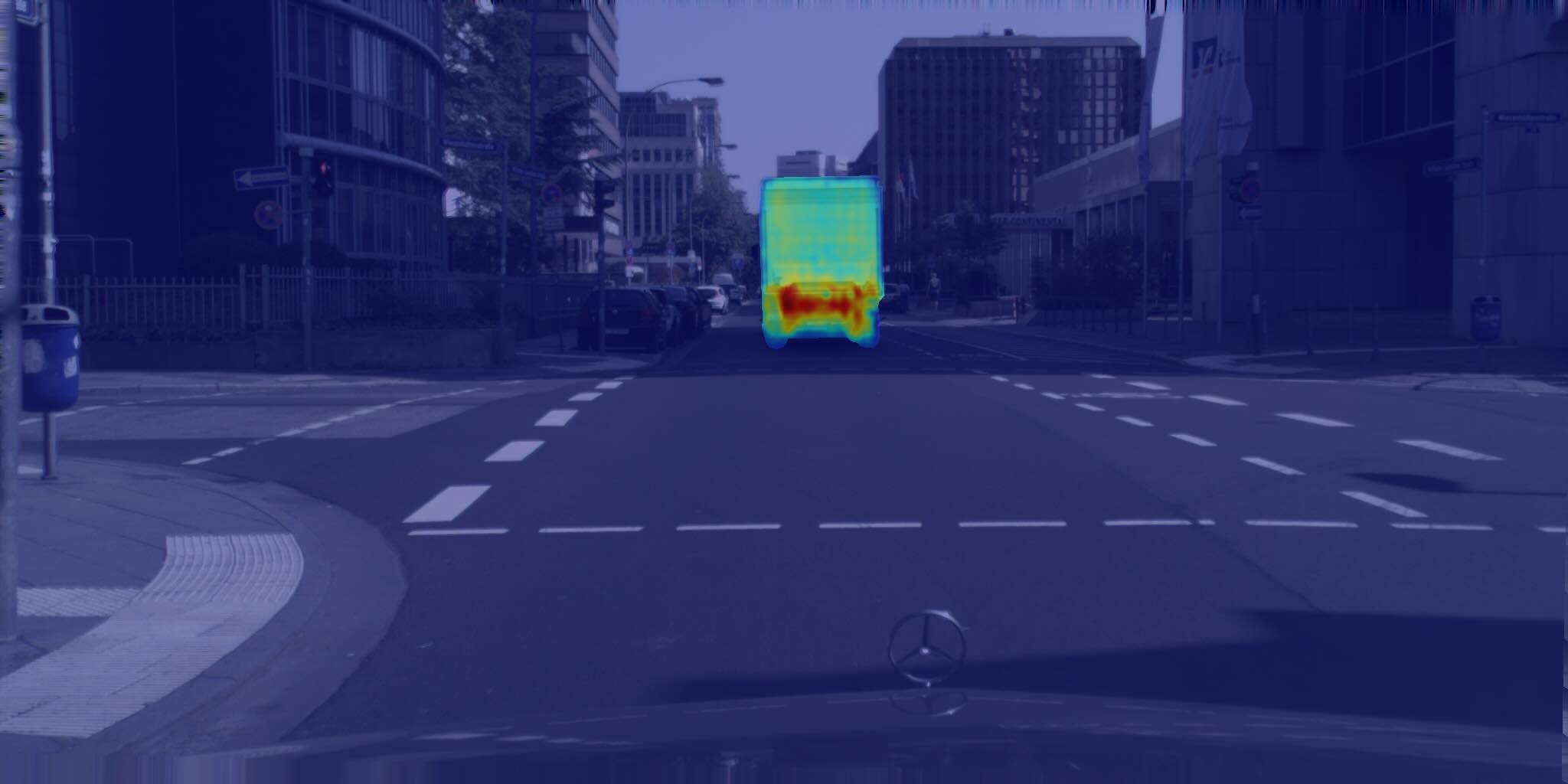}
		\includegraphics[width=\textwidth,  height=2.9cm]{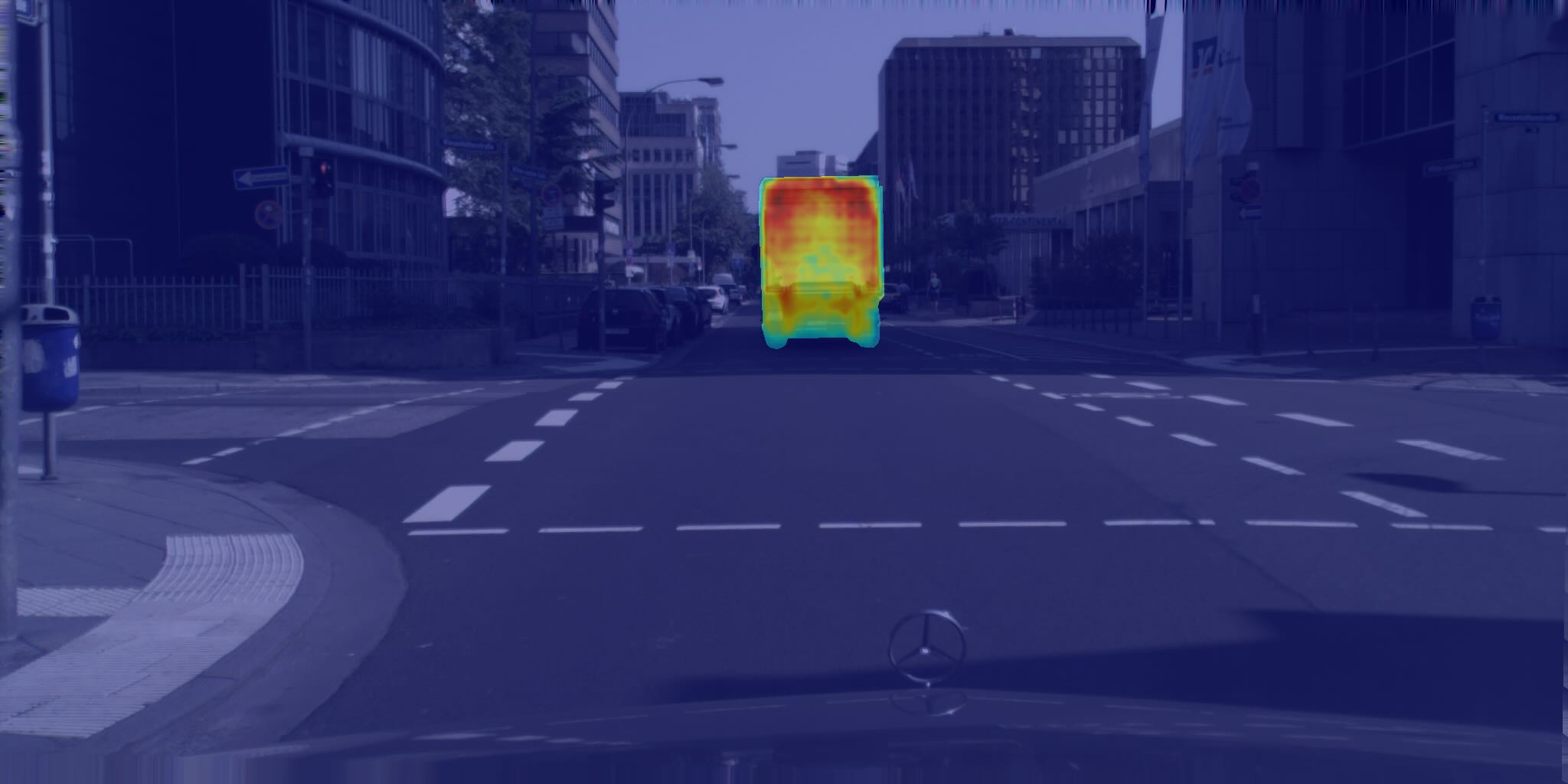}
  \subcaption{Group activations \textbf{without} entropy regularization}
\end{subfigure}
\begin{subfigure}{0.48\columnwidth}
		\includegraphics[width=\textwidth,  height=2.9cm]{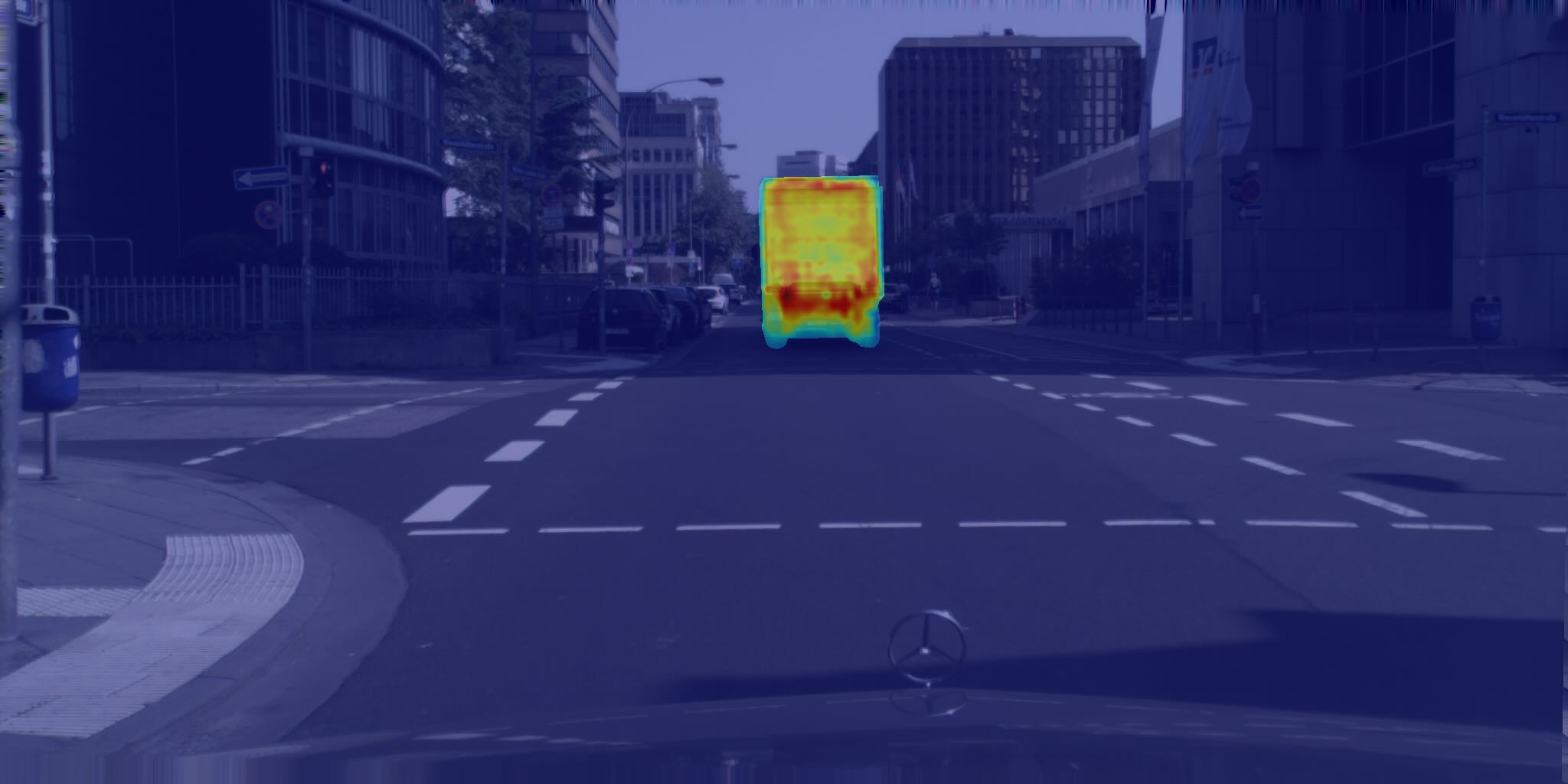}
		\includegraphics[width=\textwidth,  height=2.9cm]{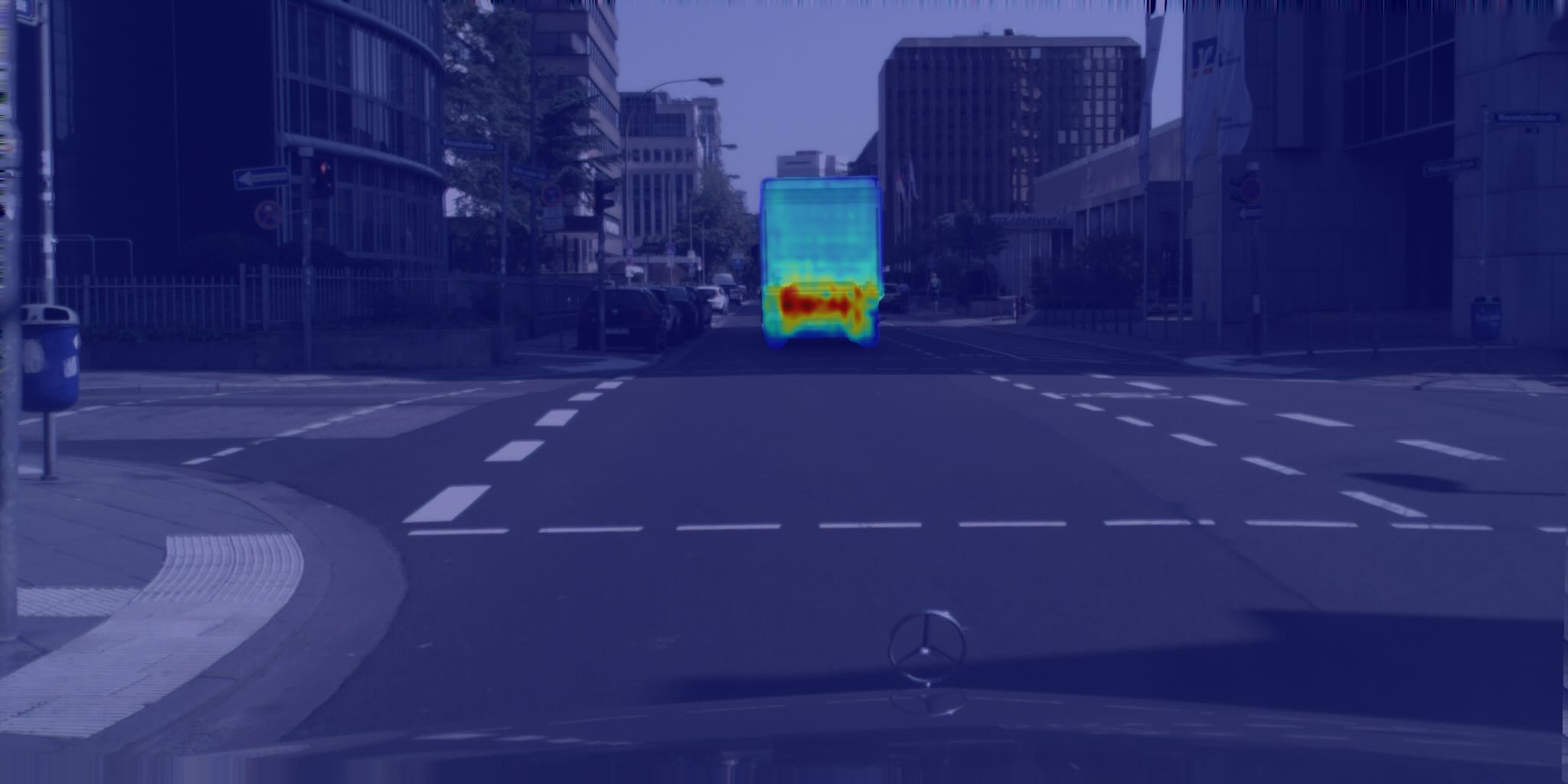}
    	\includegraphics[width=\textwidth,  height=2.9cm]{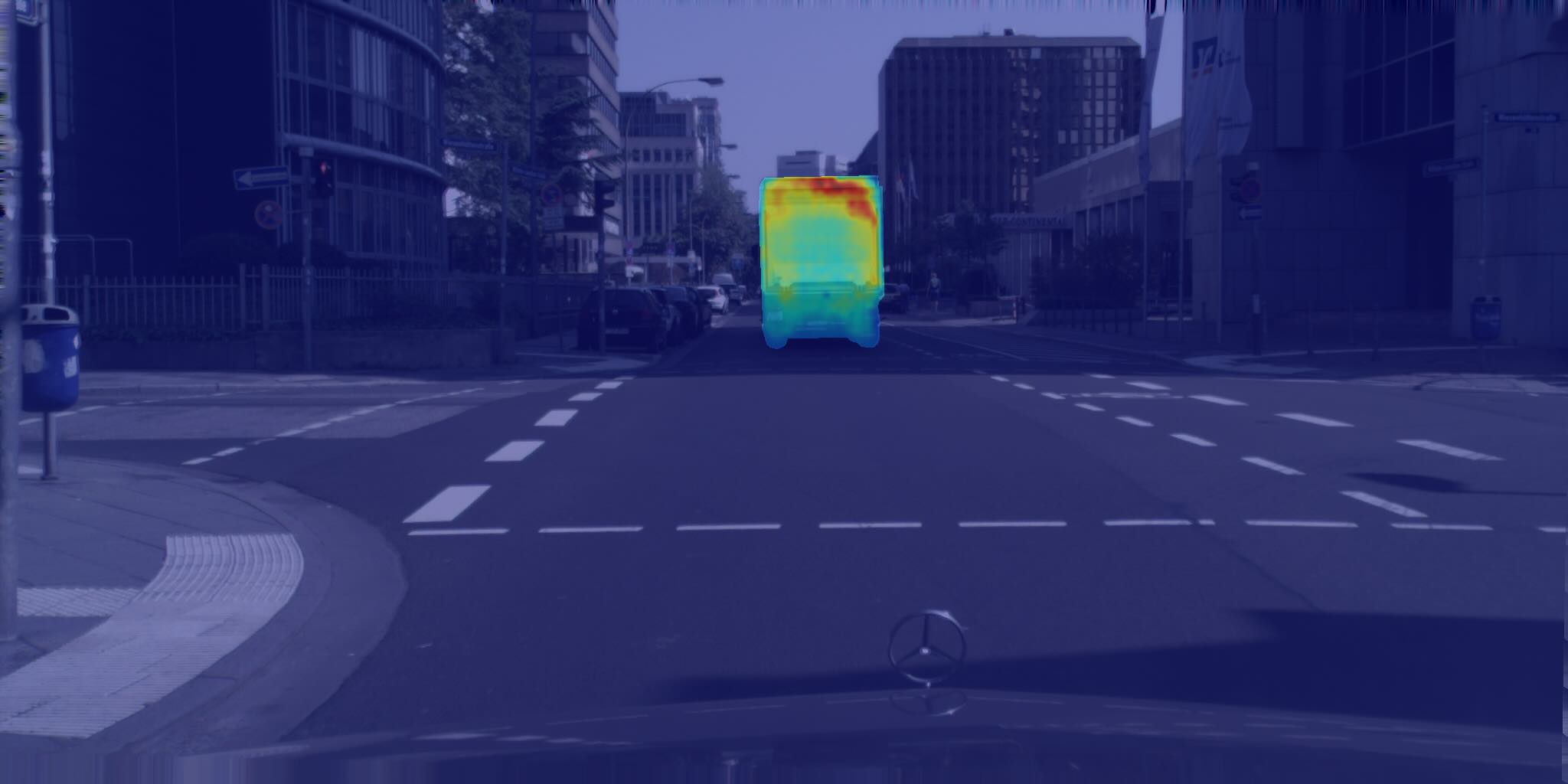}
  \subcaption{Group activations \textbf{with} entropy regularization}
    \end{subfigure}
  \caption{Example of group activations with or without entropy regularization for the class \textit{truck} on Cityscapes.}
		\label{fig:group-act}
	\end{figure}

We observe that besides reducing the number of prototypes used in the groups as shown in Table~\ref{tab:ablation}, the entropy loss also encourages diversity in the activations of the groups assigned to the same class as shown in Figure~\ref{fig:group-act} for Cityscapes. In Table~\ref{tab:overlap}, we present a quantitative analysis of this phenomenon. Firstly, on the validation sets of Pascal, Cityscapes, and ADE20K, we binarize all the group activations using the $95^{th}$ percentile. Then we compute on those validation sets the mIoU between the binarized group activation maps assigned to the same class, as a measure of overlap in the group activations. We observe that the entropy regularization on all three datasets decreases the overlap between groups of up to $19\%$ for Pascal. A low overlap between group activations enables the model to focus on different prototypical parts of the object and avoid groups all focusing on the whole object. Those results can be explained by the increased sparsity of the grouping functions, which ultimately leads to more variation in prototype assignment between the groups.

\section{Multi-scale prototype analysis}
\label{sec:scale-analysis}

In the proposed ScaleProtoSeg method, each set of prototypes assigned to a scale $s \in S$ corresponds to a specific field of view (FOV) from the ASPP layer in~\cite{chen2017deeplab}. We hypothesize that certain class-specific prototypes form pairs or groups of quasi-equivariant prototype activations between scales. Our definition of quasi-equivariance across scales is as follows. For a class $c \in \mathcal{C}$, let us consider $\forall \mathbf{x} \in \mathbb{R}^{H \times W \times 3}$ and its downscaled version $\mathbf{x'} \in \mathbb{R}^{\frac{H}{2} \times \frac{W}{2} \times 3}$ (the factor of 2 here matched the atrous rate increase from scale 1 to 2). Let us select the candidate pair of quasi-equivariant prototypes $\mathbf{p}_{1, i} \in P_{1, c}$ and $\mathbf{p}_{2, j} \in P_{2, c}$ and compute their respective activation maps $\mathbf{A}_{1, i}$ from all $\mathbf{z}_{1} \in f_{1}(\mathbf{x'})$ with $g_\text{proto}(\mathbf{z}_{1}, \mathbf{p}_{1, i})$, and $\mathbf{A}_{2, j}$ from all $\mathbf{z}_{2} \in f_{2}(\mathbf{x})$ with $g_\text{proto}(\mathbf{z}_{2}, \mathbf{p}_{2, j})$. The pair of prototypes is considered as quasi-equivariant if:
\begin{equation}
    f_\text{upsample}(\mathbf{A}_{1, i}) \sim \mathbf{A}_{2, j}
\end{equation}
with $f_\text{upsample}$ the upsampling function for $\mathbf{A}_{1, i}$ to the original size of $\mathbf{A}_{2, j}$. The similarity measure between activations is defined below.

In order to identify the sets of prototypes with quasi-equivariant activations for a specific class $c \in \mathcal{C}$ we first specify the increasing atrous rates ratio of each ASPP output feature maps with respect to the smallest one: $\{ r_{1 \rightarrow 1}, r_{1 \rightarrow 2}, r_{1 \rightarrow 3}, r_{1 \rightarrow 4}\}$ in the case of $S = 4$. The objective is to compute for all training images $\mathbf{x} \in \mathbb{R}^{H \times W \times 3}$, downscaled versions of the original image $\mathbf{x}_{1 \rightarrow s}$  such that $\mathbf{x}_{1 \rightarrow s} \in \mathbb{R}^{\frac{H}{r_{1 \rightarrow s}} \times \frac{W}{r_{1 \rightarrow s}} \times 3}$. Then, for all distinct pairs of feature map scales such that $(s, s') \in S^{2}$ and $s < s'$, we select $\mathbf{x}_{1 \rightarrow s}$ and $\mathbf{x}_{1 \rightarrow s'}$ to compute $f_{s}(\mathbf{x}_{1 \rightarrow s'})$ and $f_{s'}(\mathbf{x}_{1 \rightarrow s})$. The objective is to align, through the downscaling of the image, the semantic parts covered by the FOV of each ASPP scale $s$ and $s'$. Then we compare all pairs of prototypes from the scales $s$ and $s'$. For a specific pair $(\mathbf{p}_{s, i}, \mathbf{p}_{s', j})$ we compute from the feature maps $f_{s}(\mathbf{x}_{1 \rightarrow s'})$ and $f_{s'}(\mathbf{x}_{1 \rightarrow s})$, the activation maps $f_\text{upsample}(\mathbf{A}_{s, i})$ and $\mathbf{A}_{s', j}$, as described above. The similarity measure is then computed as follows, as we want to focus on the most activated parts for the objects assigned to $c$ we threshold the activation maps to a percentile $p_{th} \in \{0.6, 0.7 \}$ only considering the positions where the ground truth label $y_{\mathbf{z}} = c$. Then we derive the mIoU between the binarized activation maps across all the training set images, defining our similarity measure. Pairs of quasi-equivariant prototypes are identified when the mIoU is above a fixed threshold $\text{IoU}_{th} = 0.5$ on the training set. Lastly, pairs of quasi-equivariant prototypes are merged into groups if they overlap across scales.

\begin{figure}[h]
\centering
 \captionsetup[subfigure]{justification=centering}
  \begin{subfigure}{0.32\textwidth}
		\includegraphics[width=\textwidth]{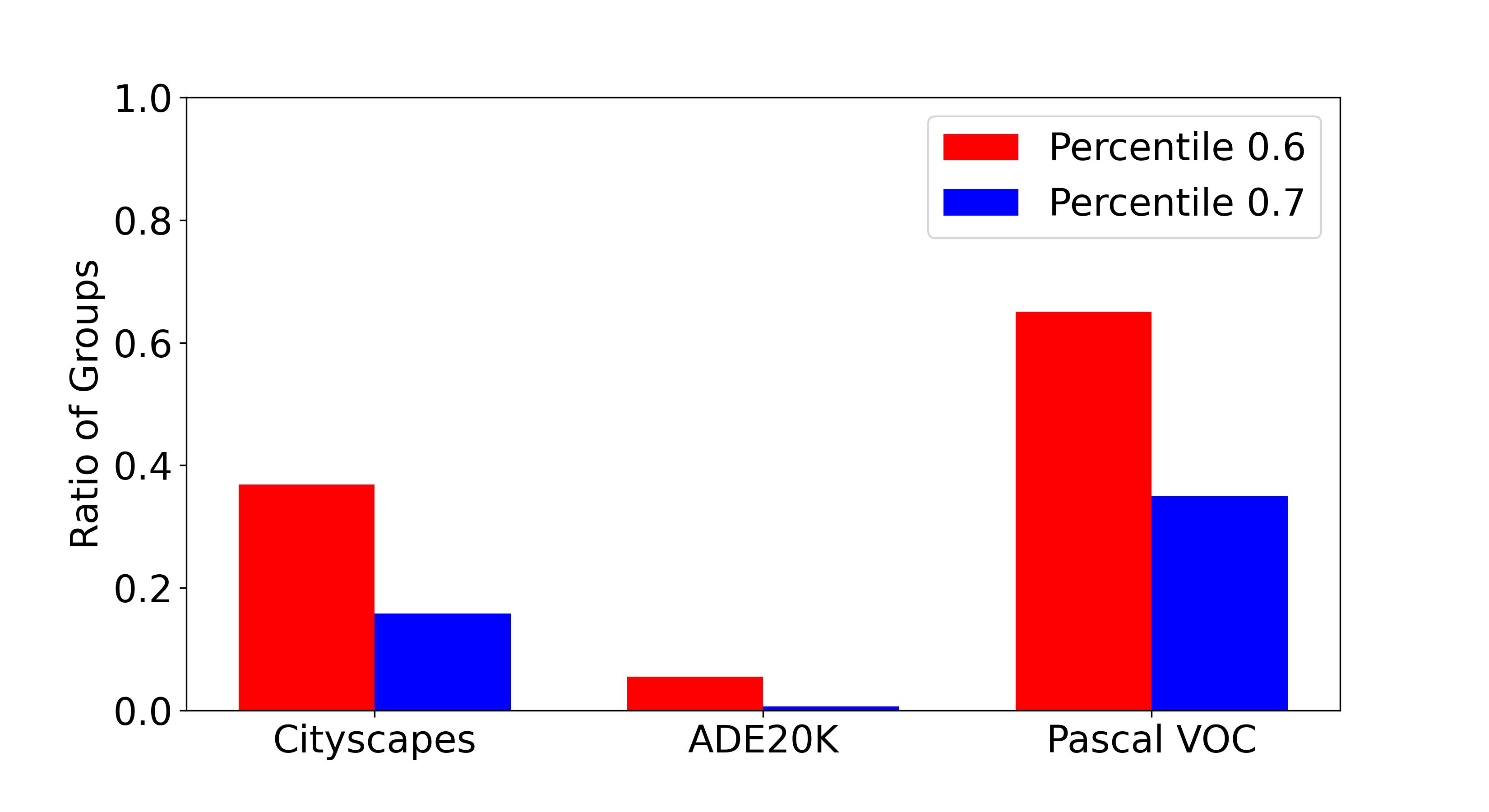}
  \subcaption{Ratio of quasi-equivariant groups identified.}
\end{subfigure}
\begin{subfigure}{0.32\textwidth}
		\includegraphics[width=\textwidth]{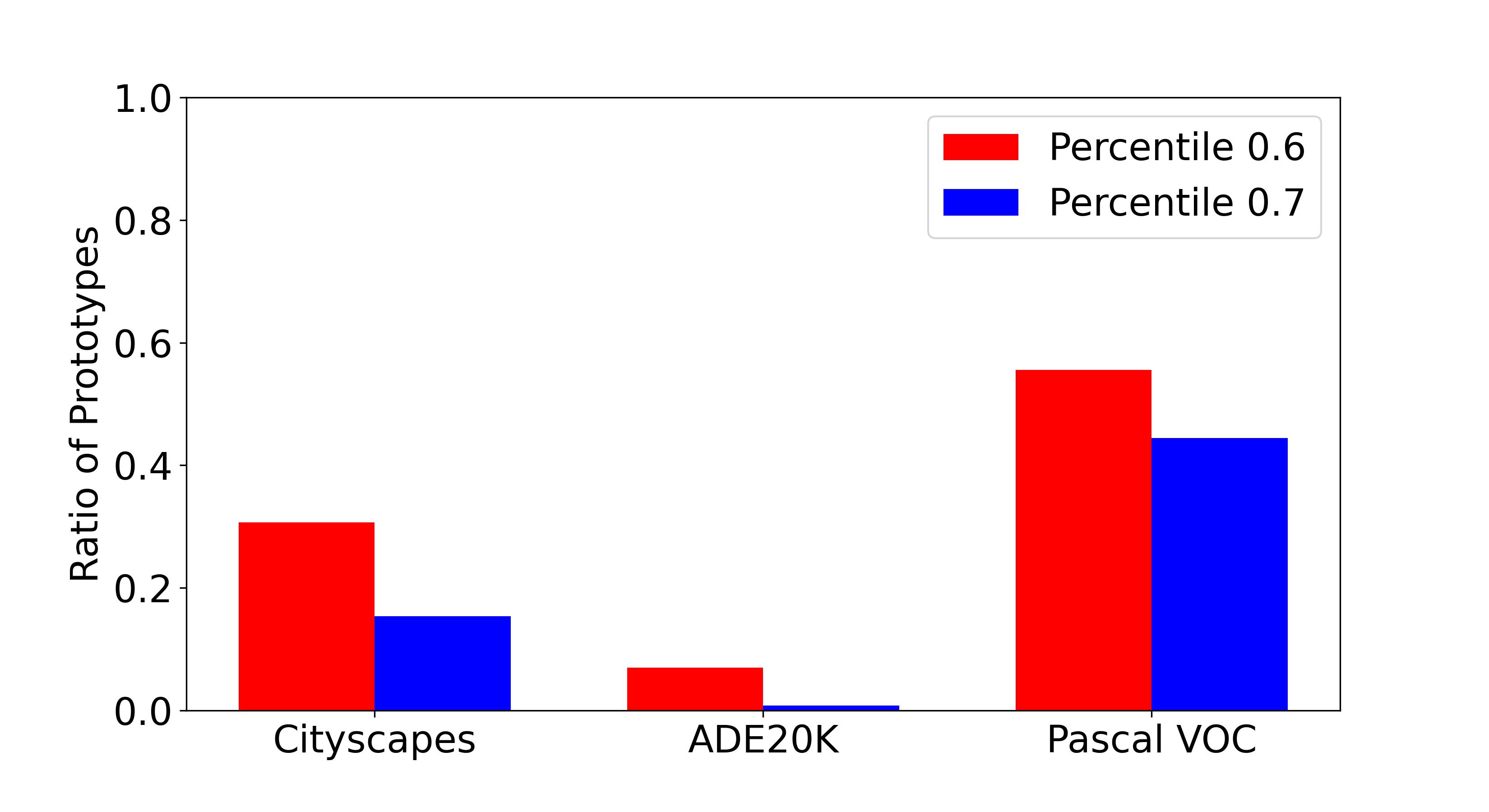}
  \subcaption{Ratio of prototypes in the quasi-equivariant groups.}
    \end{subfigure}
\begin{subfigure}{0.32\textwidth}
		\includegraphics[width=\textwidth]{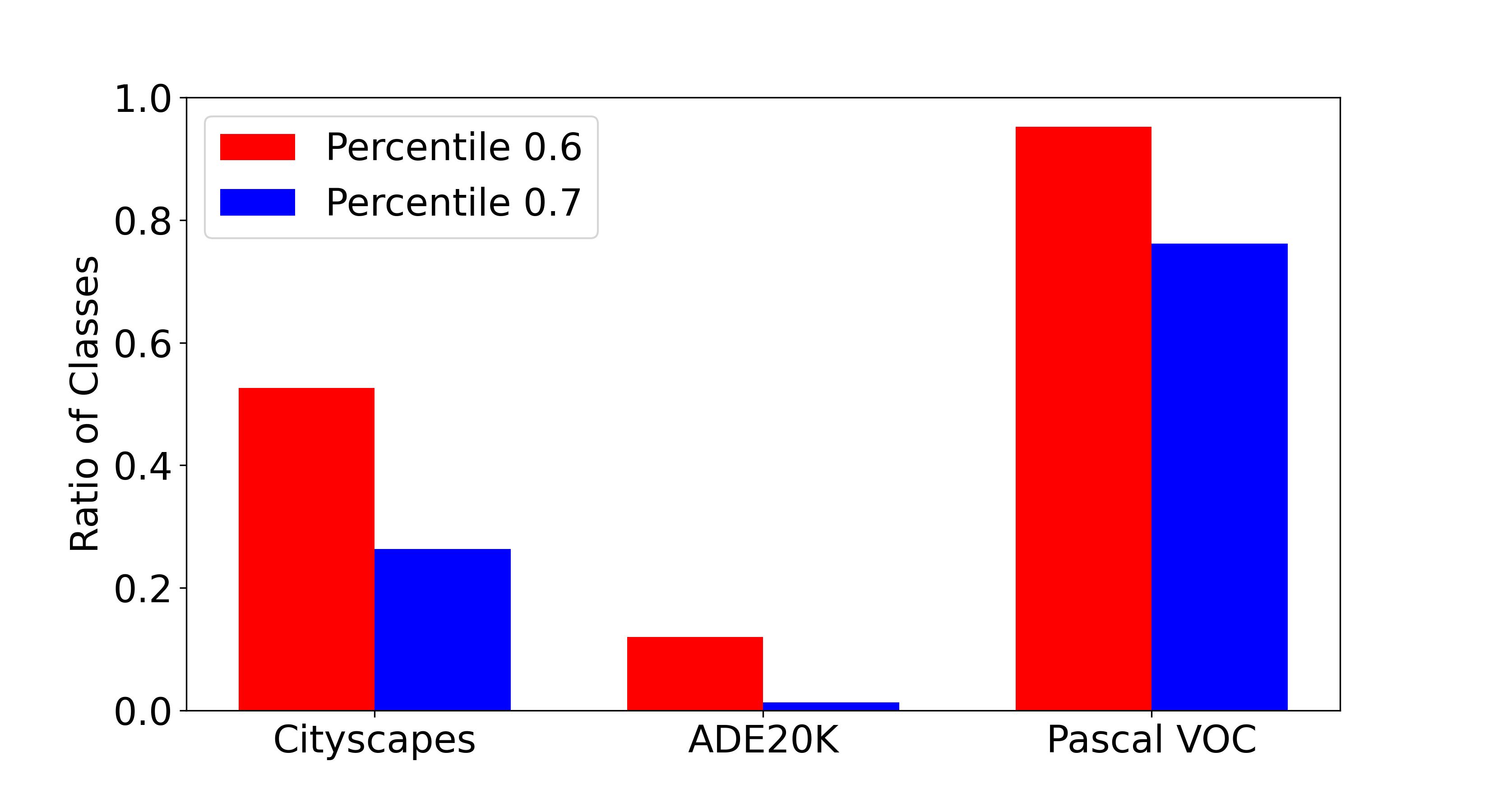}
  \subcaption{Ratio of classes with quasi-equivariant groups.}
    \end{subfigure}
  \caption{Analysis of the presence of quasi-equivariant groups in all three datasets for $p_{th} \in \{0.6, 0.7 \}$ on our best performing ScaleProtoSeg run. In \textbf{(a)}, the number of quasi-equivariant groups is compared to the total number of groups defined in the sparse grouping mechanism.}
		\label{fig:equiv}
	\end{figure}

\begin{figure*}[h!]
\vspace{-1em}
\centering
 \captionsetup[subfigure]{justification=centering}
  \begin{subfigure}{\textwidth}
	\includegraphics[width=0.48\textwidth, height=2.6cm]{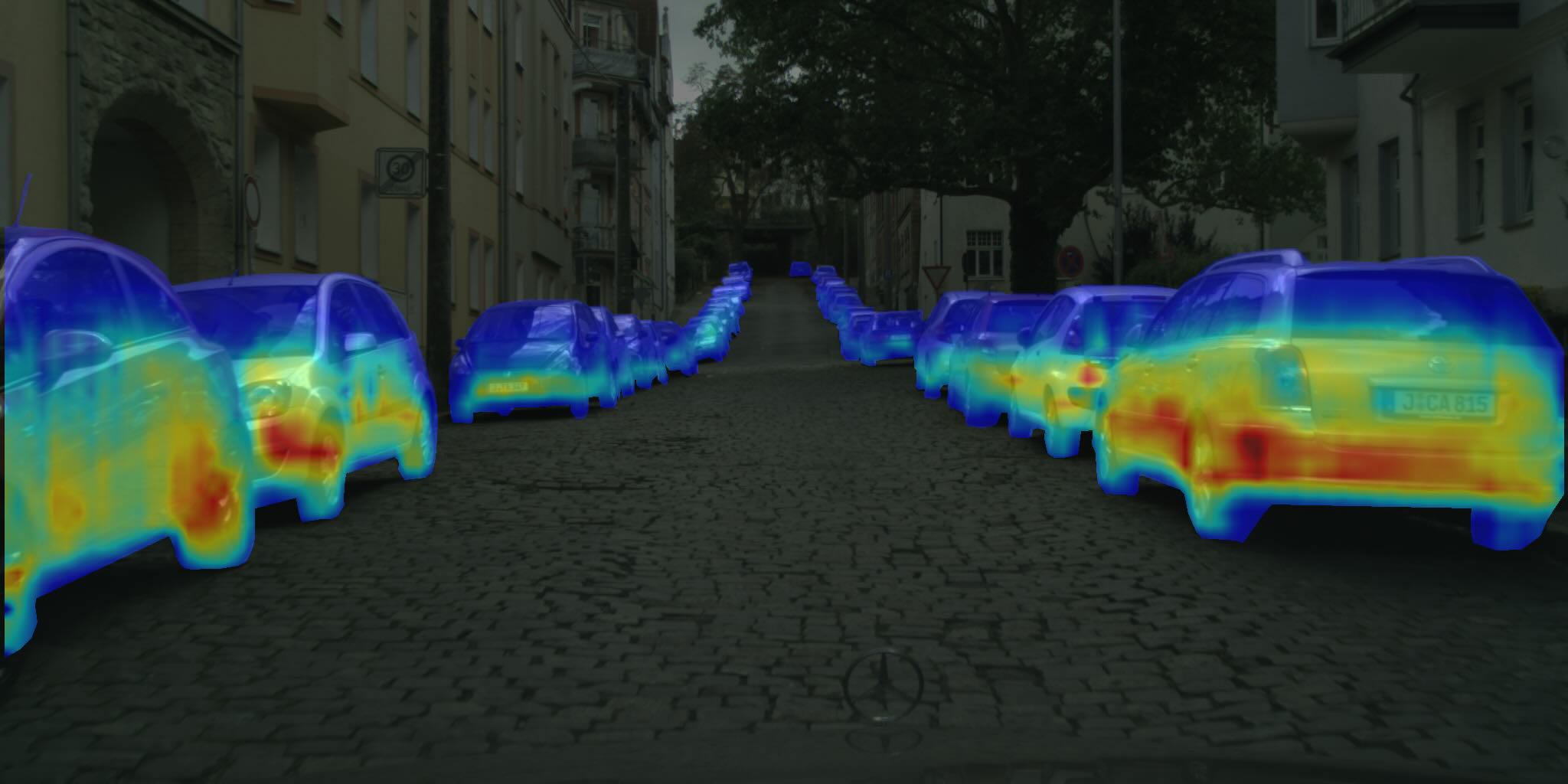}
	\includegraphics[width=0.48\textwidth, height=2.6cm]{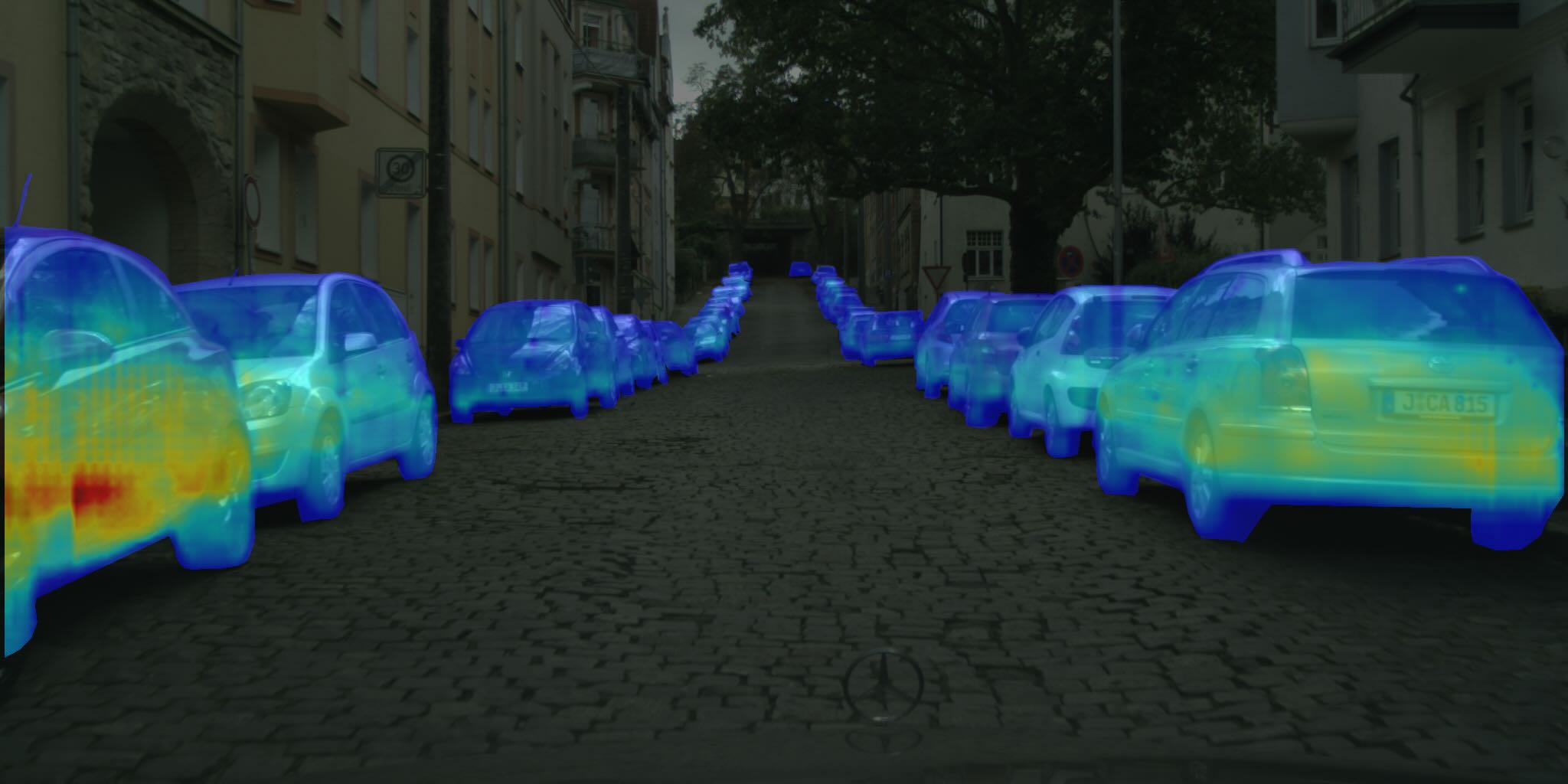}
    \subcaption{Prototype activations on a pair of quasi-equivariant prototypes for an input image downscaled by a ratio of $2$ and $1$ respectively.}
    \end{subfigure}
\begin{subfigure}{\textwidth}
		\includegraphics[width=0.48\textwidth, height=2.6cm]{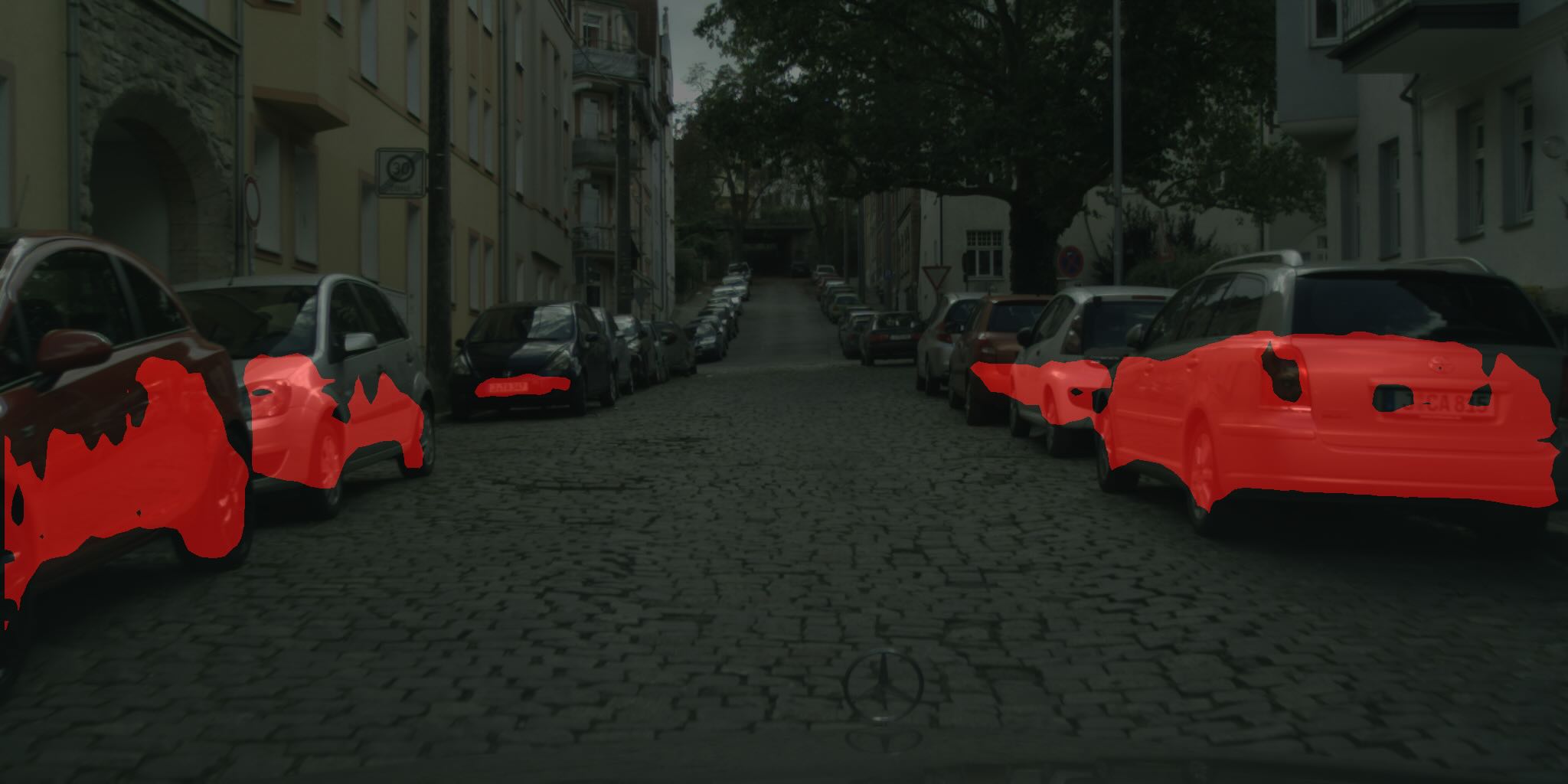}
	\includegraphics[width=0.48\textwidth, height=2.6cm]{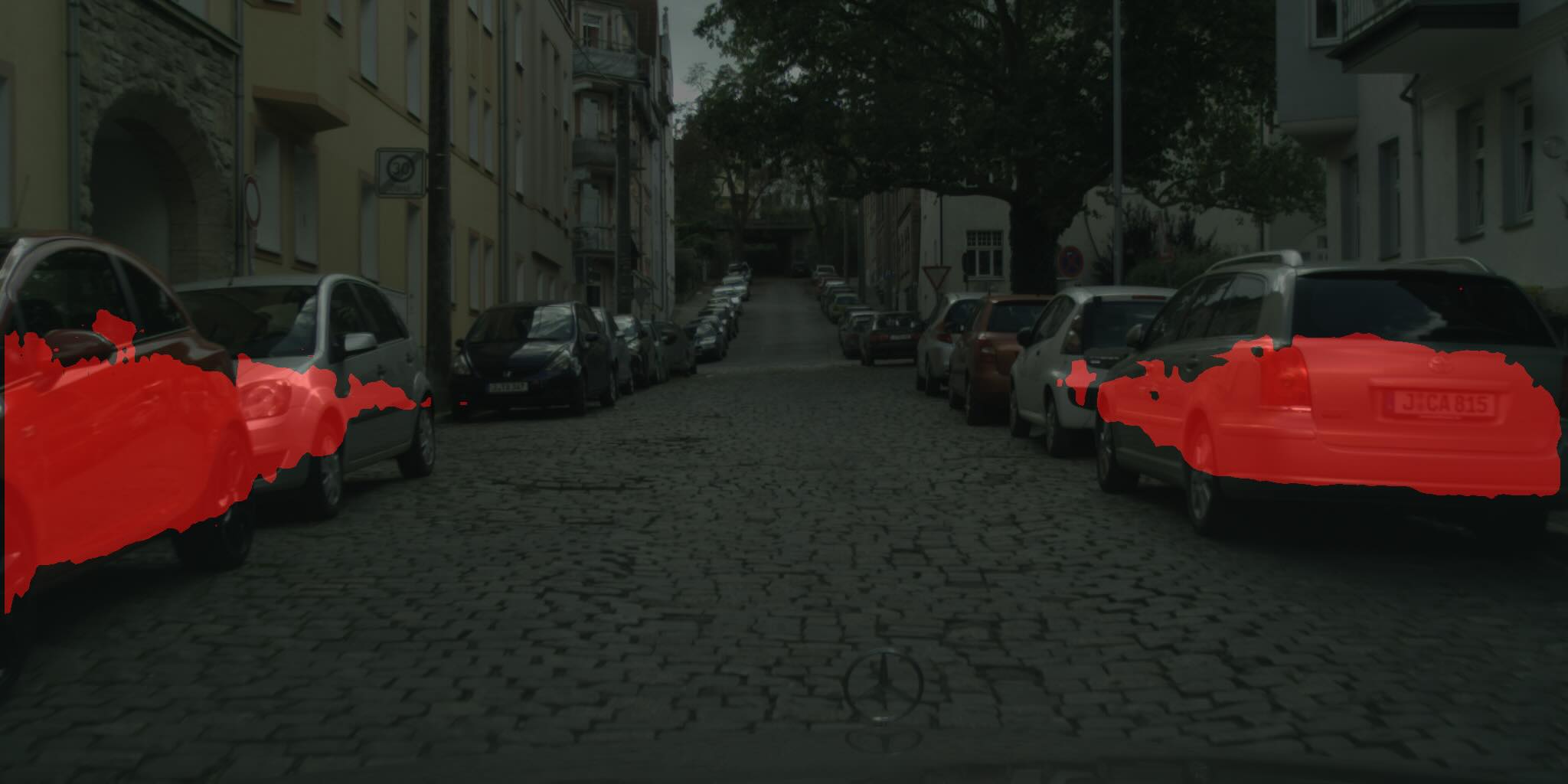}
    \subcaption{Binarized prototype activations on a pair of quasi-equivariant prototypes for an input image downscaled by a ratio of $2$ and $1$ respectively.}
    \end{subfigure}
  \caption{Example of quasi-equivariant pair of prototypes for the class \textit{car} on Cityscapes. Prototype on the left is from \textit{Scale 1} and on the right is from \textit{Scale 2}.}
		\label{fig:car-equiv}
	\end{figure*}

\begin{figure*}[h!]
\centering
 \captionsetup[subfigure]{justification=centering}
  \begin{subfigure}{\textwidth}
	\includegraphics[width=0.48\textwidth, height=2.6cm]{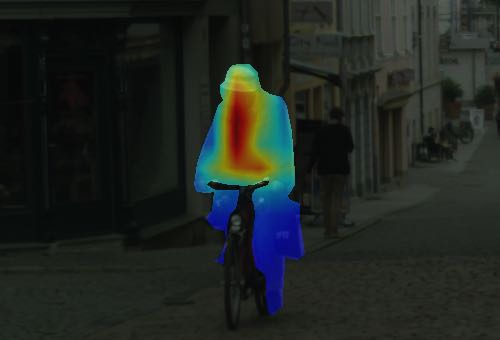}
	\includegraphics[width=0.48\textwidth, height=2.6cm]{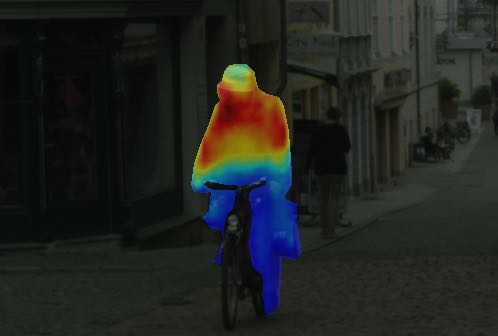}
    \subcaption{Prototype activations on a pair of quasi-equivariant prototypes for an input image downscaled by a ratio of $3$ and $1$ respectively.}
    \end{subfigure}
\begin{subfigure}{\textwidth}
		\includegraphics[width=0.48\textwidth, height=2.6cm]{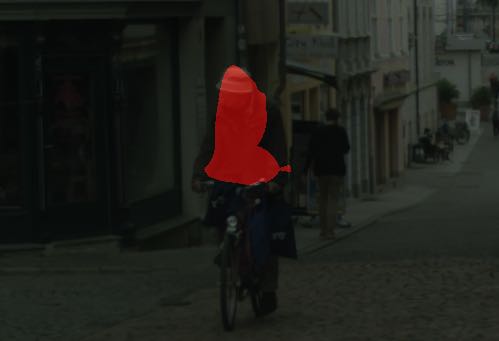}
	\includegraphics[width=0.48\textwidth, height=2.6cm]{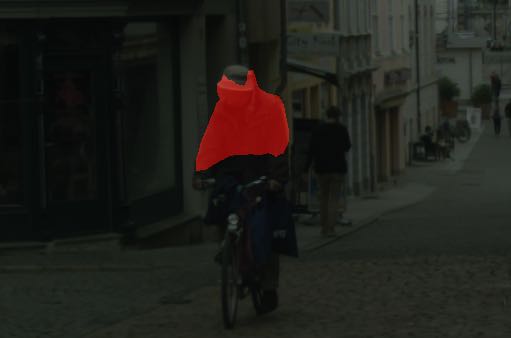}
    \subcaption{Binarized prototype activations on a pair of quasi-equivariant prototypes for an input image downscaled by a ratio of $3$ and $1$ respectively.}
    \end{subfigure}
  \caption{Example of quasi-equivariant pair of prototypes for the class \textit{rider} on Cityscapes. Prototype on the left is from \textit{Scale 1} and on the right is from \textit{Scale 3}.}
		\label{fig:rider-equiv}
	\end{figure*}

\begin{figure*}[h!]
\centering
 \captionsetup[subfigure]{justification=centering}
  \begin{subfigure}{\textwidth}
	\includegraphics[width=0.48\textwidth, height=2.6cm]{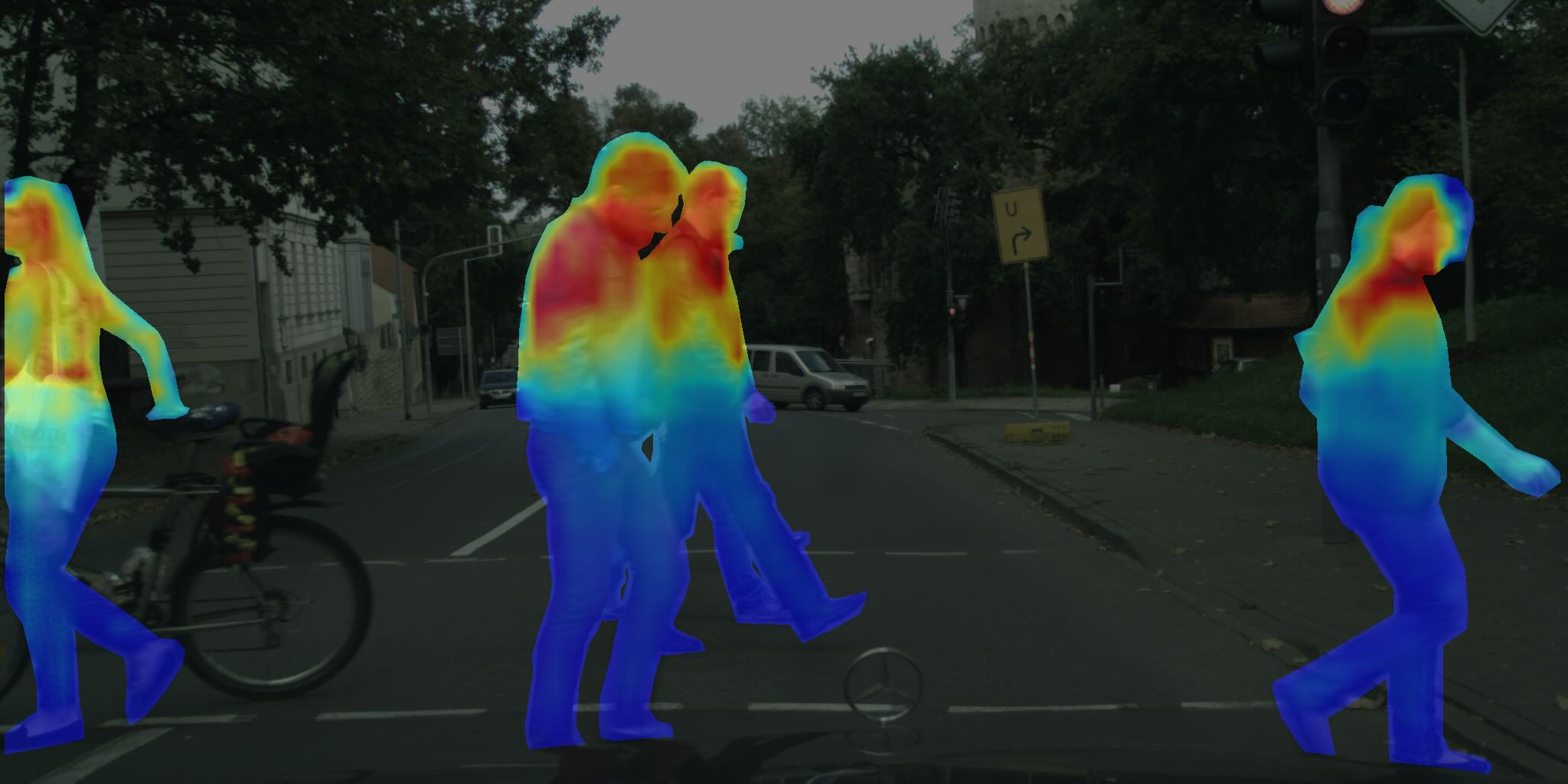}
	\includegraphics[width=0.48\textwidth, height=2.6cm]{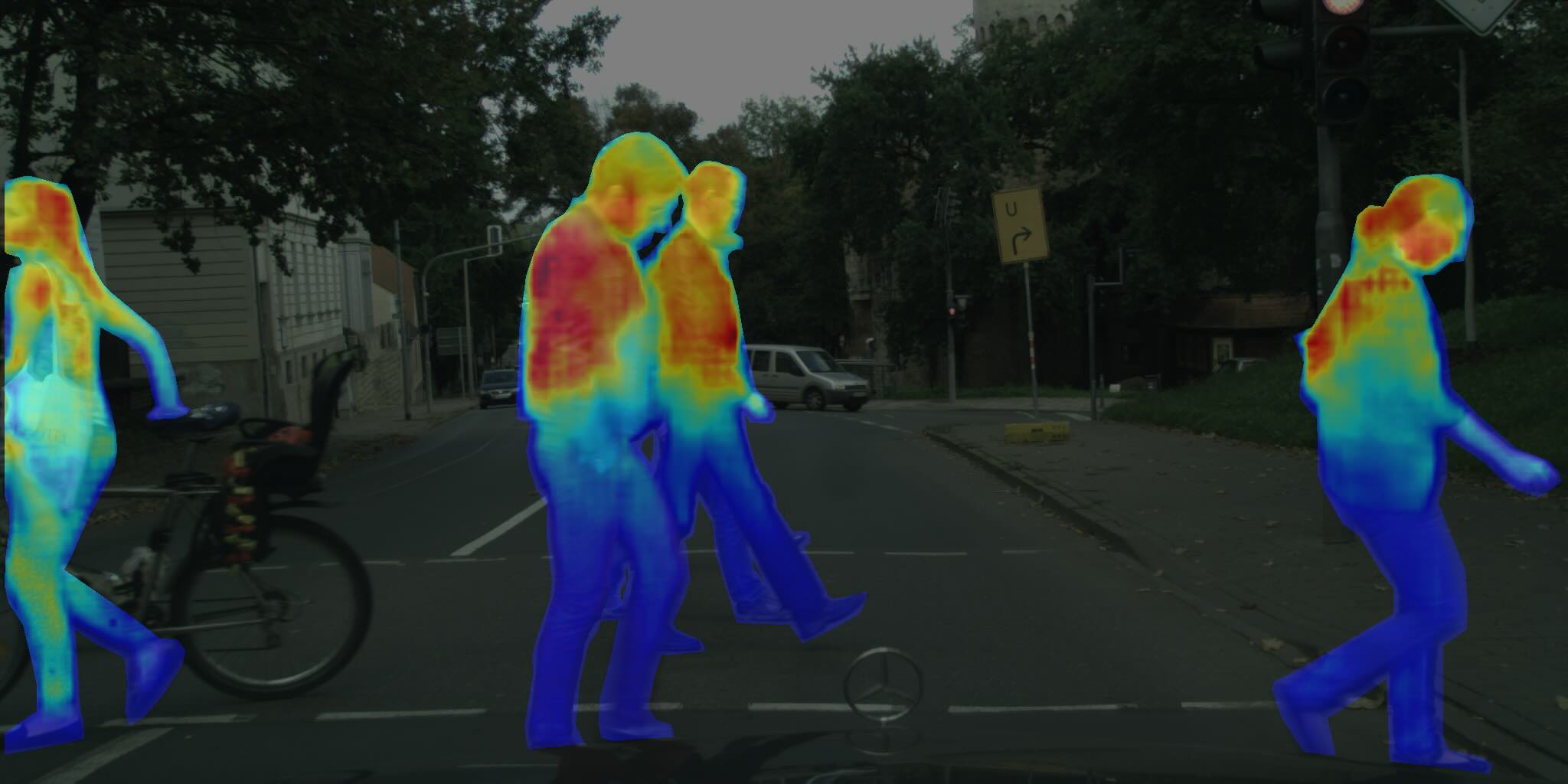}
    \subcaption{Prototype activations on a pair of quasi-equivariant prototypes for an input image downscaled by a ratio of $3$ and $1$ respectively.}
    \end{subfigure}
\begin{subfigure}{\textwidth}
		\includegraphics[width=0.48\textwidth, height=2.6cm]{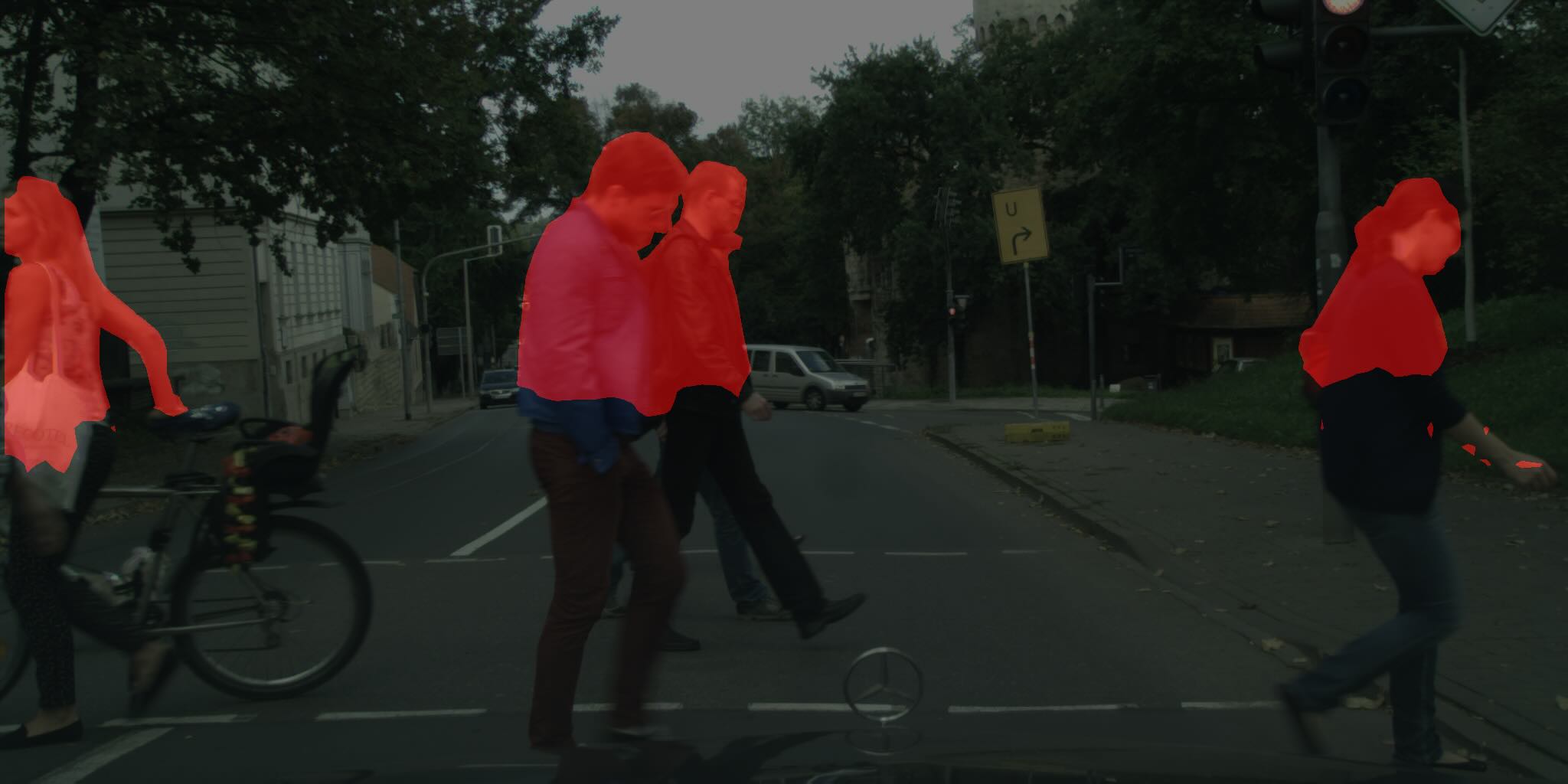}
	\includegraphics[width=0.48\textwidth, height=2.6cm]{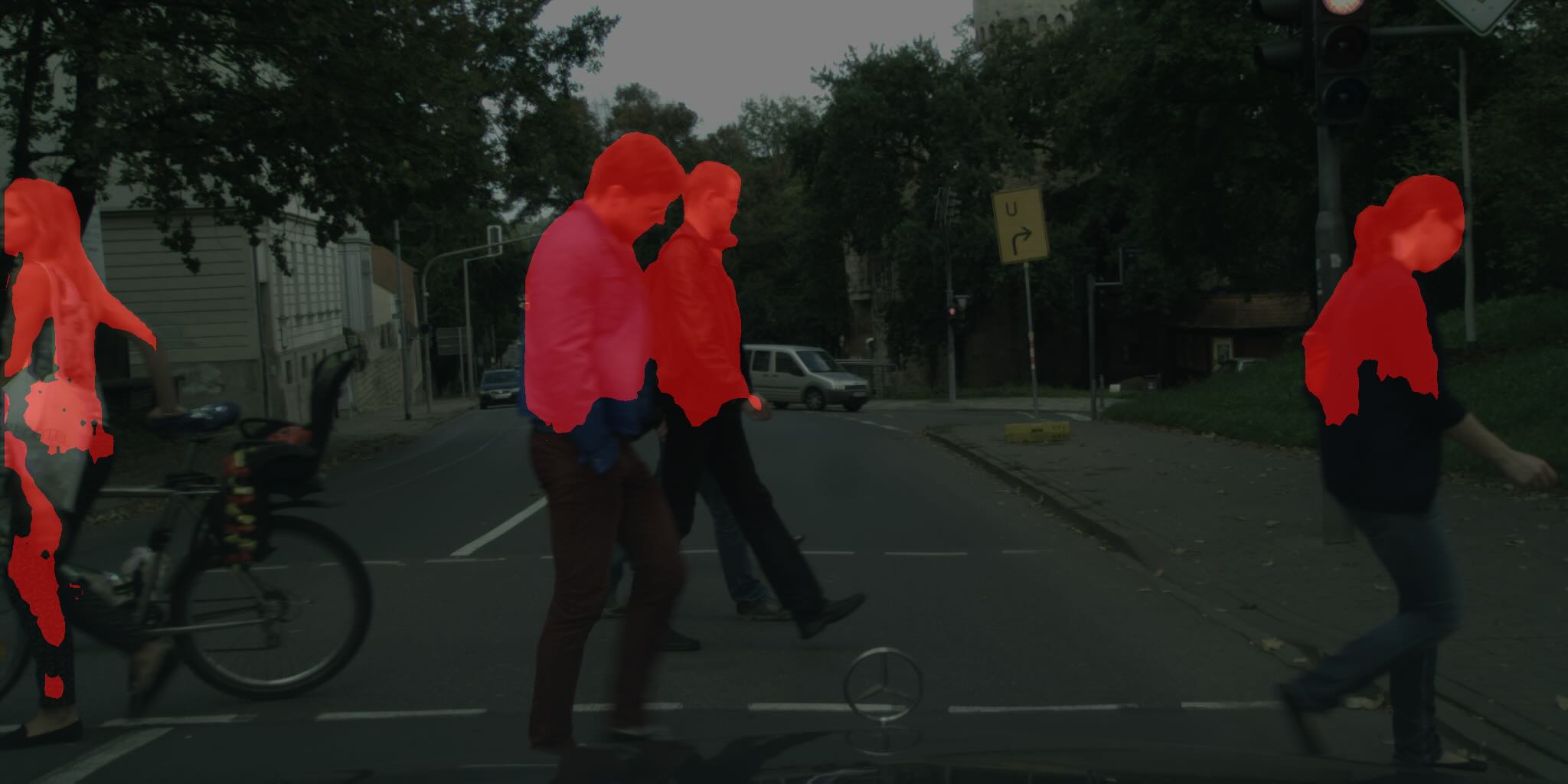}
    \subcaption{Binarized prototype activations on a pair of quasi-equivariant prototypes for an input image downscaled by a ratio of $3$ and $1$ respectively.}
    \end{subfigure}
  \caption{Example of quasi-equivariant pair of prototypes for the class \textit{person} on Cityscapes. Prototype on the left is from \textit{Scale 1} and on the right is from \textit{Scale 3}.}
		\label{fig:person-equiv}
	\end{figure*}

In Figure~\ref{fig:equiv}, we present the results of the equivariance analysis on three datasets. Interestingly, we observe that for both Cityscapes and Pascal VOC there are more than $50\%$ of classes with quasi-equivariant groups for $p_{th} = 0.6$, demonstrating that the network learns through its multi-scale prototypes similar activations for similar prototypical parts in the image at different scales. For ADE20K, we observe that only $12\%$ of classes have quasi-equivariant groups. We argue that since ADE20K is characterized by many complex scenes, as we constrain the representation space to the learning of only 12 prototypes across scales, all the learned prototypes represent scale-specific contextual information. We suppose that if we were to increase the number of prototypes per scale, redundant information would start to appear across scales and more prototypes would be activated on similar prototypical parts at different scales, leading to an increase in the number of classes with quasi-equivariant groups. In Figure~\ref{fig:car-equiv},~\ref{fig:rider-equiv}, and~\ref{fig:person-equiv} we show examples of pair of quasi-equivariant prototypes for $p_{th} = 0.6$.

\section{Sparsity and Computation Overhead}
\label{sec:overhead}

\begin{table}[h!]
\centering
\resizebox{\columnwidth}{!}{
\begin{tabular}{|l|l|c|c|c|}
\hline
\textbf{Dataset}            & \textbf{Method} & \textbf{Avg Weight} & \textbf{Parameters} & \textbf{It/s}\\ \hline
\multirow{2}{*}{Pascal}     & ProtoSeg        & 0.527 $\pm 0.004$                    & 12.1K $\pm 0.3$   & \textbf{20.7} $\pm 0.2$                          \\\cline{2-2}
                            & ScaleProtoSeg   & \textbf{0.063} $\pm 0.001$         & \textbf{9.1K} $\pm 0.0$   &  15.6 $\pm 0.5$                      \\ \hline
\multirow{2}{*}{Cityscapes} & ProtoSeg        & 0.523 $\pm 0.001$                   & 9.9K $\pm 0.2$  &  \textbf{52.5} $\pm 0.2$                   \\ \cline{2-2}
                            & ScaleProtoSeg   & \textbf{0.071} $\pm 0.001$           & \textbf{7.6K}  $\pm 0.1$ &  5.1   $\pm 0.0$                \\ \hline
\multirow{2}{*}{ADE20K}     & ProtoSeg        & 0.503 $\pm 0.001$                   & 238.2K $\pm 2.2$    &  \textbf{51.9} $\pm 2.0$                       \\ \cline{2-2}
                            & ScaleProtoSeg   & \textbf{0.437} $\pm 0.001$          & \textbf{135.4K} $\pm 0.3$   &   14.5 $\pm 0.8$                  \\ \hline
\end{tabular}}
\caption{Analysis of the sparsity of the final classification layer via the average absolute weight in $\mathbf{w}_{h_\text{proto}}$ for ProtoSeg and $\mathbf{w}_{h_\text{group}}$ for ScaleProtoSeg, and the computation overhead with respect to DeepLabv2 induced by the interpretable methods.}
\label{tab:overhead}
\end{table}

In the sparse grouping mechanism proposed in our method, we also enforce a strong sparsity regularization on the last layer $h_\text{proto}$ which enables the model to constrain the negative effect of the prototypes not assigned to a class in the final decision process, as shown in Section~\ref{sec:training-proc}. We see in Table~\ref{tab:overhead} that our method presents a small average absolute weight on $\mathbf{w}_{h_\text{group}}$ and as a consequence a strong sparsity for both Pascal VOC and Cityscapes, especially compared to ProtoSeg. Moreover, despite the increased complexity of ADE20K, our method presents still a smaller average absolute weight than ProtoSeg.

We also analyze in Table~\ref{tab:overhead} the number of extra-parameters necessary in our method and ProtoSeg compared to the black-box baseline: DeepLabv2, after pruning at inference. We observe that, due to the limited number of active prototypes and the group projection, ScaleProtoSeg requires less computational overhead, especially for a large number of classes like in ADE20K, where our method uses $43 \%$ less extra parameters. Those results are computed on 3 local runs per method, and it is important to highlight that ProtoSeg on Pascal despite similar performance presents $\sim 25$ more prototypes after pruning compared to~\cite{sacha2023protoseg}, which also directly impacts the sparsity metric in Table~\ref{tab:perf-ext}.

It is also important to consider the computational overhead during training. First, we showed that the added number of parameters compared to DeepLabv2 is minimal. However, some overhead complexity arises in ProtoSeg and ScaleProto from the KLD regularisation loss and the prototype projection. In particular, the impact on training time of the KLD loss is much more important in the case of ProtoSeg at a high number of prototypes compared to ScaleProtoSeg as in our method we parallelize the loss across scales. Moreover, at training time the fine-tuning of the classification layer and groups after the prototype projection of our method adds only a limited overhead to the training as we tune only a handful of parameters for a few epochs.

Finally as shown in Table~\ref{tab:overhead} at inference our method is slower than ProtoSeg due to the grouping process but we believe that is likely due to a lack of parallelization in our code.

\section{Group Representations and Assignments}

\begin{figure*}[h]
    \centering
    \includegraphics[width=\textwidth]{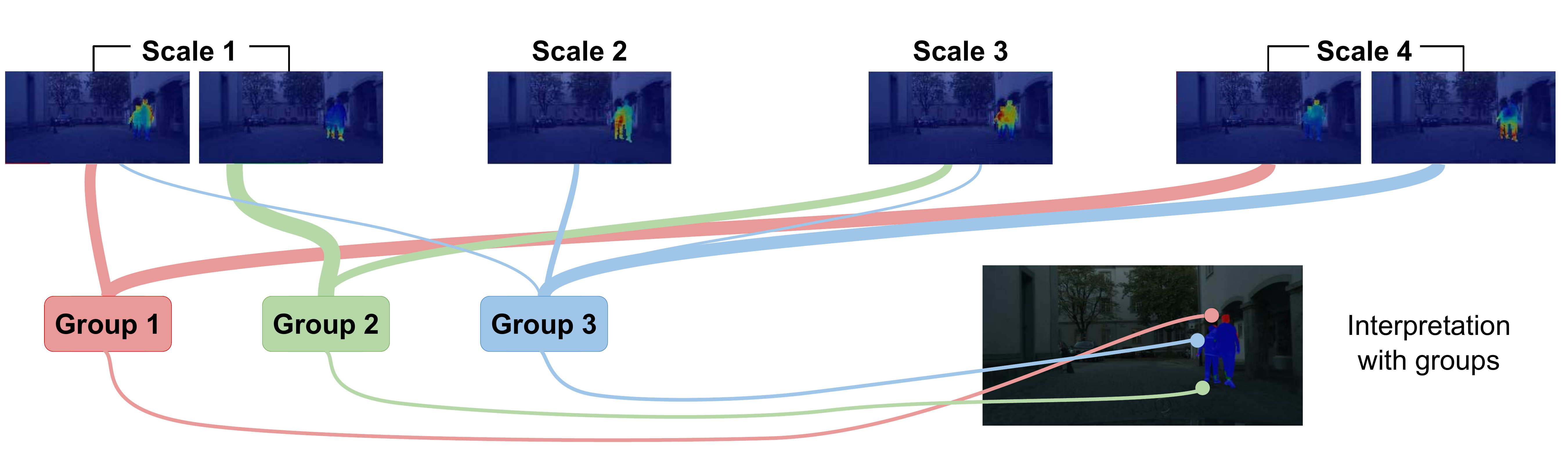}
    \caption{ScaleProtoSeg provides an interpretation of the resulting segmentation through groups of prototypes. For an example of the class \textit{person} on Cityscapes, $2$ prototypes per scale (whose activations are displayed \textbf{at the top} of the figure) are used by the model across the $3$ learned groups shown \textbf{at the bottom right}. For this class, groups can correspond to the feet, the main body or the head of the person.}
    \label{fig:group-pers}
\end{figure*}

In this section, we aim to present more representative examples of our results in terms of interpretability. First, we show in Figure~\ref{fig:group-pers} an example of group activations for the class \textit{person} in Cityscapes, where groups 1 and 2 identify the head and feet respectively of a \textit{person}, while group 3 identifies the main body of the \textit{persons}. Interestingly, similarly to the class \textit{car} in the main paper, group 2 is mainly activated by a prototype at scale 1 and as a consequence seems to be more present on \textit{person} further away in the image as seen in \textbf{(b)}. This level of understanding of the scale effect on the model representation is a key contribution of our method compared to ProtoSeg in terms of interpretability.

\begin{figure*}[h]
\centering
\setlength\tabcolsep{1.5pt} 
 \begin{tabular}{lccc}
 \rotatebox{90}{\quad\quad Original} & 
 \includegraphics[width=0.3\textwidth,  height=2.3cm]{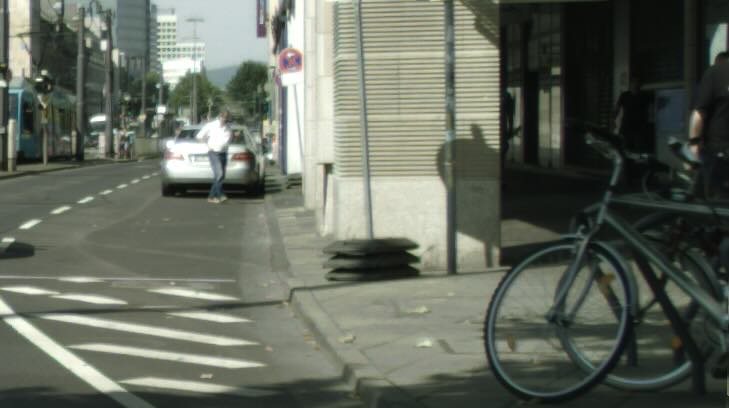} & 
 \includegraphics[width=0.3\textwidth, height=2.3cm]{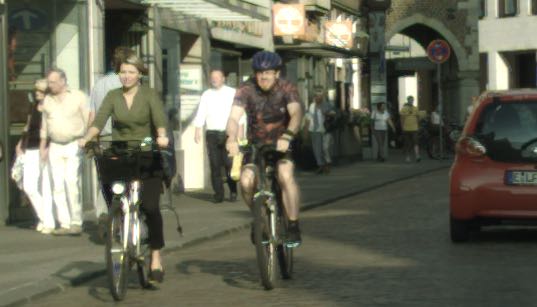} & 
 \includegraphics[width=0.3\textwidth,height=2.3cm]{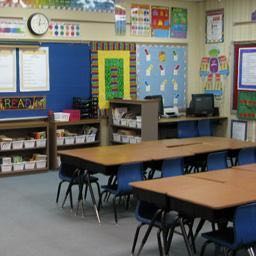} \\
 \rotatebox{90}{\parbox{2.3cm}{\centering ProtoSeg \cite{sacha2023protoseg} prototypes}} & 
 \includegraphics[width=0.3\textwidth,height=2.3cm]{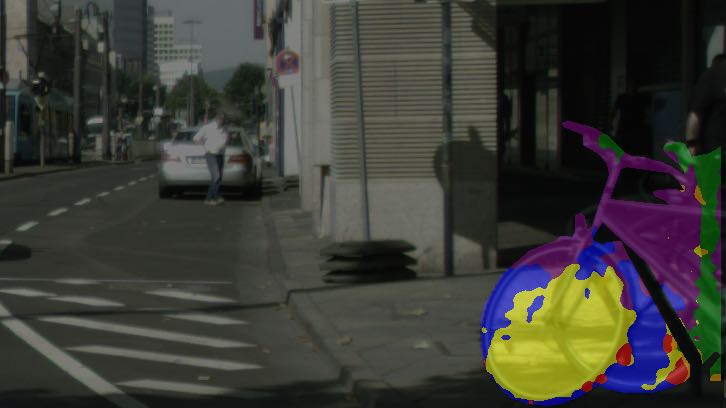} & 
 \includegraphics[width=0.3\textwidth,height=2.3cm]{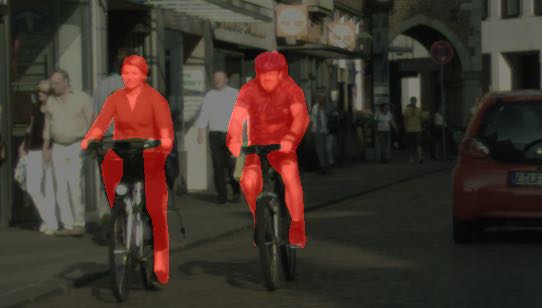} & 
 \includegraphics[width=0.3\textwidth, height=2.3cm]{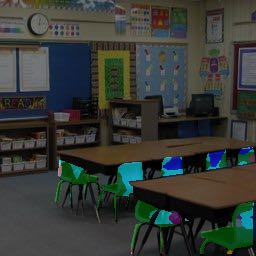} \\
 \rotatebox{90}{\parbox{2.3cm}{\centering ScaleProtoSeg groups}} & 
 \includegraphics[width=0.3\textwidth, height=2.3cm]{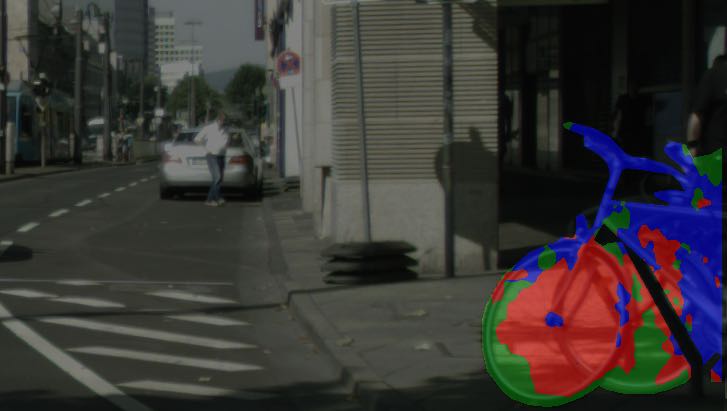} & 
 \includegraphics[width=0.3\textwidth,height=2.3cm]{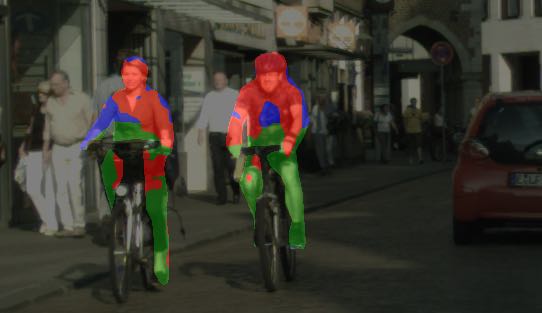} & 
 \includegraphics[width=0.3\textwidth, height=2.3cm]{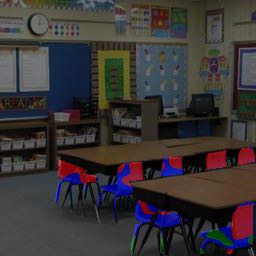} \\
 \rotatebox{90}{\parbox{2.3cm}{\centering ProtoSeg \cite{sacha2023protoseg} output}} & 
 \includegraphics[width=0.3\textwidth,height=2.3cm]{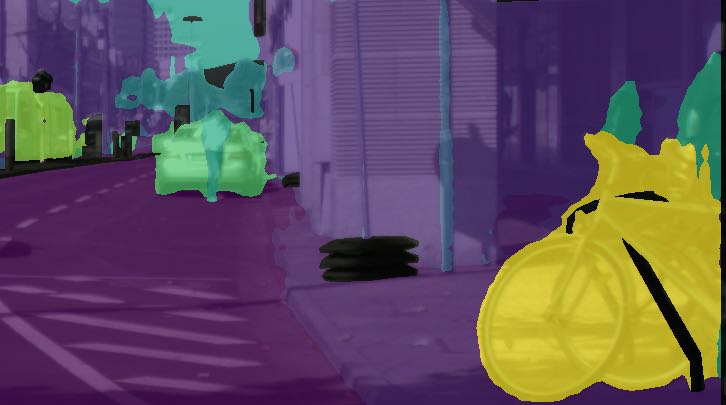} &
  \includegraphics[width=0.3\textwidth,height=2.3cm]{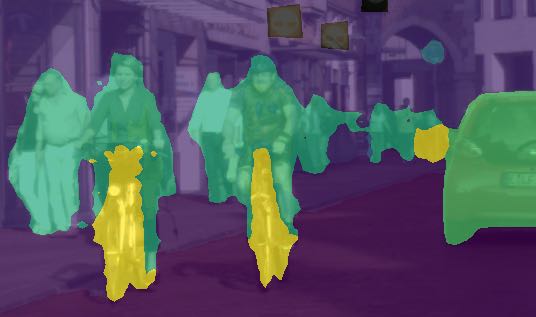} & 
 \includegraphics[width=0.3\textwidth,height=2.3cm]{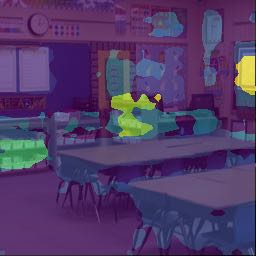} \\
 \rotatebox{90}{{\parbox{2.3cm}{\centering ScaleProtoSeg output}}} & 
 \includegraphics[width=0.3\textwidth,height=2.3cm]{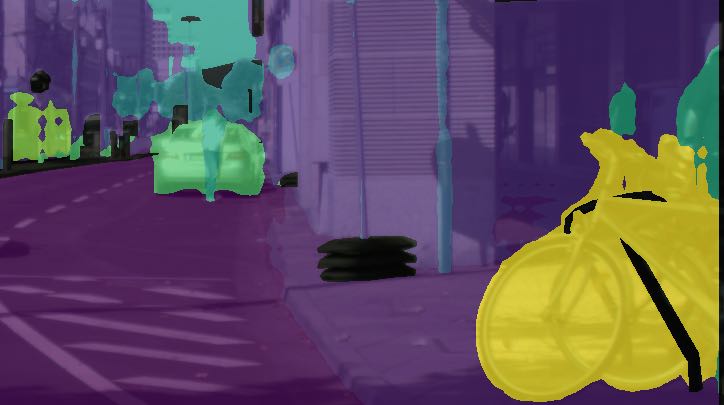} & 
 \includegraphics[width=0.3\textwidth,height=2.3cm]{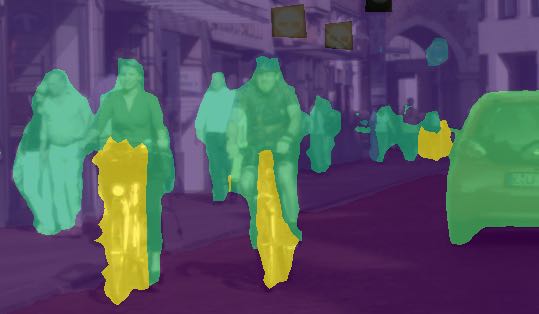} & 
 \includegraphics[width=0.3\textwidth,height=2.3cm]{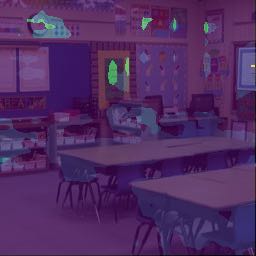}
 
 \end{tabular}
 \caption{Model prototype, group assignments and prediction for the class \textit{bicycle}, \textit{rider}, and \textit{chair} on Cityscapes, and ADE20K.\vspace{-2em}}
 \label{fig:output-bis}
\end{figure*}

\begin{figure*}[h]
\centering
\setlength\tabcolsep{1.5pt} 
 \begin{tabular}{lccc}
 \rotatebox{90}{\quad\quad Original} & 
 \includegraphics[width=0.3\textwidth,  height=2.3cm]{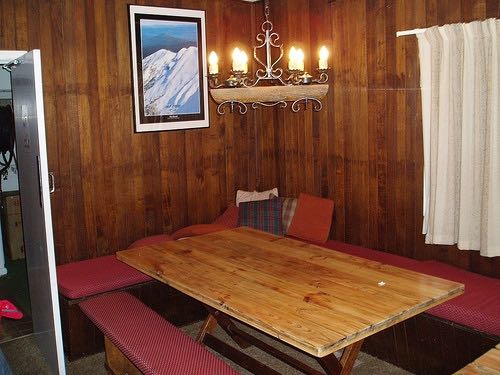} & 
 \includegraphics[width=0.3\textwidth, height=2.3cm]{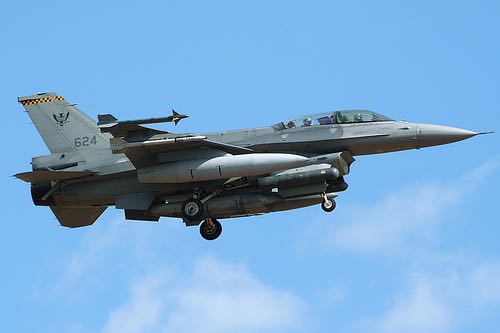} & 
 \includegraphics[width=0.3\textwidth,height=2.3cm]{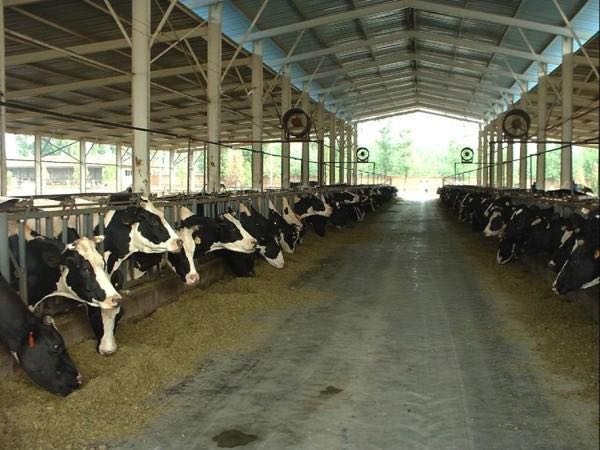} \\
 \rotatebox{90}{\parbox{2.3cm}{\centering ProtoSeg \cite{sacha2023protoseg} prototypes}} & 
 \includegraphics[width=0.3\textwidth,height=2.3cm]{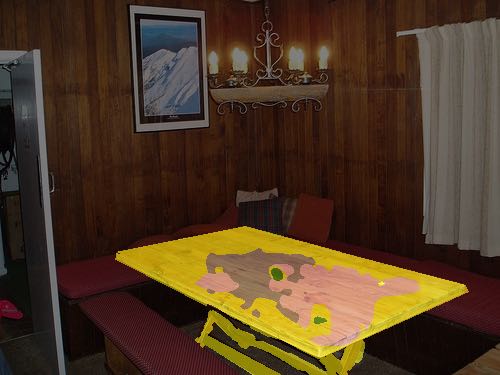} & 
 \includegraphics[width=0.3\textwidth,height=2.3cm]{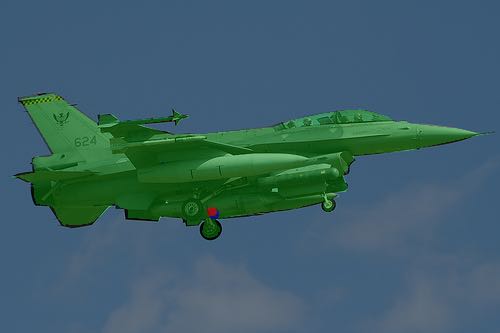} & 
 \includegraphics[width=0.3\textwidth, height=2.3cm]{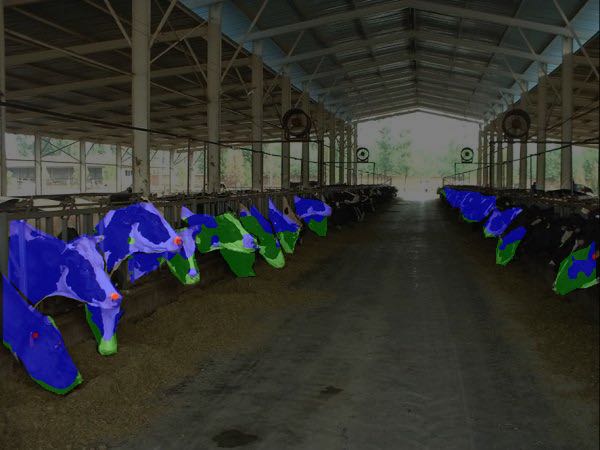} \\
 \rotatebox{90}{\parbox{2.3cm}{\centering ScaleProtoSeg groups}} & 
 \includegraphics[width=0.3\textwidth, height=2.3cm]{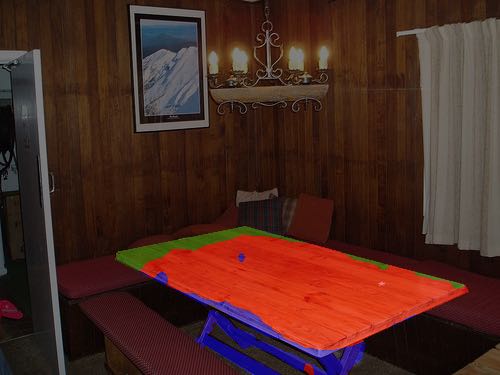} & 
 \includegraphics[width=0.3\textwidth,height=2.3cm]{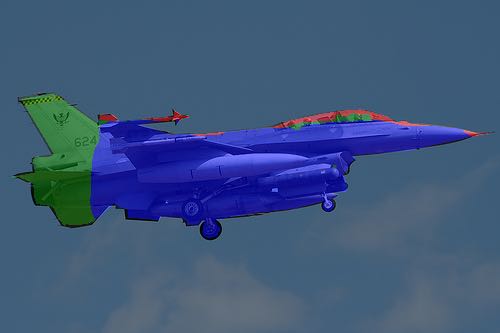} & 
 \includegraphics[width=0.3\textwidth, height=2.3cm]{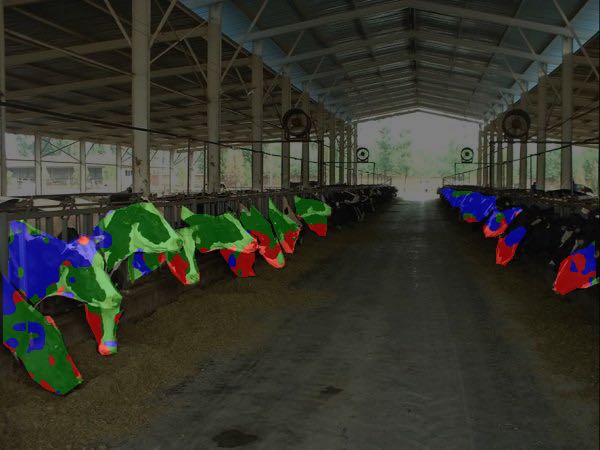} \\
 \rotatebox{90}{\parbox{2.3cm}{\centering ProtoSeg \cite{sacha2023protoseg} output}} & 
 \includegraphics[width=0.3\textwidth,height=2.3cm]{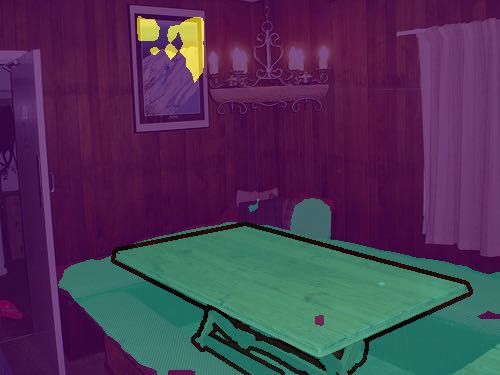} &
  \includegraphics[width=0.3\textwidth,height=2.3cm]{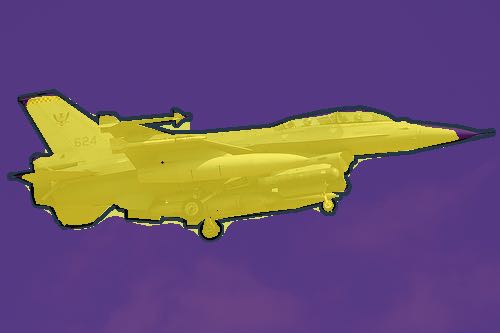} & 
 \includegraphics[width=0.3\textwidth,height=2.3cm]{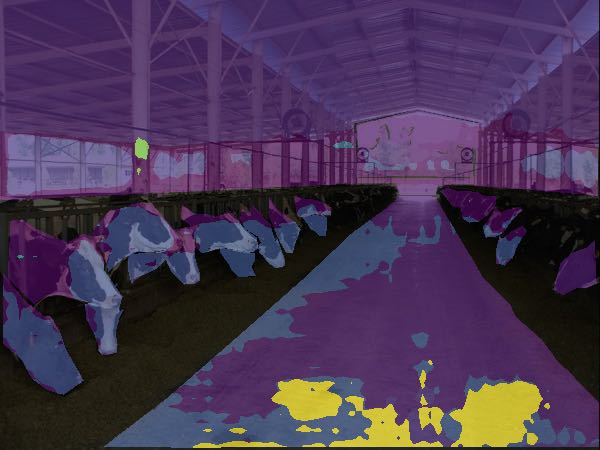} \\
 \rotatebox{90}{{\parbox{2.3cm}{\centering ScaleProtoSeg output}}} & 
 \includegraphics[width=0.3\textwidth,height=2.3cm]{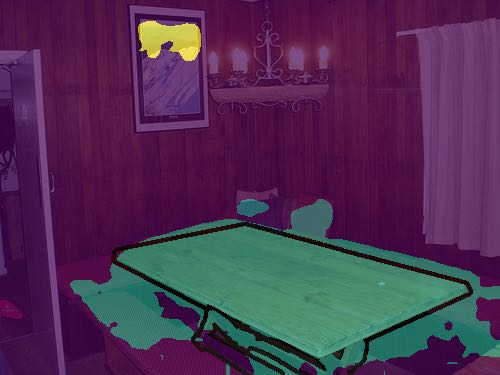} & 
 \includegraphics[width=0.3\textwidth,height=2.3cm]{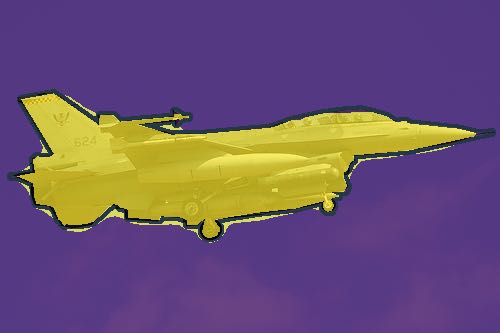} & 
 \includegraphics[width=0.3\textwidth,height=2.3cm]{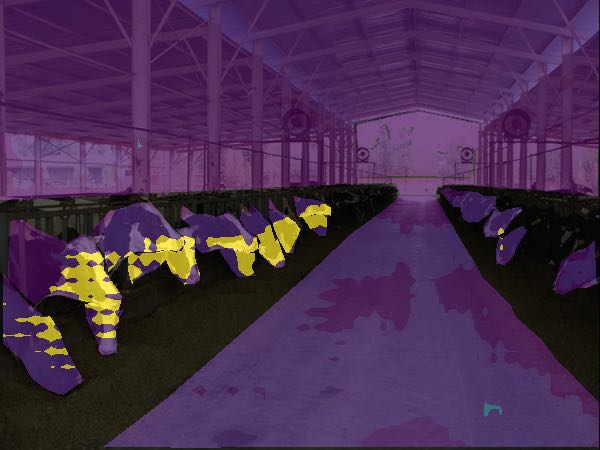}
 
 \end{tabular}
 \caption{Model prototype, group assignments and prediction for the class \textit{dinning table}, \textit{aeroplane}, and \textit{animal} on Pascal, and ADE20K.}
 \label{fig:output-tri}
\end{figure*}

Moreover, we show in Figure~\ref{fig:output-bis} and~\ref{fig:output-tri} more comparative results between our method and ProtoSeg, where again the limitations on the number of groups simplify the analysis of the decision process for our method as seen for the class \textit{bicycle} and \textit{chair}. In the analysis of the class \textit{rider} and \textit{aeroplane}, we can observe another advantage of our method as ProtoSeg presents a failure case of prototype pruning where only one prototype specific to the class \textit{rider} or \textit{aeroplane} is left after post-processing.

\section{Nearest Image Patches to Prototypes}

\begin{figure*}[htbp]
    \centering
    \begin{tabular}{ccc}
        \rotatebox{90}{{\parbox{4cm}{\centering Prototypes}}} &
        \begin{subfigure}[b]{0.45\linewidth}
            \centering
            \includegraphics[width=\linewidth, height=4cm]{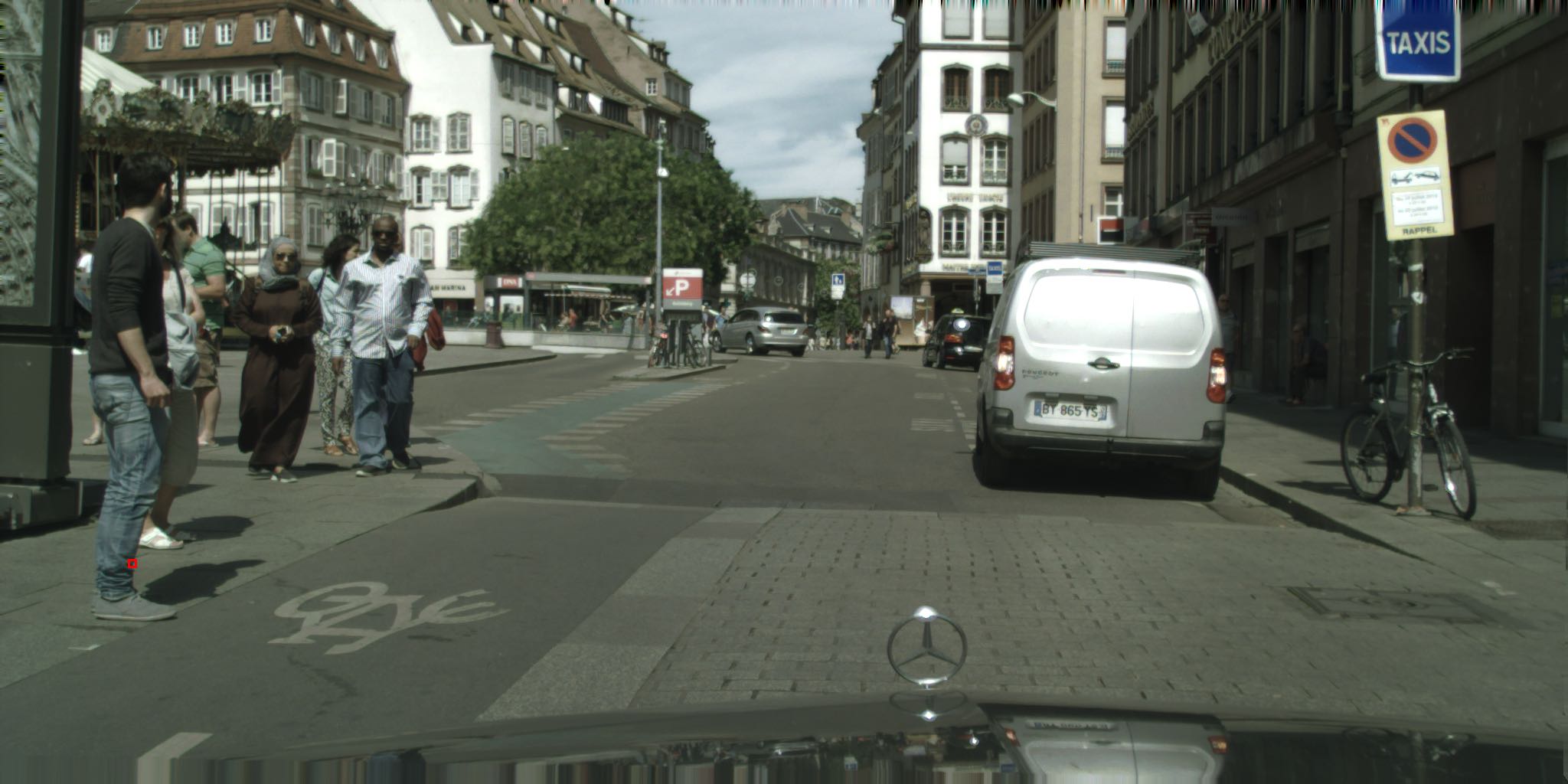}
        \end{subfigure} &
        \begin{subfigure}[b]{0.45\linewidth}
            \centering
            \includegraphics[width=\linewidth, height=4cm]{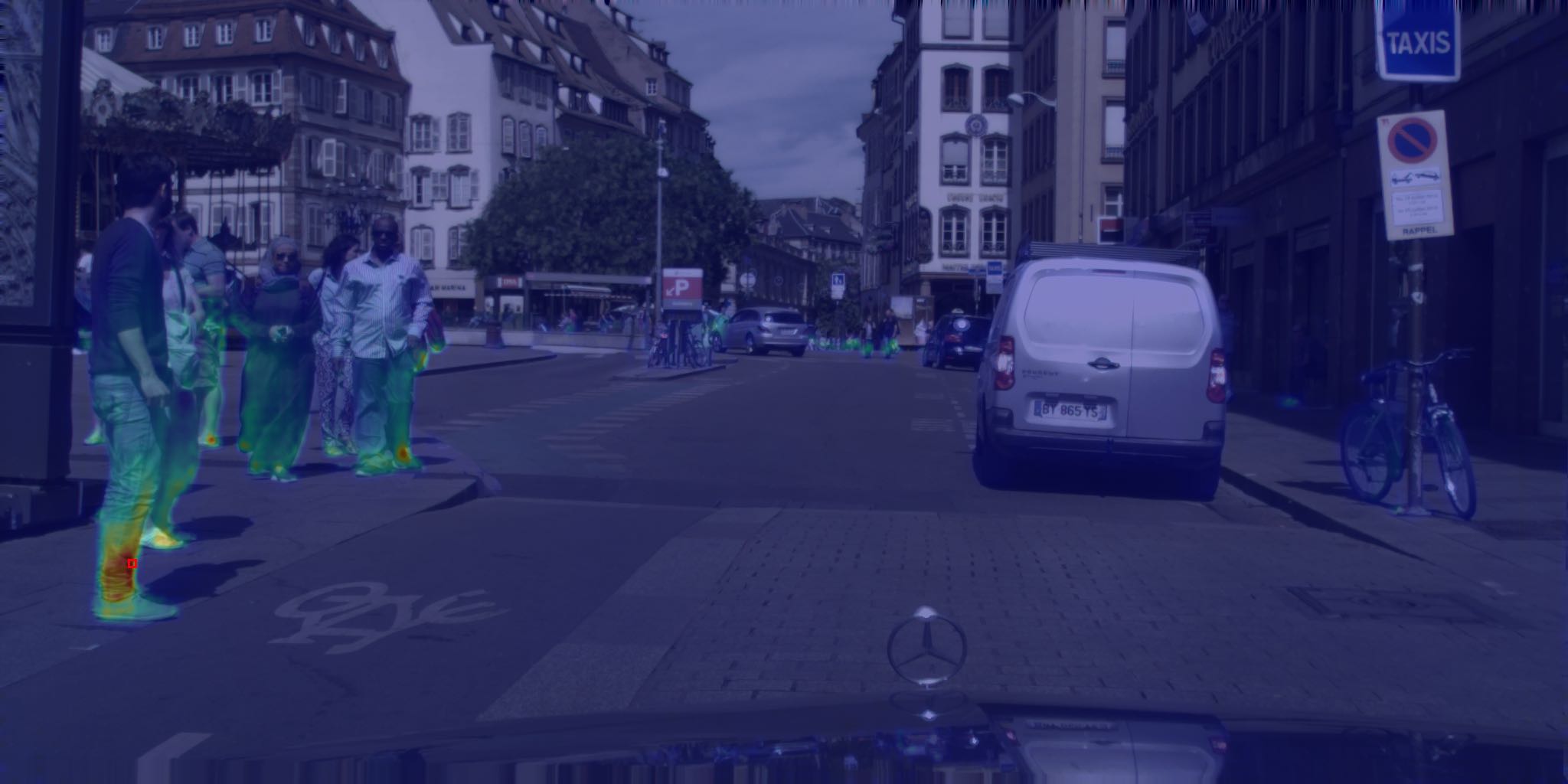}
        \end{subfigure}
    \end{tabular}
    
    \vspace{2mm}
    
    \rule{\linewidth}{0.5pt}
    
    \vspace{5mm}
    
    \begin{tabular}{ccc}
        \rotatebox{90}{{\parbox{4cm}{\centering Top 1 Patch}}} &
        \begin{subfigure}[b]{0.45\linewidth}
            \centering
            \includegraphics[width=\linewidth, height=4cm]{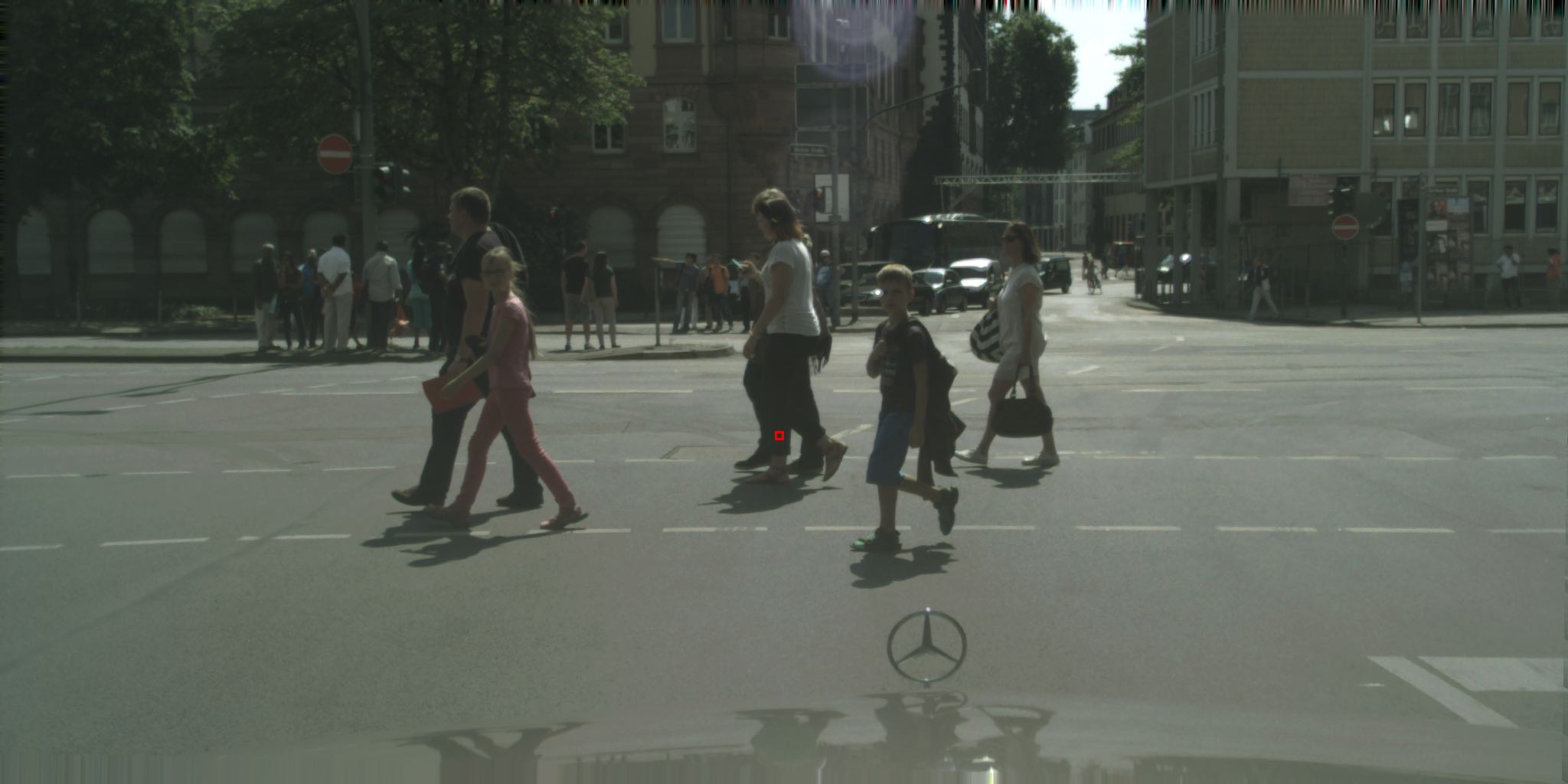}
        \end{subfigure} &
        \begin{subfigure}[b]{0.45\linewidth}
            \centering
            \includegraphics[width=\linewidth, height=4cm]{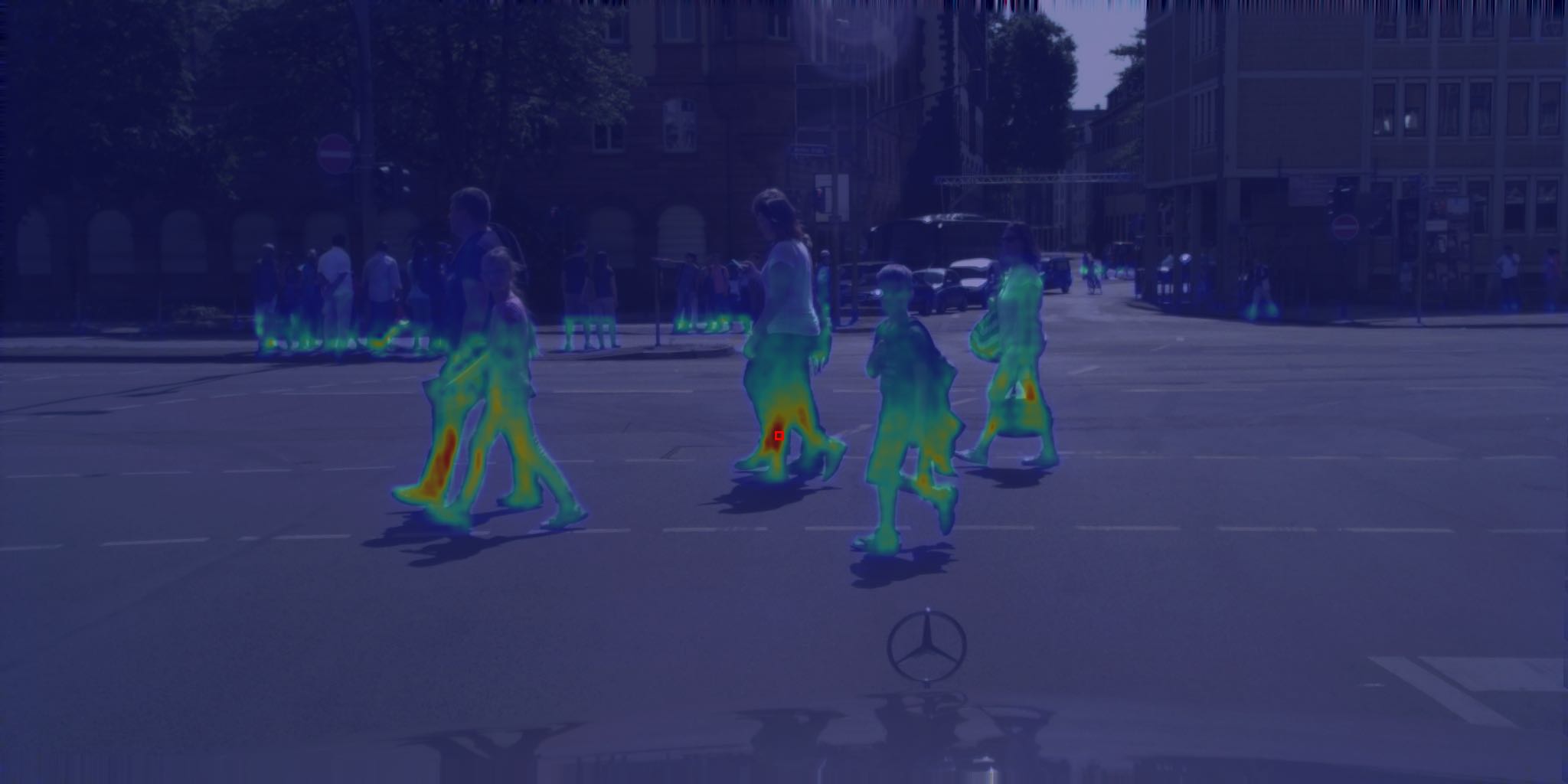}
        \end{subfigure} \\
        \rotatebox{90}{{\parbox{4cm}{\centering Top 2 Patch}}} &
        \begin{subfigure}[b]{0.45\linewidth}
            \centering
            \includegraphics[width=\linewidth, height=4cm]{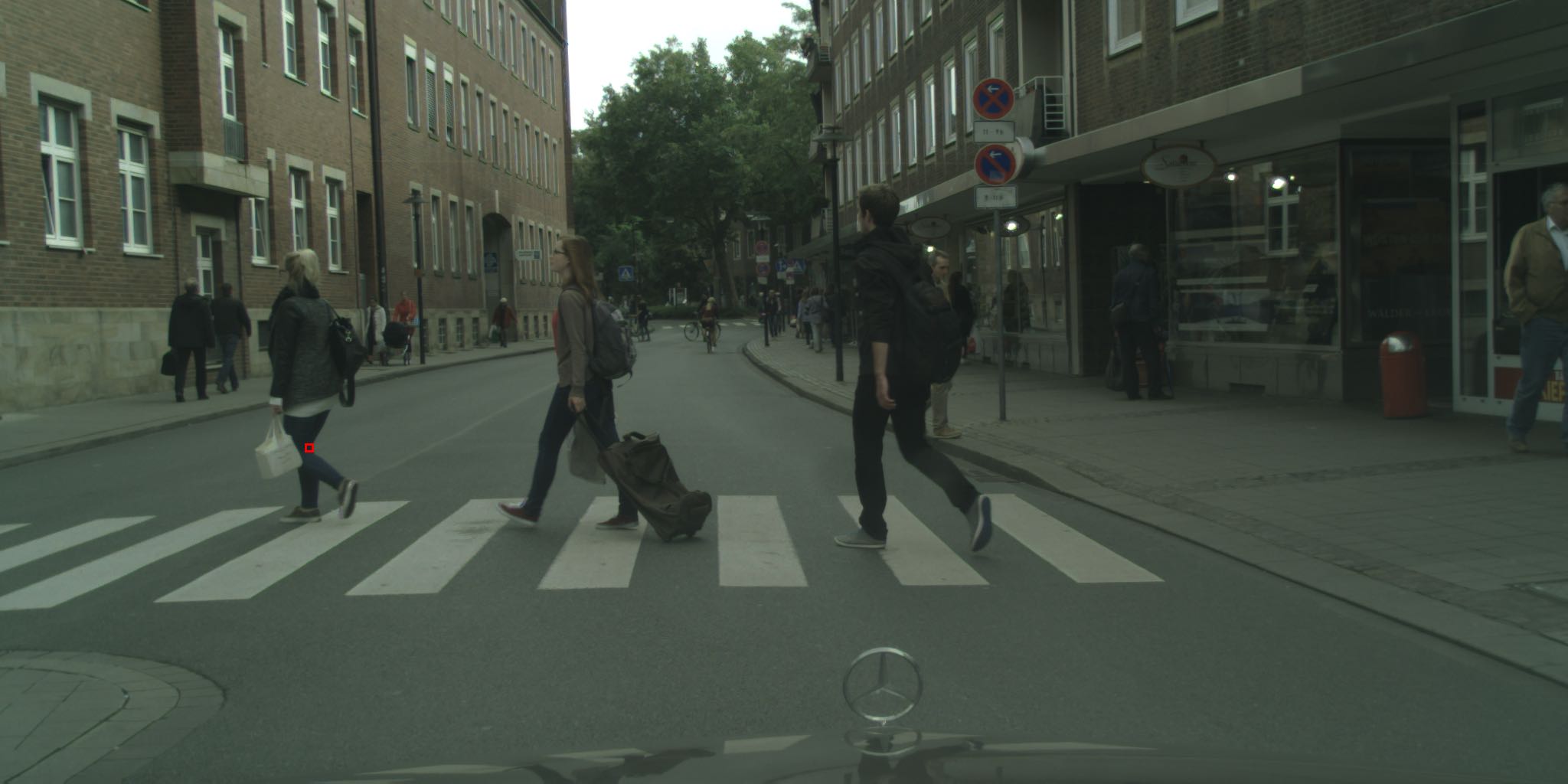}
        \end{subfigure} &
        \begin{subfigure}[b]{0.45\linewidth}
            \centering
            \includegraphics[width=\linewidth, height=4cm]{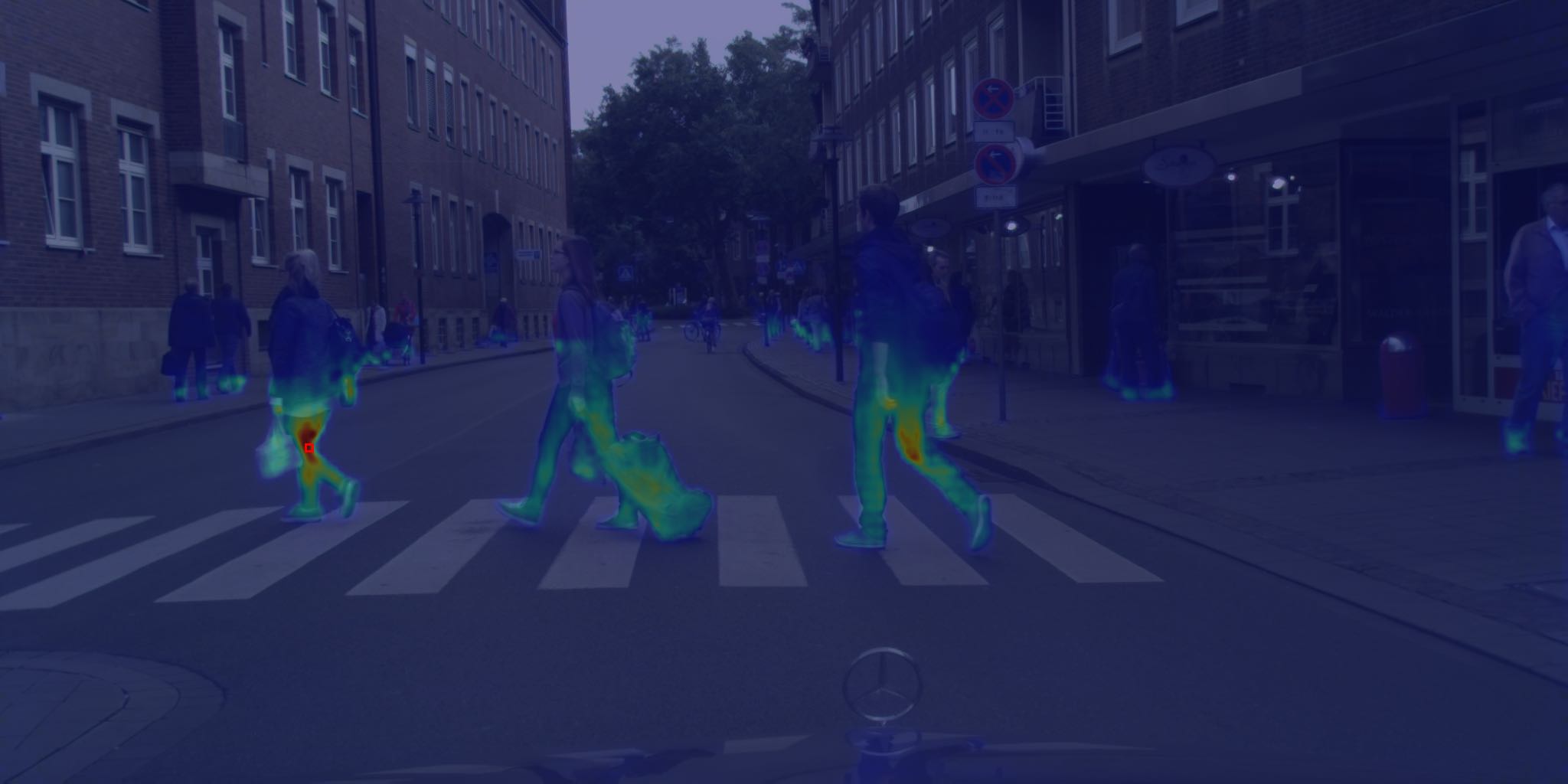}
        \end{subfigure} \\
        \rotatebox{90}{{\parbox{4cm}{\centering Top 3 Patch}}} &
        \begin{subfigure}[b]{0.45\linewidth}
            \centering
            \includegraphics[width=\linewidth, height=4cm]{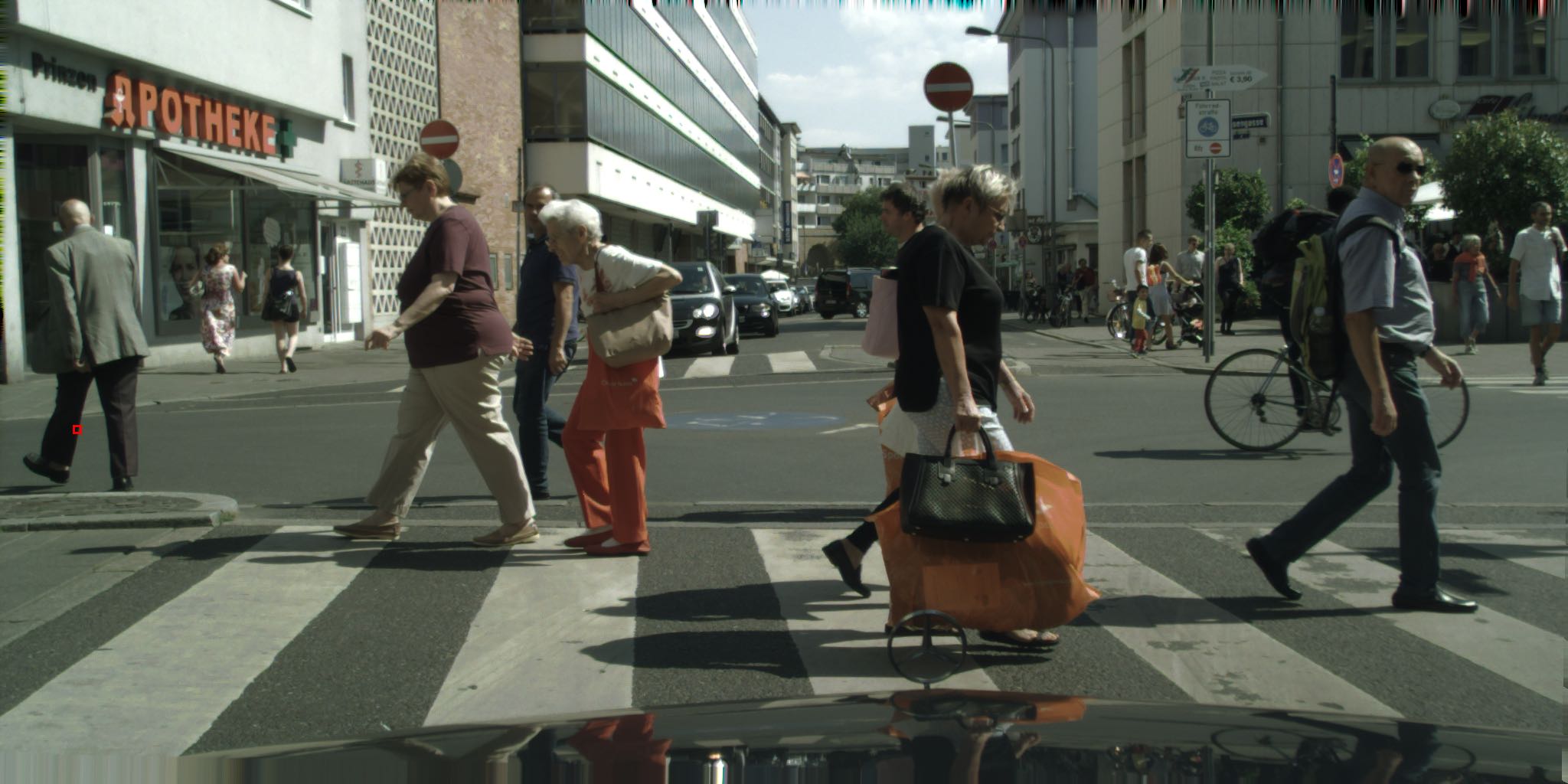}
        \end{subfigure} &
        \begin{subfigure}[b]{0.45\linewidth}
            \centering
            \includegraphics[width=\linewidth, height=4cm]
            {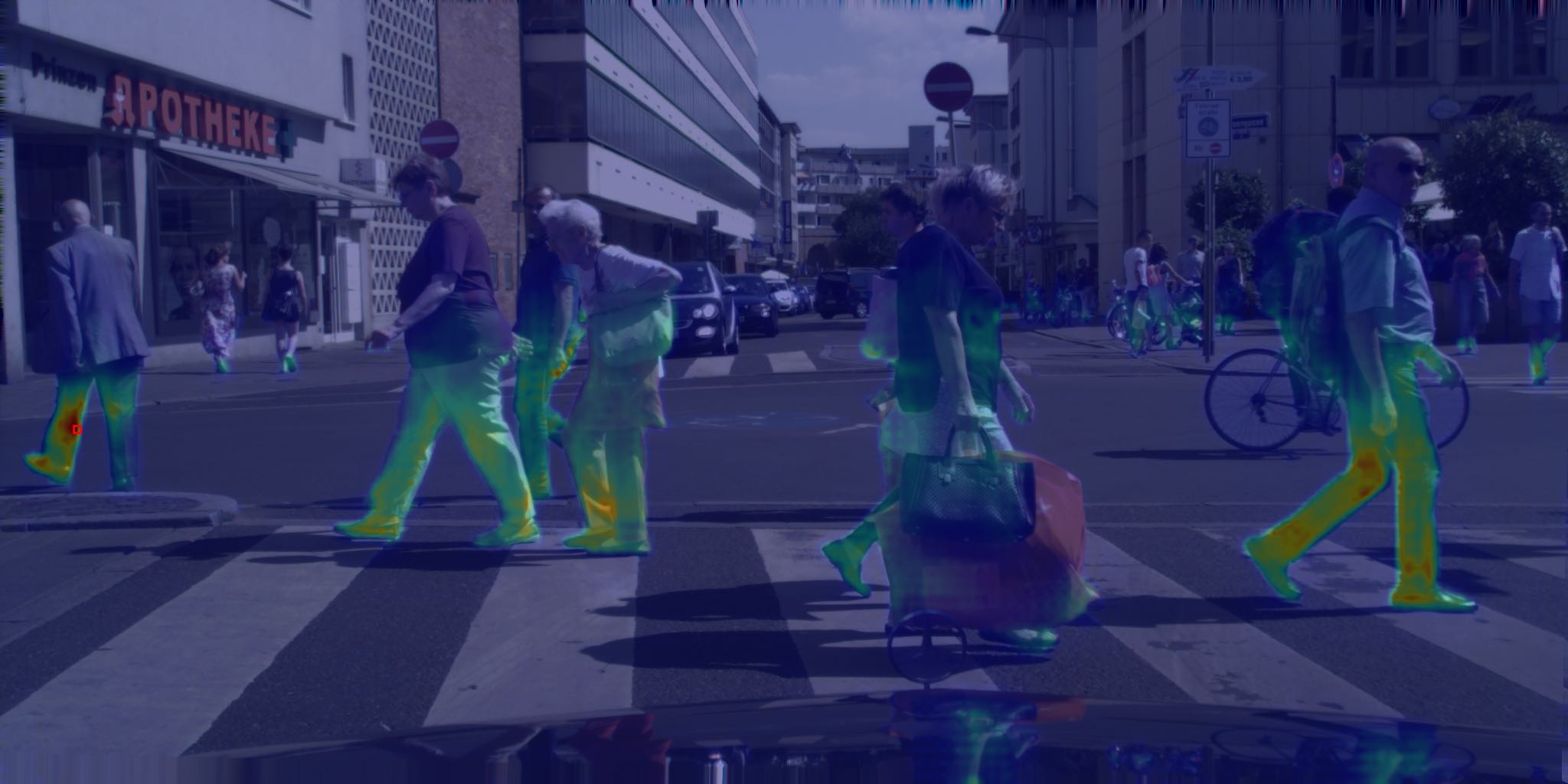}
        \end{subfigure}
    \end{tabular}
    
    \caption{The first row represents a prototype for the class \textit{person} in Cityscapes marked by a red box and the activation of the prototype on its image. The second to fourth rows show the closest patches to the prototype and its activation on the images containing the patches.}
    \label{fig:person-proto}
\end{figure*}

\begin{figure*}[htbp]
    \centering
    \begin{tabular}{ccc}
        \rotatebox{90}{{\parbox{4cm}{\centering Prototypes}}} &
        \begin{subfigure}[b]{0.45\linewidth}
            \centering
            \includegraphics[width=\linewidth, height=4cm]{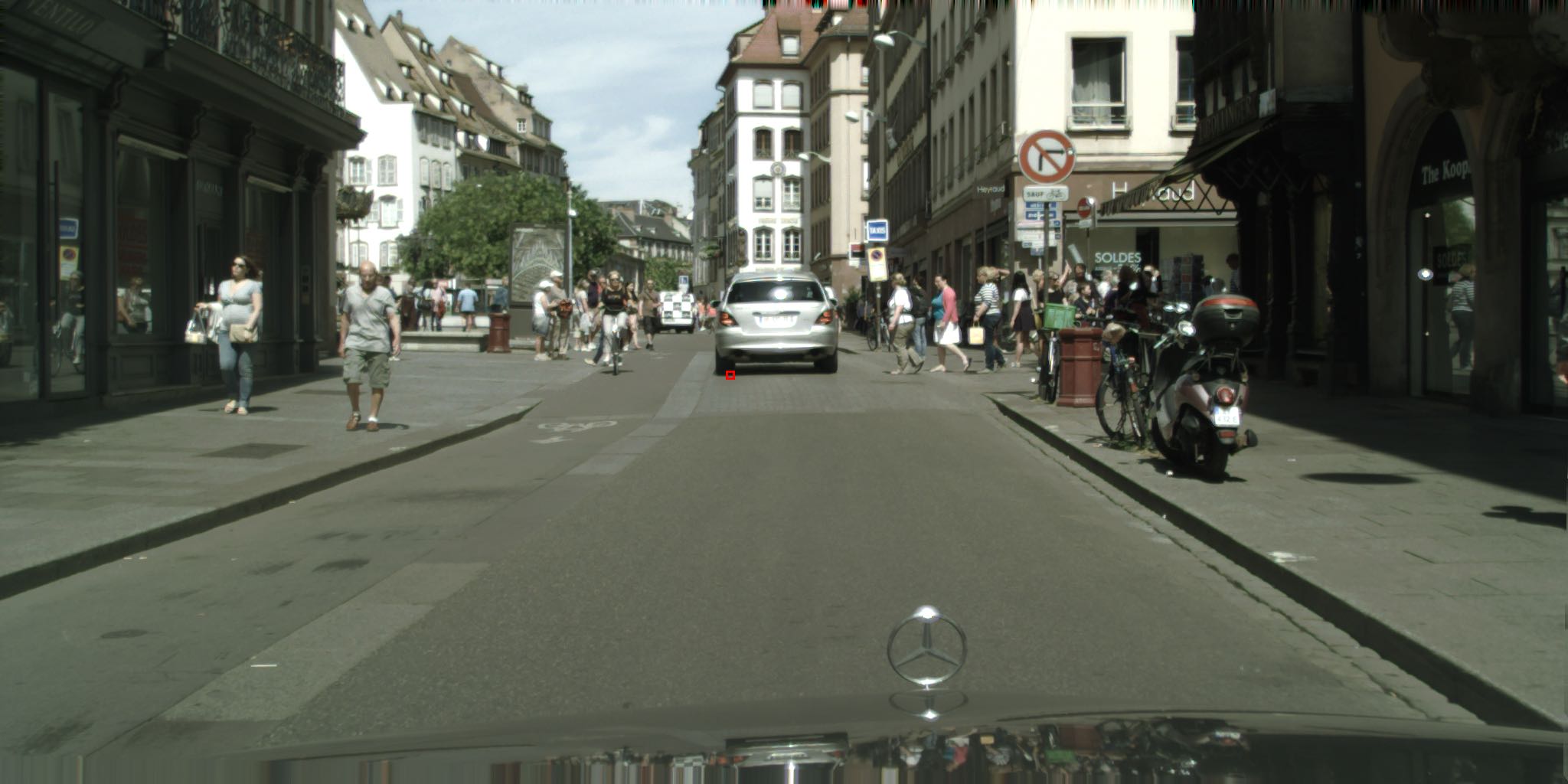}
        \end{subfigure} &
        \begin{subfigure}[b]{0.45\linewidth}
            \centering
            \includegraphics[width=\linewidth, height=4cm]{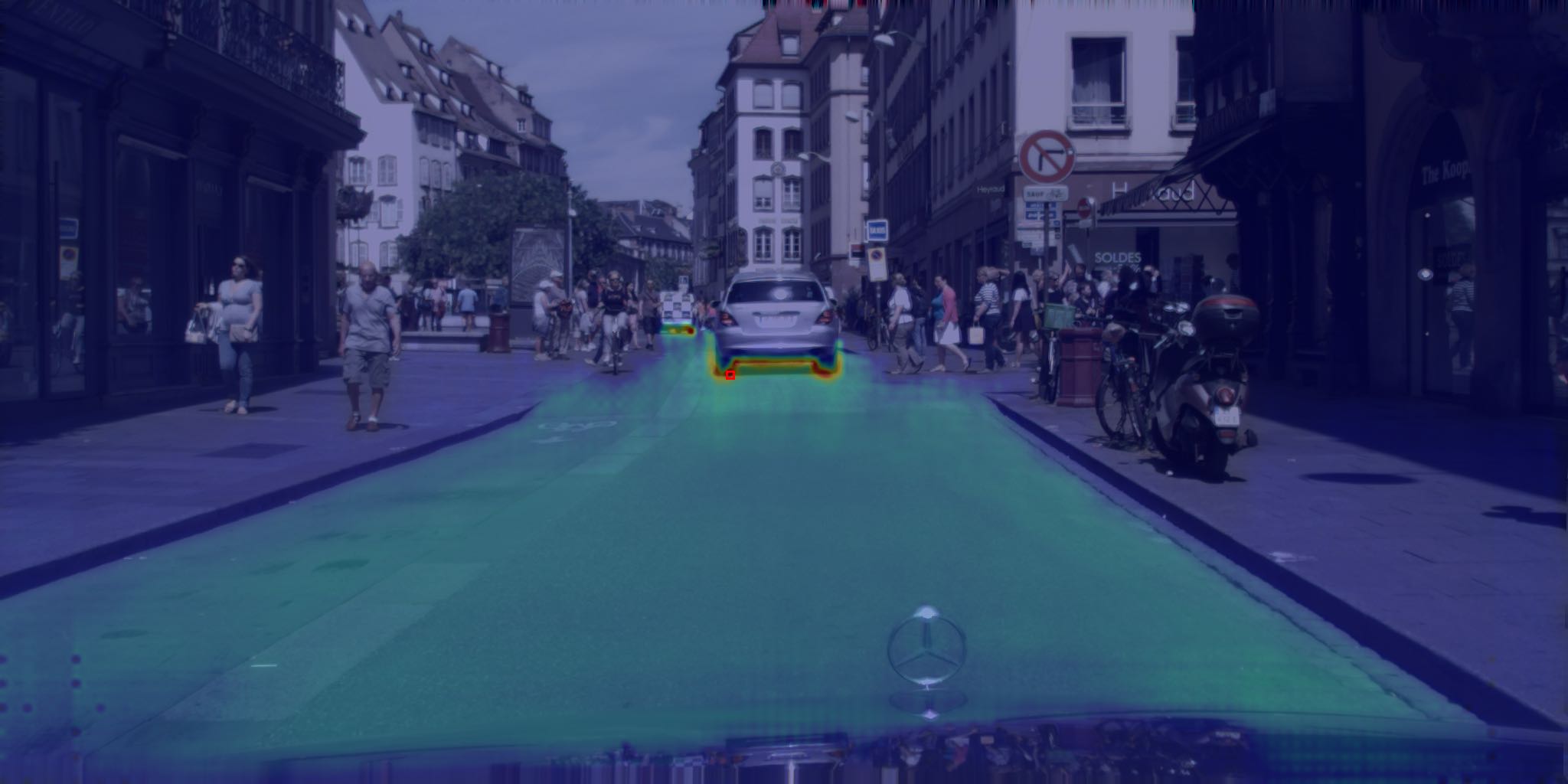}
        \end{subfigure}
    \end{tabular}
    
    \vspace{2mm}

    \rule{\linewidth}{0.5pt}
    
    \vspace{5mm}
    
    \begin{tabular}{ccc}
        \rotatebox{90}{{\parbox{4cm}{\centering Top 1 Patch}}} &
        \begin{subfigure}[b]{0.45\linewidth}
            \centering
            \includegraphics[width=\linewidth, height=4cm]{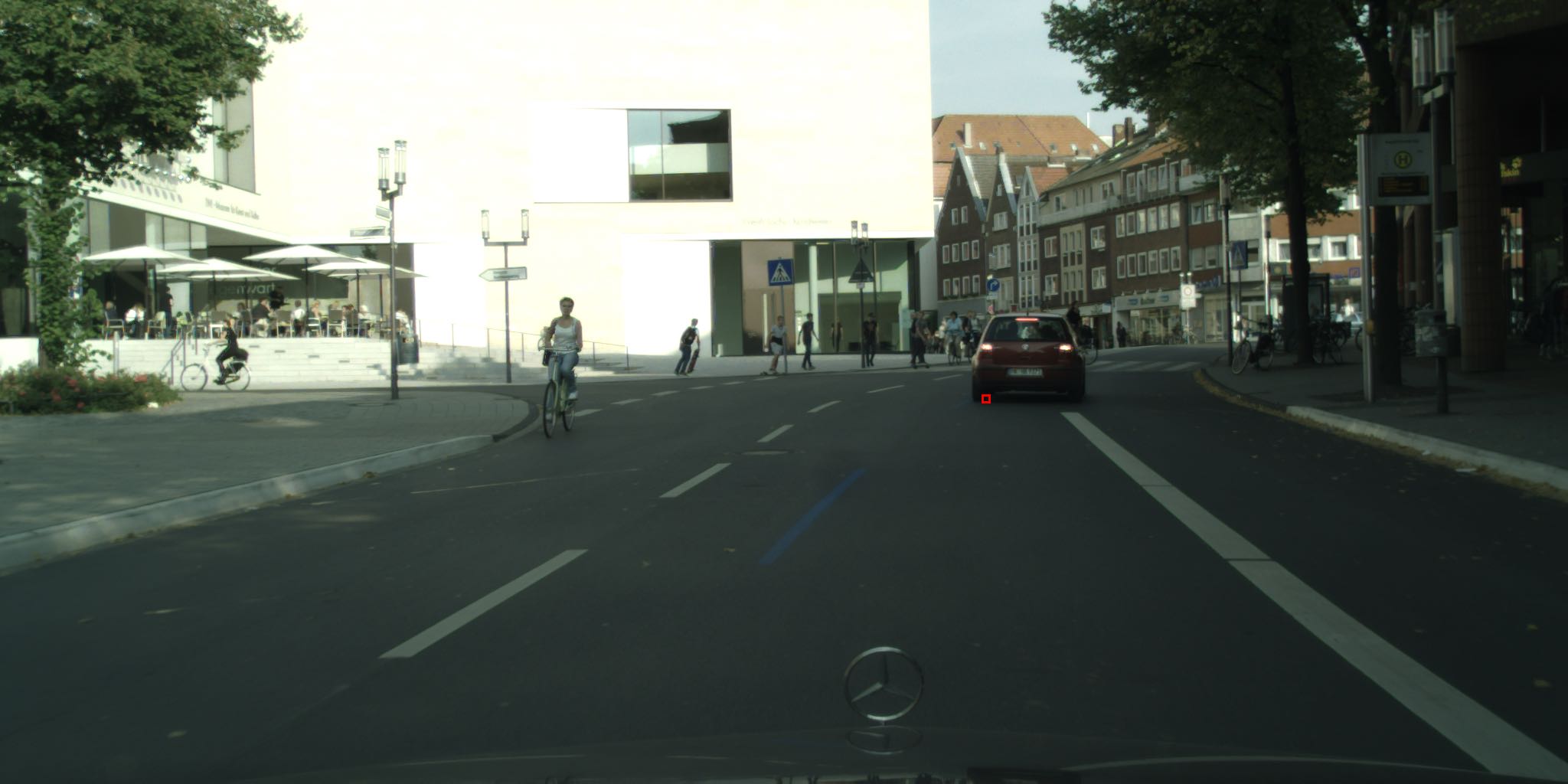}
        \end{subfigure} &
        \begin{subfigure}[b]{0.45\linewidth}
            \centering
            \includegraphics[width=\linewidth, height=4cm]{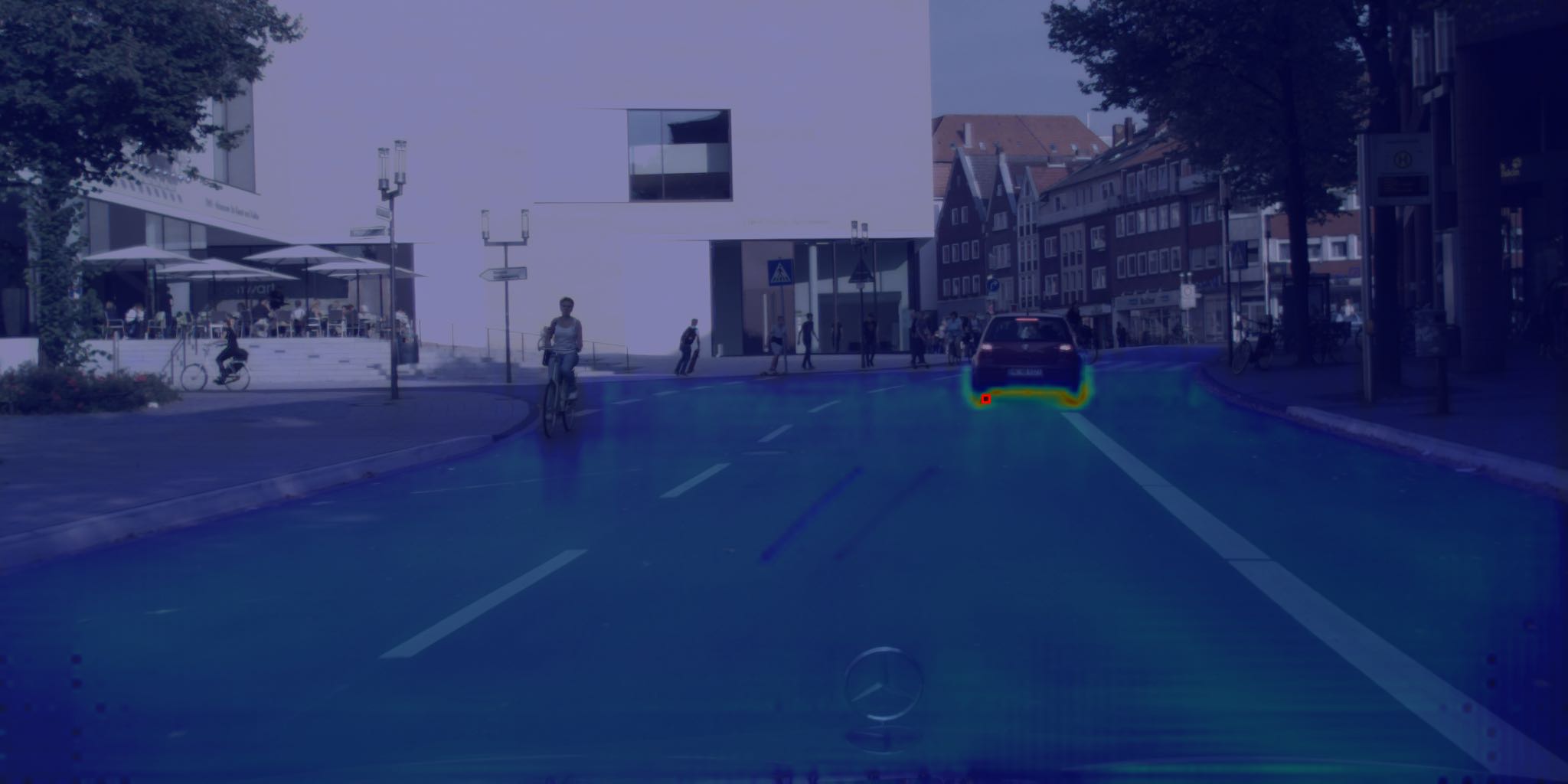}
        \end{subfigure} \\
        \rotatebox{90}{{\parbox{4cm}{\centering Top 2 Patch}}} &
        \begin{subfigure}[b]{0.45\linewidth}
            \centering
            \includegraphics[width=\linewidth, height=4cm]{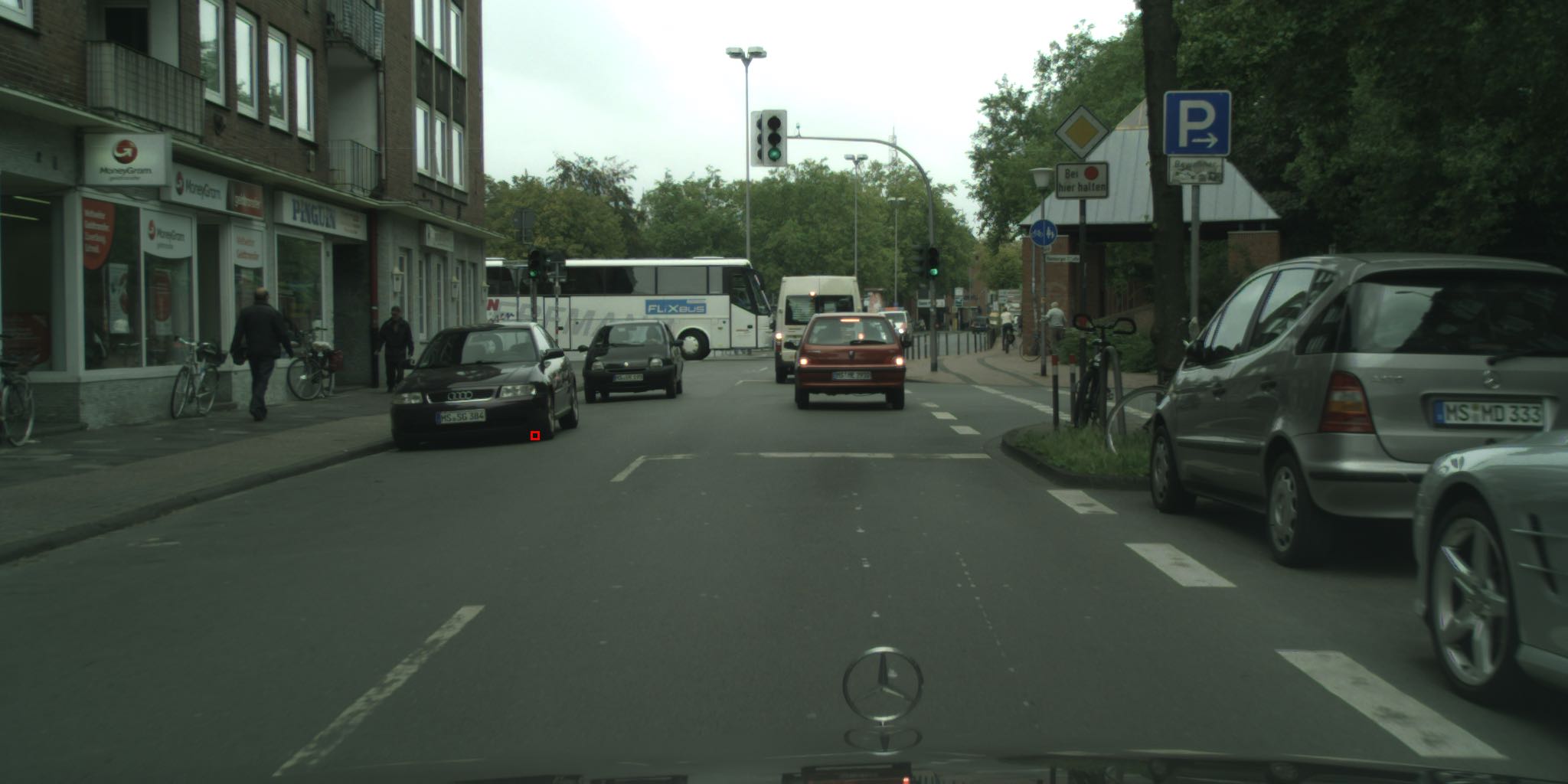}
        \end{subfigure} &
        \begin{subfigure}[b]{0.45\linewidth}
            \centering
            \includegraphics[width=\linewidth, height=4cm]{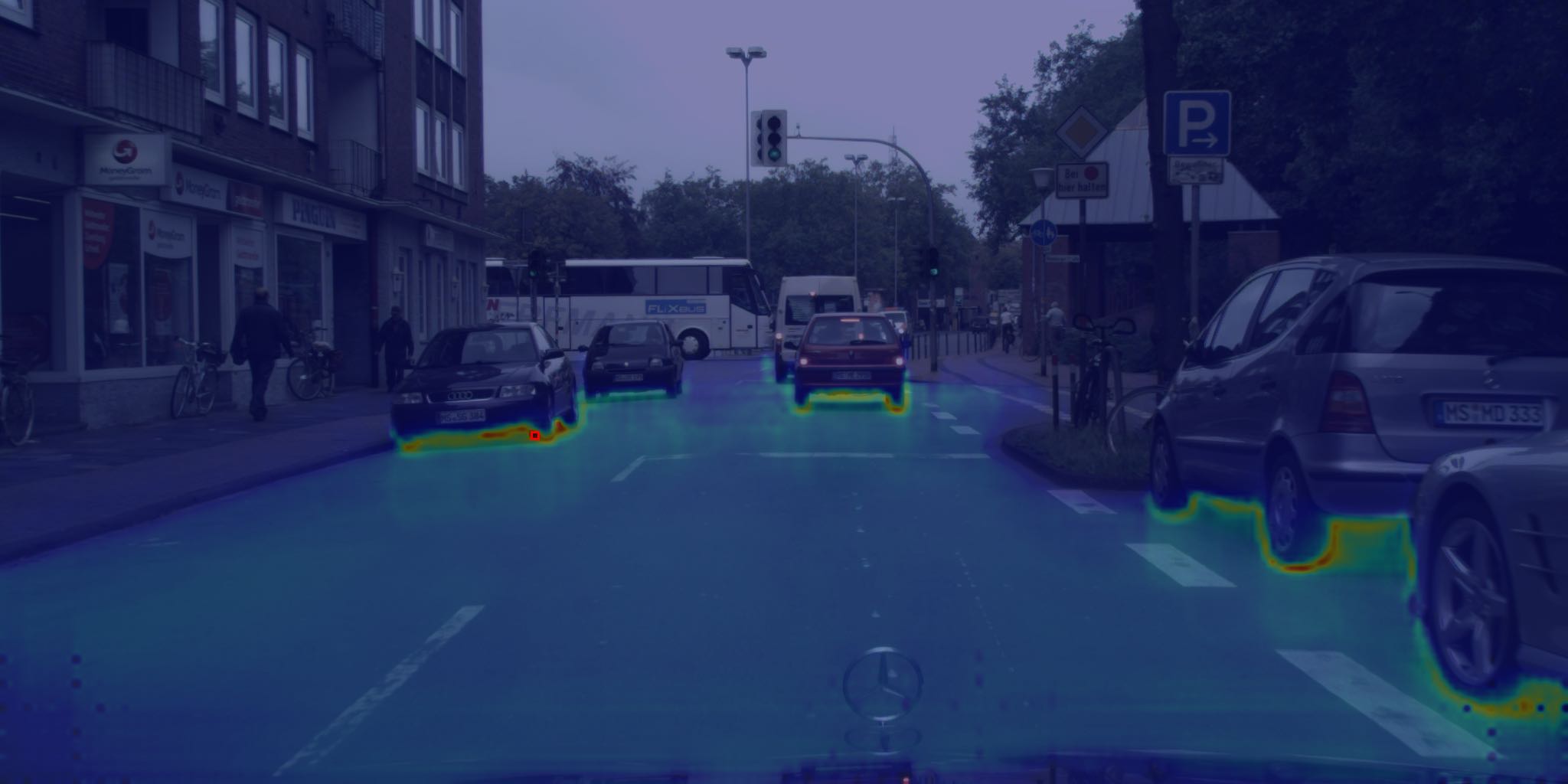}
        \end{subfigure} \\
        \rotatebox{90}{{\parbox{4cm}{\centering Top 3 Patch}}} &
        \begin{subfigure}[b]{0.45\linewidth}
            \centering
            \includegraphics[width=\linewidth, height=4cm]{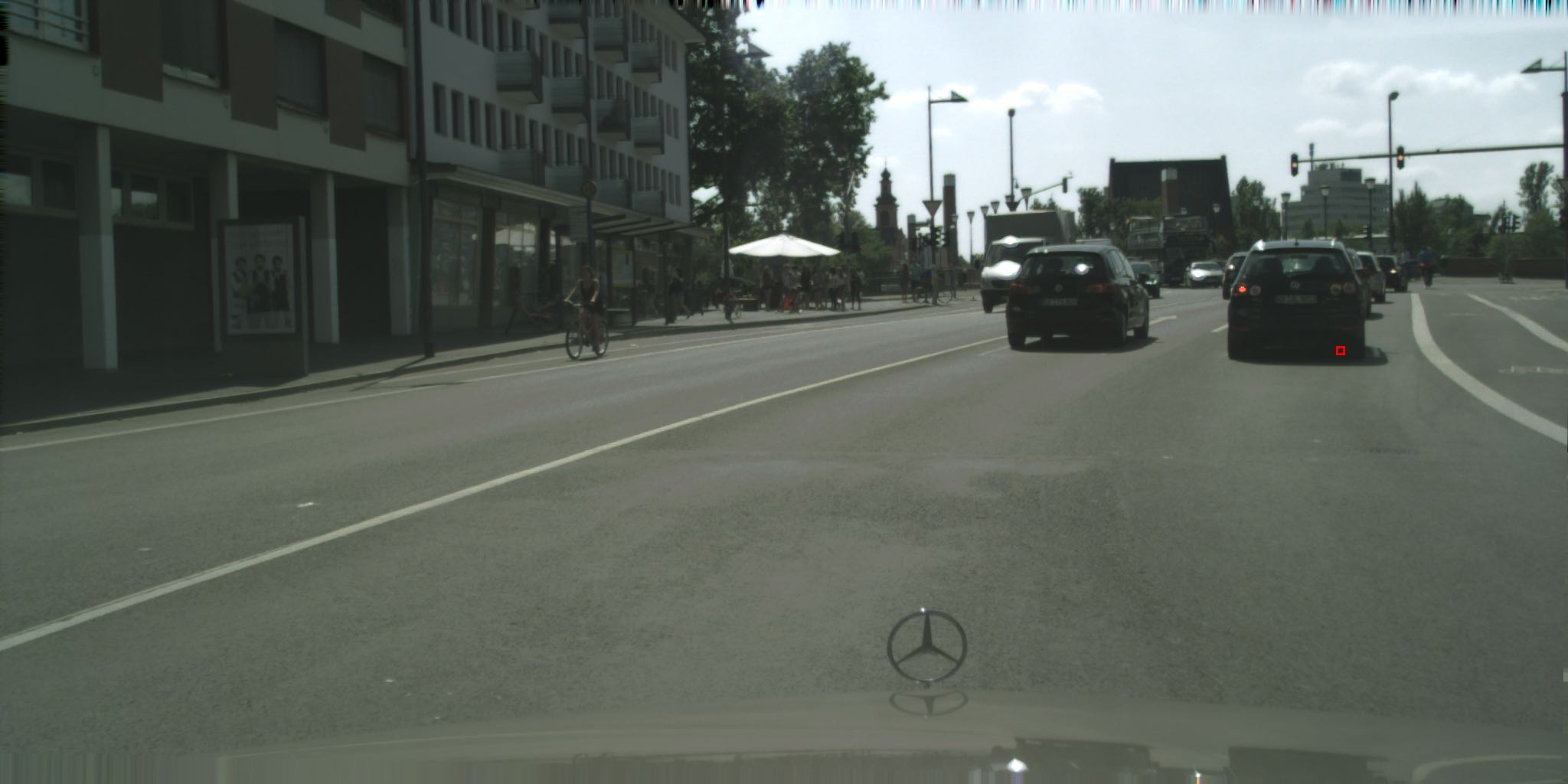}
        \end{subfigure} &
        \begin{subfigure}[b]{0.45\linewidth}
            \centering
            \includegraphics[width=\linewidth, height=4cm]
            {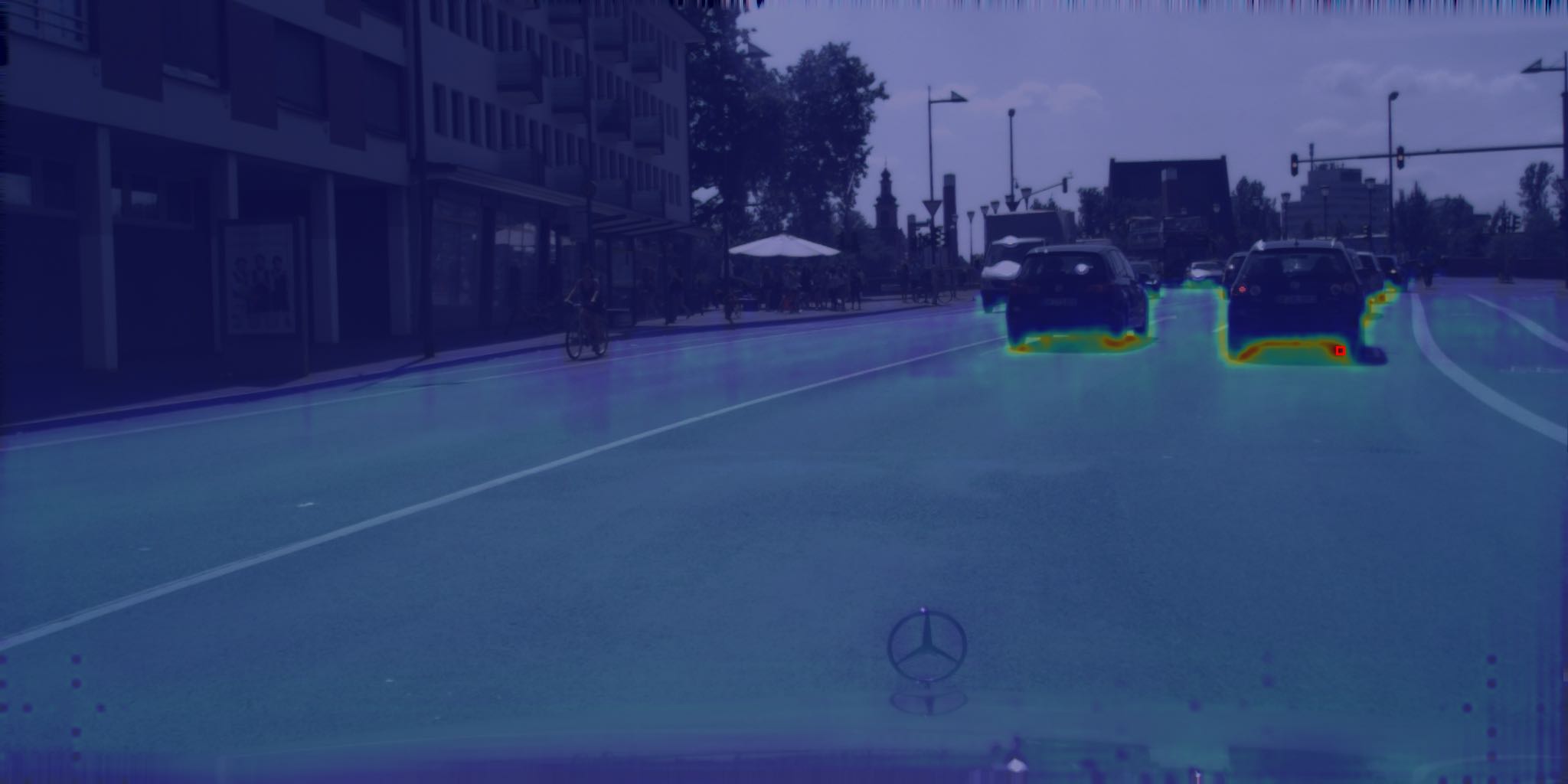}
        \end{subfigure}
    \end{tabular}
    
    \caption{The first row represents a prototype for the class \textit{car} in Cityscapes marked by a red box and the activation of the prototype on its image. The second to fourth rows show the closest patches to the prototype and its activation on the images containing the patches.}
    \label{fig:car-proto}
\end{figure*}

\begin{figure*}[htbp]
    \centering
    \begin{tabular}{ccc}
        \rotatebox{90}{{\parbox{4cm}{\centering Prototypes}}} &
        \begin{subfigure}[b]{0.45\linewidth}
            \centering
            \includegraphics[width=\linewidth, height=4cm]{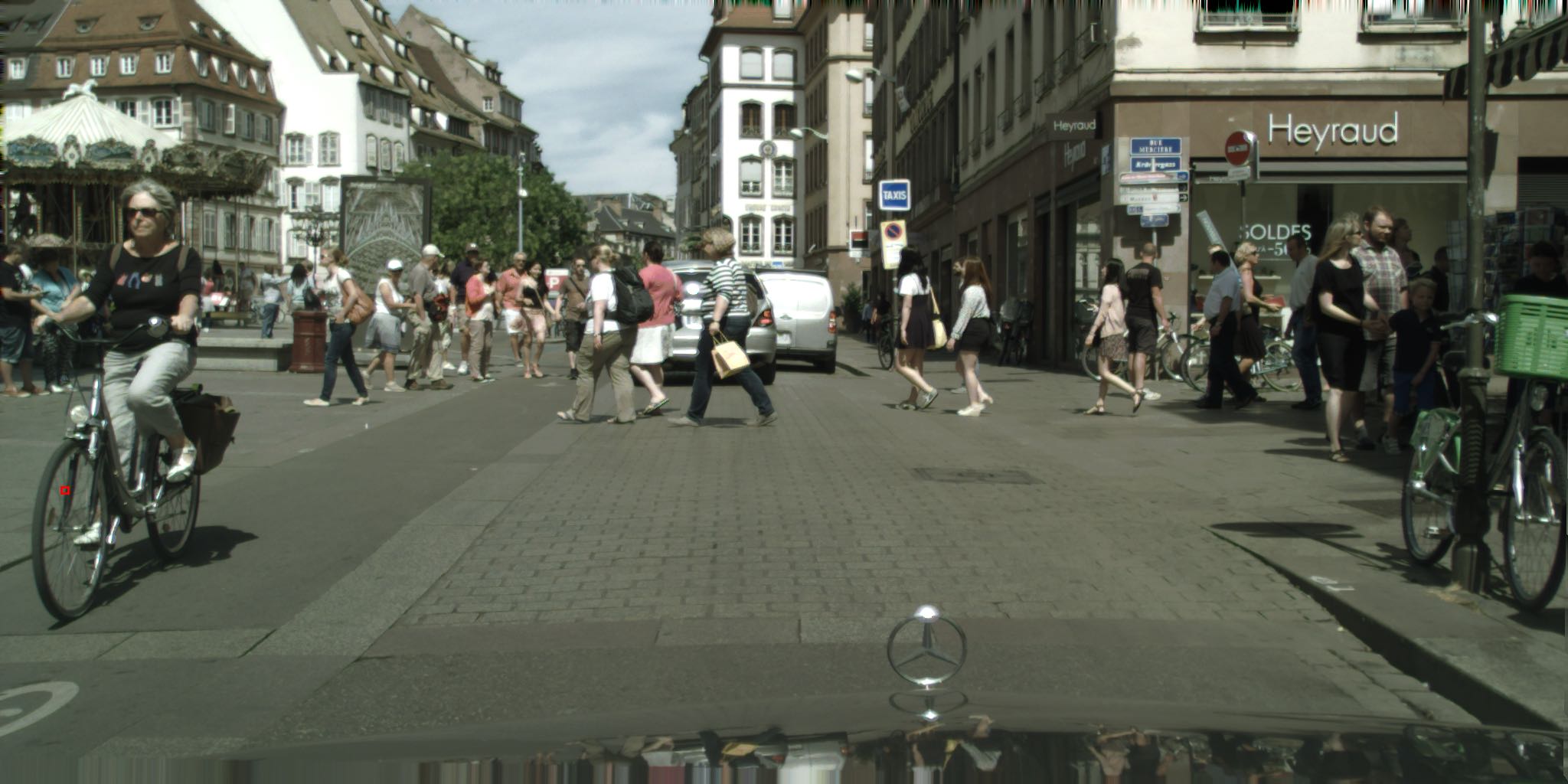}
        \end{subfigure} &
        \begin{subfigure}[b]{0.45\linewidth}
            \centering
            \includegraphics[width=\linewidth, height=4cm]{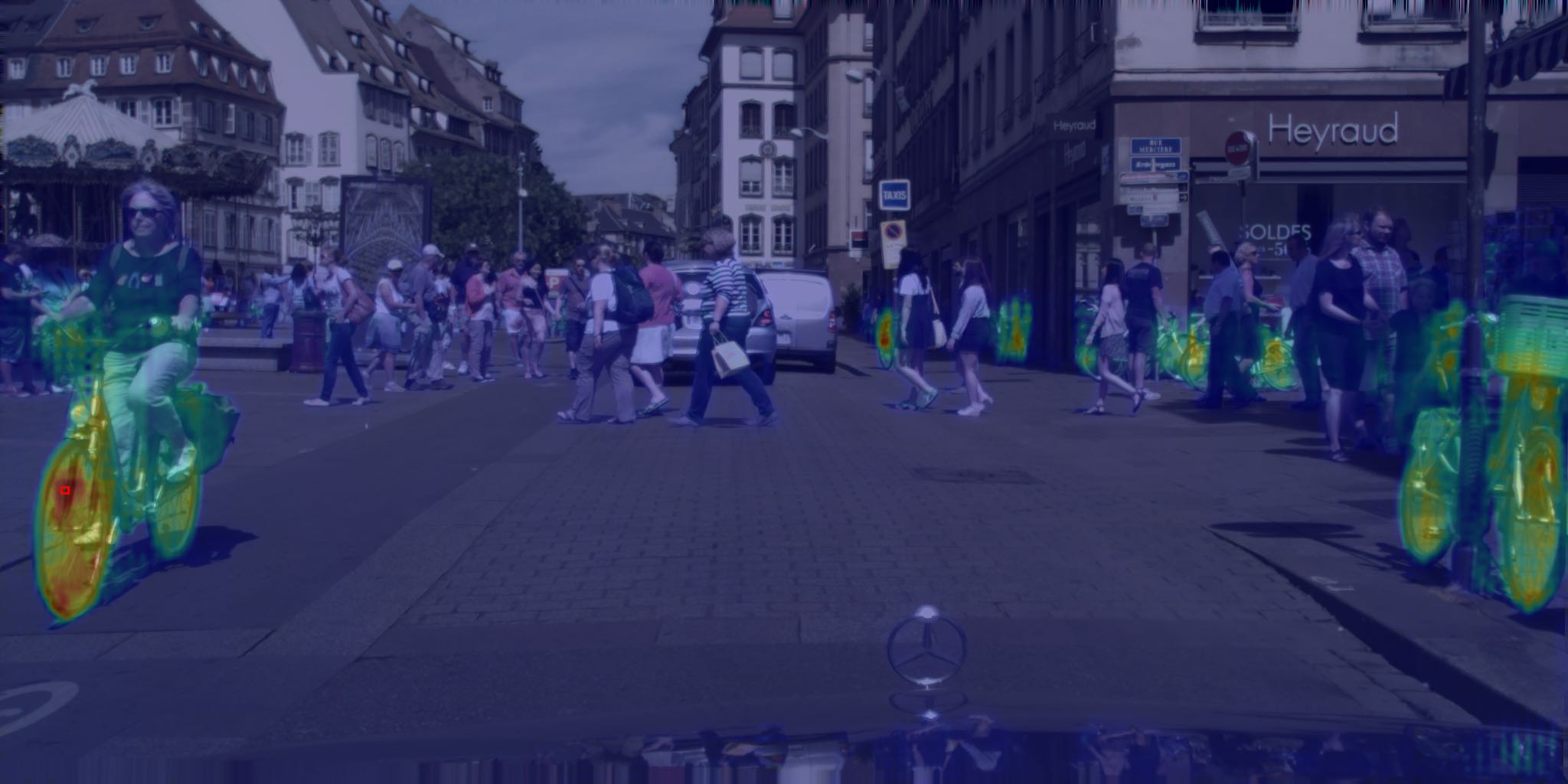}
        \end{subfigure}
    \end{tabular}
    
    \vspace{2mm}
    
    \rule{\linewidth}{0.5pt} 
    
    \vspace{5mm}
    
    \begin{tabular}{ccc}
        \rotatebox{90}{{\parbox{4cm}{\centering Top 1 Patch}}} &
        \begin{subfigure}[b]{0.45\linewidth}
            \centering
            \includegraphics[width=\linewidth, height=4cm]{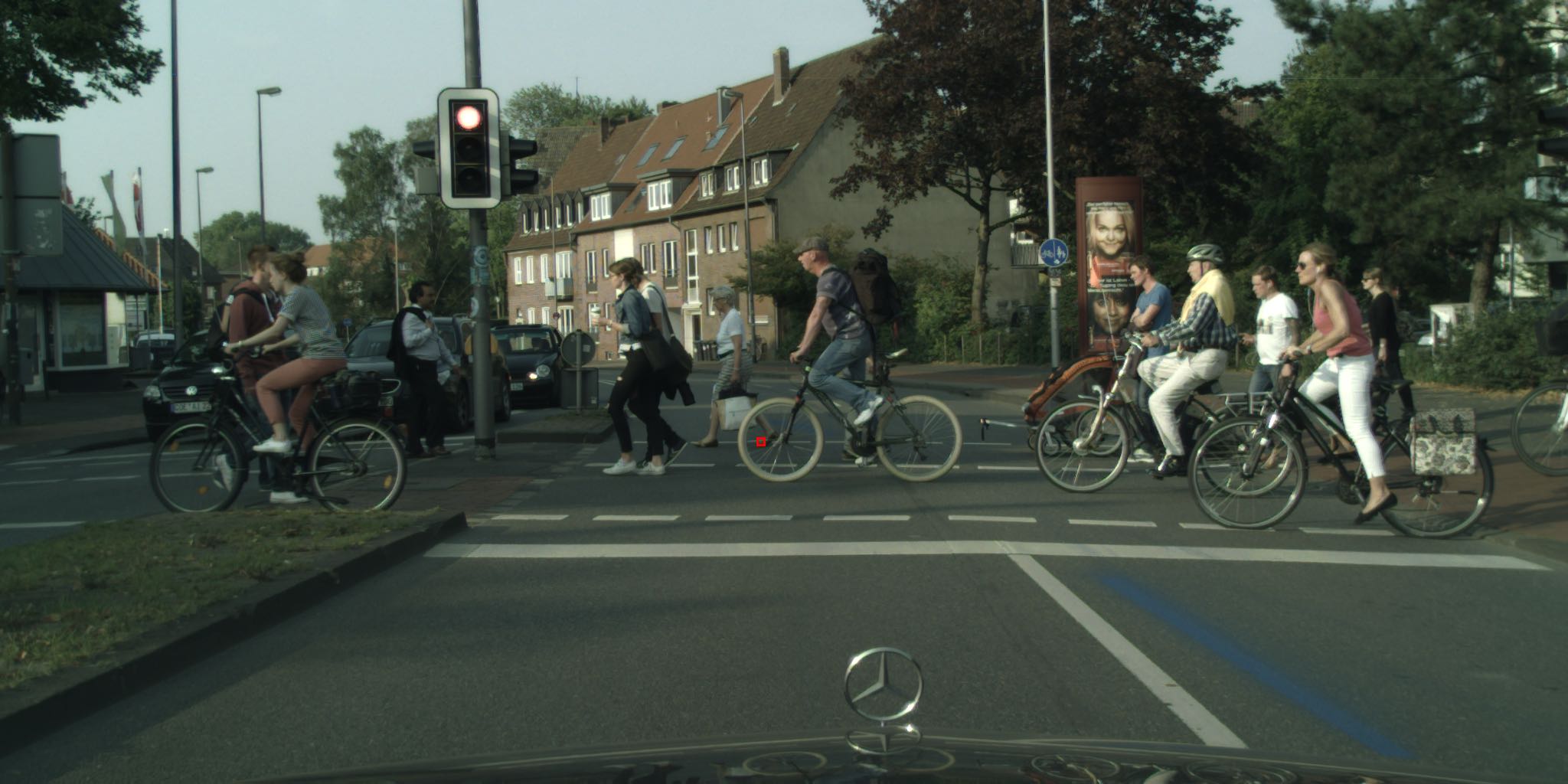}
        \end{subfigure} &
        \begin{subfigure}[b]{0.45\linewidth}
            \centering
            \includegraphics[width=\linewidth, height=4cm]{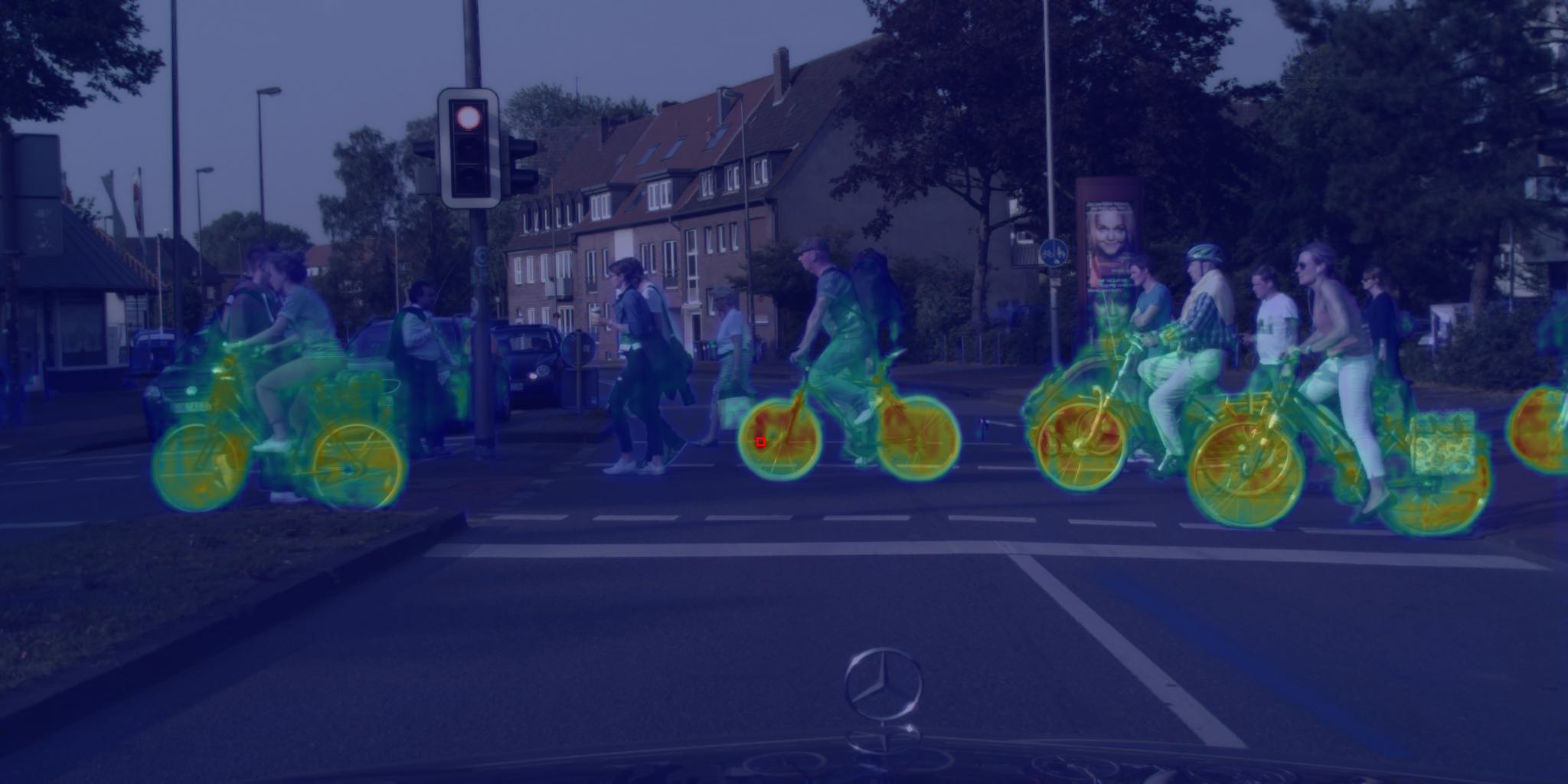}
        \end{subfigure} \\
        \rotatebox{90}{{\parbox{4cm}{\centering Top 2 Patch}}} &
        \begin{subfigure}[b]{0.45\linewidth}
            \centering
            \includegraphics[width=\linewidth, height=4cm]{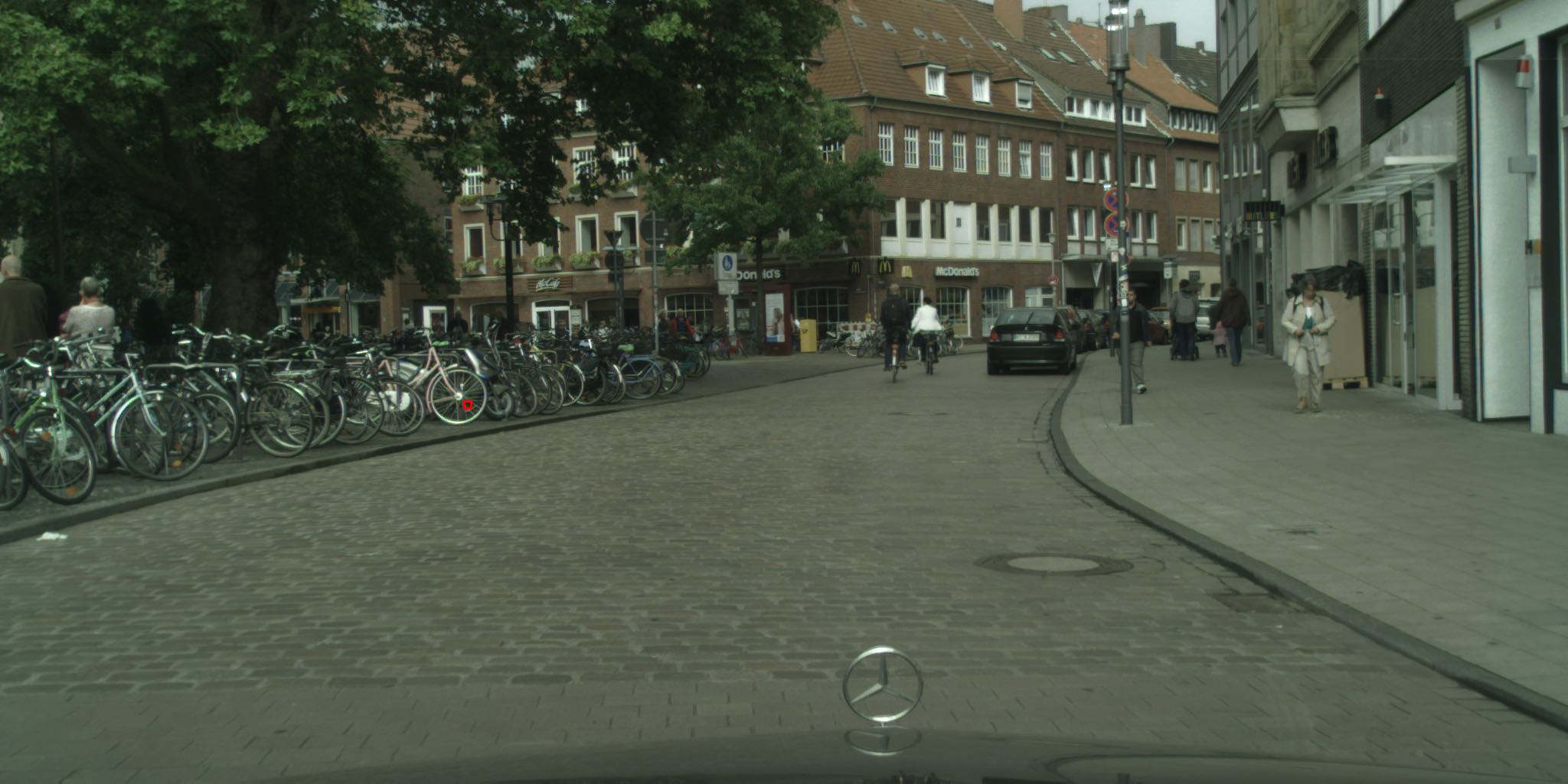}
        \end{subfigure} &
        \begin{subfigure}[b]{0.45\linewidth}
            \centering
            \includegraphics[width=\linewidth, height=4cm]{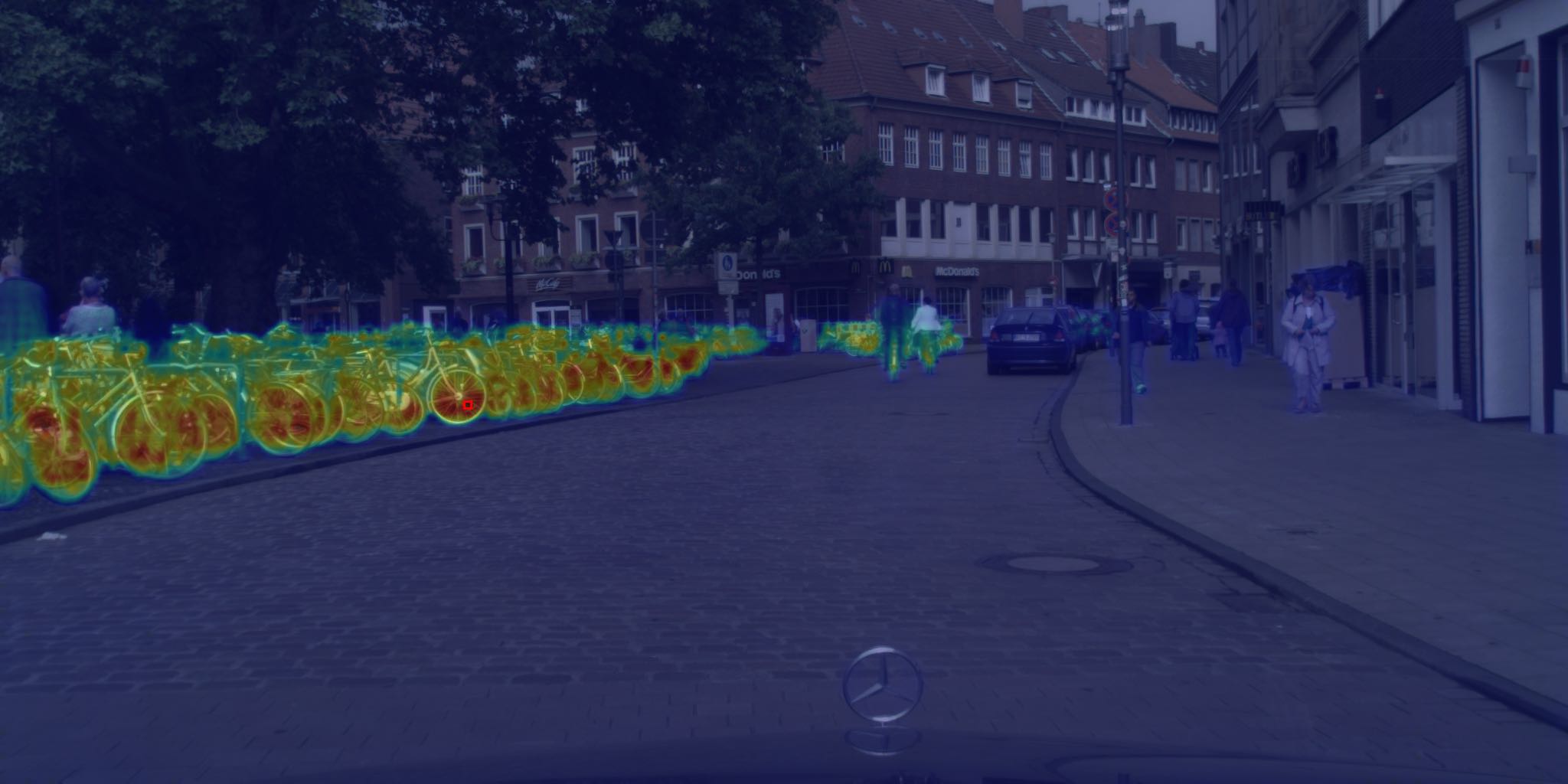}
        \end{subfigure} \\
        \rotatebox{90}{{\parbox{4cm}{\centering Top 3 Patch}}} &
        \begin{subfigure}[b]{0.45\linewidth}
            \centering
            \includegraphics[width=\linewidth, height=4cm]{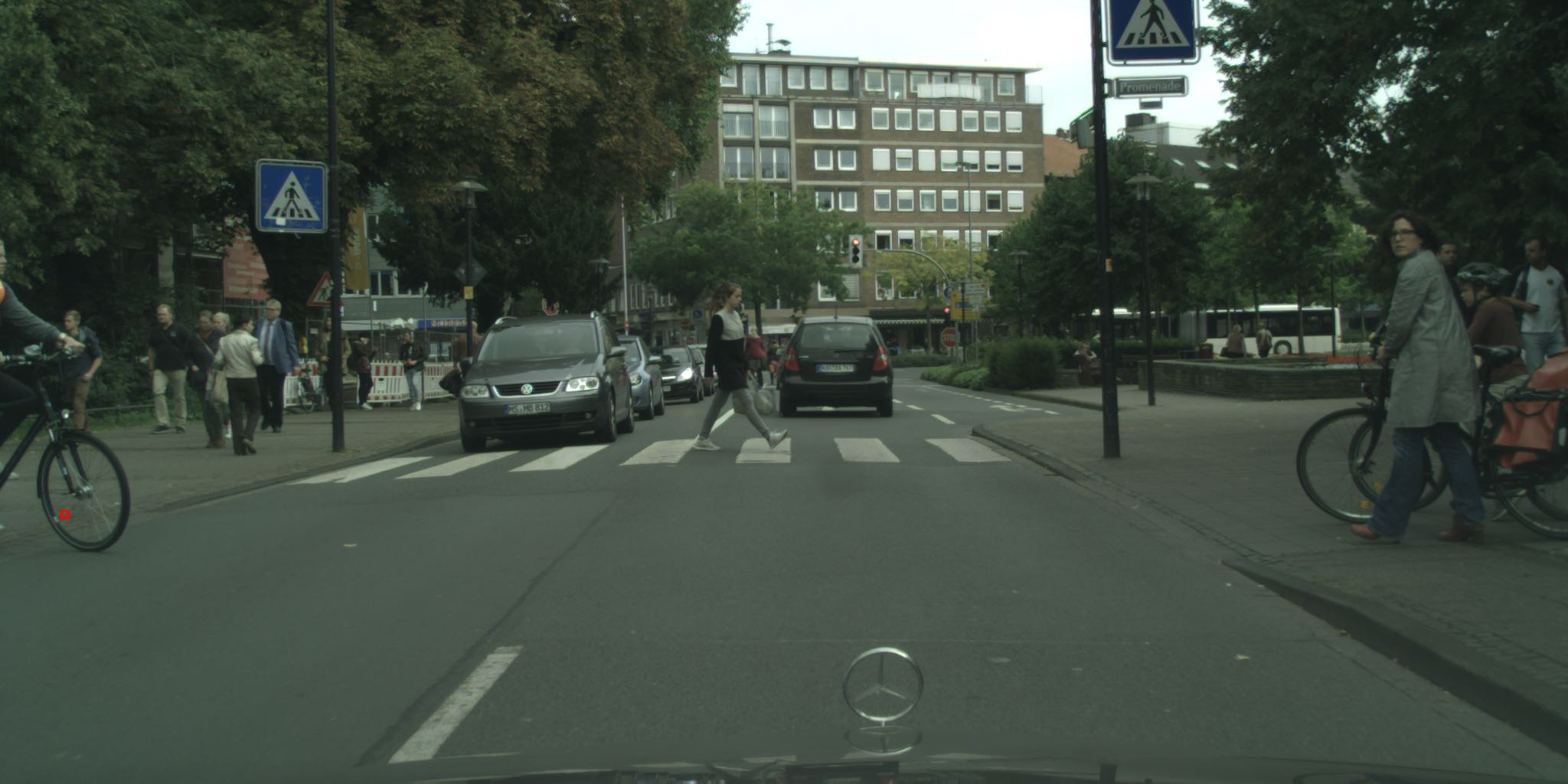}
        \end{subfigure} &
        \begin{subfigure}[b]{0.45\linewidth}
            \centering
            \includegraphics[width=\linewidth, height=4cm]
            {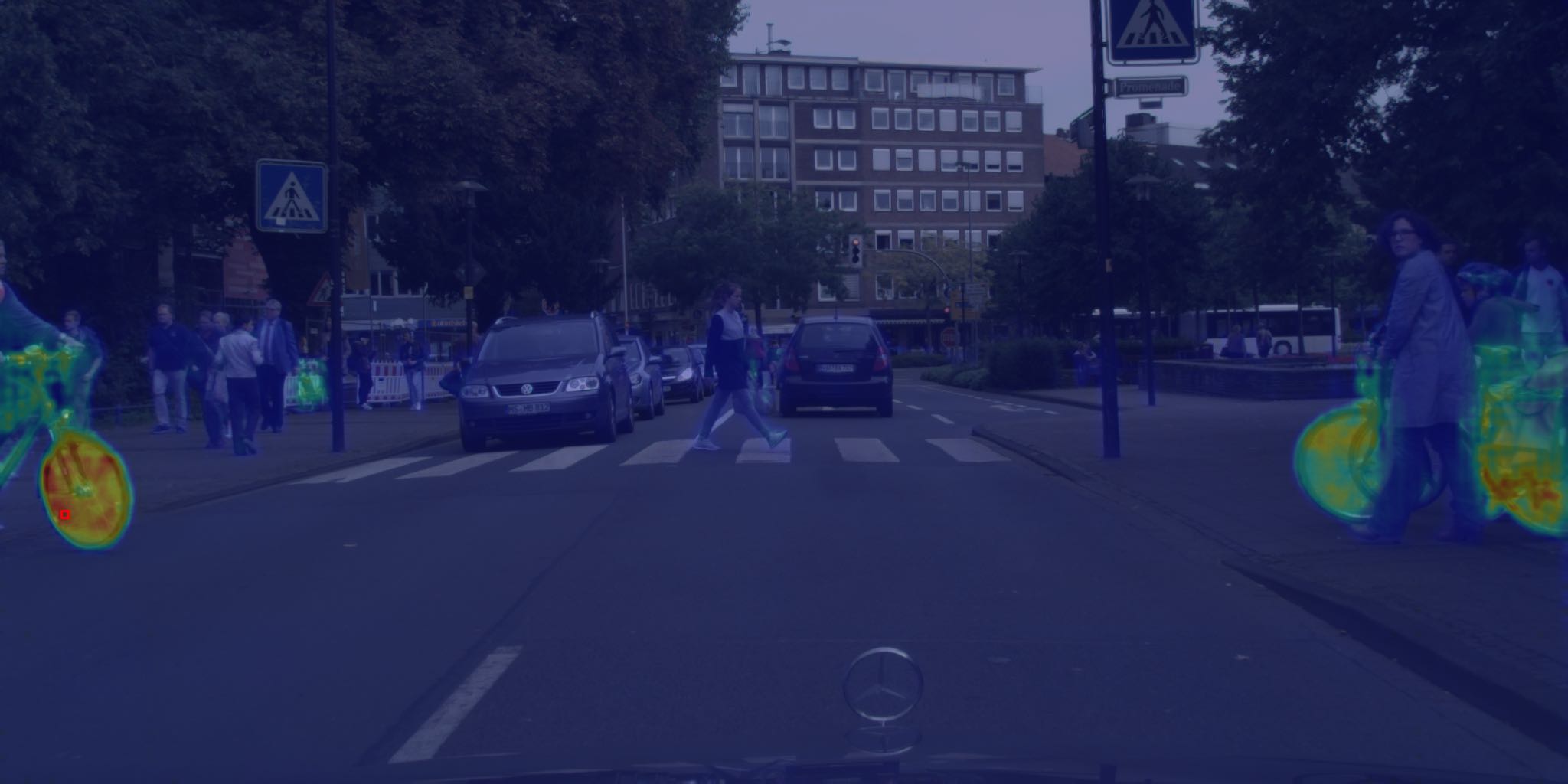}
        \end{subfigure}
    \end{tabular}
    
    \caption{The first row represents a prototype for the class \textit{bicycle} in Cityscapes marked by a red box and the activation of the prototype on its image. The second to fourth rows show the closest patches to the prototype and its activation on the images containing the patches.}
    \label{fig:bike-proto}
\end{figure*}

\begin{figure*}[htbp]
    \centering
    
    \begin{tabular}{ccc}
        \rotatebox{90}{{\parbox{4cm}{\centering Prototypes}}} &
        \begin{subfigure}[b]{0.45\linewidth}
            \centering
            \includegraphics[width=\linewidth, height=4cm]{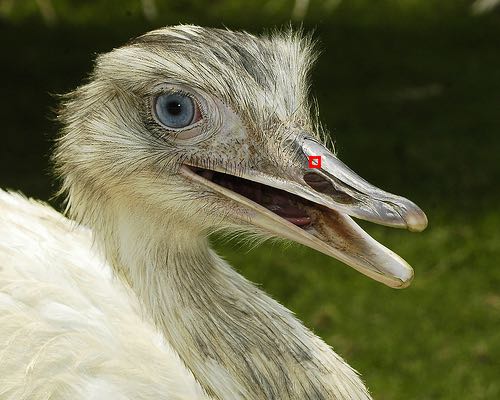}
        \end{subfigure} &
        \begin{subfigure}[b]{0.45\linewidth}
            \centering
            \includegraphics[width=\linewidth, height=4cm]{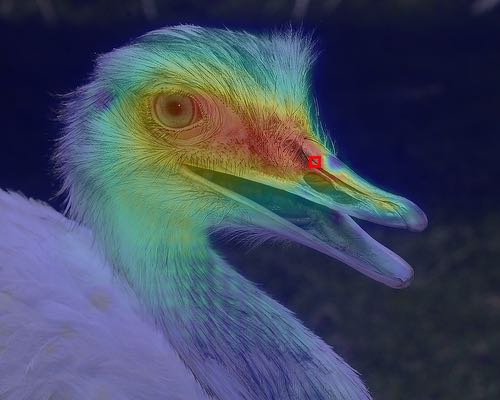}
        \end{subfigure}
    \end{tabular}
    
    \vspace{2mm} 
    
    \rule{\linewidth}{0.5pt} 
    
    \vspace{5mm}

    \begin{tabular}{ccc}
        \rotatebox{90}{{\parbox{4cm}{\centering Top 1 Patch}}} &
        \begin{subfigure}[b]{0.45\linewidth}
            \centering
            \includegraphics[width=\linewidth, height=4cm]{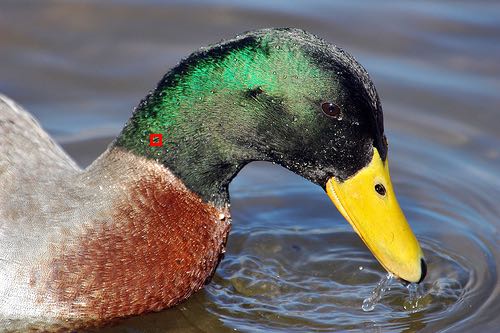}
        \end{subfigure} &
        \begin{subfigure}[b]{0.45\linewidth}
            \centering
            \includegraphics[width=\linewidth, height=4cm]{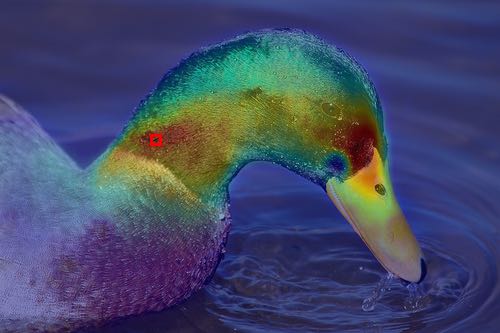}
        \end{subfigure} \\
        \rotatebox{90}{{\parbox{4cm}{\centering Top 2 Patch}}} &
        \begin{subfigure}[b]{0.45\linewidth}
            \centering
            \includegraphics[width=\linewidth, height=4cm]{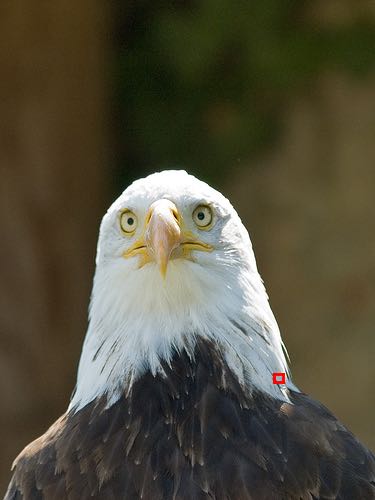}
        \end{subfigure} &
        \begin{subfigure}[b]{0.45\linewidth}
            \centering
            \includegraphics[width=\linewidth, height=4cm]{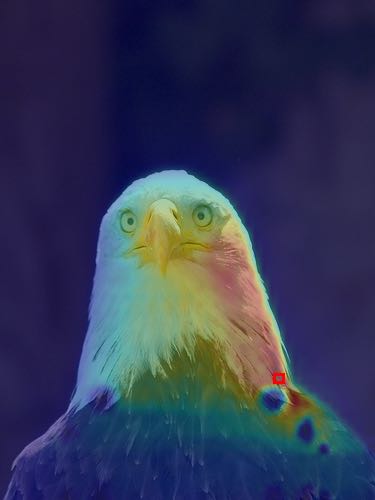}
        \end{subfigure} \\
        \rotatebox{90}{{\parbox{4cm}{\centering Top 3 Patch}}} &
        \begin{subfigure}[b]{0.45\linewidth}
            \centering
            \includegraphics[width=\linewidth, height=4cm]{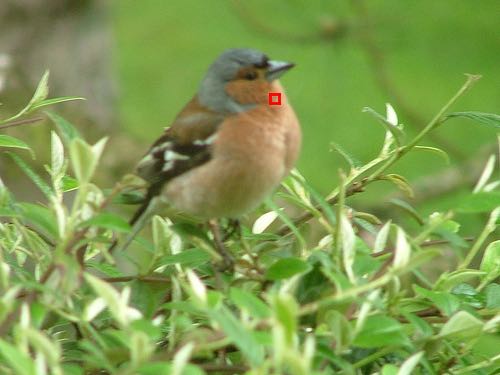}
        \end{subfigure} &
        \begin{subfigure}[b]{0.45\linewidth}
            \centering
            \includegraphics[width=\linewidth, height=4cm]
            {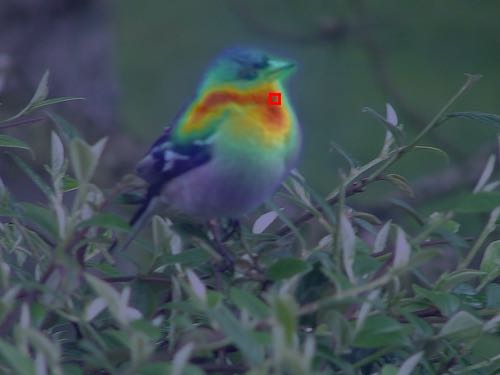}
        \end{subfigure}
    \end{tabular}
    
    \caption{The first row represents a prototype for the class \textit{bird} in PASCAL VOC marked by a red box and the activation of the prototype on its image. The second to fourth rows show the closest patches to the prototype and its activation on the images containing the patches.}
    \label{fig:bird-proto}
\end{figure*}

\begin{figure*}[htbp]
    \centering
    
    \begin{tabular}{ccc}
        \rotatebox{90}{{\parbox{4cm}{\centering Prototypes}}} &
        \begin{subfigure}[b]{0.45\linewidth}
            \centering
            \includegraphics[width=\linewidth, height=4cm]{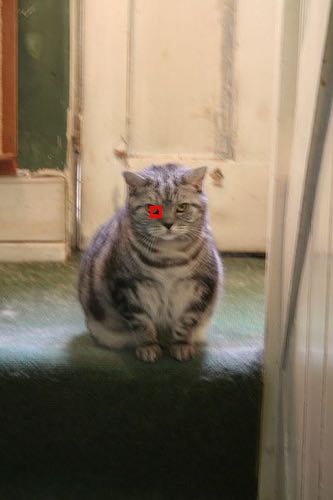}
        \end{subfigure} &
        \begin{subfigure}[b]{0.45\linewidth}
            \centering
            \includegraphics[width=\linewidth, height=4cm]{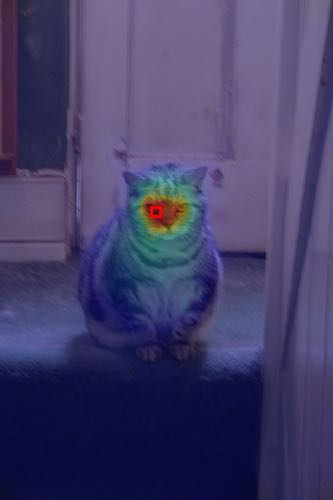}
        \end{subfigure}
    \end{tabular}
    
    \vspace{2mm} 
    
    \rule{\linewidth}{0.5pt} 
    
    \vspace{5mm}

    \begin{tabular}{ccc}
        \rotatebox{90}{{\parbox{4cm}{\centering Top 1 Patch}}} &
        \begin{subfigure}[b]{0.45\linewidth}
            \centering
            \includegraphics[width=\linewidth, height=4cm]{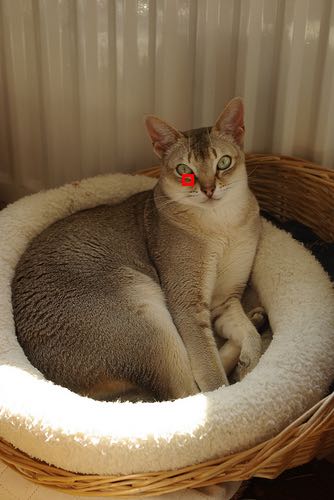}
        \end{subfigure} &
        \begin{subfigure}[b]{0.45\linewidth}
            \centering
            \includegraphics[width=\linewidth, height=4cm]{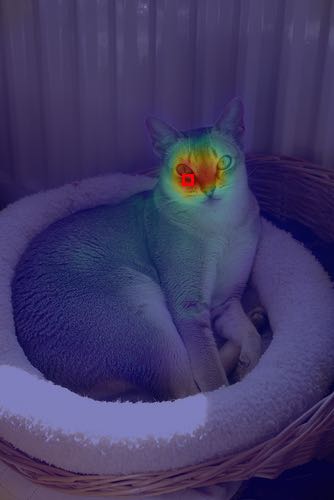}
        \end{subfigure} \\
        \rotatebox{90}{{\parbox{4cm}{\centering Top 2 Patch}}} &
        \begin{subfigure}[b]{0.45\linewidth}
            \centering
            \includegraphics[width=\linewidth, height=4cm]{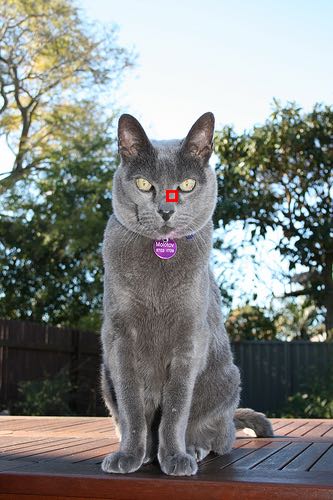}
        \end{subfigure} &
        \begin{subfigure}[b]{0.45\linewidth}
            \centering
            \includegraphics[width=\linewidth, height=4cm]{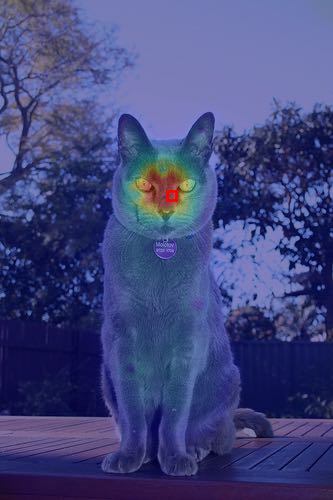}
        \end{subfigure} \\
        \rotatebox{90}{{\parbox{4cm}{\centering Top 3 Patch}}} &
        \begin{subfigure}[b]{0.45\linewidth}
            \centering
            \includegraphics[width=\linewidth, height=4cm]{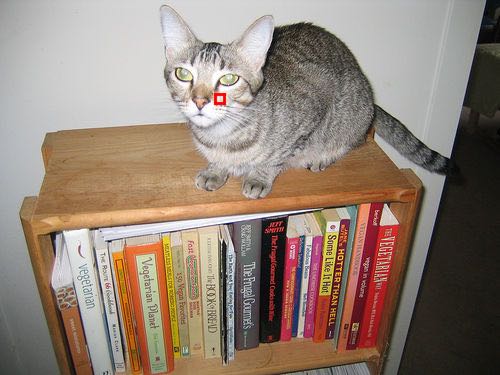}
        \end{subfigure} &
        \begin{subfigure}[b]{0.45\linewidth}
            \centering
            \includegraphics[width=\linewidth, height=4cm]
            {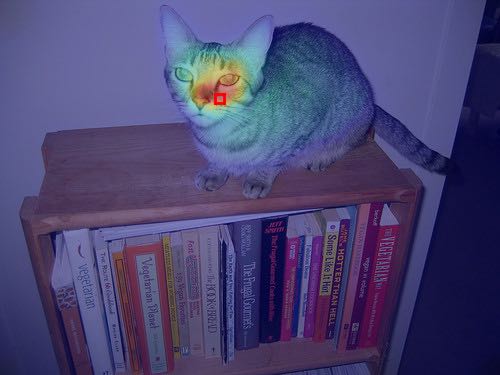}
        \end{subfigure}
    \end{tabular}
    
    \caption{The first row represents a prototype for the class \textit{cat} in PASCAL VOC marked by a red box and the activation of the prototype on its image. The second to fourth rows show the closest patches to the prototype and its activation on the images containing the patches.}
    \label{fig:cat-proto}
\end{figure*}

\begin{figure*}[htbp]
    \centering
    
    \begin{tabular}{ccc}
        \rotatebox{90}{{\parbox{4cm}{\centering Prototypes}}} &
        \begin{subfigure}[b]{0.45\linewidth}
            \centering
            \includegraphics[width=\linewidth, height=4cm]{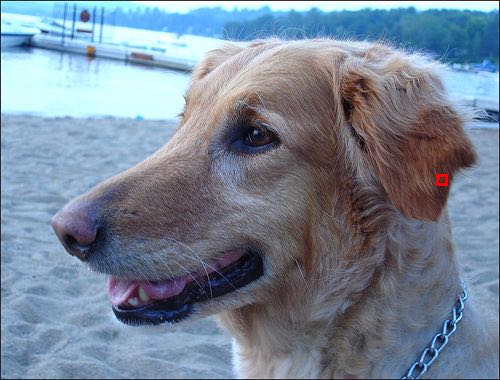}
        \end{subfigure} &
        \begin{subfigure}[b]{0.45\linewidth}
            \centering
            \includegraphics[width=\linewidth, height=4cm]{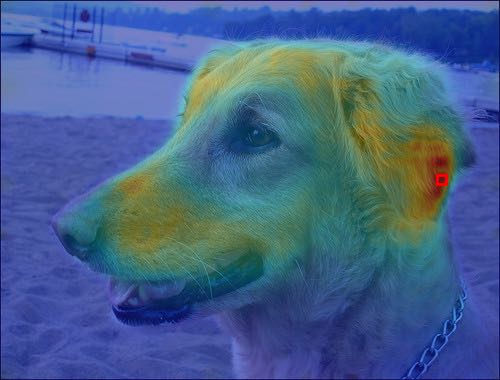}
        \end{subfigure}
    \end{tabular}
    
    \vspace{2mm} 
    
    \rule{\linewidth}{0.5pt} 
    
    \vspace{5mm}

    \begin{tabular}{ccc}
        \rotatebox{90}{{\parbox{4cm}{\centering Top 1 Patch}}} &
        \begin{subfigure}[b]{0.45\linewidth}
            \centering
            \includegraphics[width=\linewidth, height=4cm]{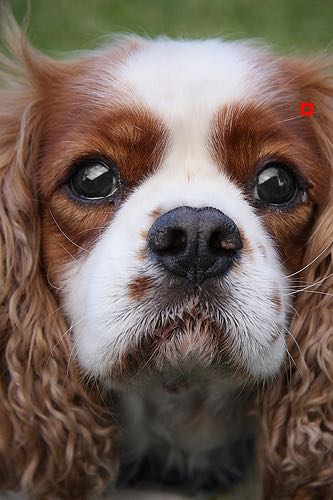}
        \end{subfigure} &
        \begin{subfigure}[b]{0.45\linewidth}
            \centering
            \includegraphics[width=\linewidth, height=4cm]{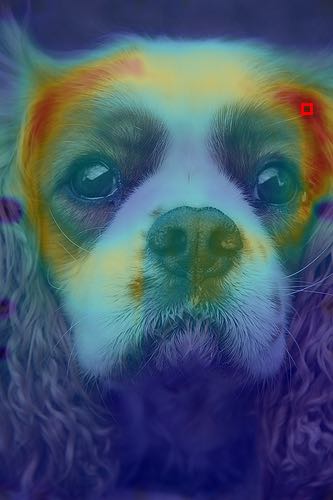}
        \end{subfigure} \\
        \rotatebox{90}{{\parbox{4cm}{\centering Top 2 Patch}}} &
        \begin{subfigure}[b]{0.45\linewidth}
            \centering
            \includegraphics[width=\linewidth, height=4cm]{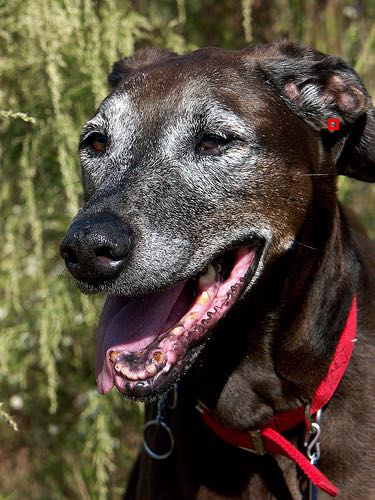}
        \end{subfigure} &
        \begin{subfigure}[b]{0.45\linewidth}
            \centering
            \includegraphics[width=\linewidth, height=4cm]{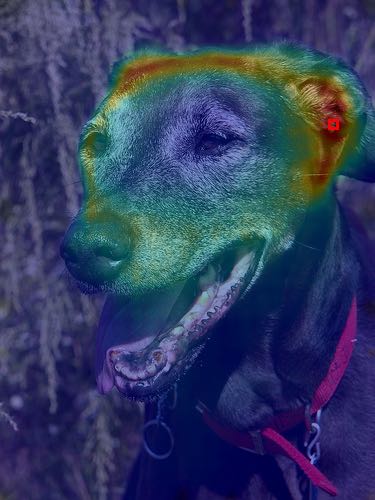}
        \end{subfigure} \\
        \rotatebox{90}{{\parbox{4cm}{\centering Top 3 Patch}}} &
        \begin{subfigure}[b]{0.45\linewidth}
            \centering
            \includegraphics[width=\linewidth, height=4cm]{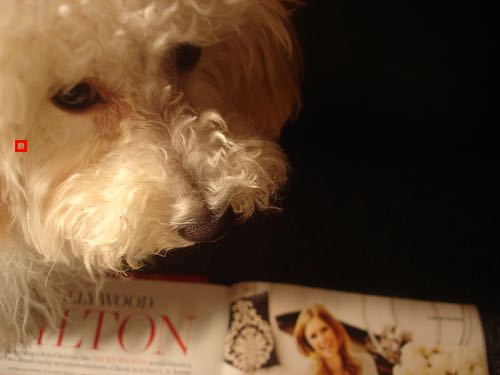}
        \end{subfigure} &
        \begin{subfigure}[b]{0.45\linewidth}
            \centering
            \includegraphics[width=\linewidth, height=4cm]
            {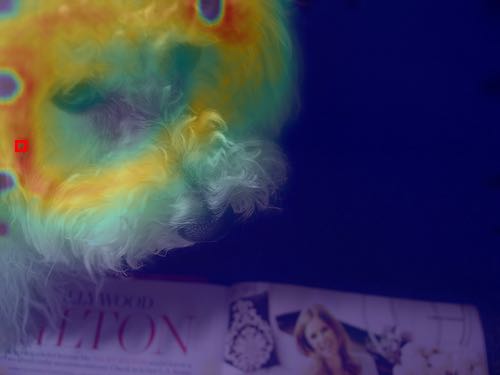}
        \end{subfigure}
    \end{tabular}
    
    \caption{The first row represents a prototype for the class \textit{dog} in PASCAL VOC marked by a red box and the activation of the prototype on its image. The second to fourth rows show the closest patches to the prototype and its activation on the images containing the patches.}
    \label{fig:dog-proto}
\end{figure*}

For the learned prototypes to be interpretable to the users, the semantic parts that they represent should be similar across images. To analyze whether our method is consistent across images, we extend the consistency metric from Section~\ref{sec:result} with visualizations of the nearest images in the validation set from a few prototypes in Cityscapes and Pascal VOC. Those visualizations are shown in Figures [\ref{fig:person-proto} - \ref{fig:dog-proto}], and showcase strong semantic correspondences between the prototypes and their closest patches.

\section{Nearest Prototypes to Image Patches}

\begin{figure*}[htbp]
    \centering
    
    \begin{tabular}{cc}
        \rotatebox{90}{{\parbox{4cm}{\centering Image Patches}}} &
        \begin{subfigure}[b]{0.45\linewidth}
            \centering
            \includegraphics[width=\linewidth, height=4cm]{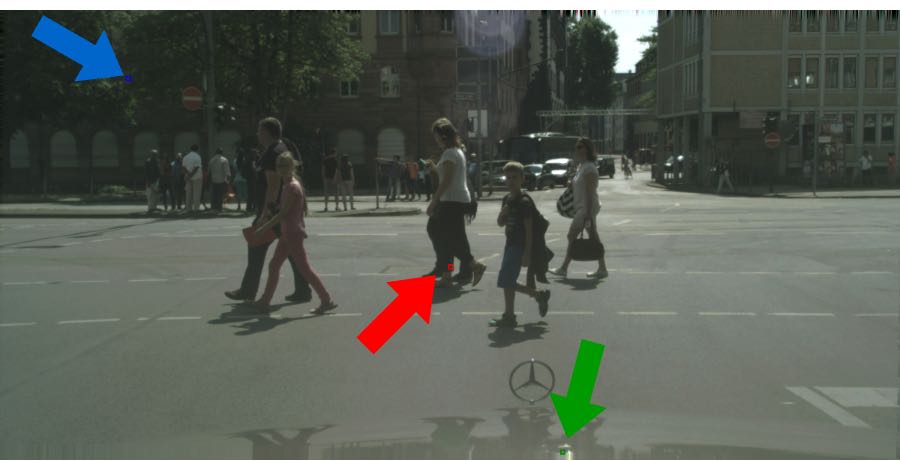}
        \end{subfigure}
    \end{tabular}
    
    \vspace{2mm} 
    
    \textcolor{red}{\rule{\linewidth}{0.5pt}}
    
    \vspace{5mm}

    \begin{tabular}{ccc}
        \rotatebox{90}{{\parbox{4cm}{\centering Top 1 Prototype}}} &
        \begin{subfigure}[b]{0.45\linewidth}
            \centering
            \includegraphics[width=\linewidth, height=4cm]{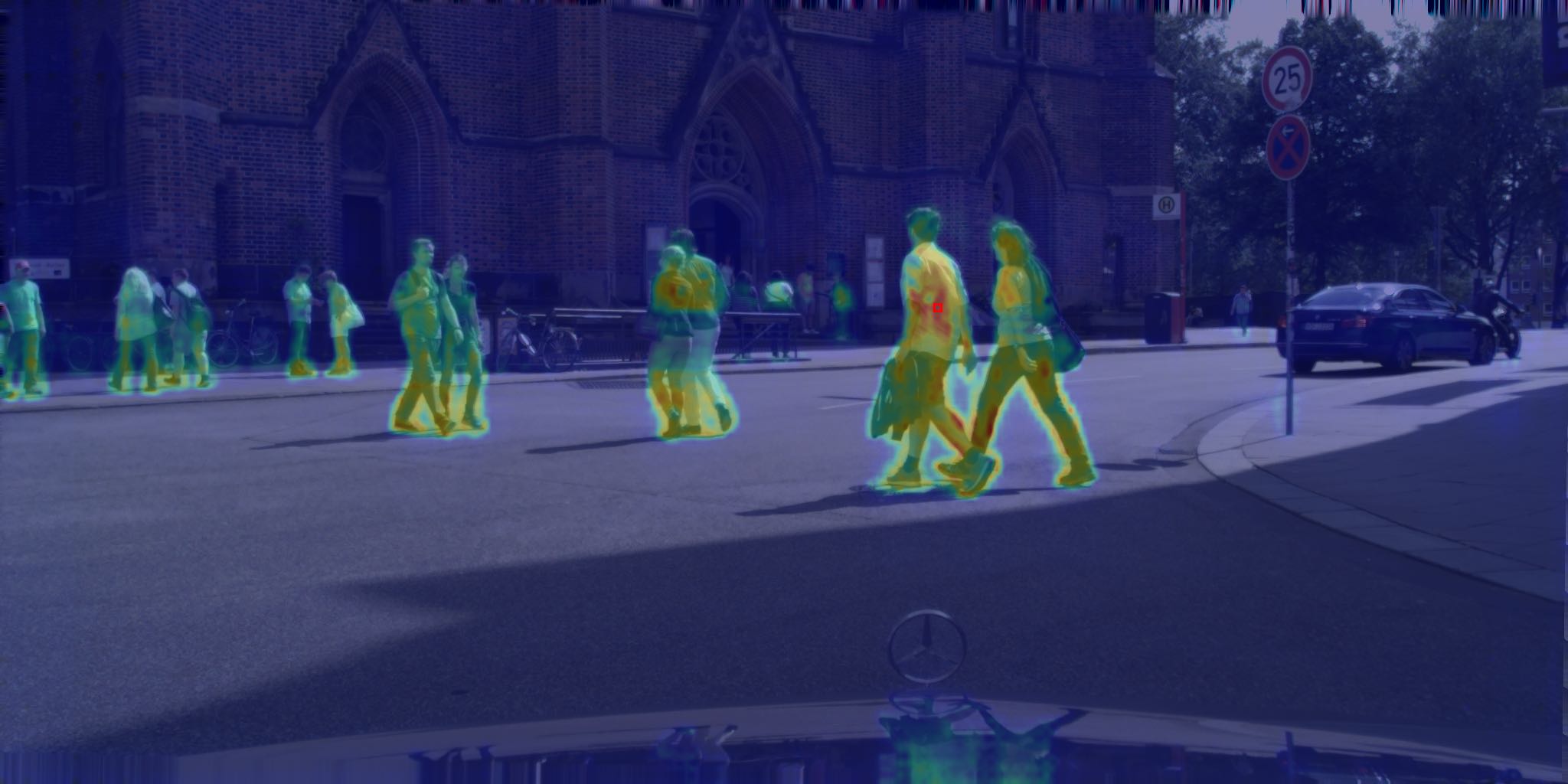}
        \end{subfigure} &
        \begin{subfigure}[b]{0.45\linewidth}
            \centering
            \includegraphics[width=\linewidth, height=4cm]{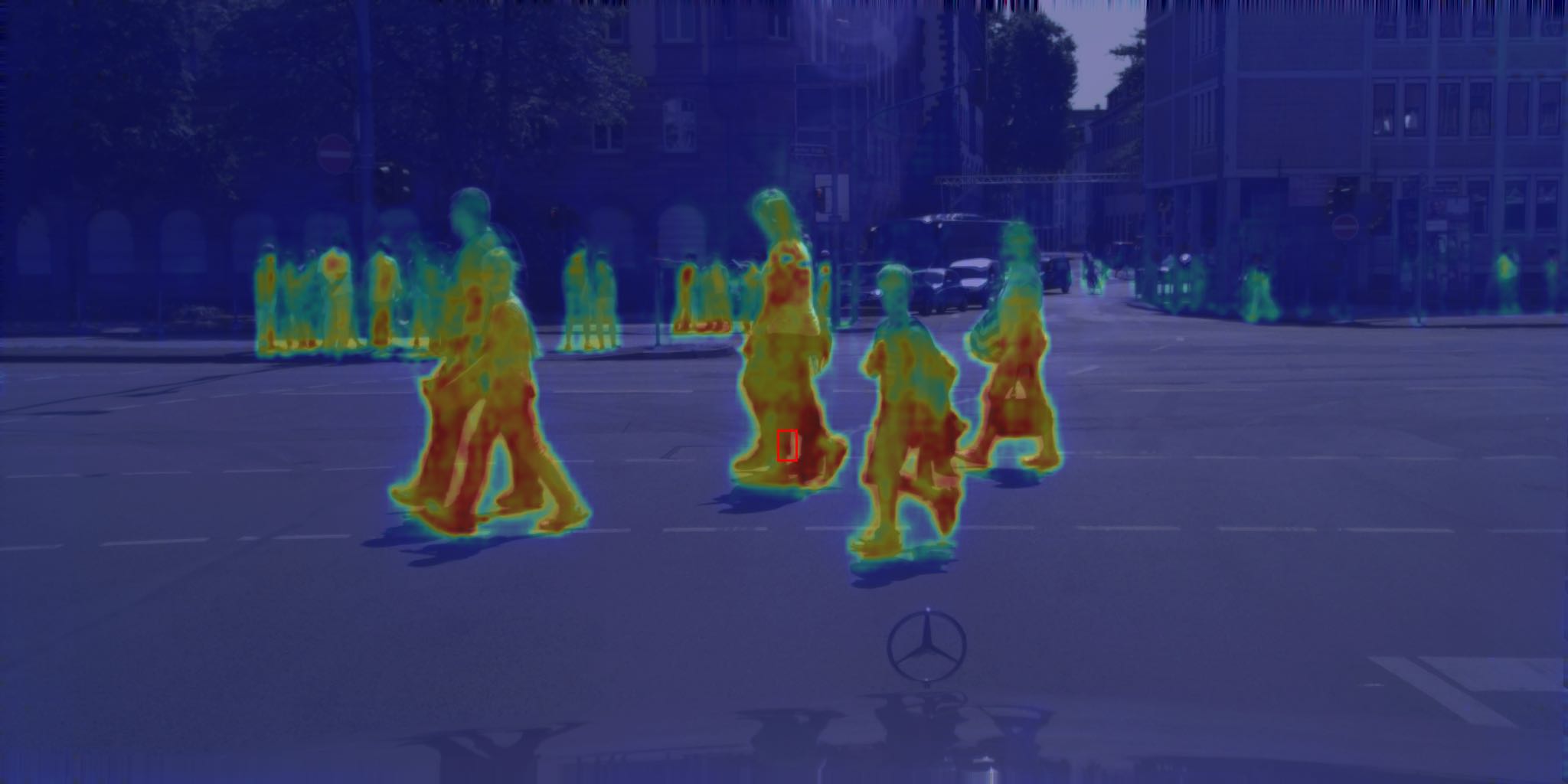}
        \end{subfigure} \\
        \rotatebox{90}{{\parbox{4cm}{\centering Top 2 Prototype}}} &
        \begin{subfigure}[b]{0.45\linewidth}
            \centering
            \includegraphics[width=\linewidth, height=4cm]{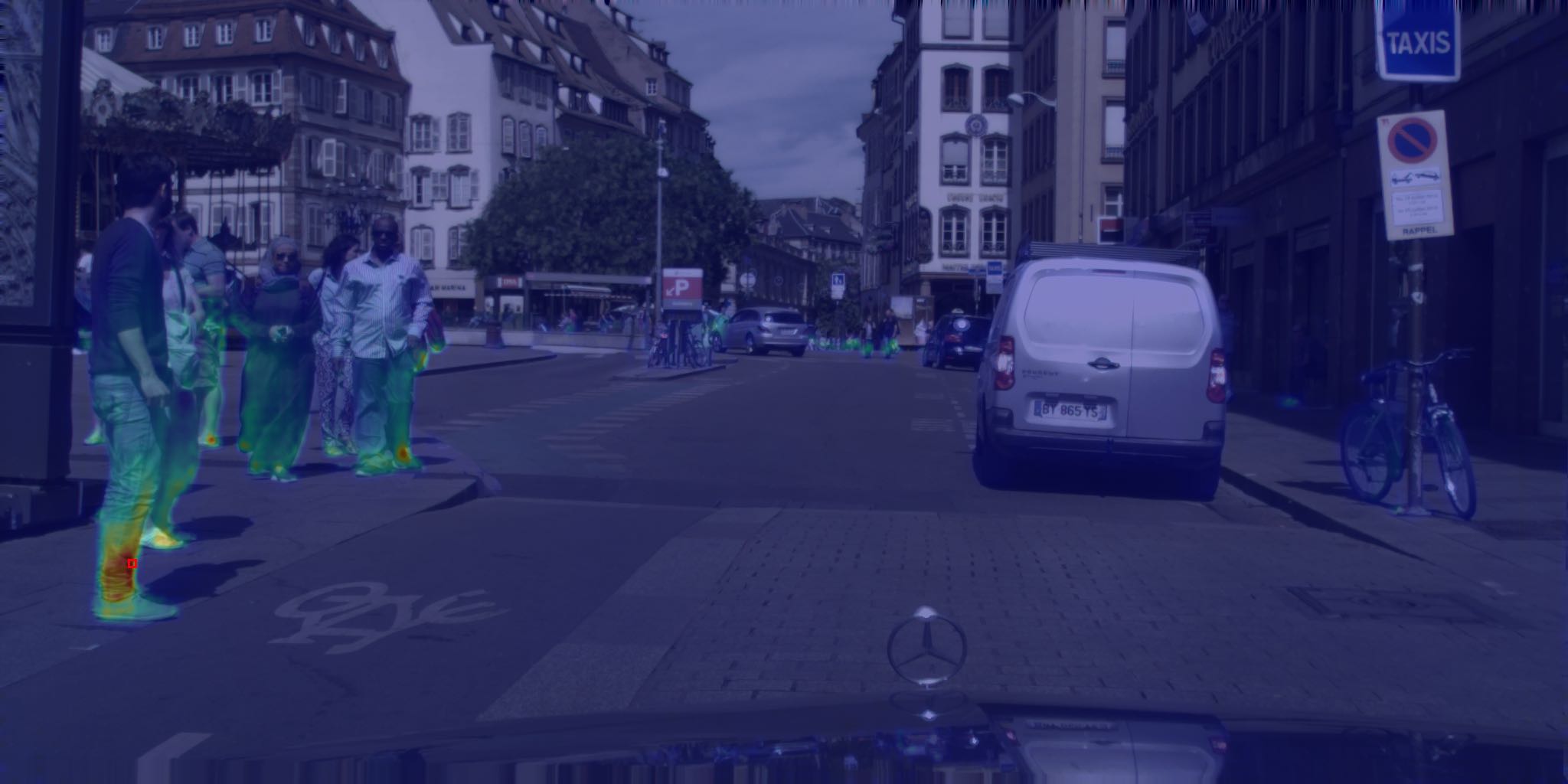}
        \end{subfigure} &
        \begin{subfigure}[b]{0.45\linewidth}
            \centering
            \includegraphics[width=\linewidth, height=4cm]{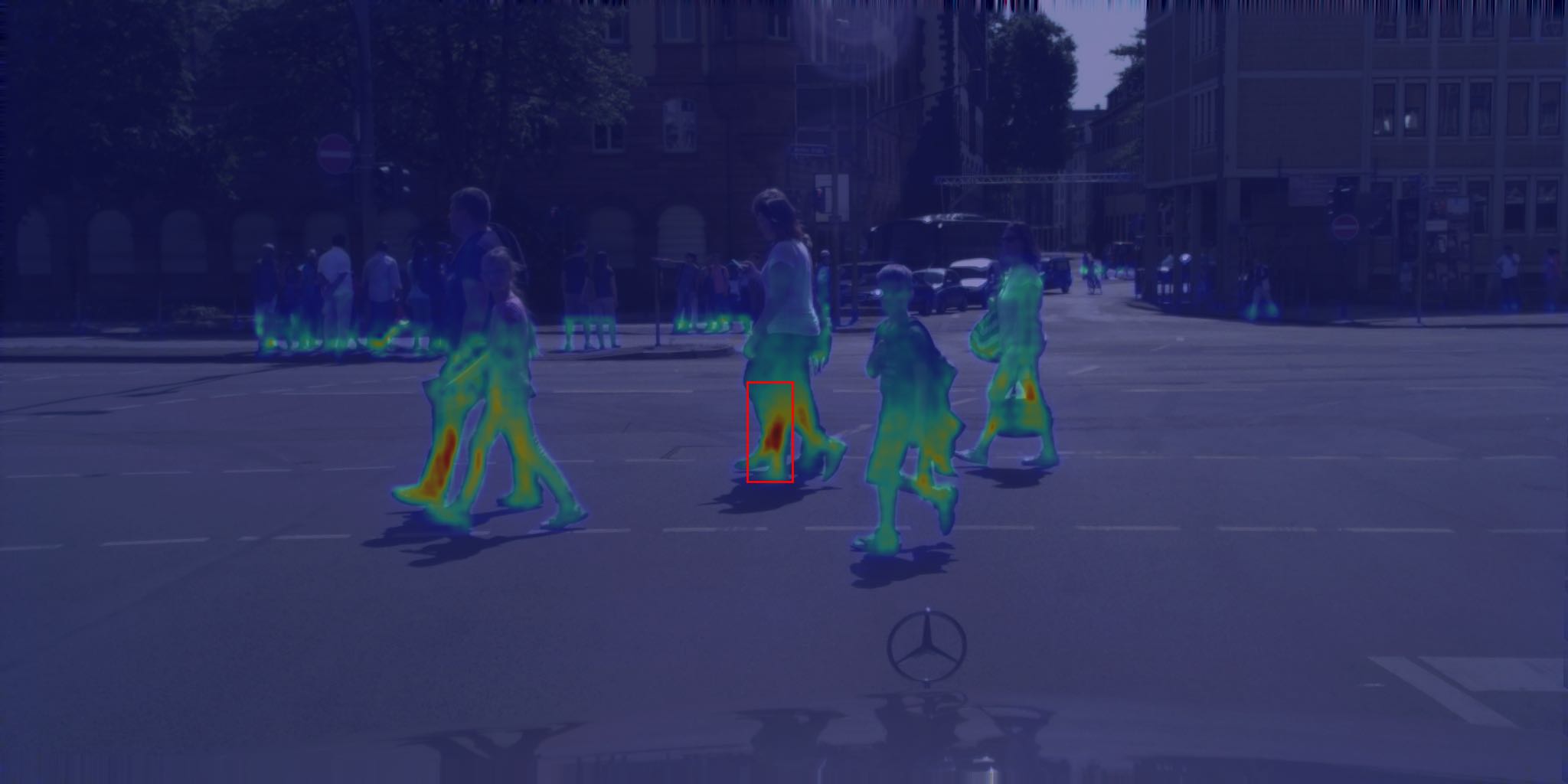}
        \end{subfigure} \\
        \rotatebox{90}{{\parbox{4cm}{\centering Top 3 Prototype}}} &
        \begin{subfigure}[b]{0.45\linewidth}
            \centering
            \includegraphics[width=\linewidth, height=4cm]{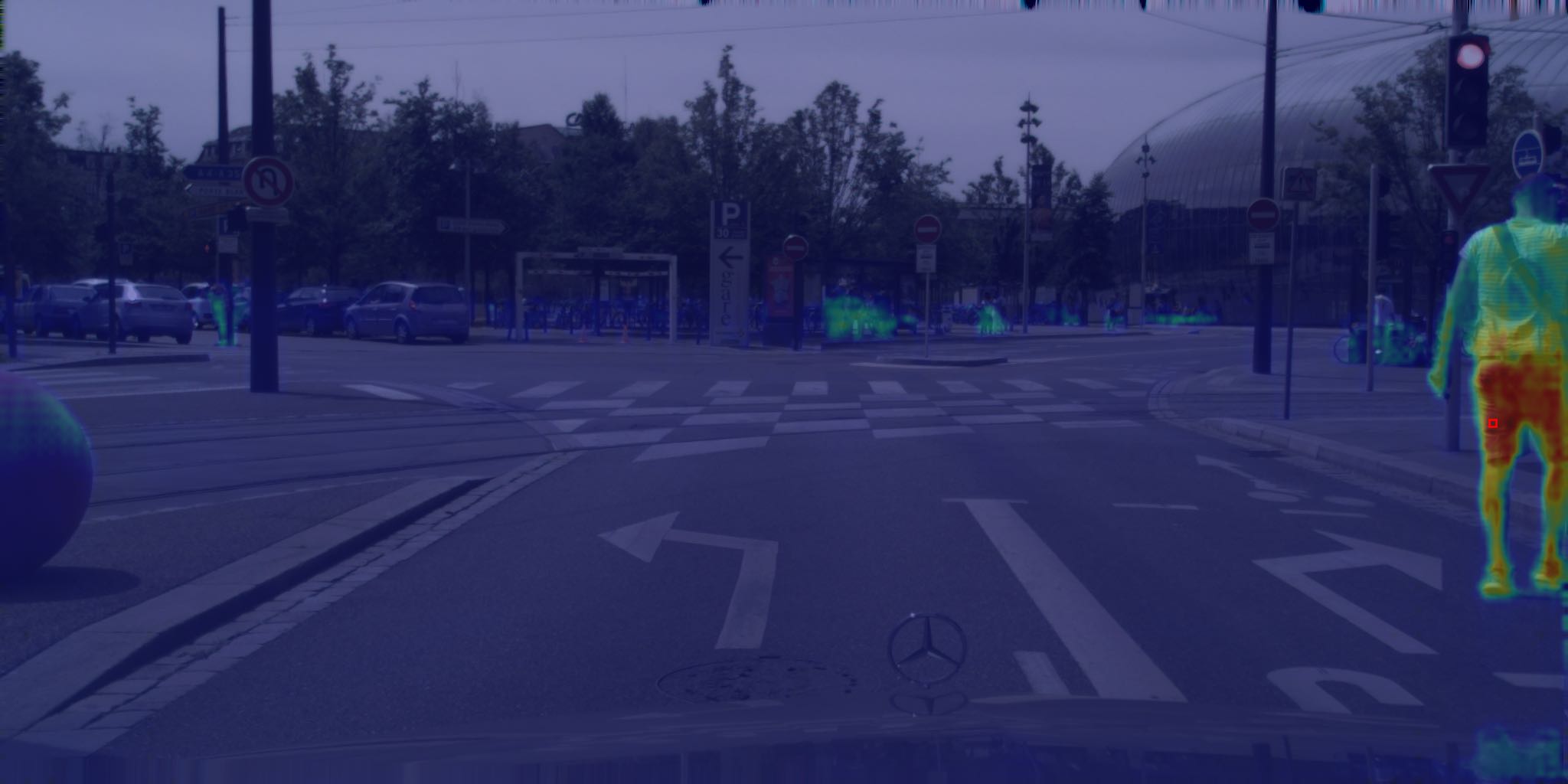}
        \end{subfigure} &
        \begin{subfigure}[b]{0.45\linewidth}
            \centering
            \includegraphics[width=\linewidth, height=4cm]{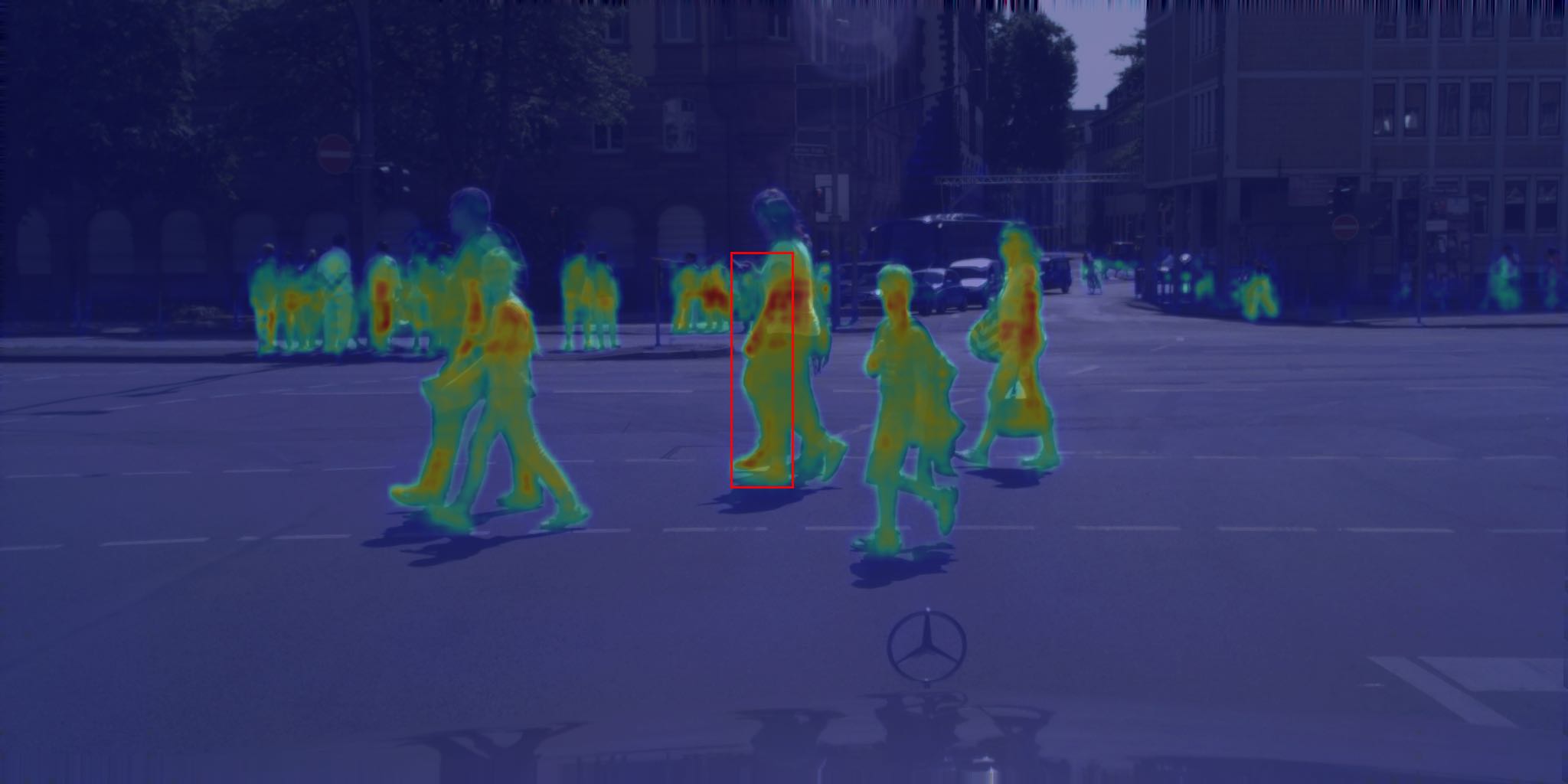}
        \end{subfigure}
    \end{tabular}
    
    \caption{The first row represents a validation image from Cityscapes with three randomly selected patches \textcolor{red}{red}, \textcolor{green}{green}, \textcolor{blue}{blue}. The second to fourth rows show the closest prototypes to the \textcolor{red}{red patch} via its activation on itself and on the image containing the \textcolor{red}{red patch}. In the second column, the bounding boxes correspond to the similarly activated area centered around the random patch.}
    \label{fig:person-img}
\end{figure*}

\begin{figure*}[htbp]
    \centering
    
    \begin{tabular}{cc}
        \rotatebox{90}{{\parbox{4cm}{\centering Image Patches}}} &
        \begin{subfigure}[b]{0.45\linewidth}
            \centering
            \includegraphics[width=\linewidth, height=4cm]{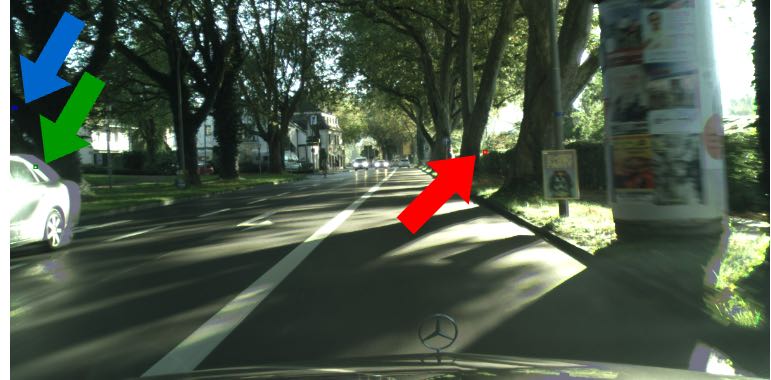}
        \end{subfigure}
    \end{tabular}
    
    \vspace{2mm} 
    
    \textcolor{green}{\rule{\linewidth}{0.5pt}}
    
    \vspace{5mm}

    \begin{tabular}{ccc}
        \rotatebox{90}{{\parbox{4cm}{\centering Top 1 Prototype}}} &
        \begin{subfigure}[b]{0.45\linewidth}
            \centering
            \includegraphics[width=\linewidth, height=4cm]{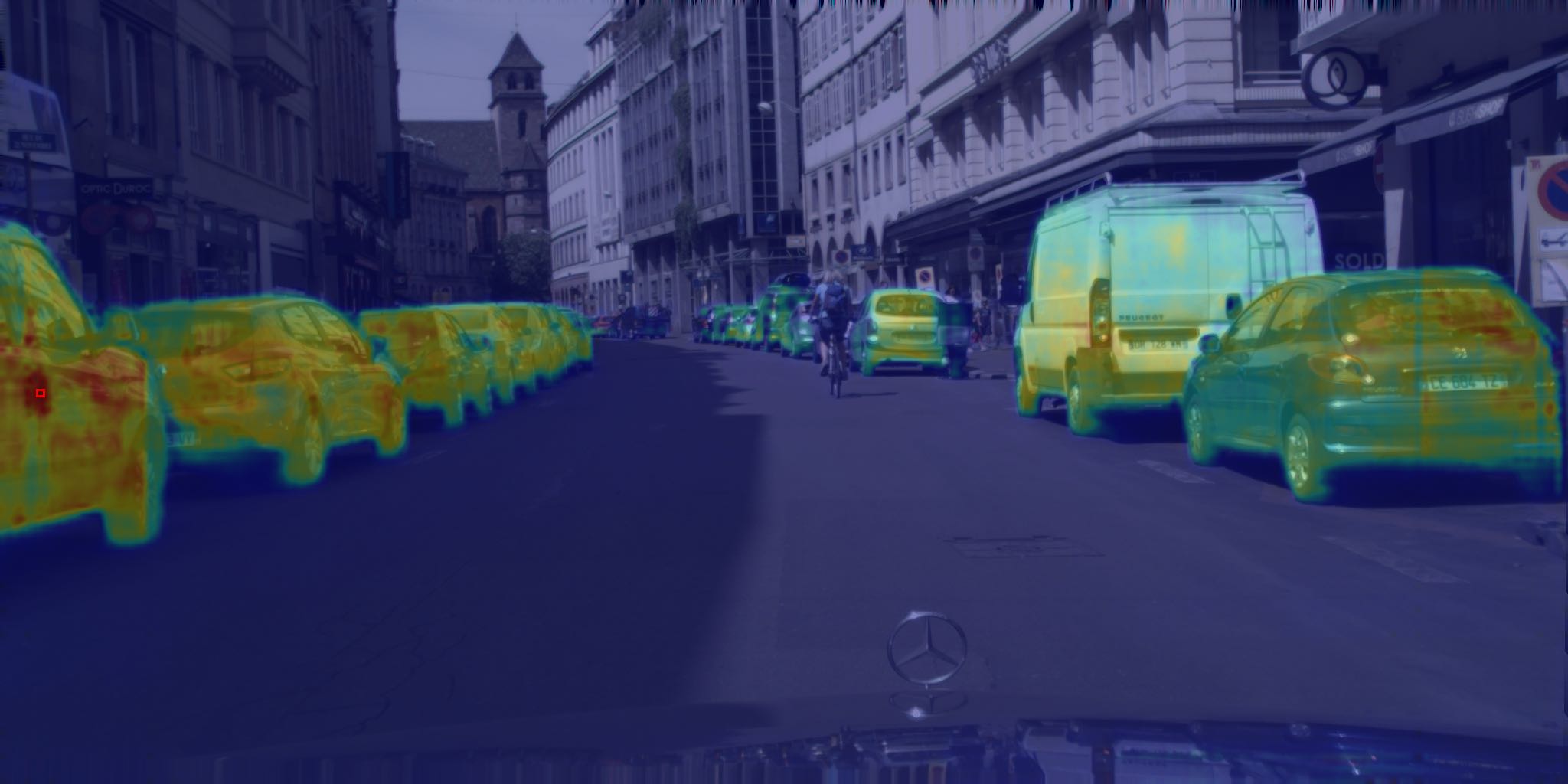}
        \end{subfigure} &
        \begin{subfigure}[b]{0.45\linewidth}
            \centering
            \includegraphics[width=\linewidth, height=4cm]{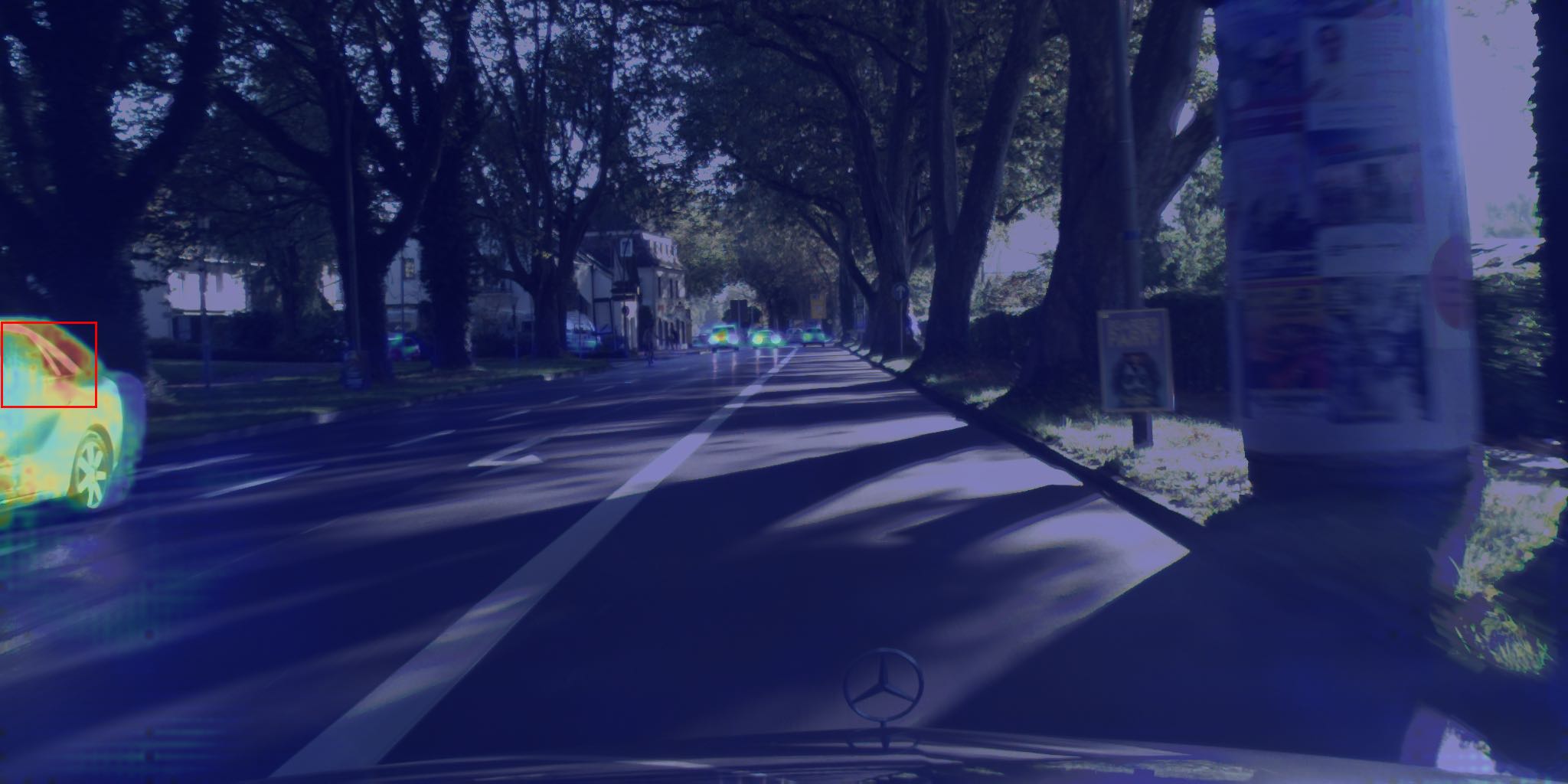}
        \end{subfigure} \\
        \rotatebox{90}{{\parbox{4cm}{\centering Top 2 Prototype}}} &
        \begin{subfigure}[b]{0.45\linewidth}
            \centering
            \includegraphics[width=\linewidth, height=4cm]{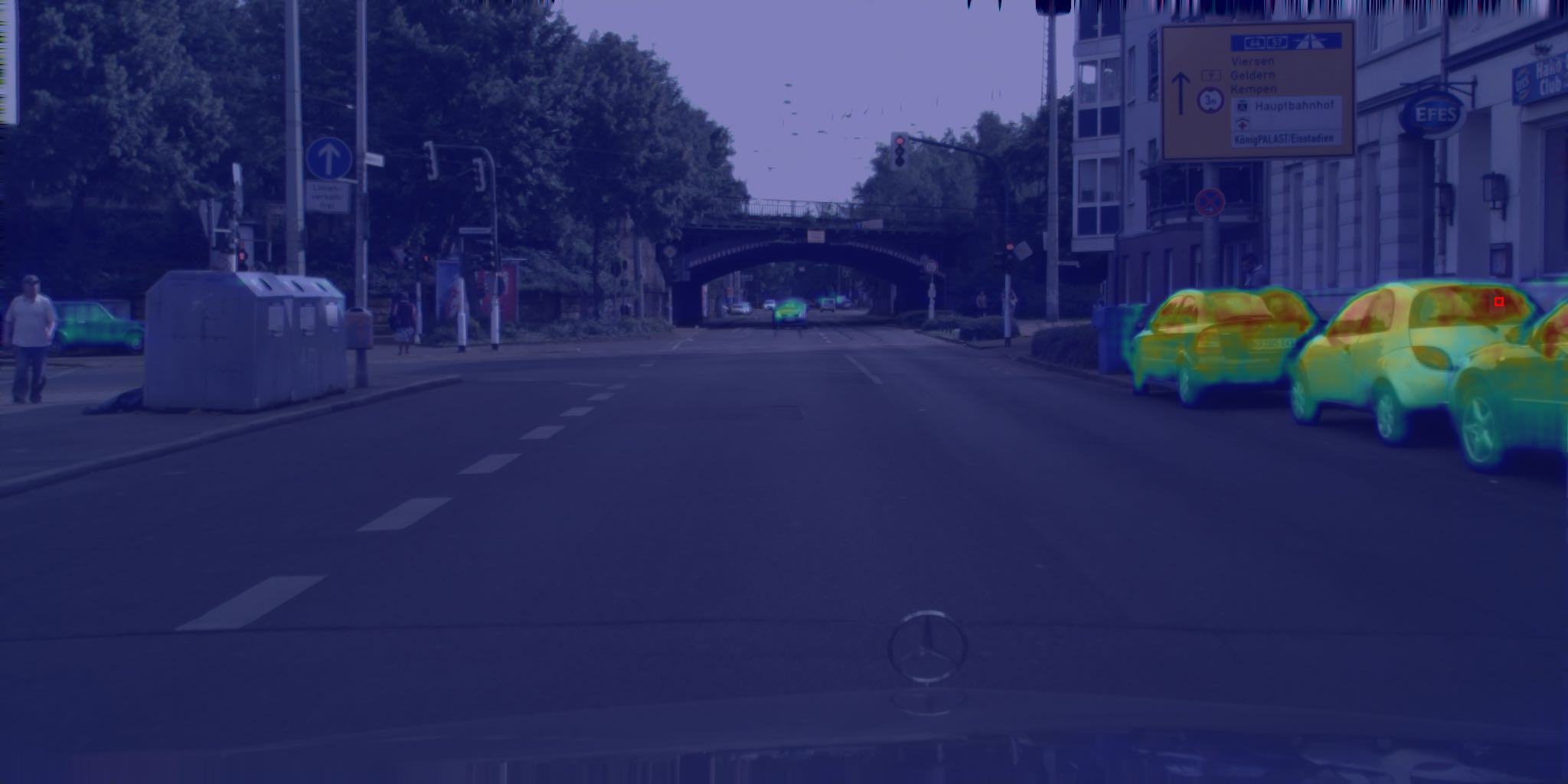}
        \end{subfigure} &
        \begin{subfigure}[b]{0.45\linewidth}
            \centering
            \includegraphics[width=\linewidth, height=4cm]{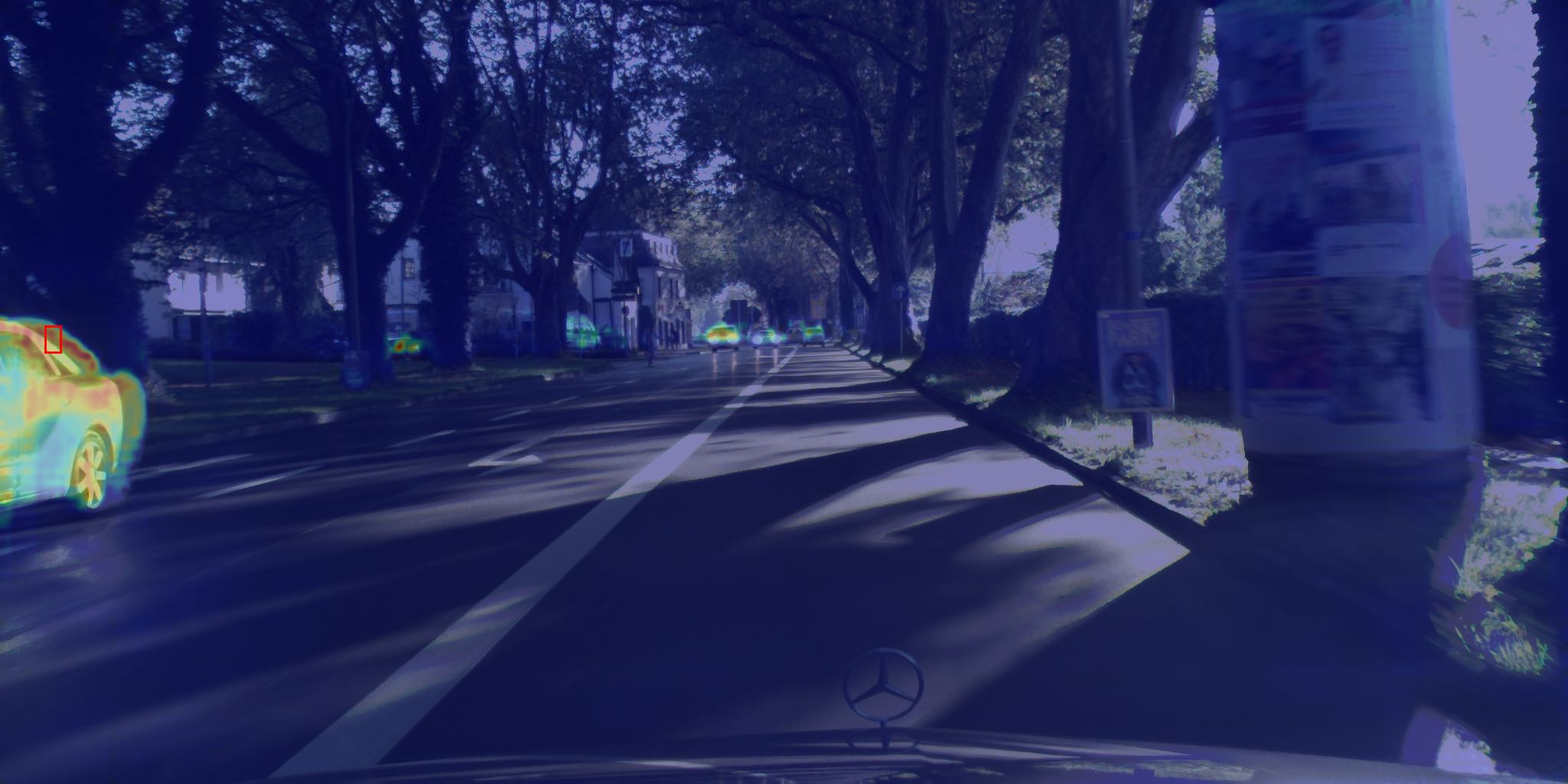}
        \end{subfigure} \\
        \rotatebox{90}{{\parbox{4cm}{\centering Top 3 Prototype}}} &
        \begin{subfigure}[b]{0.45\linewidth}
            \centering
            \includegraphics[width=\linewidth, height=4cm]{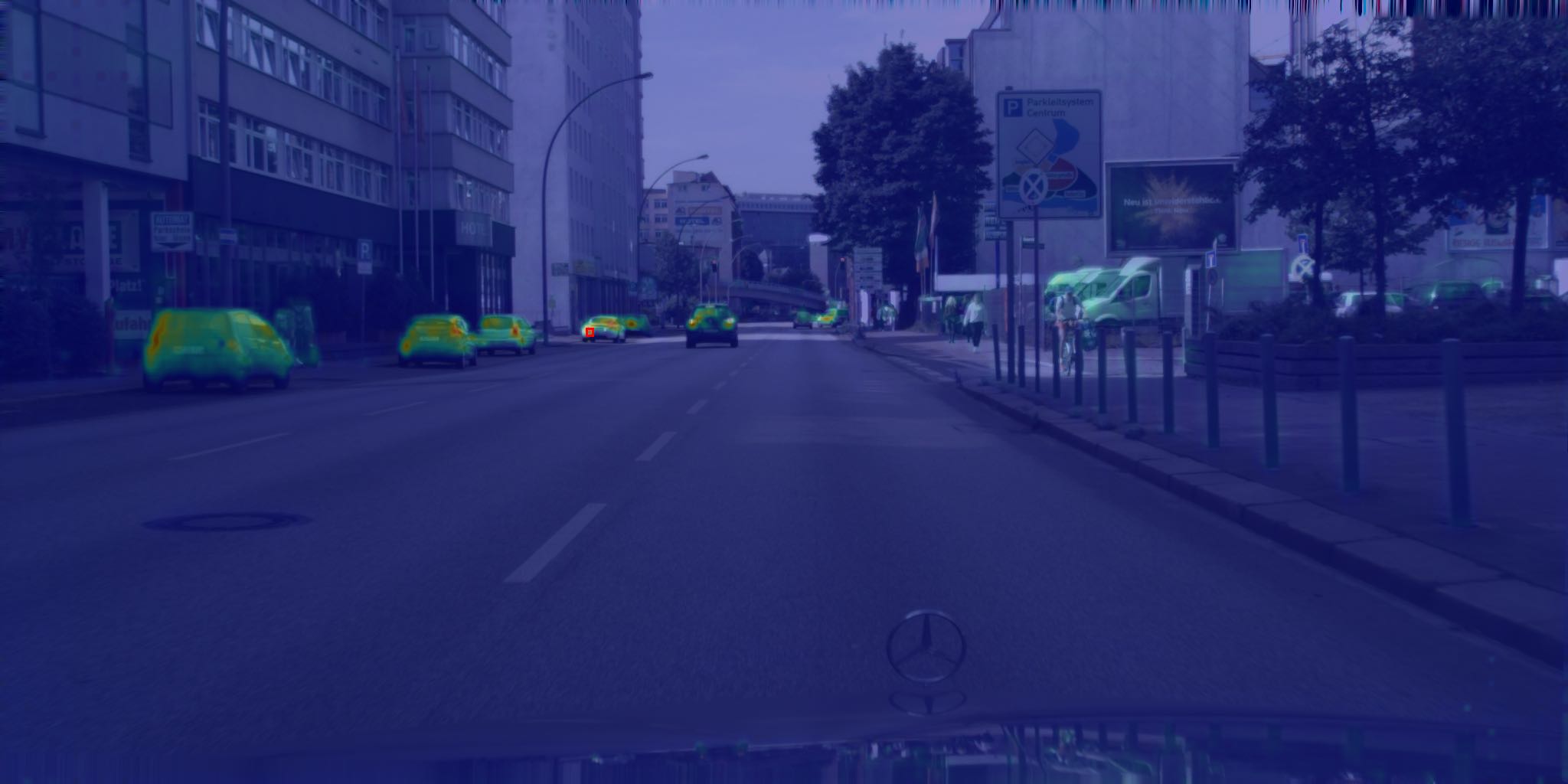}
        \end{subfigure} &
        \begin{subfigure}[b]{0.45\linewidth}
            \centering
            \includegraphics[width=\linewidth, height=4cm]{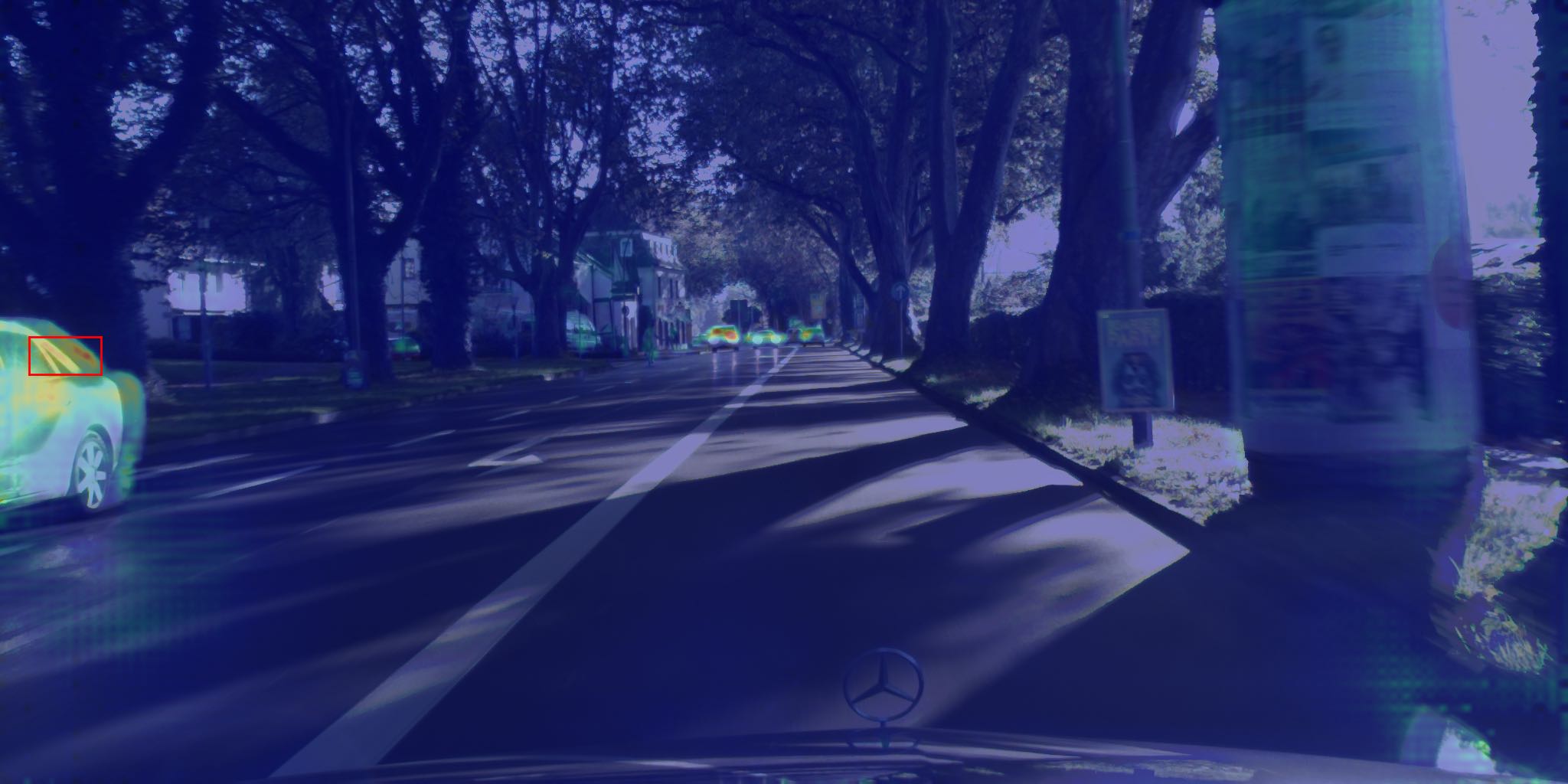}
        \end{subfigure}
    \end{tabular}
    
    \caption{The first row represents a validation image from Cityscapes with three randomly selected patches \textcolor{red}{red}, \textcolor{green}{green}, \textcolor{blue}{blue}. The second to fourth rows show the closest prototypes to the \textcolor{green}{green patch} via its activation on itself and on the image containing the \textcolor{green}{green patch}. In the second column, the bounding boxes correspond to the similarly activated area centered around the random patch.}
    \label{fig:car-img}
\end{figure*}

\begin{figure*}[htbp]
    \centering
    
    \begin{tabular}{cc}
        \rotatebox{90}{{\parbox{4cm}{\centering Image Patches}}} &
        \begin{subfigure}[b]{0.45\linewidth}
            \centering
            \includegraphics[width=\linewidth, height=4cm]{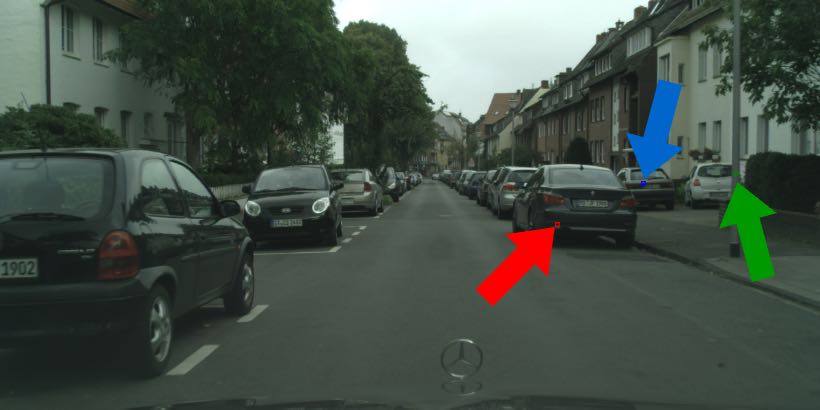}
        \end{subfigure}
    \end{tabular}
    
    \vspace{2mm} 
    
    \textcolor{green}{\rule{\linewidth}{0.5pt}}
    
    \vspace{5mm}

    \begin{tabular}{ccc}
        \rotatebox{90}{{\parbox{4cm}{\centering Top 1 Prototype}}} &
        \begin{subfigure}[b]{0.45\linewidth}
            \centering
            \includegraphics[width=\linewidth, height=4cm]{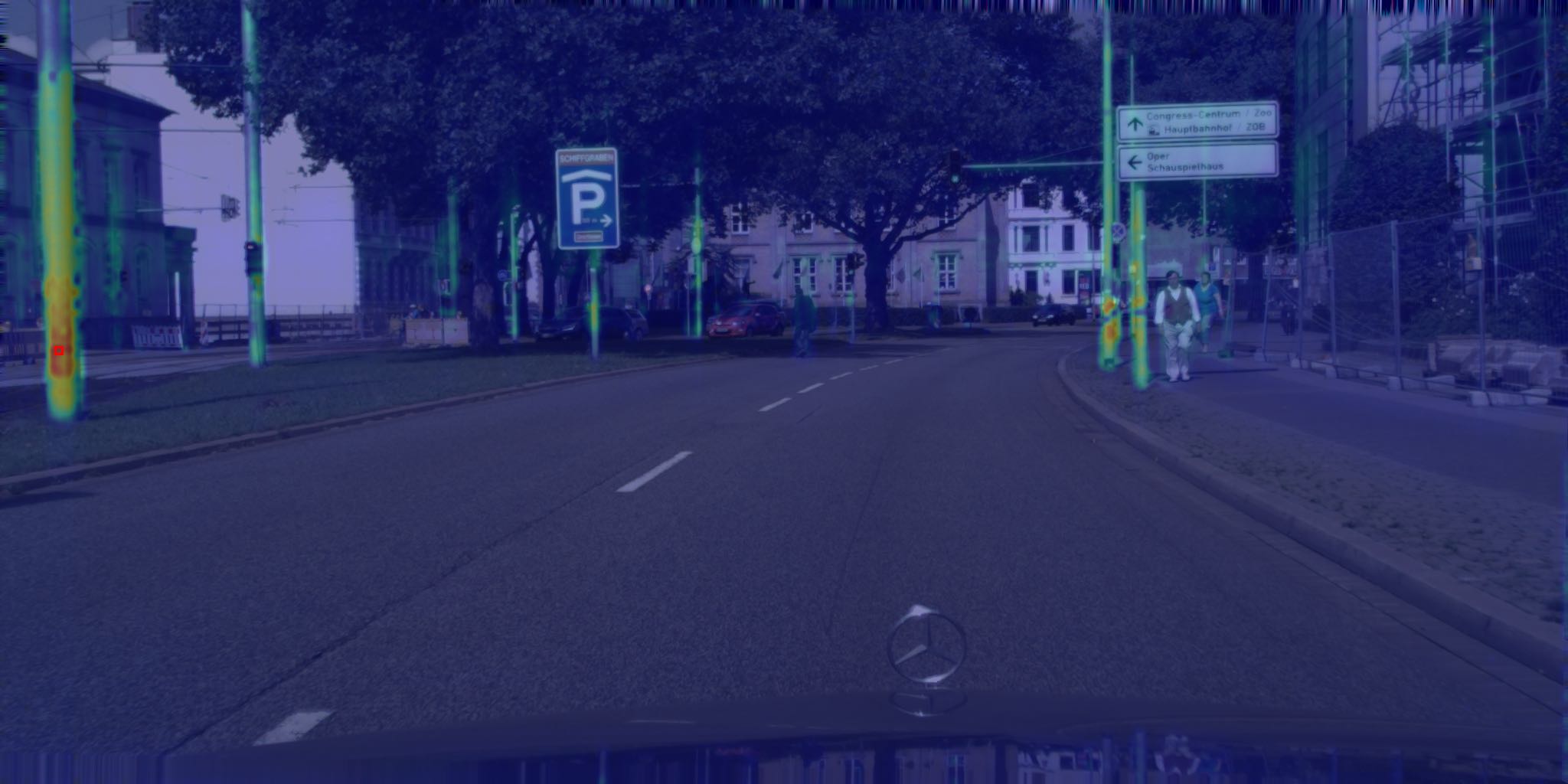}
        \end{subfigure} &
        \begin{subfigure}[b]{0.45\linewidth}
            \centering
            \includegraphics[width=\linewidth, height=4cm]{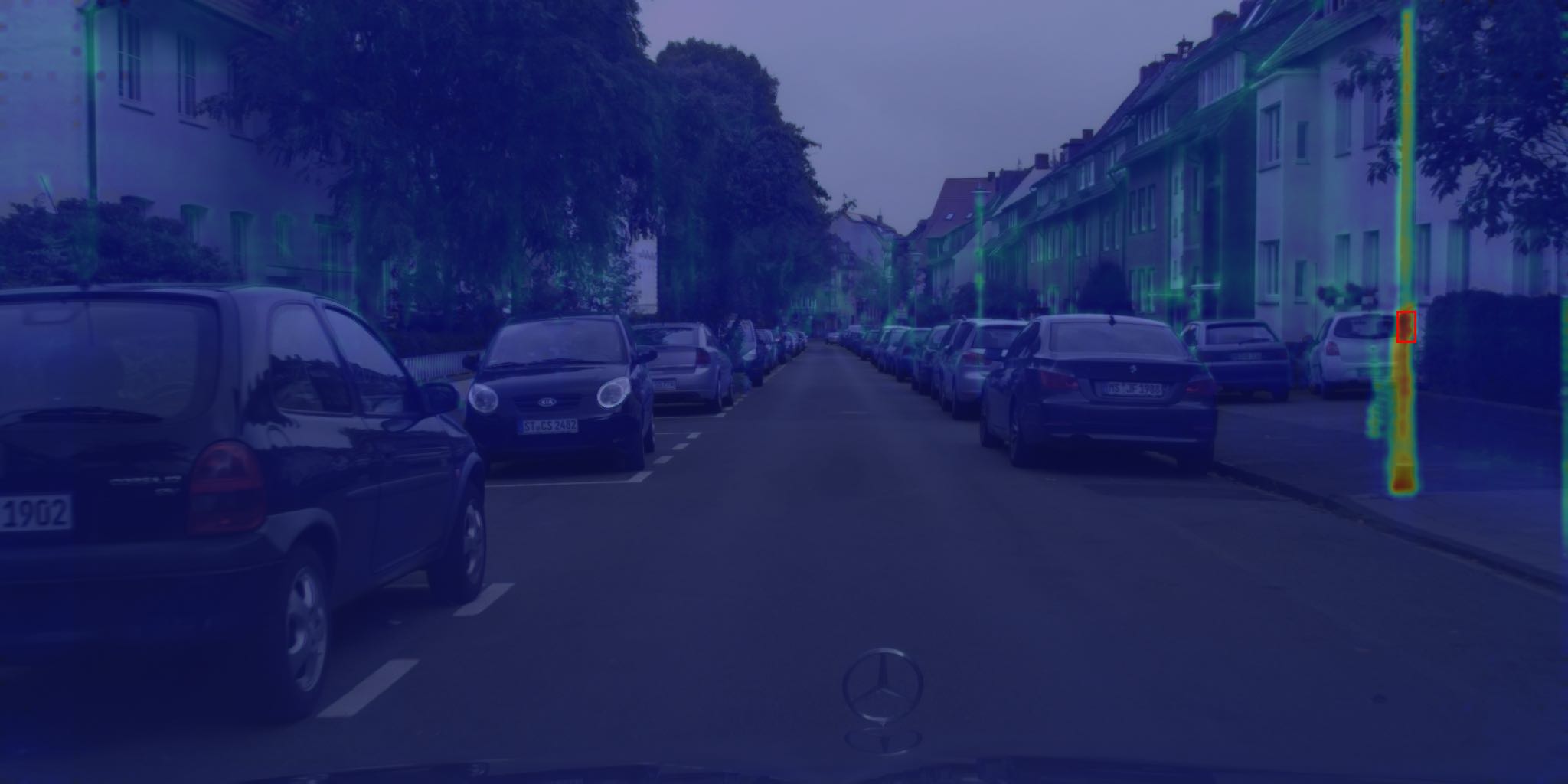}
        \end{subfigure} \\
        \rotatebox{90}{{\parbox{4cm}{\centering Top 2 Prototype}}} &
        \begin{subfigure}[b]{0.45\linewidth}
            \centering
            \includegraphics[width=\linewidth, height=4cm]{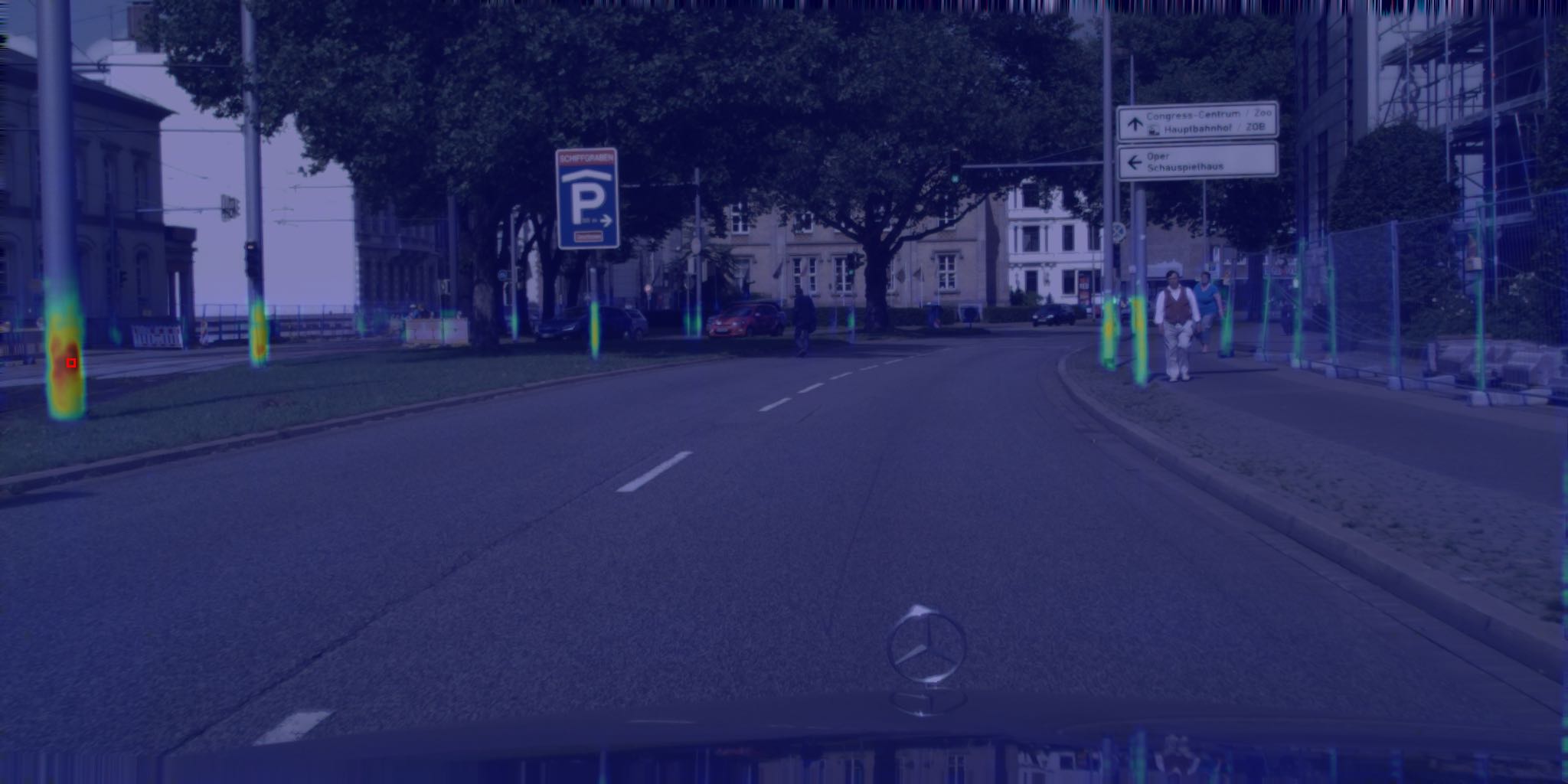}
        \end{subfigure} &
        \begin{subfigure}[b]{0.45\linewidth}
            \centering
            \includegraphics[width=\linewidth, height=4cm]{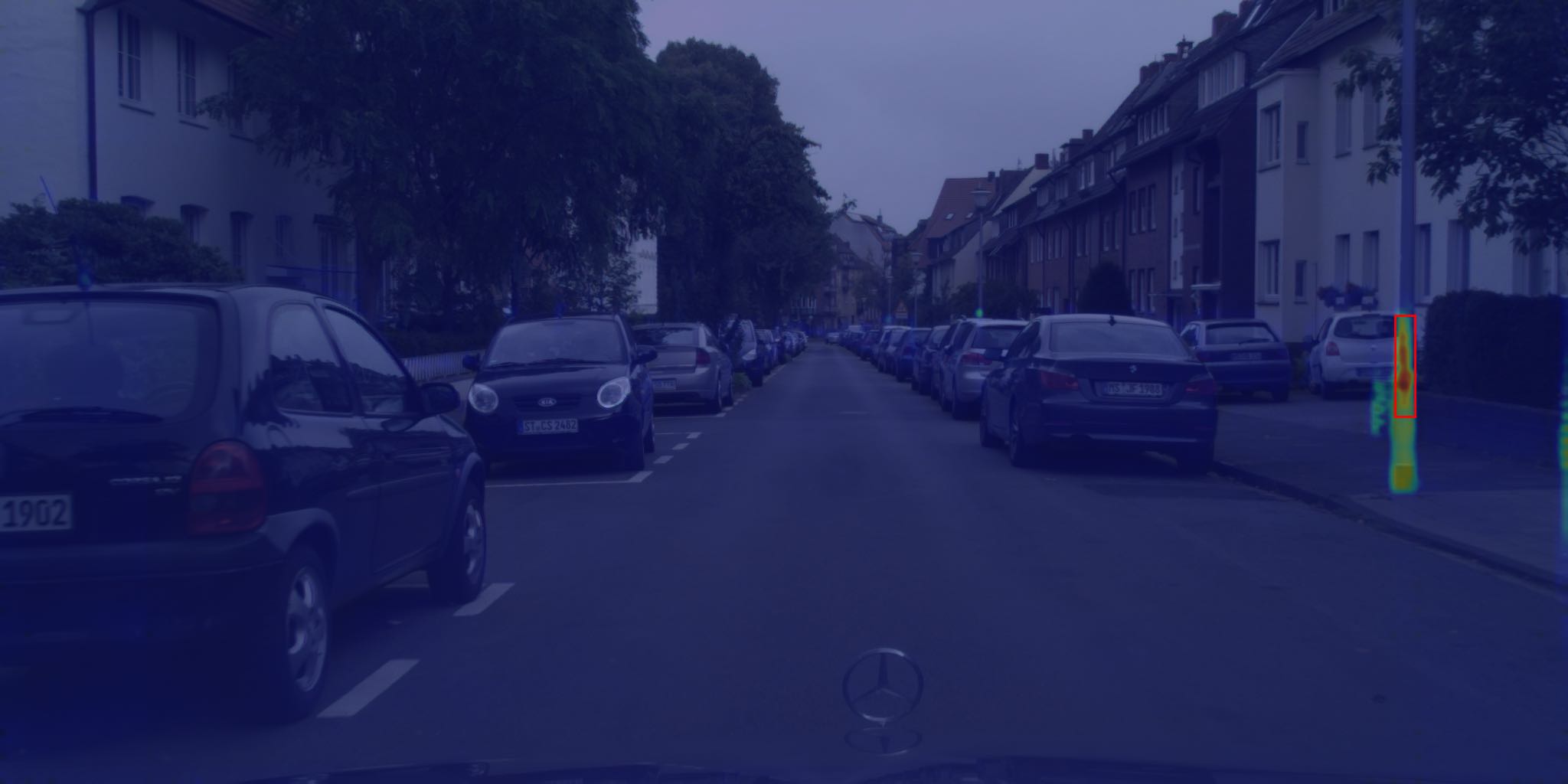}
        \end{subfigure} \\
        \rotatebox{90}{{\parbox{4cm}{\centering Top 3 Prototype}}} &
        \begin{subfigure}[b]{0.45\linewidth}
            \centering
            \includegraphics[width=\linewidth, height=4cm]{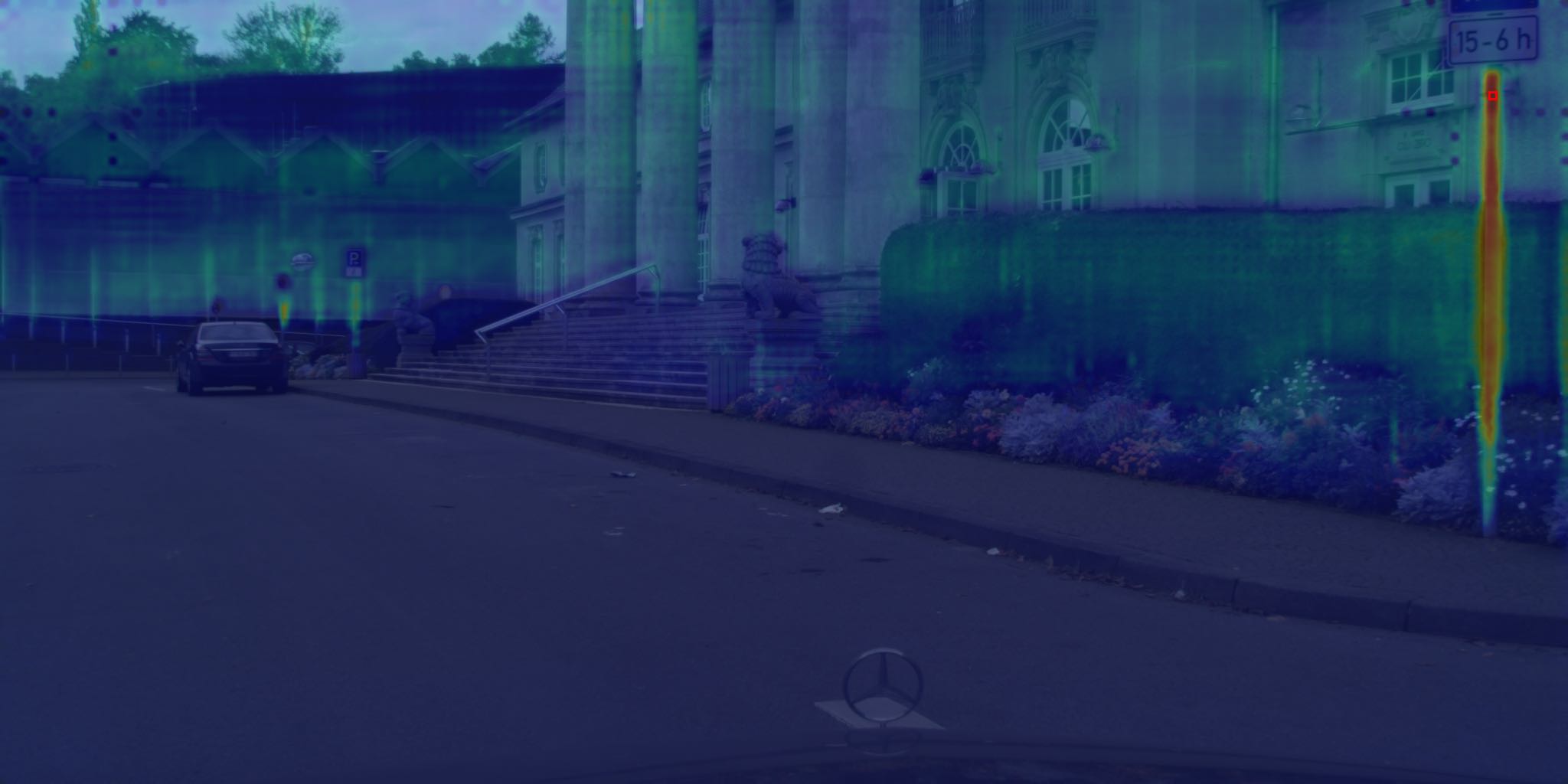}
        \end{subfigure} &
        \begin{subfigure}[b]{0.45\linewidth}
            \centering
            \includegraphics[width=\linewidth, height=4cm]{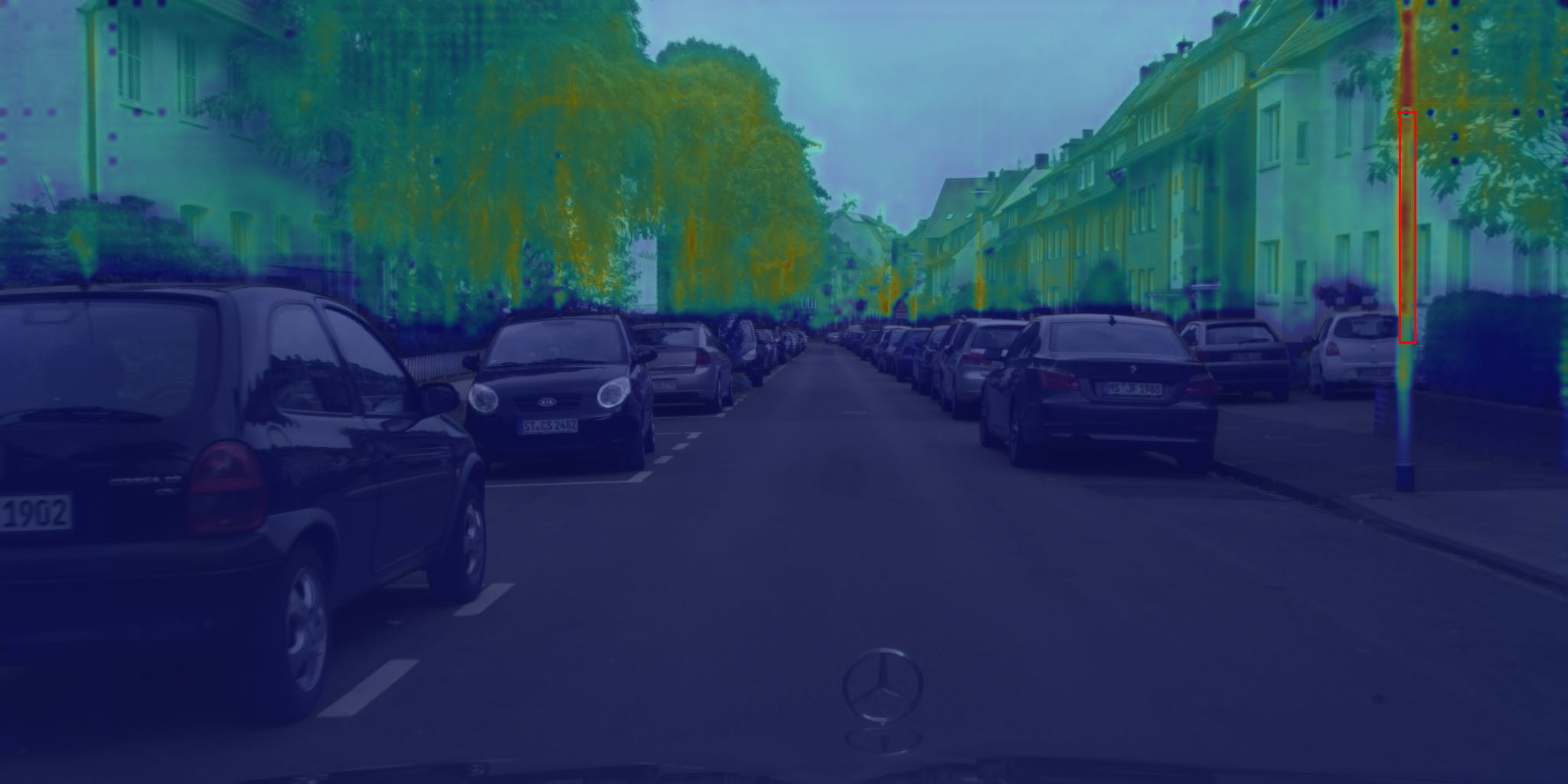}
        \end{subfigure}
    \end{tabular}
    
    \caption{The first row represents a validation image from Cityscapes with three randomly selected patches \textcolor{red}{red}, \textcolor{green}{green}, \textcolor{blue}{blue}. The second to fourth rows show the closest prototypes to the \textcolor{green}{green patch} via its activation on itself and on the image containing the \textcolor{green}{green patch}. In the second column, the bounding boxes correspond to the similarly activated area centered around the random patch.}
    \label{fig:pole-img}
\end{figure*}

\begin{figure*}[htbp]
    \centering
    
    \begin{tabular}{cc}
        \rotatebox{90}{{\parbox{4cm}{\centering Image Patches}}} &
        \begin{subfigure}[b]{0.45\linewidth}
            \centering
            \includegraphics[width=\linewidth, height=4cm]{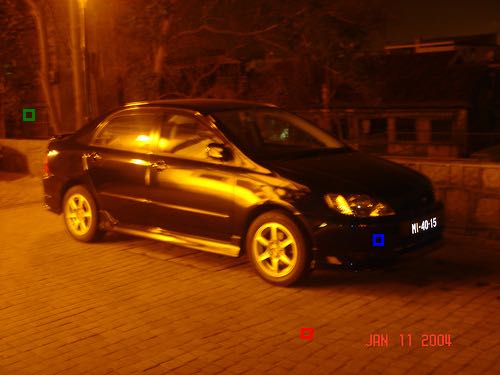}
        \end{subfigure}
    \end{tabular}
    
    \vspace{2mm} 
    
    \textcolor{blue}{\rule{\linewidth}{0.5pt}}
    
    \vspace{5mm}

    \begin{tabular}{ccc}
        \rotatebox{90}{{\parbox{4cm}{\centering Top 1 Prototype}}} &
        \begin{subfigure}[b]{0.45\linewidth}
            \centering
            \includegraphics[width=\linewidth, height=4cm]{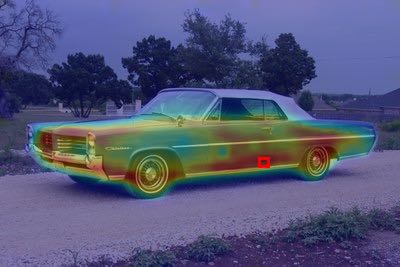}
        \end{subfigure} &
        \begin{subfigure}[b]{0.45\linewidth}
            \centering
            \includegraphics[width=\linewidth, height=4cm]{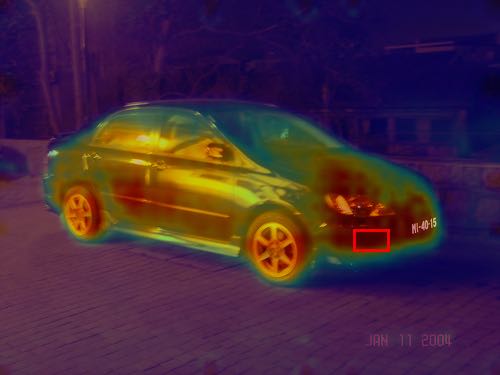}
        \end{subfigure} \\
        \rotatebox{90}{{\parbox{4cm}{\centering Top 2 Prototype}}} &
        \begin{subfigure}[b]{0.45\linewidth}
            \centering
            \includegraphics[width=\linewidth, height=4cm]{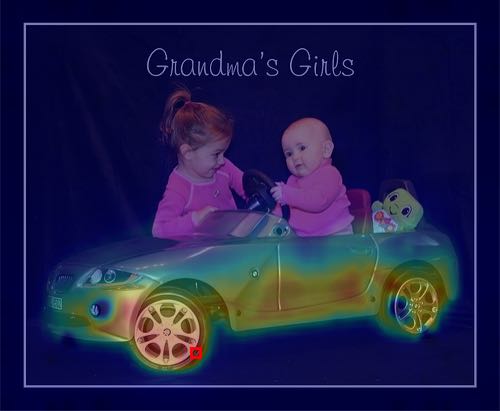}
        \end{subfigure} &
        \begin{subfigure}[b]{0.45\linewidth}
            \centering
            \includegraphics[width=\linewidth, height=4cm]{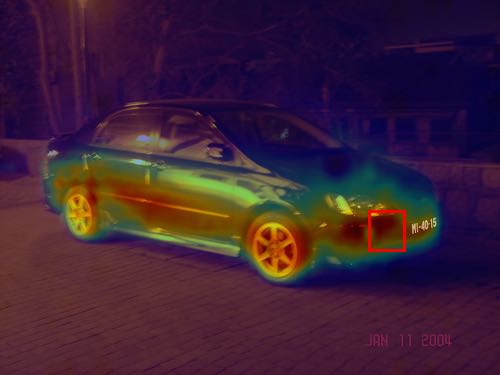}
        \end{subfigure} \\
        \rotatebox{90}{{\parbox{4cm}{\centering Top 3 Prototype}}} &
        \begin{subfigure}[b]{0.45\linewidth}
            \centering
            \includegraphics[width=\linewidth, height=4cm]{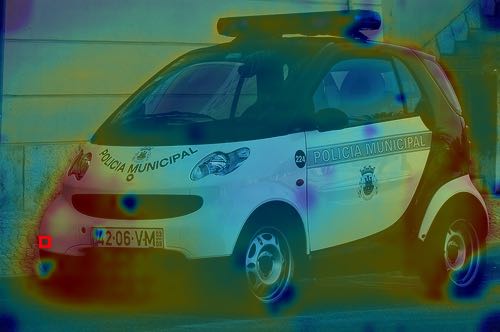}
        \end{subfigure} &
        \begin{subfigure}[b]{0.45\linewidth}
            \centering
            \includegraphics[width=\linewidth, height=4cm]{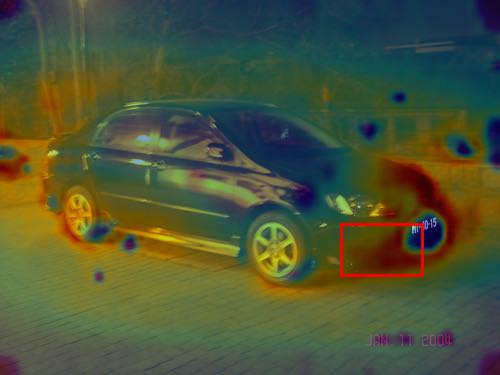}
        \end{subfigure}
    \end{tabular}
    
    \caption{The first row represents a validation image from PASCAL VOC with three randomly selected patches \textcolor{red}{red}, \textcolor{green}{green}, \textcolor{blue}{blue}. The second to fourth rows show the closest prototypes to the \textcolor{blue}{blue patch} via its activation on itself and on the image containing the \textcolor{blue}{blue patch}. In the second column, the bounding boxes correspond to the similarly activated area centered around the random patch.}
    \label{fig:car-pascal-img}
\end{figure*}

\begin{figure*}[htbp]
    \centering
    
    \begin{tabular}{cc}
        \rotatebox{90}{{\parbox{4cm}{\centering Image Patches}}} &
        \begin{subfigure}[b]{0.45\linewidth}
            \centering
            \includegraphics[width=\linewidth, height=4cm]{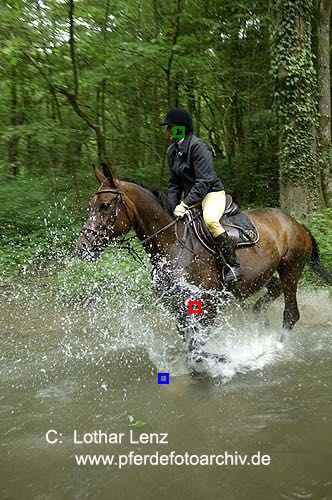}
        \end{subfigure}
    \end{tabular}
    
    \vspace{2mm} 
    
    \textcolor{green}{\rule{\linewidth}{0.5pt}}
    
    \vspace{5mm}

    \begin{tabular}{ccc}
        \rotatebox{90}{{\parbox{4cm}{\centering Top 1 Prototype}}} &
        \begin{subfigure}[b]{0.45\linewidth}
            \centering
            \includegraphics[width=\linewidth, height=4cm]{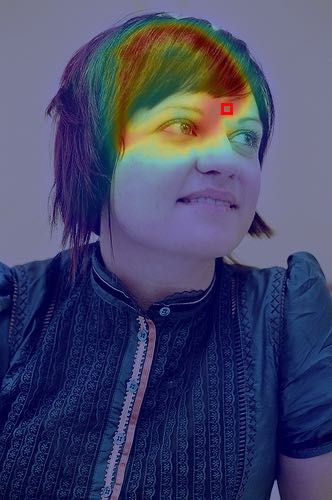}
        \end{subfigure} &
        \begin{subfigure}[b]{0.45\linewidth}
            \centering
            \includegraphics[width=\linewidth, height=4cm]{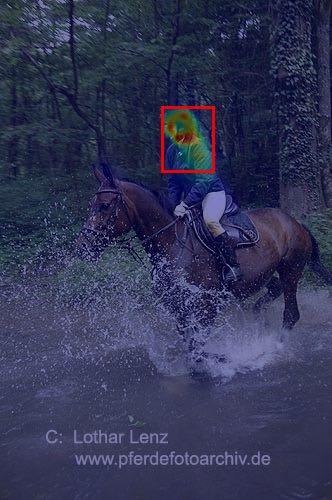}
        \end{subfigure} \\
        \rotatebox{90}{{\parbox{4cm}{\centering Top 2 Prototype}}} &
        \begin{subfigure}[b]{0.45\linewidth}
            \centering
            \includegraphics[width=\linewidth, height=4cm]{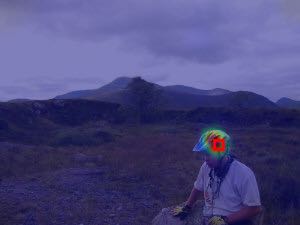}
        \end{subfigure} &
        \begin{subfigure}[b]{0.45\linewidth}
            \centering
            \includegraphics[width=\linewidth, height=4cm]{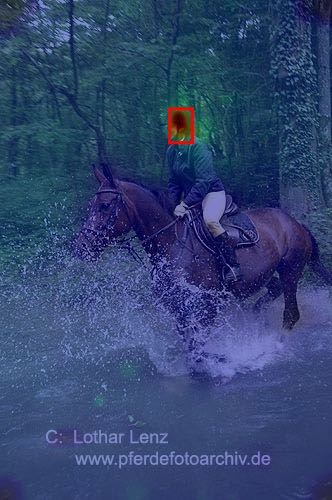}
        \end{subfigure} \\
        \rotatebox{90}{{\parbox{4cm}{\centering Top 3 Prototype}}} &
        \begin{subfigure}[b]{0.45\linewidth}
            \centering
            \includegraphics[width=\linewidth, height=4cm]{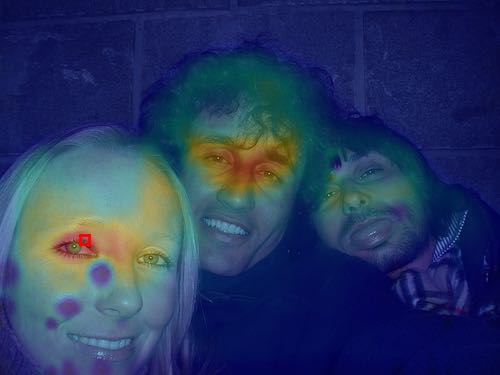}
        \end{subfigure} &
        \begin{subfigure}[b]{0.45\linewidth}
            \centering
            \includegraphics[width=\linewidth, height=4cm]{images/247/person_loc_1/sample_247_loc_1_closest_0_scale_2_172-original_with_self_act_and_box.jpg}
        \end{subfigure}
    \end{tabular}
    
    \caption{The first row represents a validation image from PASCAL VOC with three randomly selected patches \textcolor{red}{red}, \textcolor{green}{green}, \textcolor{blue}{blue}. The second to fourth rows show the closest prototypes to the \textcolor{green}{green patch} via its activation on itself and on the image containing the \textcolor{green}{green patch}. In the second column, the bounding boxes correspond to the similarly activated area centered around the random patch.}
    \label{fig:person-pascal-img}
\end{figure*}

\begin{figure*}[htbp]
    \centering
    
    \begin{tabular}{cc}
        \rotatebox{90}{{\parbox{4cm}{\centering Image Patches}}} &
        \begin{subfigure}[b]{0.45\linewidth}
            \centering
            \includegraphics[width=\linewidth, height=4cm]{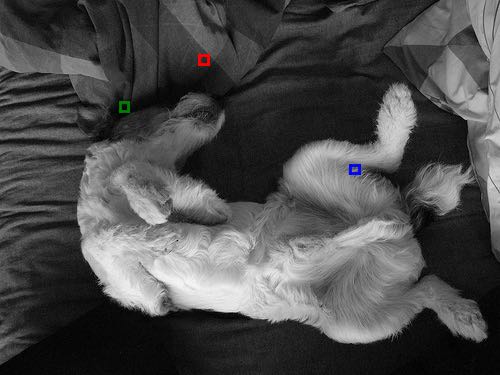}
        \end{subfigure}
    \end{tabular}
    
    \vspace{2mm} 
    
    \textcolor{blue}{\rule{\linewidth}{0.5pt}}
    
    \vspace{5mm}

    \begin{tabular}{ccc}
        \rotatebox{90}{{\parbox{4cm}{\centering Top 1 Prototype}}} &
        \begin{subfigure}[b]{0.45\linewidth}
            \centering
            \includegraphics[width=\linewidth, height=4cm]{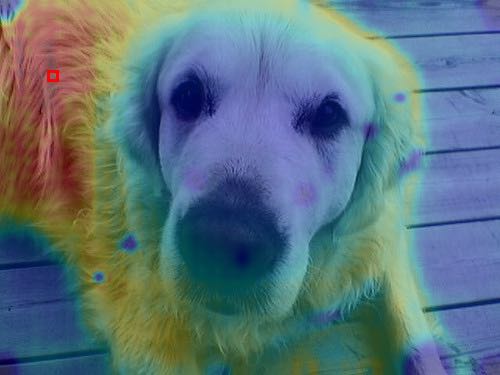}
        \end{subfigure} &
        \begin{subfigure}[b]{0.45\linewidth}
            \centering
            \includegraphics[width=\linewidth, height=4cm]{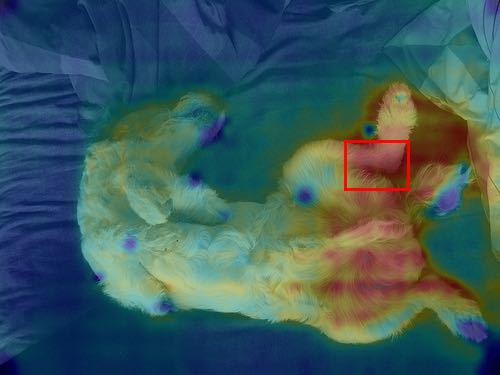}
        \end{subfigure} \\
        \rotatebox{90}{{\parbox{4cm}{\centering Top 2 Prototype}}} &
        \begin{subfigure}[b]{0.45\linewidth}
            \centering
            \includegraphics[width=\linewidth, height=4cm]{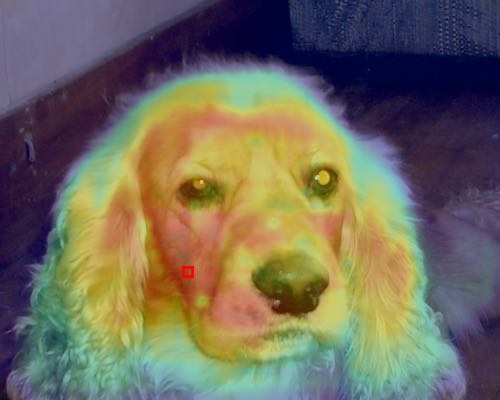}
        \end{subfigure} &
        \begin{subfigure}[b]{0.45\linewidth}
            \centering
            \includegraphics[width=\linewidth, height=4cm]{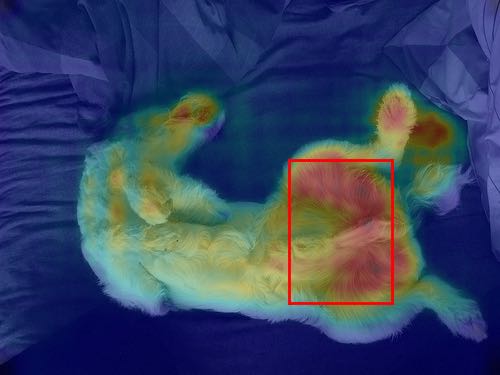}
        \end{subfigure} \\
        \rotatebox{90}{{\parbox{4cm}{\centering Top 3 Prototype}}} &
        \begin{subfigure}[b]{0.45\linewidth}
            \centering
            \includegraphics[width=\linewidth, height=4cm]{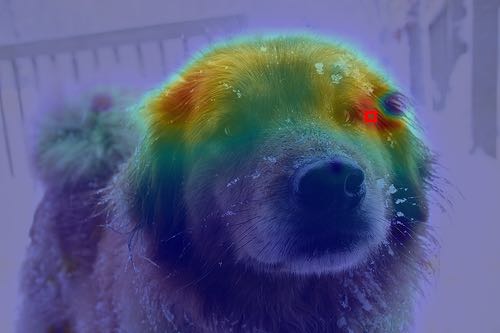}
        \end{subfigure} &
        \begin{subfigure}[b]{0.45\linewidth}
            \centering
            \includegraphics[width=\linewidth, height=4cm]{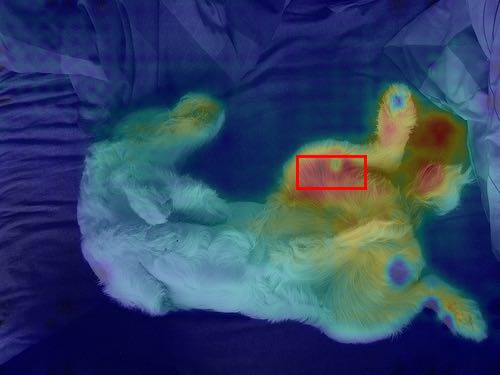}
        \end{subfigure}
    \end{tabular}
    
    \caption{The first row represents a validation image from PASCAL VOC with three randomly selected patches \textcolor{red}{red}, \textcolor{green}{green}, \textcolor{blue}{blue}. The second to fourth rows show the closest prototypes to the \textcolor{blue}{blue patch} via its activation on itself and on the image containing the \textcolor{blue}{blue patch}. In the second column, the bounding boxes correspond to the similarly activated area centered around the random patch.}
    \label{fig:dog-img}
\end{figure*}

To further analyze the semantic correspondences in the prototype layer of the model, we study the nearest prototypes to all image patches during the decision process. This is also directly linked to the consistency metric. To this purpose, in Figure [\ref{fig:person-img} - \ref{fig:dog-img}] we visualize the three closest prototypes for random patches in an image. The visualizations again showcase good semantic matches for our proposed method.

\section{Semantic meaning of groups}
\begin{figure*}
    \centering
    \includegraphics[width=0.8\linewidth]{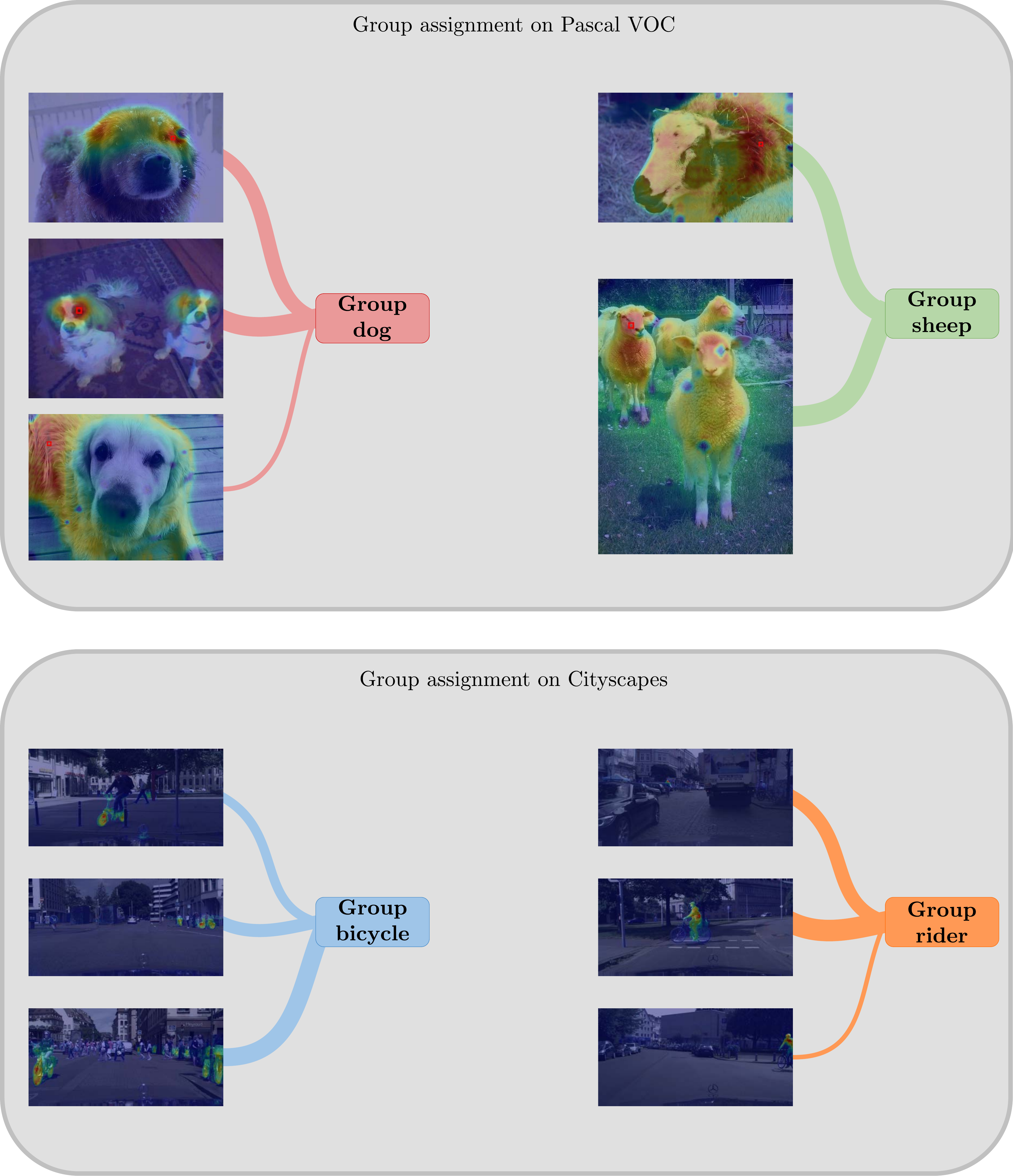}
    \caption{\textbf{Top:} for the classes \textit{dog} and \textit{sheep} on Pascal VOC, prototypes with the highest activation for the considered group represent the head. \textbf{Bottom}: for the class \textit{bicycle} of Cityscapes, prototypes with the highest activation for the considered group represent the bicycle wheels, while for the class \textit{rider} they mainly highlight the person's upper body.}
    \label{fig:semantic_groups}
\end{figure*}

We provide a clearer understanding of the grouping mechanism by showing, in Figure~\ref{fig:semantic_groups}, the activations of prototypes assigned to the same group for a given class. It can be observed that prototypes are grouped into semantically meaningful groups.

\section{Failure case analysis}
\begin{figure*}[htbp]
    \centering
    
    \begin{tabular}{ccc}
        \begin{subfigure}[b]{0.3\linewidth}
            \centering
            \includegraphics[width=\linewidth]{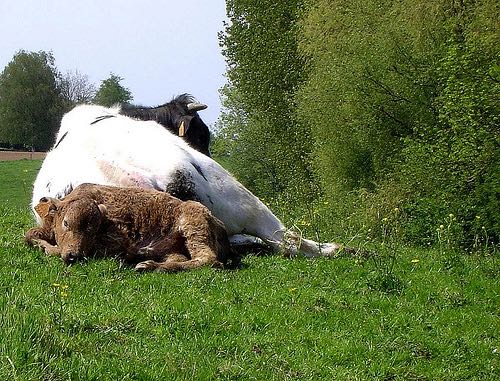}
        \end{subfigure} &
        \begin{subfigure}[b]{0.3\linewidth}
            \centering
            \includegraphics[width=\linewidth]{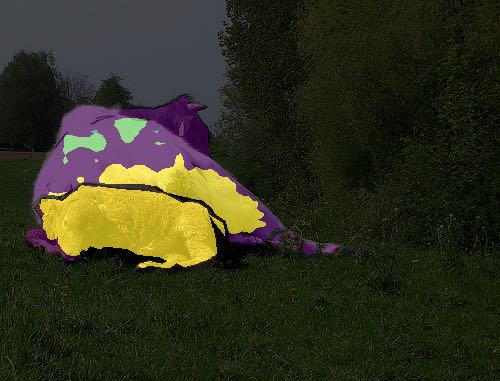}
        \end{subfigure} &
        \begin{subfigure}[b]{0.3\linewidth}
            \centering
            \includegraphics[width=\linewidth]{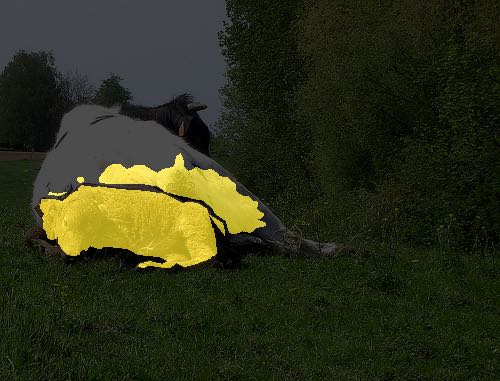}
        \end{subfigure}
    \end{tabular}
    \caption{PASCAL VOC image with its predictions for the class \textit{cow} and in particular the most common error for this class is with the class \textit{sheep}.}
    \label{fig:segmentation}
\end{figure*}

\begin{figure*}[htbp]
    \centering
    
    \begin{tabular}{ccc}
        \begin{subfigure}[b]{0.3\linewidth}
            \centering
            \includegraphics[width=\linewidth]{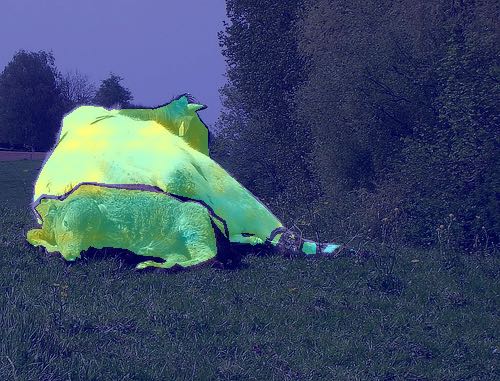}
        \end{subfigure} &
        \begin{subfigure}[b]{0.3\linewidth}
            \centering
            \includegraphics[width=\linewidth]{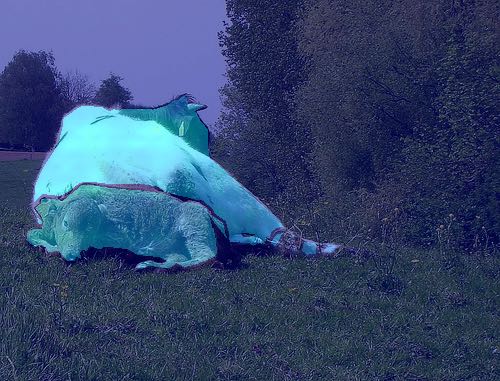}
        \end{subfigure} &
        \begin{subfigure}[b]{0.3\linewidth}
            \centering
            \includegraphics[width=\linewidth]{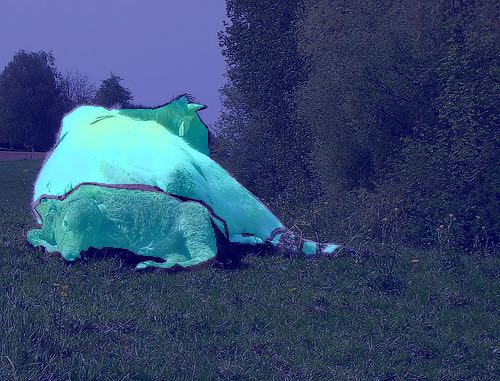}
        \end{subfigure}
    \end{tabular}
    \caption{Group activations of the class \textit{cow} for the image.}
    \label{fig:group-cow}
\end{figure*}

\begin{figure*}[htbp]
    \centering
    
    \begin{tabular}{cccc}
        \begin{subfigure}[b]{0.22\linewidth}
            \centering
            \includegraphics[width=\linewidth, height=3cm]{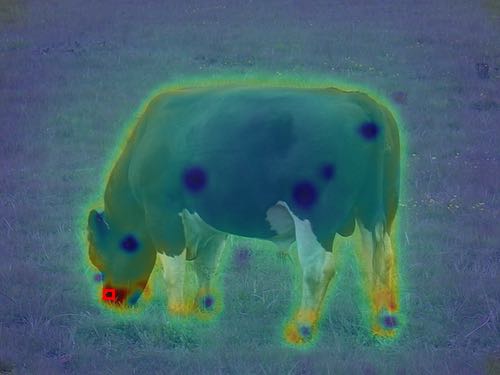}
        \end{subfigure} &
        \begin{subfigure}[b]{0.22\linewidth}
            \centering
            \includegraphics[width=\linewidth, height=3cm]{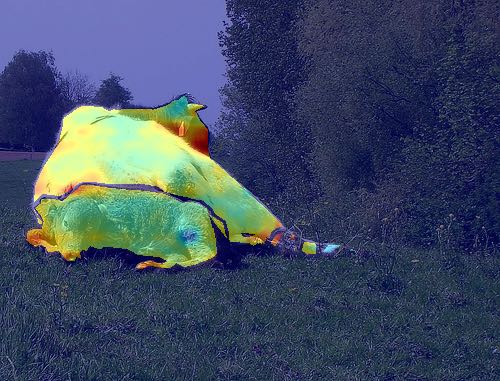}
        \end{subfigure} &
        \begin{subfigure}[b]{0.22\linewidth}
            \centering
            \includegraphics[width=\linewidth, height=3cm]{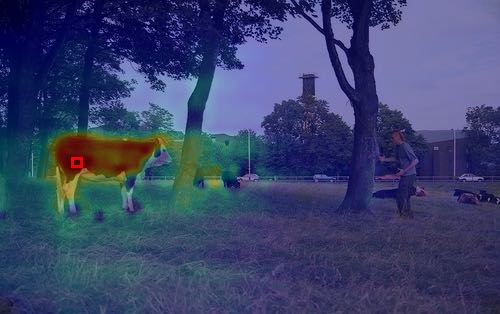}
        \end{subfigure} &
        \begin{subfigure}[b]{0.22\linewidth}
            \centering
            \includegraphics[width=\linewidth, height=3cm]{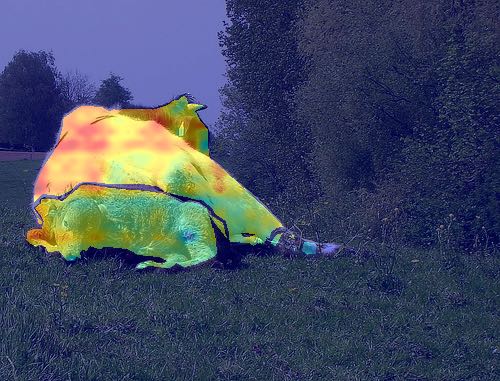}
        \end{subfigure}
    \end{tabular}
    \caption{Most activated prototypes represented by their training sample and their activation on the input image for Group 1 of class \textit{cow}.}
    \label{fig:proto-cow-1}
\end{figure*}

\begin{figure*}[htbp]
    \centering
    
    \begin{tabular}{cccc}
        \begin{subfigure}[b]{0.22\linewidth}
            \centering
            \includegraphics[width=\linewidth, height=3cm]{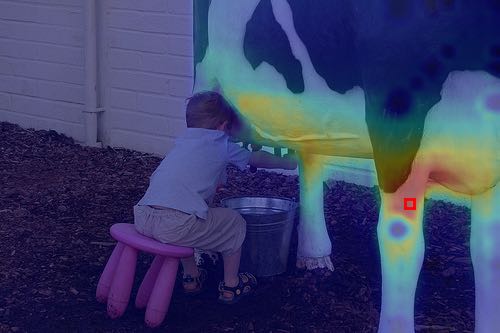}
        \end{subfigure} &
        \begin{subfigure}[b]{0.22\linewidth}
            \centering
            \includegraphics[width=\linewidth, height=3cm]{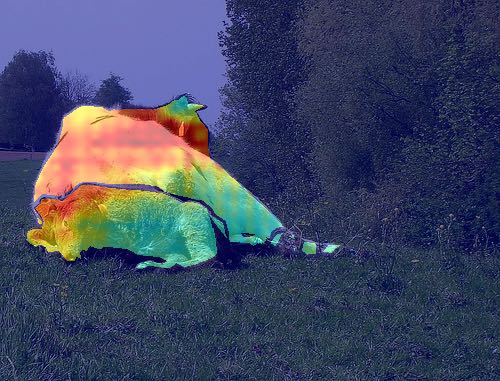}
        \end{subfigure} &
        \begin{subfigure}[b]{0.22\linewidth}
            \centering
            \includegraphics[width=\linewidth, height=3cm]{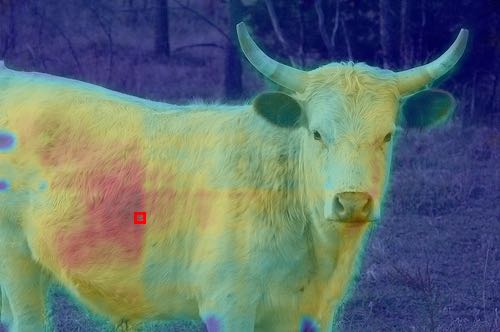}
        \end{subfigure} &
        \begin{subfigure}[b]{0.22\linewidth}
            \centering
            \includegraphics[width=\linewidth, height=3cm]{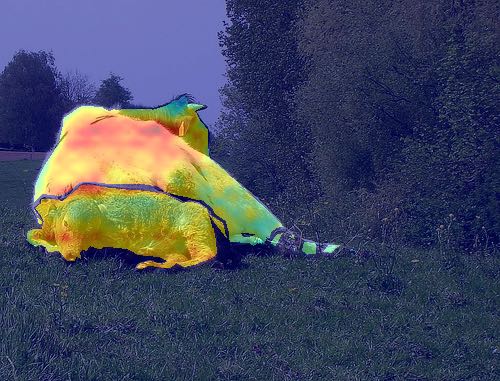}
        \end{subfigure}
    \end{tabular}
    \caption{Most activated prototypes represented by their training sample and their activation on the input image for Group 2 of class \textit{cow}.}
   \label{fig:proto-cow-2}
\end{figure*}

\begin{figure*}[htbp]
    \centering
    
    \begin{tabular}{cccc}
        \begin{subfigure}[b]{0.22\linewidth}
            \centering
            \includegraphics[width=\linewidth, height=3cm]{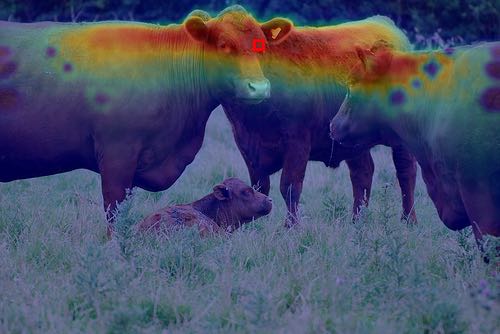}
        \end{subfigure} &
        \begin{subfigure}[b]{0.22\linewidth}
            \centering
            \includegraphics[width=\linewidth, height=3cm]{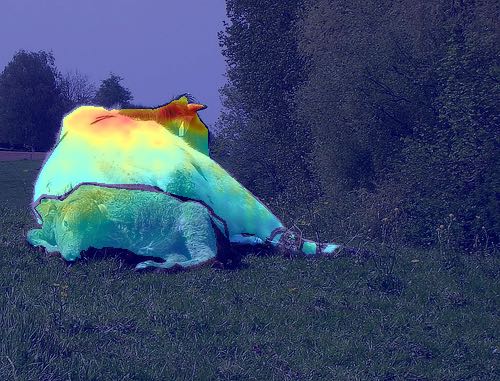}
        \end{subfigure} &
        \begin{subfigure}[b]{0.22\linewidth}
            \centering
            \includegraphics[width=\linewidth, height=3cm]{images/failure_1/Failure_1/prototype-img_scale_2_158-original_with_self_act_and_box.jpg}
        \end{subfigure} &
        \begin{subfigure}[b]{0.22\linewidth}
            \centering
            \includegraphics[width=\linewidth, height=3cm]{images/failure_1/Failure_1/plot_img_1181_proto_158_class_cow.jpg}
        \end{subfigure}
    \end{tabular}
    \caption{Most activated prototypes represented by their training sample and their activation on the input image for Group 3 of class \textit{cow}.}
   \label{fig:proto-cow-3}
\end{figure*}

\begin{figure*}[htbp]
    \centering
    
    \begin{tabular}{cccc}
        \begin{subfigure}[b]{0.22\linewidth}
            \centering
            \includegraphics[width=\linewidth, height=3cm]{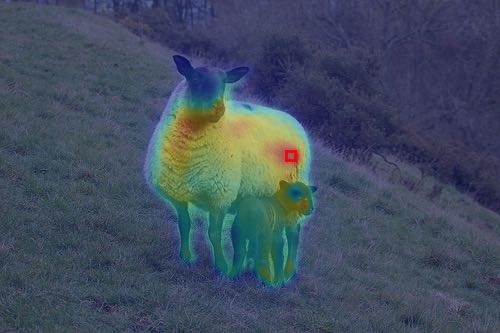}
        \end{subfigure} &
        \begin{subfigure}[b]{0.22\linewidth}
            \centering
            \includegraphics[width=\linewidth, height=3cm]{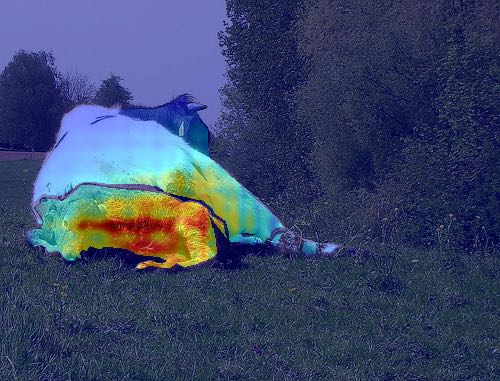}
        \end{subfigure} &
        \begin{subfigure}[b]{0.22\linewidth}
            \centering
            \includegraphics[width=\linewidth, height=3cm]{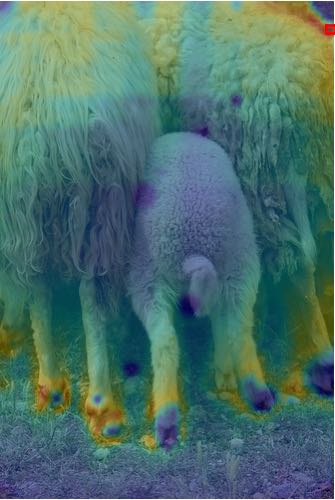}
        \end{subfigure} &
        \begin{subfigure}[b]{0.22\linewidth}
            \centering
            \includegraphics[width=\linewidth, height=3cm]{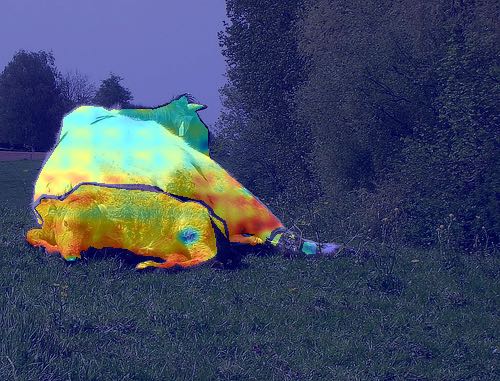}
        \end{subfigure}
    \end{tabular}
    \caption{Most activated prototypes represented by their training sample and their activation on the input image for Group 1 of class \textit{sheep}.}
    \label{fig:proto-sheep-1}
\end{figure*}

\begin{figure*}[htbp]
    \centering
    
    \begin{tabular}{cccc}
        \begin{subfigure}[b]{0.22\linewidth}
            \centering
            \includegraphics[width=\linewidth, height=3cm]{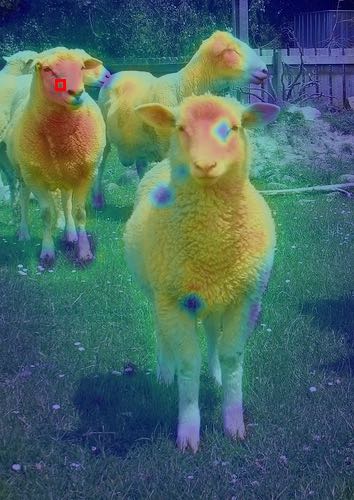}
        \end{subfigure} &
        \begin{subfigure}[b]{0.22\linewidth}
            \centering
            \includegraphics[width=\linewidth, height=3cm]{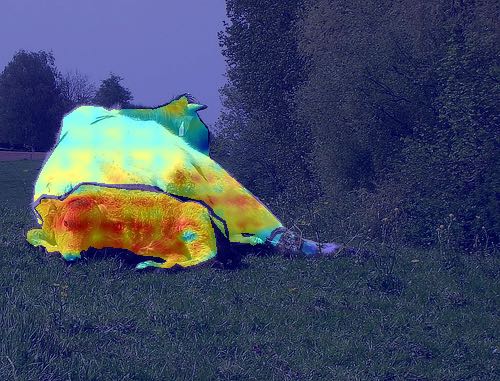}
        \end{subfigure} &
        \begin{subfigure}[b]{0.22\linewidth}
            \centering
            \includegraphics[width=\linewidth, height=3cm]{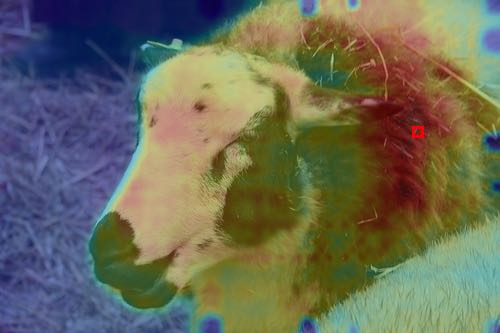}
        \end{subfigure} &
        \begin{subfigure}[b]{0.22\linewidth}
            \centering
            \includegraphics[width=\linewidth, height=3cm]{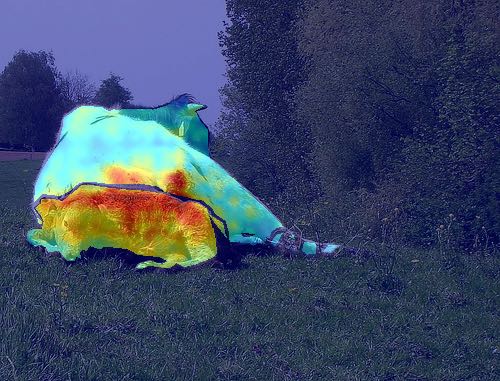}
        \end{subfigure}
    \end{tabular}
    \caption{Most activated prototypes represented by their training sample and their activation on the input image for Group 2 of class \textit{sheep}.}
    \label{fig:proto-sheep-2}
\end{figure*}

\begin{figure*}[htbp]
    \centering
    
    \begin{tabular}{cccc}
        \begin{subfigure}[b]{0.22\linewidth}
            \centering
            \includegraphics[width=\linewidth, height=3cm]{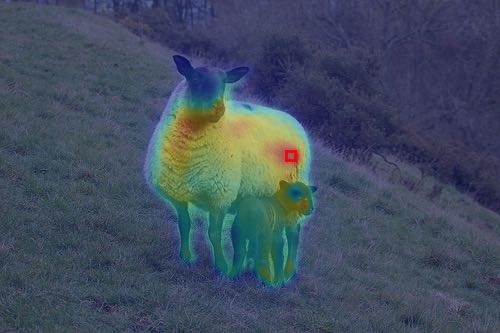}
        \end{subfigure} &
        \begin{subfigure}[b]{0.22\linewidth}
            \centering
            \includegraphics[width=\linewidth, height=3cm]{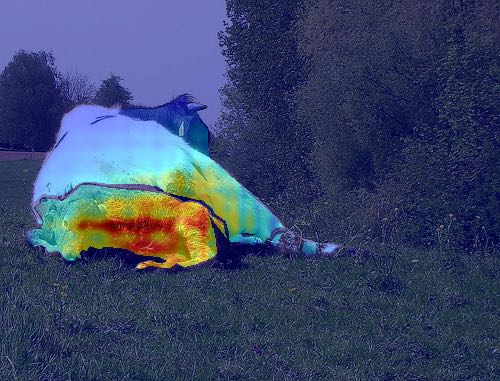}
        \end{subfigure} &
        \begin{subfigure}[b]{0.22\linewidth}
            \centering
            \includegraphics[width=\linewidth, height=3cm]{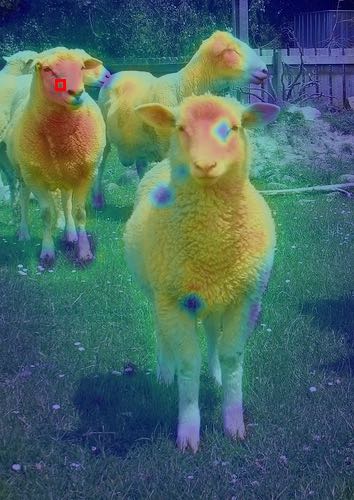}
        \end{subfigure} &
        \begin{subfigure}[b]{0.22\linewidth}
            \centering
            \includegraphics[width=\linewidth, height=3cm]{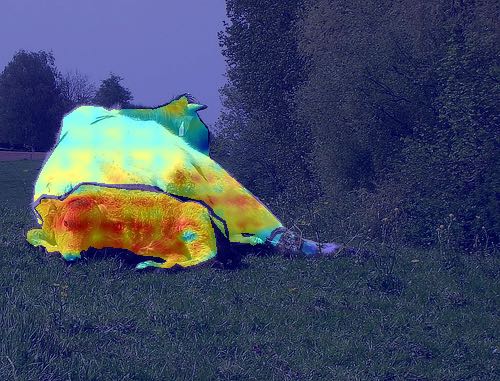}
        \end{subfigure}
    \end{tabular}
    \caption{Most activated prototypes represented by their training sample and their activation on the input image for Group 3 of class \textit{sheep}.}
    \label{fig:proto-sheep-3}
\end{figure*}

In this section, we analyze an image from PASCAL VOC where the IoU for a given class: cow, falls below a threshold set to $\epsilon_{IoU} = 0.2$, to describe how our method can be leveraged for interpretability analysis.
For this failure, we observe in Figure~\ref{fig:segmentation} that the veal in the image is classified as a sheep. Indeed, we observe in Figure~\ref{fig:group-cow} that for the group activations of the class cow, only the first group is slightly activated essentially on the second cow behind, while the 3 sheep groups activate both the top and bottom of the animal. We observe that all the main prototypes for the cow groups in Figure [~\ref{fig:proto-cow-1} - ~\ref{fig:proto-cow-3}] either activate a part not visible on the veal or seem to rather focus on the "texture" of the animal which can be misleading in this case. For the class sheep, we observed in Figure [~\ref{fig:proto-sheep-1} - ~\ref{fig:proto-sheep-3}] that the two most activated groups: 1 and 3, are driven by the same prototype which is a Scale 1 prototype focusing on texture. This seems reasonable knowing that the wool is a characteristic specific to this class and that a similar aspect can be seen for the veal.
To avoid learning misleading "texture" prototypes such as the one mentioned previously, the method presented in~\cite{bontempelli2022concept} can be leveraged to enforce forgetting such prototypes via human expert feedback.

\begin{figure*}[htbp]
    \centering
    
    \begin{tabular}{ccc}
        \begin{subfigure}[b]{0.3\linewidth}
            \centering
            \includegraphics[width=\linewidth]{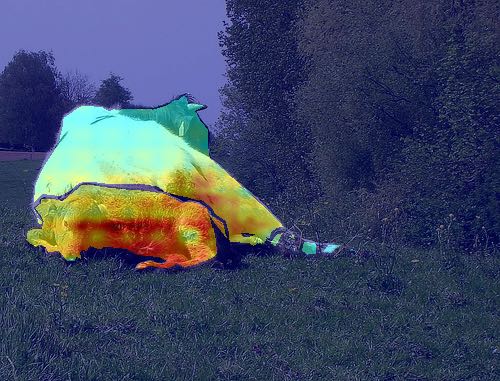}
        \end{subfigure} &
        \begin{subfigure}[b]{0.3\linewidth}
            \centering
            \includegraphics[width=\linewidth]{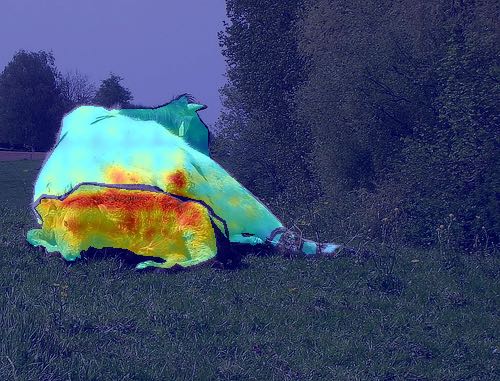}
        \end{subfigure} &
        \begin{subfigure}[b]{0.3\linewidth}
            \centering
            \includegraphics[width=\linewidth]{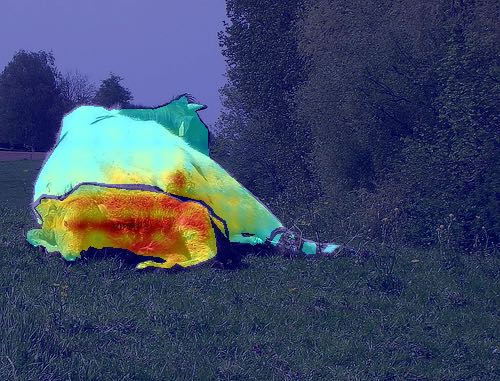}
        \end{subfigure}
    \end{tabular}
    \caption{Group activations of the class \textit{sheep} for the image.}
    \label{fig:group-sheep}
\end{figure*}

\end{document}